\documentclass[sigconf]{acmart}
\usepackage{booktabs}
\usepackage{bbm}
\usepackage{color}
\usepackage{array}

\newcommand{\PreserveBackslash}[1]{\let\temp=\\#1\let\\=\temp}
\newcolumntype{C}[1]{>{\PreserveBackslash\centering}p{#1}}
\newcolumntype{R}[1]{>{\PreserveBackslash\raggedleft}p{#1}}
\newcolumntype{L}[1]{>{\PreserveBackslash\raggedright}p{#1}}
\newcommand{\rotatetitle}[2]{\rotatebox[origin=l]{90}{\small\begin{tabular}{C{#1}}#2\end{tabular}}}

\AtBeginDocument{%
  \providecommand\BibTeX{{%
    \normalfont B\kern-0.5em{\scshape i\kern-0.25em b}\kern-0.8em\TeX}}}

\setcopyright{rightsretained}
\copyrightyear{2021}
\acmYear{2021}
\acmDOI{10.1145/3474085.3475589}

\acmConference[MM '21] {Proceedings of the 29th ACM International Conference on Multimedia}{October 20--24, 2021}{Virtual Event, China}
\acmBooktitle{Proceedings of the 29th ACM International Conference on Multimedia (MM '21), October 20--24, 2021, Virtual Event, China}
\acmPrice{15.00}
\acmISBN{978-1-4503-8651-7/21/10}

\begin{document}

\title{ChartPointFlow for Topology-Aware 3D Point Cloud Generation}

\author{Takumi Kimura}
\affiliation{%
  \institution{Graduate School of System Informatics, Kobe University}
  \city{Kobe}
  \country{Japan}
}
\email{kimura@ai.cs.kobe-u.ac.jp}

\author{Takashi Matsubara}
\affiliation{%
  \institution{Graduate School of Engineering Sciences, Osaka University}
  \city{Toyonaka}
  \country{Japan}
}
\email{matsubara@sys.es.osaka-u.ac.jp}

\author{Kuniaki Uehara}
\affiliation{%
  \institution{Faculty of Business Administration, Osaka Gakuin University}
  \city{Suita}
  \country{Japan}
}
\email{kuniaki.uehara@ogu.ac.jp}


\begin{abstract}
  A point cloud serves as a representation of the surface of a three-dimensional (3D) shape.
  Deep generative models have been adapted to model their variations typically using a map from a ball-like set of latent variables.
  However, previous approaches did not pay much attention to the topological structure of a point cloud, despite that a continuous map cannot express the varying numbers of holes and intersections.
  Moreover, a point cloud is often composed of multiple subparts, and it is also difficult to express.
  In this study, we propose ChartPointFlow, a flow-based generative model with multiple latent labels for 3D point clouds.
  Each label is assigned to points in an unsupervised manner.
  Then, a map conditioned on a label is assigned to a continuous subset of a point cloud, similar to a chart of a manifold.
  This enables our proposed model to preserve the topological structure with clear boundaries, whereas previous approaches tend to generate blurry point clouds and fail to generate holes.
  The experimental results demonstrate that ChartPointFlow achieves state-of-the-art performance in terms of generation and reconstruction compared with other point cloud generators.
  Moreover, ChartPointFlow divides an object into semantic subparts using charts, and it demonstrates  superior performance in case of unsupervised segmentation.
\end{abstract}



\maketitle


\section{Introduction}
A three-dimensional (3D) point cloud, which is a set of 3D locations in a Euclidean space, has gained popularity as a representation of a geometric shape~\cite{Klokov2017,Qi2017,Qi2017a,Su2018,Sun2019,Wang2019,Wang2020,Yan2020,Zaheer2017,Zhao2018} (see the survey~\cite{Guo2019a} for more details).
Specifically, the point cloud of an object's surface is easily acquired using sensors such as LiDARs and Kinects.
Point clouds can capture a much higher resolution than voxels, and can be processed using simpler manipulations than meshes.
By leveraging the flexibility of deep learning, a deep generative model of point clouds enables a variety of synthesis tasks such as generation, reconstruction, and super-resolution~\cite{Achlioptas2018,Arshad2020,Hui2020,Kim2020,Li2019,Ramasinghe2020,Shu2019,Valsesia2019,Yang2019}.
Because it is difficult to measure the quality of a generated point cloud numerically, most studies employ flow-based generative models~\cite{Dinh2017,grathwohl2019,Kingma2018} or generative adversarial networks (GANs)~\cite{Goodfellow2014}.
These methods learn a map that transforms a latent distribution into an object in the data space, and then they evaluate the object without a heuristic distance.

\begin{figure}[t]
  \centering
  \hspace*{1mm}\small\textbf{latent space}\hspace*{2.2cm}\textbf{data space}\\
  \rotatetitle{5cm}{\textbf{proposed}\hspace*{1.5cm}\textbf{existing}\\[2mm]}
  \includegraphics[width=6.5cm]{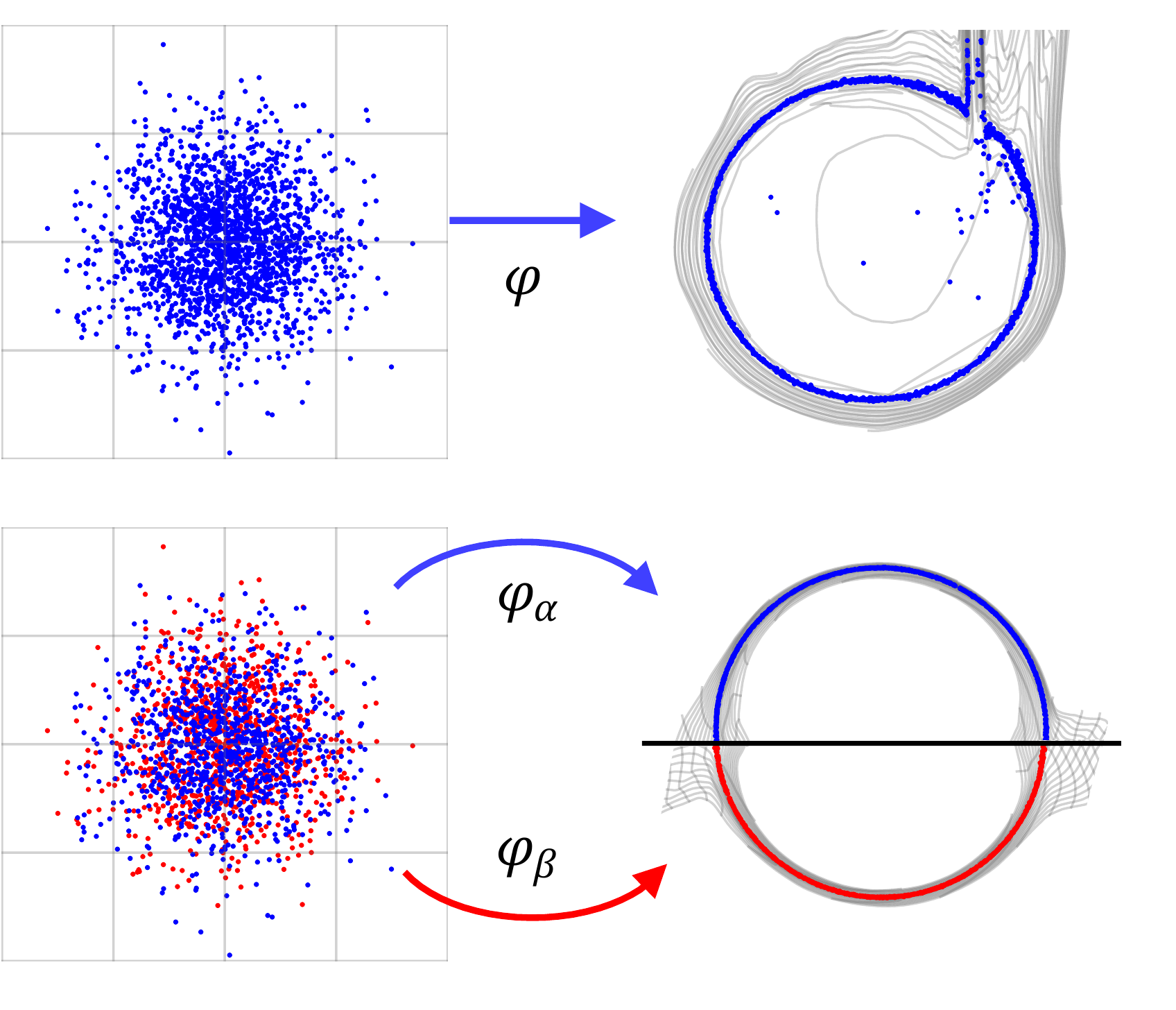}\\[-5mm]
  \caption{Conceptual comparison of existing methods (top) and the proposed method (bottom).}
  \label{fig:visualize_charts}
  \vspace*{-2mm}
\end{figure}

\begin{figure}
  \centering
  \hspace*{-3mm}
  \includegraphics[scale=0.25]{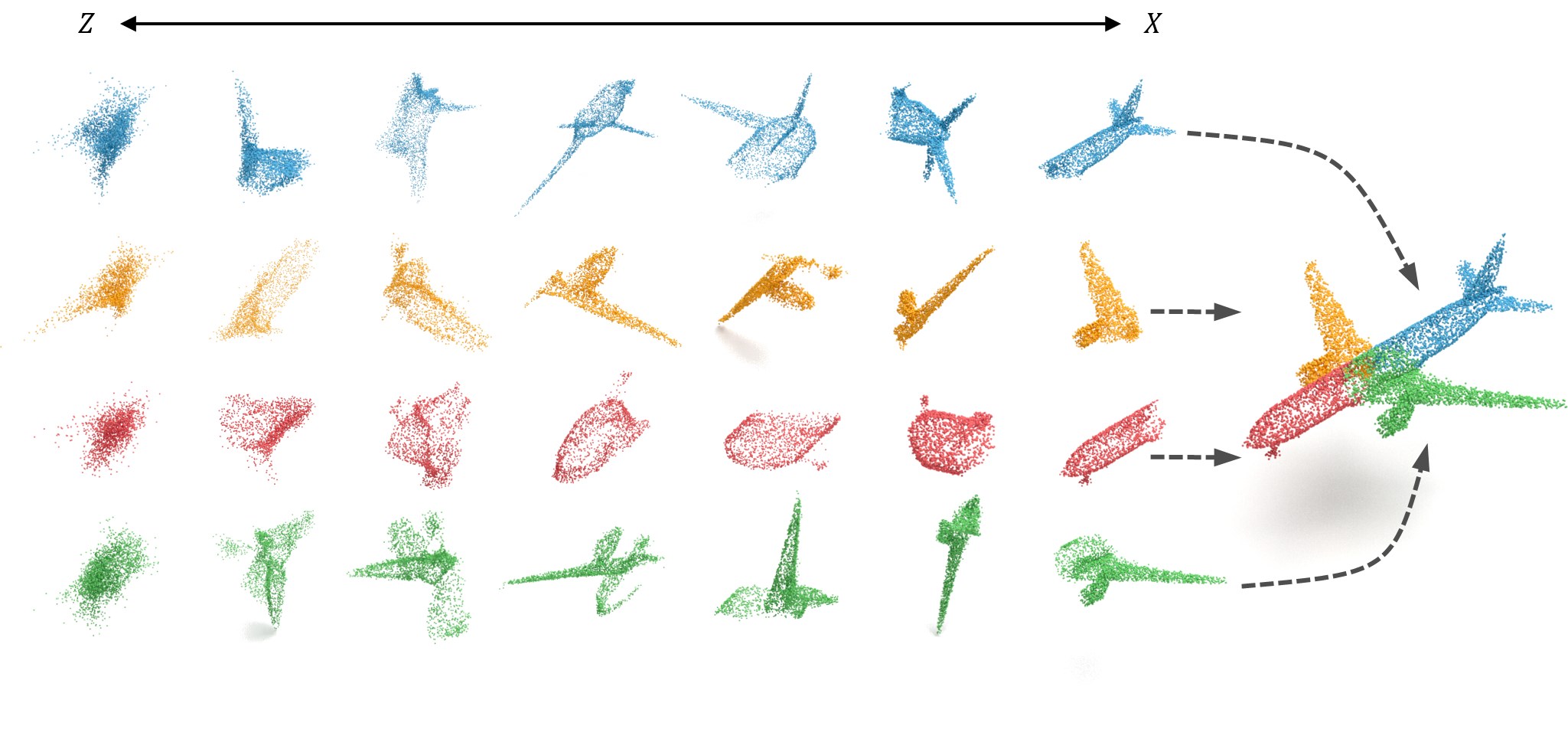}
  \vspace*{-12mm}
  \caption{Transformation from simple latent distributions (left) into an object (right).
    Each row corresponds to a chart.}
  \label{fig:sequence}
  \vspace*{-3mm}
\end{figure}

As a representation of an object's surface, a point cloud often has a thin, circular, or hollow structure~\cite{Luo2020}.
Flow-based generative models encounter a difficulty in expressing such manifold-like structures because a bijective map that is necessary for these models does not exist between a Euclidean space and a manifold with holes, as shown in the top panel of Fig.~\ref{fig:visualize_charts}.
To express a point cloud $X$ lying on the one-dimensional (1D) circle $S^{1}$, a map $\varphi$ modeled using a neural network squashes a two-dimensional (2D) ball in a latent space and stretches it to trace an arc, resulting in a discontinuity and outliers.
Several existing methods address a similar issue using a flow on a manifold or a dynamic chart method~\cite{Lou2020,Rezende2020}.
However, such methods are applicable only when the geometric property of the target manifold is known and fixed.
This assumption does not always hold for point cloud datasets of a variety of shapes.
Moreover, a point cloud is often composed of multiple subparts, some of which can be disconnected; additionally, it is difficult to express.
This is true for methods based on GANs and autoencoders (AEs) as well, as long as their neural networks are continuous.

Considering these drawbacks, we propose \emph{ChartPointFlow}, a generative model for 3D point clouds with latent labels.
Each label is assigned to points in unsupervised manner.
Then, a map conditioned on a label is assigned to a continuous subset of a given point cloud, similar to a chart of a manifold, and a set of charts forms an atlas that covers the entire point cloud.
Taking Fig.~\ref{fig:visualize_charts} as an example, ChartPointFlow with two charts, namely, $\varphi_\alpha$ and $\varphi_\beta$, generates two arcs separately and concatenates them in the data space, thereby generating a continuous and hollow circle.
For a more complex object, each chart is assigned to a semantic subpart, e.g., the airframe, right wing, nose, and left wing of an airplane, as shown in Fig.~\ref{fig:sequence}.
From the perspective of the generative model, ChartPointFlow with $n$ labels provides a mixture of $n$ distributions.

Furthermore, we evaluate ChartPointFlow through its performance on synthetic datasets and ShapeNet dataset~\cite{Chang2015} of point clouds.
The experiments demonstrate that ChartPointFlow preserves the topological structure in detail, whereas previous approaches tend to generate blurry point clouds and fail to generate holes.
Numerical results demonstrate that ChartPointFlow outperforms other state-of-the-art point cloud generators, such as r-GAN~\cite{Achlioptas2018}, l-GAN~\cite{Achlioptas2018}, PC-GAN~\cite{Li2019}, ShapeGF~\cite{Cai2020} , PointFlow~\cite{Yang2019}, SoftFlow~\cite{Kim2020}, AtlasNet~\cite{Groueix2018}, AtlasNet V2~\cite{Deprelle2019}, tree-GAN~\cite{Shu2019}, and GCN-GAN~\cite{Valsesia2019}.
In terms of reconstruction and unsupervised semantic segmentation, ChartPointFlow outperforms AtlasNet~\cite{Groueix2018} and AtlasNet V2~\cite{Deprelle2019}, which are based on AEs and share the concept of charts and atlases.


\section{Related Work}\label{sec:related_work}
\noindent\textbf{Deep Learning on Point Clouds:}\
A point cloud is composed of points in no particular order.
PointNet takes each point separately and performs a permutation-invariant operation (max-pooling), thereby obtaining the global feature~\cite{Qi2017}.
Following PointNet, many studies focused on classification and segmentation tasks~\cite{Klokov2017,Qi2017,Qi2017a,Su2018,Sun2019,Wang2019,Wang2020,Yan2020,Zaheer2017,Zhao2018}.

\vspace*{1mm}\noindent\textbf{Likelihood-based Point Cloud Generation:}\
One of the earliest models for point cloud generation is MR-VAE~\cite{Gadelha2018}, which is based on a variational AE (VAE).
A VAE is a probabilistic model that is implemented using two neural networks, namely a decoder that generates a sample and an encoder that performs the variational inference of the latent variable~\cite{Rezende2015}.
MR-VAE was trained to minimize a heuristic distance between real and generated point clouds.
Zamorski et al.~\cite{Zamorski2020} employed an adversarial AE to regularize the latent variables.
Liu et al.~\cite{Liu2019} employed a recurrent neural network to generate a point cloud step-by-step.
Instead of an AE, Cai et al.~\cite{Cai2020} proposed ShapeGF, which used an implicit function defined using a neural network.

Yang et al.~\cite{Yang2019} proposed PointFlow, which is a combination of a permutation-invariant encoder and a point-wise flow-based generative model.
A flow-based generative model is a neural network that forms a bijective map and obtains a likelihood using the change of variables without a heuristic distance~\cite{Dinh2017,grathwohl2019,Kingma2018}.
Moreover, this model can accept and generate an arbitrary number of points.

Because no bijective map exists between manifolds of different topologies, a flow-based generative model tends to be destabilized when modeling zero-width structures, such as a surface.
This is often the case with point clouds.
Kim et al.~\cite{Kim2020} proposed SoftFlow to address this issue by adding perturbations to points at the training phase.
SoftFlow emphasizes the importance of the topology, but remains inapplicable to general topological structures, such as holes, intersections, and disconnections.
ChartPointFlow addresses this problem by using charts.

\vspace*{1mm}\noindent\textbf{Likelihood-free Point Cloud Generation:}\
Another group of models for point cloud generation involves those based on GANs.
A GAN comprises a pair of neural networks, namely, a generator that outputs artificial samples and a discriminator that evaluates their similarity to real samples without a heuristic distance or an explicit likelihood~\cite{Goodfellow2014}.
r-GAN generates all the points of a point cloud simultaneously~\cite{Achlioptas2018}.
l-GAN applies a GAN to the feature vector extracted by a pretrained AE~\cite{Achlioptas2018}.
PC-GAN employs a permutation-invariant generator~\cite{Li2019}.
Spectral-GAN handles point clouds in the spectral domain~\cite{Ramasinghe2020}.

Other GAN-based approaches can be regarded as recursive super-resolutions.
Each model first generates a sparse point cloud, and then it adds more points to interpolate the existing ones repeatedly~\cite{Arshad2020,Hui2020,Shu2019,Valsesia2019}.
Valsesia et al.~\cite{Valsesia2019} found that the points close to each other have similar feature vectors.
Shu et al.~\cite{Shu2019} also found that each point generated at the first step may be associated with a semantic subpart of the point cloud.
These results demonstrate the importance of semantic subparts.
However, the above-mentioned studies do not deal with subparts explicitly.

\vspace*{1mm}\noindent\textbf{Generative Model with Labels:}\
For modeling samples of multiple categories, deep learning-based generative models have been extended to mixture distributions, such as conditional VAEs~\cite{Kingma2014a}, conditional GANs~\cite{Mirza2014}, and conditional flow-based generative models~\cite{Dinh2017,Kingma2018,Klokov2020}.
The condition represents the class label that an image or object belongs to.
In contrast, ChartPointFlow divides each point in a single object into a class.
As shown in Fig.~\ref{fig:visualize_charts}, existing generative models encounter a difficulty in expressing a single cluster if the cluster has a different topology.

AtlasNet~\cite{Groueix2018} and AtlasNet V2~\cite{Deprelle2019} share the concept of charts and atlases with ChartPointFlow.
However, they assume to express all objects in the same category using a fixed number of fixed-size charts.
This assumption is unnatural when the objects' shapes vary widely.
For example, the topology of a chair with armrests is different from that of a chair without armrests.
In contrast, ChartPointFlow resizes charts and discards unnecessary charts by inferring the occurrence probability of each chart from a given object shape.
Although AltasNets aim to reconstruct point clouds, they cannot generate point clouds without modification.
AtlasNets are based on ordinary neural networks, which approximate arbitrary functions.
ChartPointFlow employs a flow-based generative model, which approximates only bijective functions~\cite{Teshima2020}.
Compared with AtlasNets, ChartPointFlow has an architecture that is more consistent with the definition of charts.
Luo and Hu~\cite{Luo2020} also introduced a similar concept for denoising.


\begin{figure}[t]
  \setlength{\tabcolsep}{0mm}
  \hspace*{-1.5mm}\begin{tabular}{L{2.5mm}C{13.9mm}C{13.9mm}C{13.9mm}C{13.9mm}C{13.9mm}C{13.9mm}}
    & \scriptsize\textbf{Data} & \scriptsize\textbf{PointFlow} {\scriptsize(Glow)} & \scriptsize\textbf{SoftFlow} \hspace*{0.5mm}{\scriptsize(Glow)}& \scriptsize\textbf{PointFlow} \scriptsize{(FFJORD)}& \scriptsize\textbf{SoftFlow} {\scriptsize(FFJORD)}& \scriptsize\textbf{ChartPointFlow} {\scriptsize\mbox{(proposed,~Glow)}}\\[-3mm]
    \rotatetitle{18mm}{\hspace*{-4mm}\footnotesize\textbf{circle}} & \includegraphics[width=15mm]{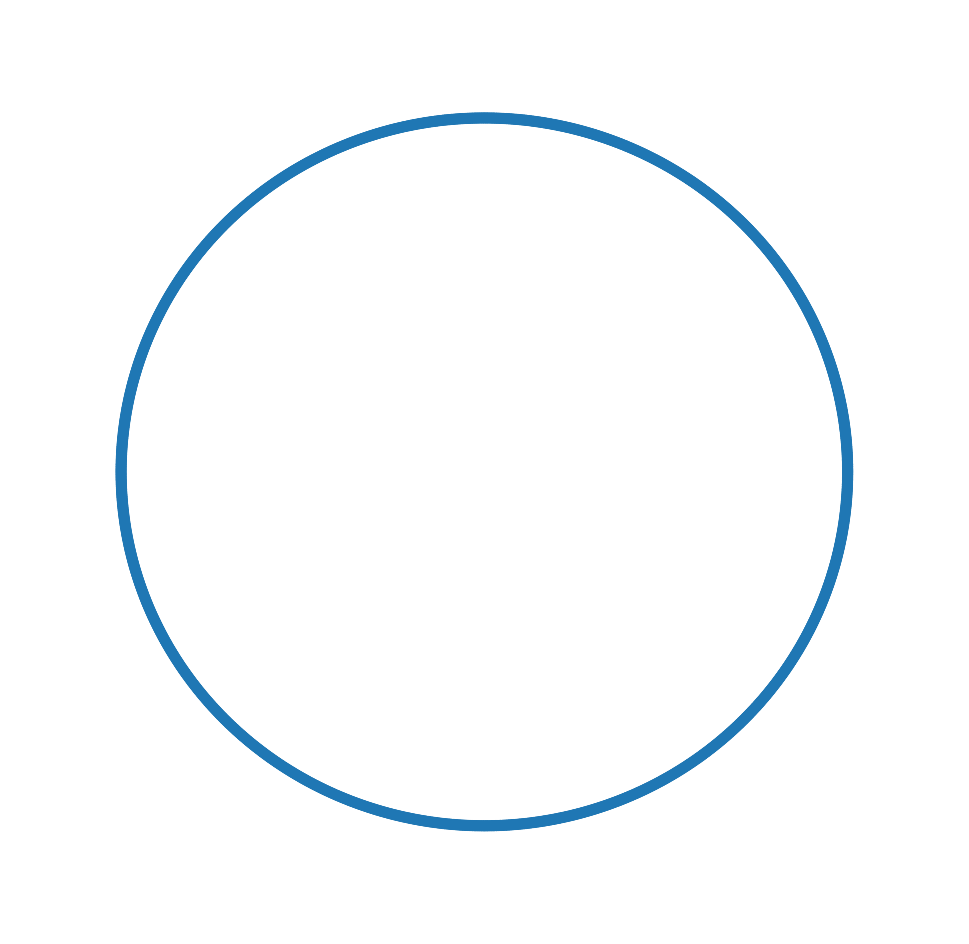} & \includegraphics[width=15mm]{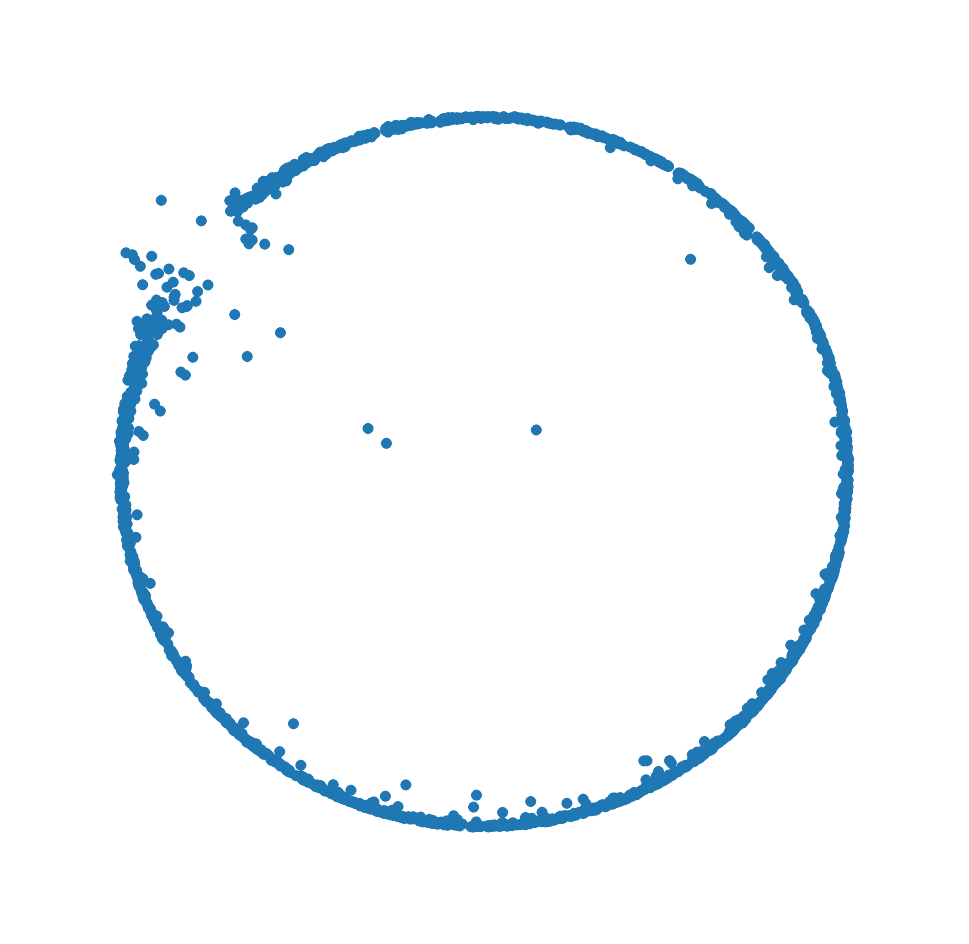} & \includegraphics[width=15mm]{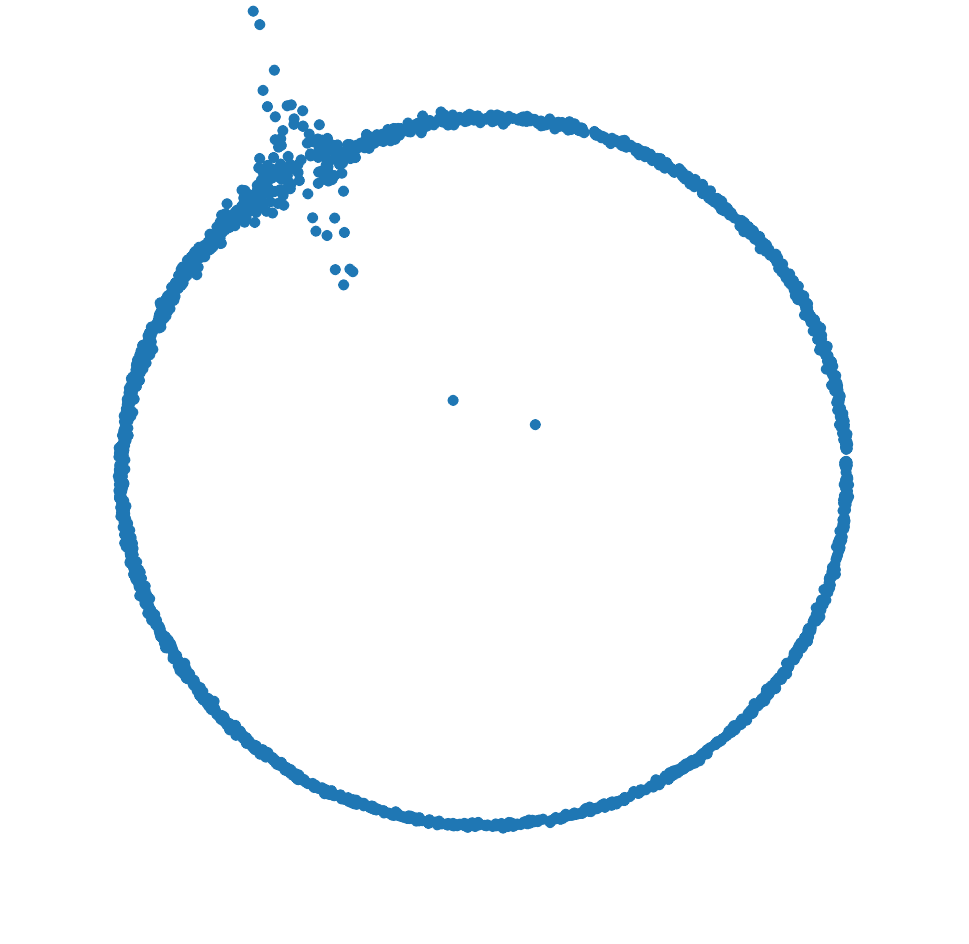} & \includegraphics[width=15mm]{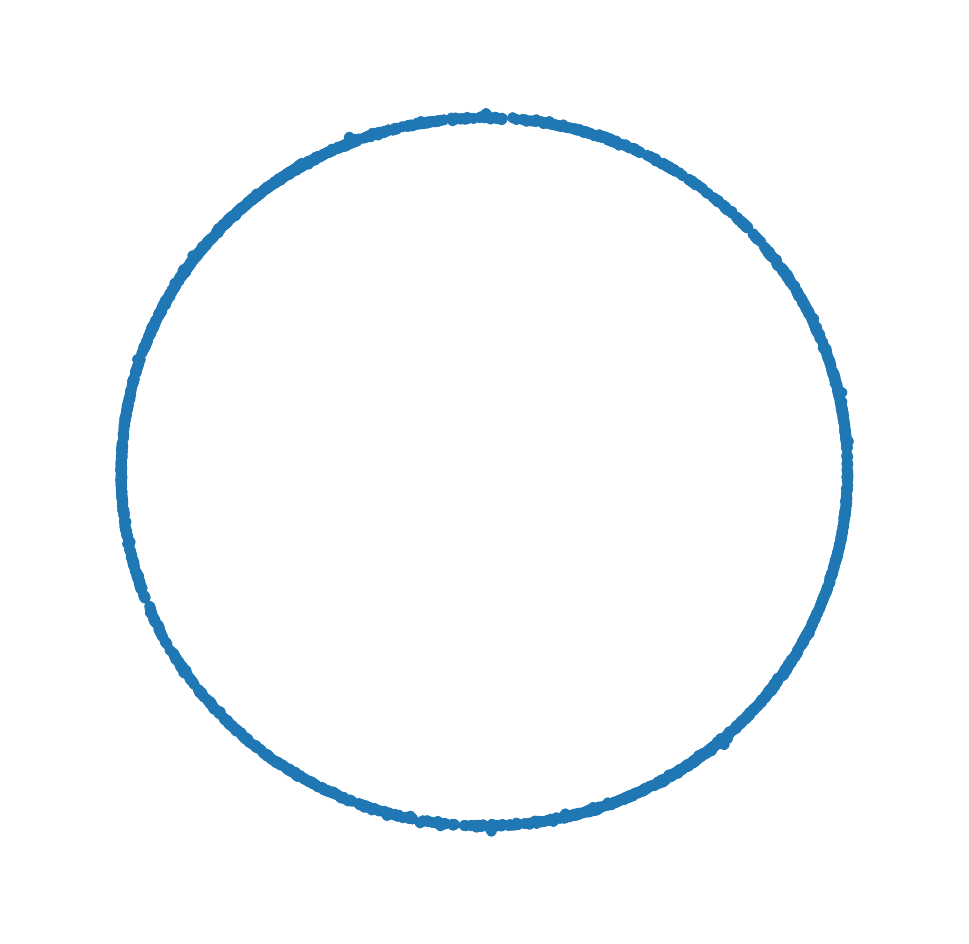} & \includegraphics[width=15mm]{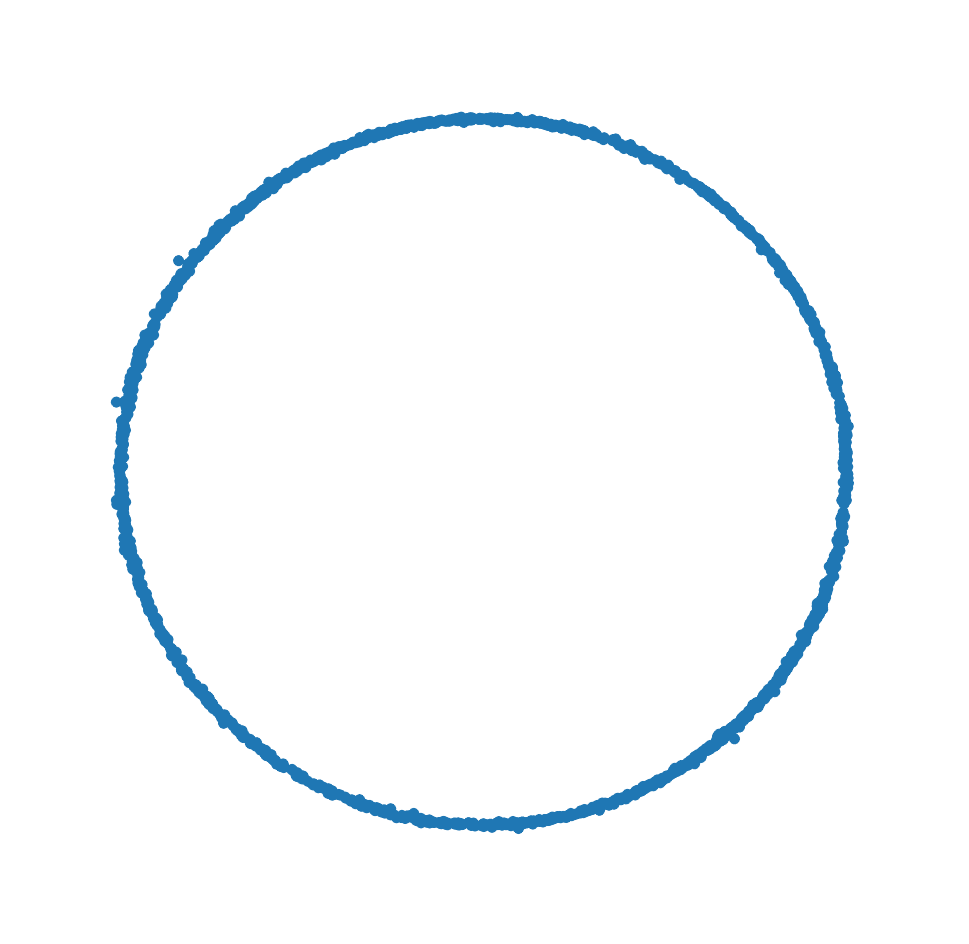} & \includegraphics[width=15mm]{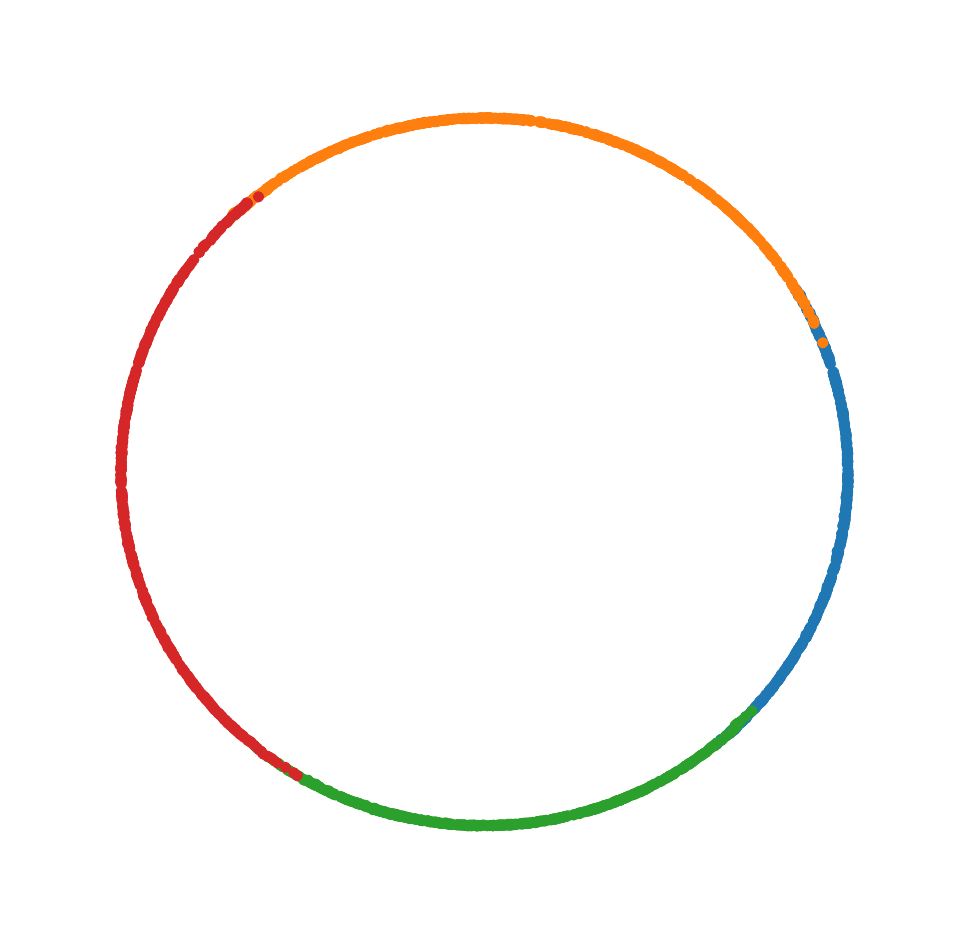} \\[-3mm]
    \rotatetitle{18mm}{\hspace*{-4mm}\footnotesize\textbf{2sines}} & \includegraphics[width=15mm]{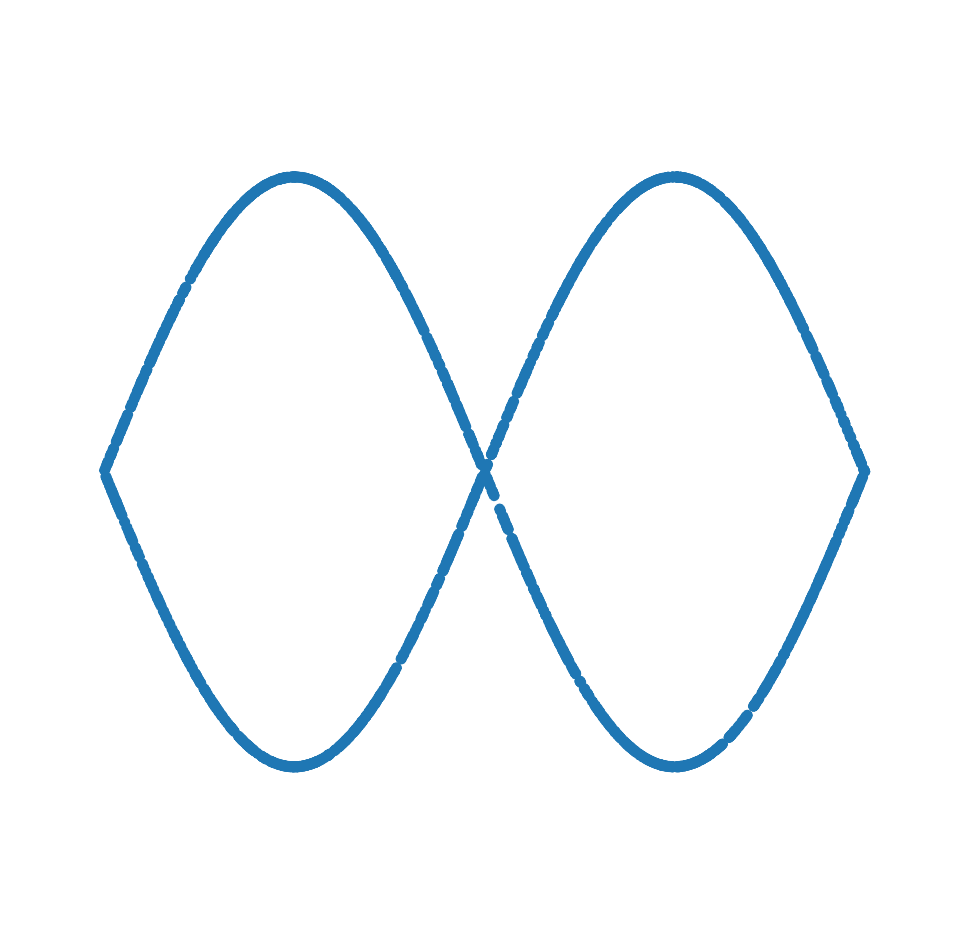} & \includegraphics[width=15mm]{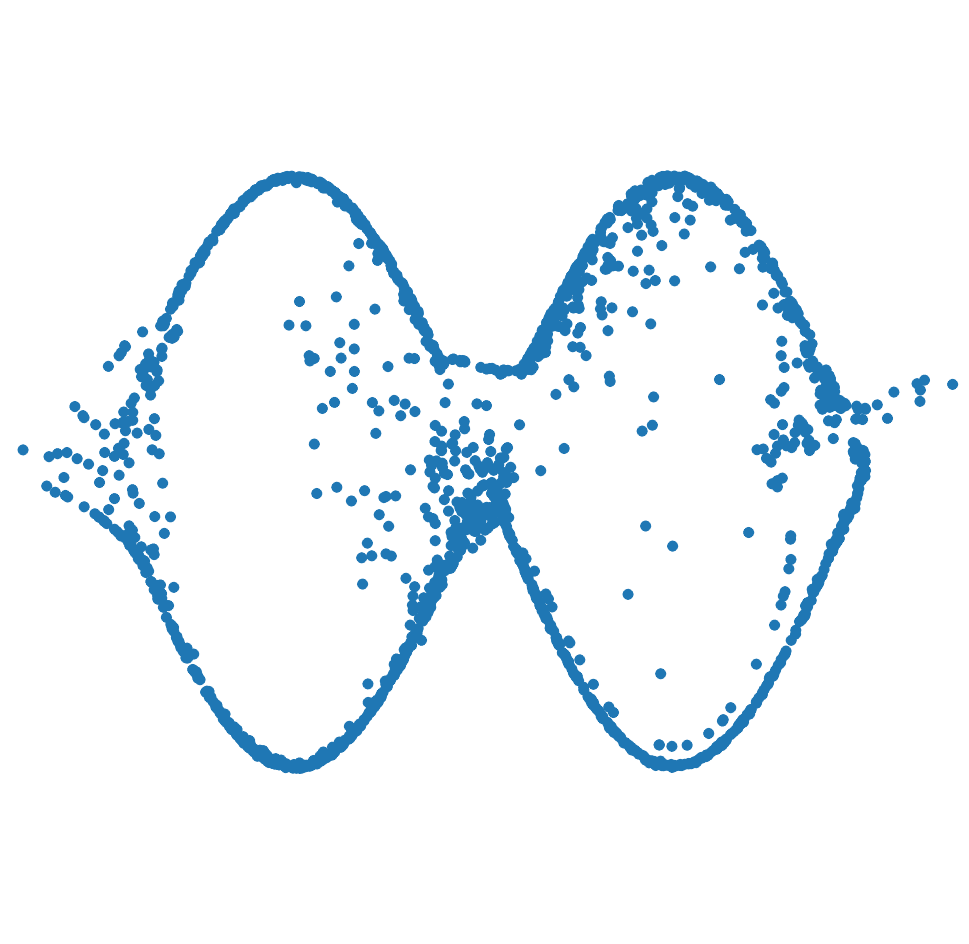} & \includegraphics[width=15mm]{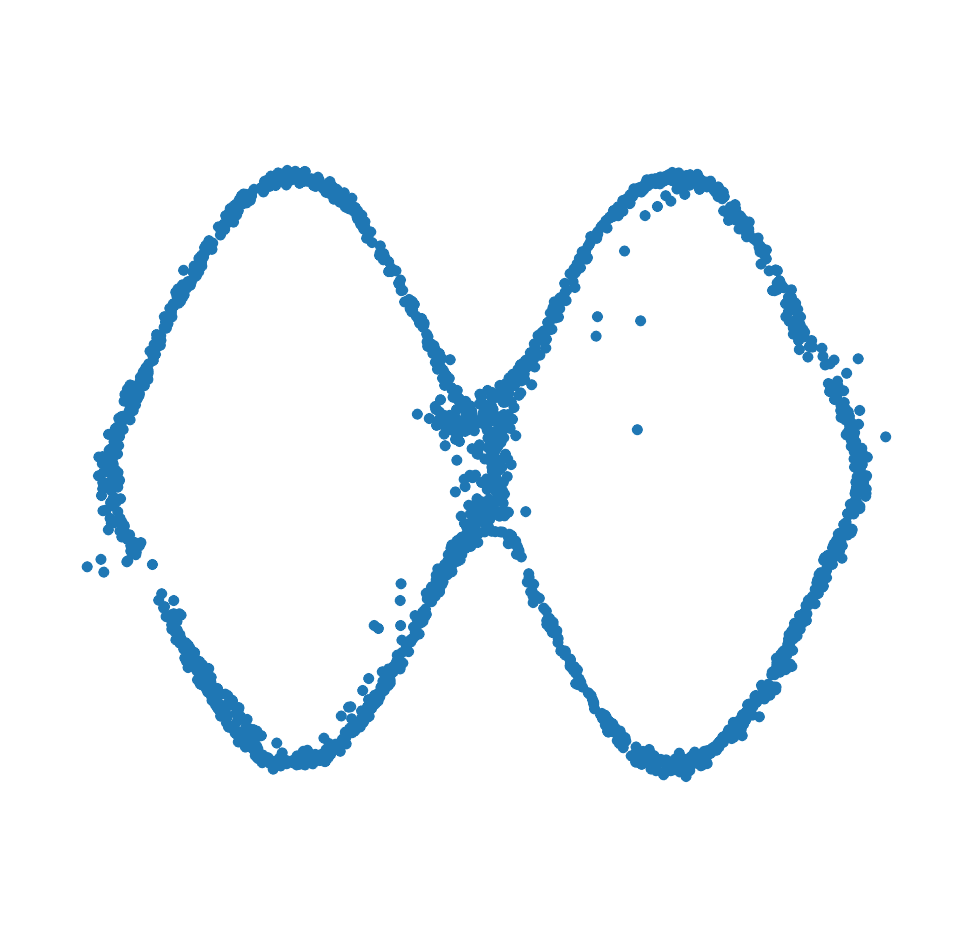} & \includegraphics[width=15mm]{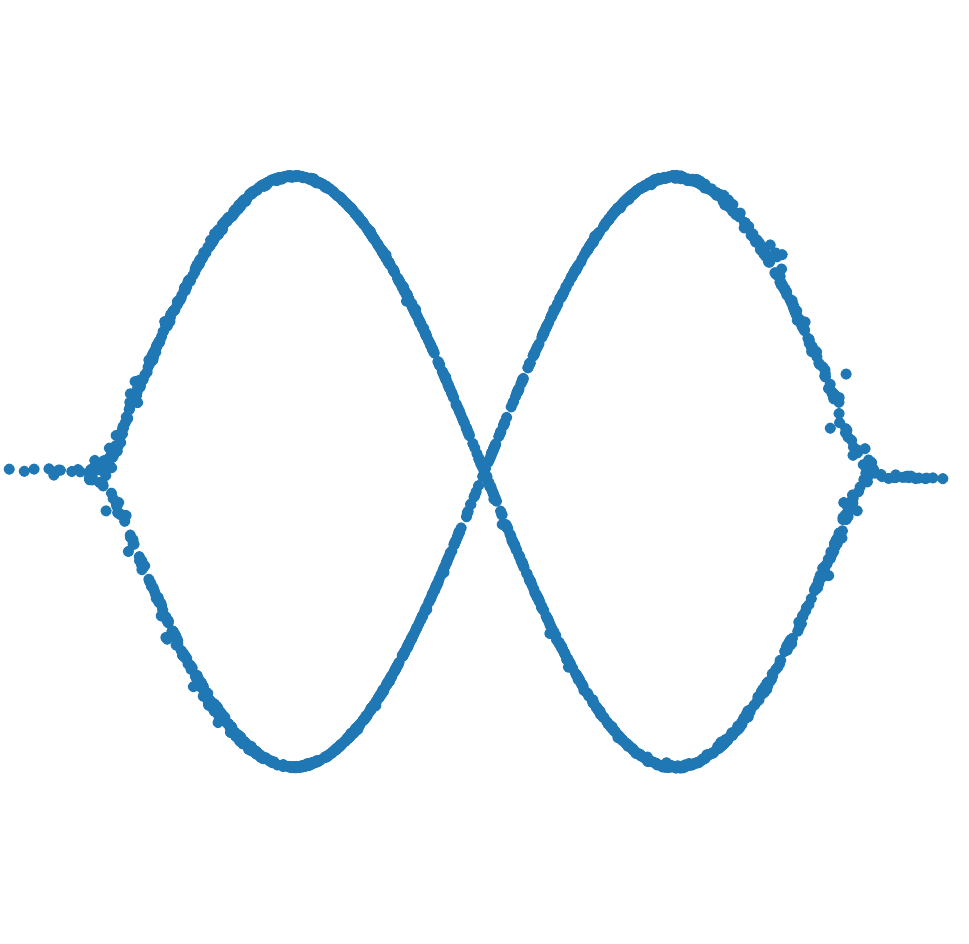} & \includegraphics[width=15mm]{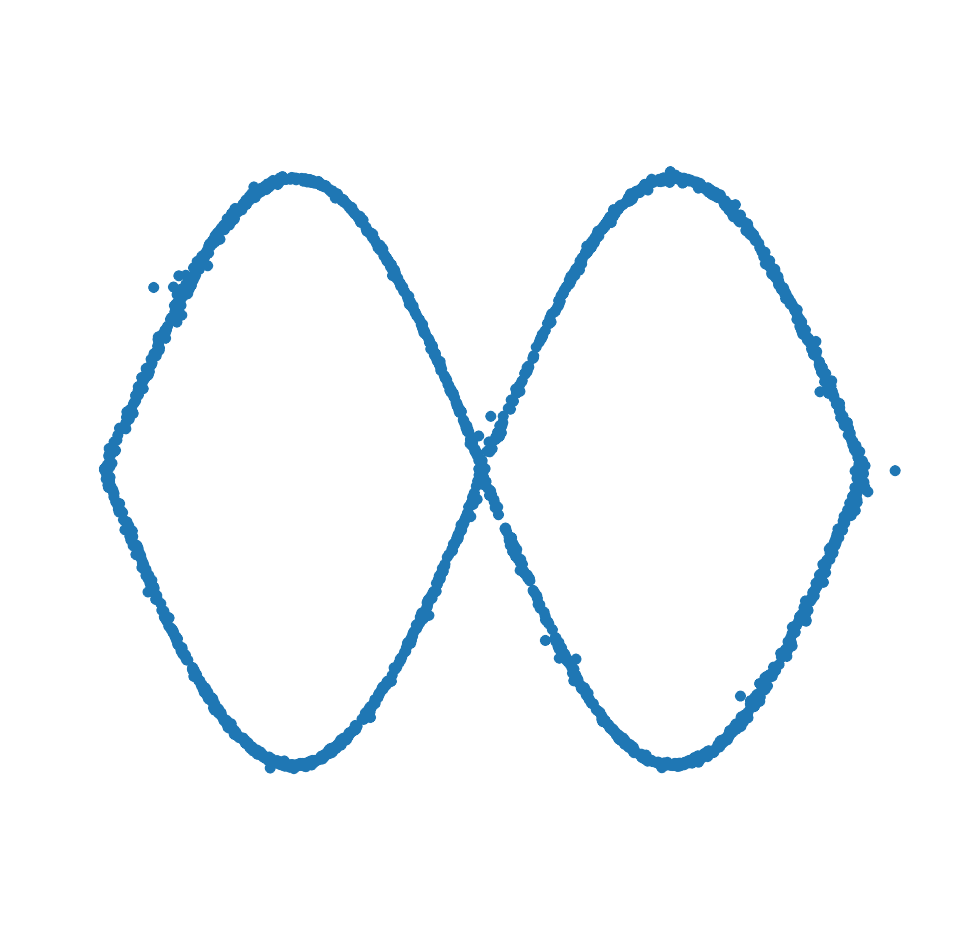} & \includegraphics[width=15mm]{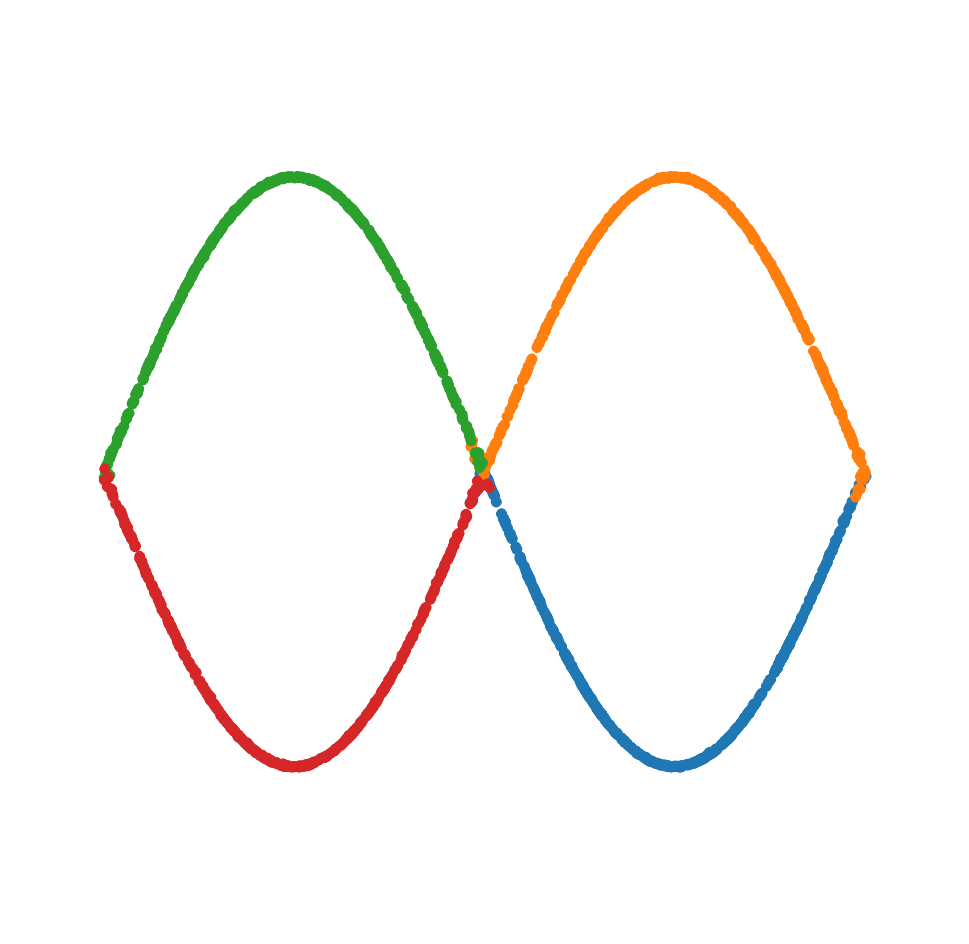} \\[-3mm]
    \rotatetitle{18mm}{\hspace*{-4mm}\footnotesize\textbf{four-circle}} & \includegraphics[width=15mm]{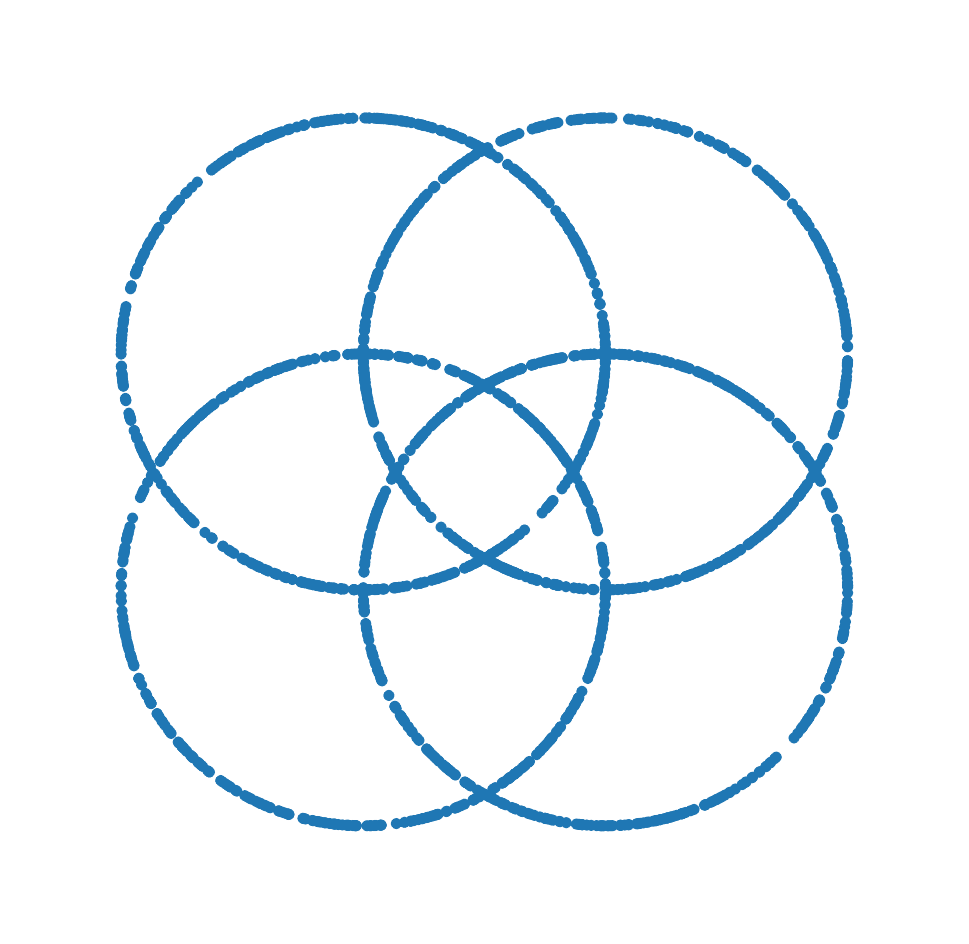} & \includegraphics[width=15mm]{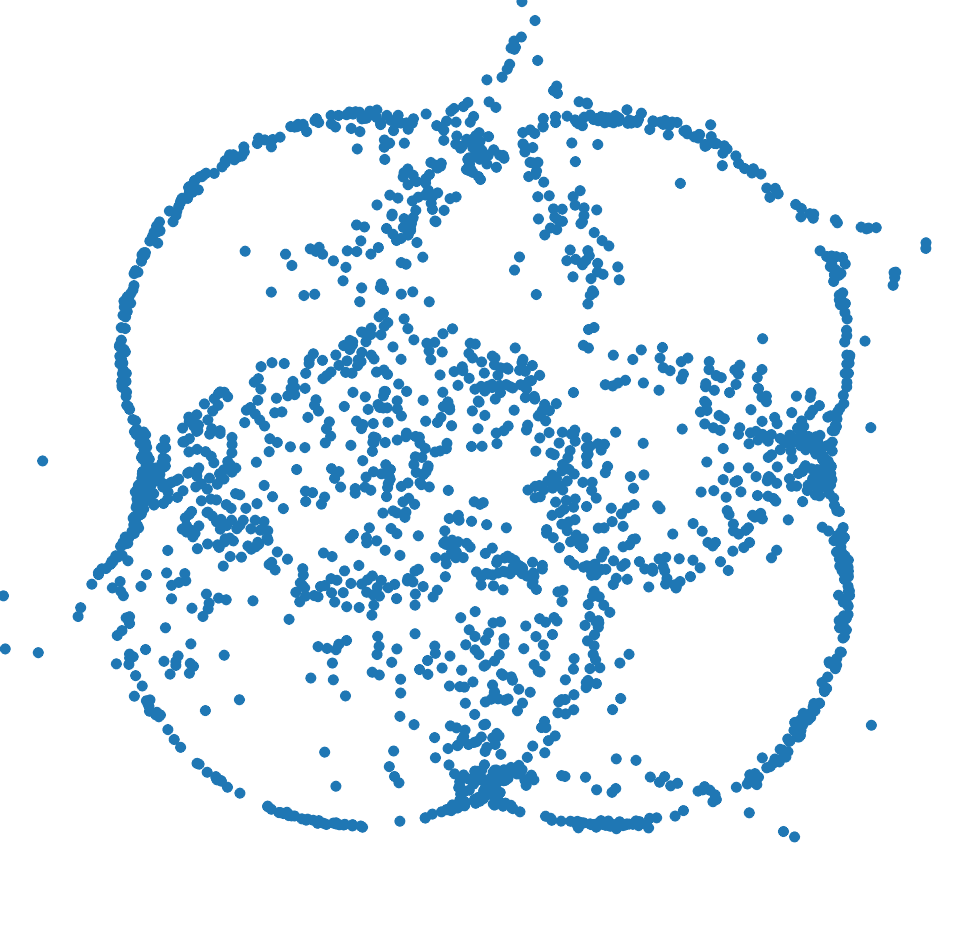} & \includegraphics[width=15mm]{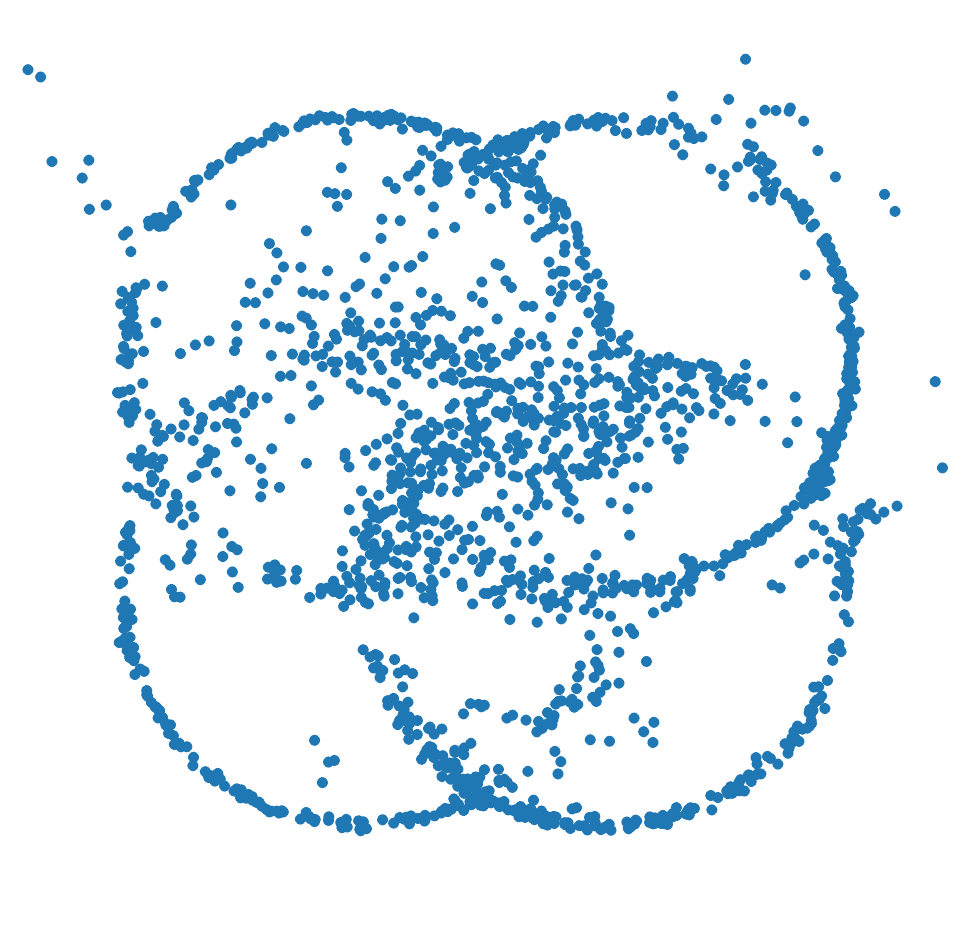} & \includegraphics[width=15mm]{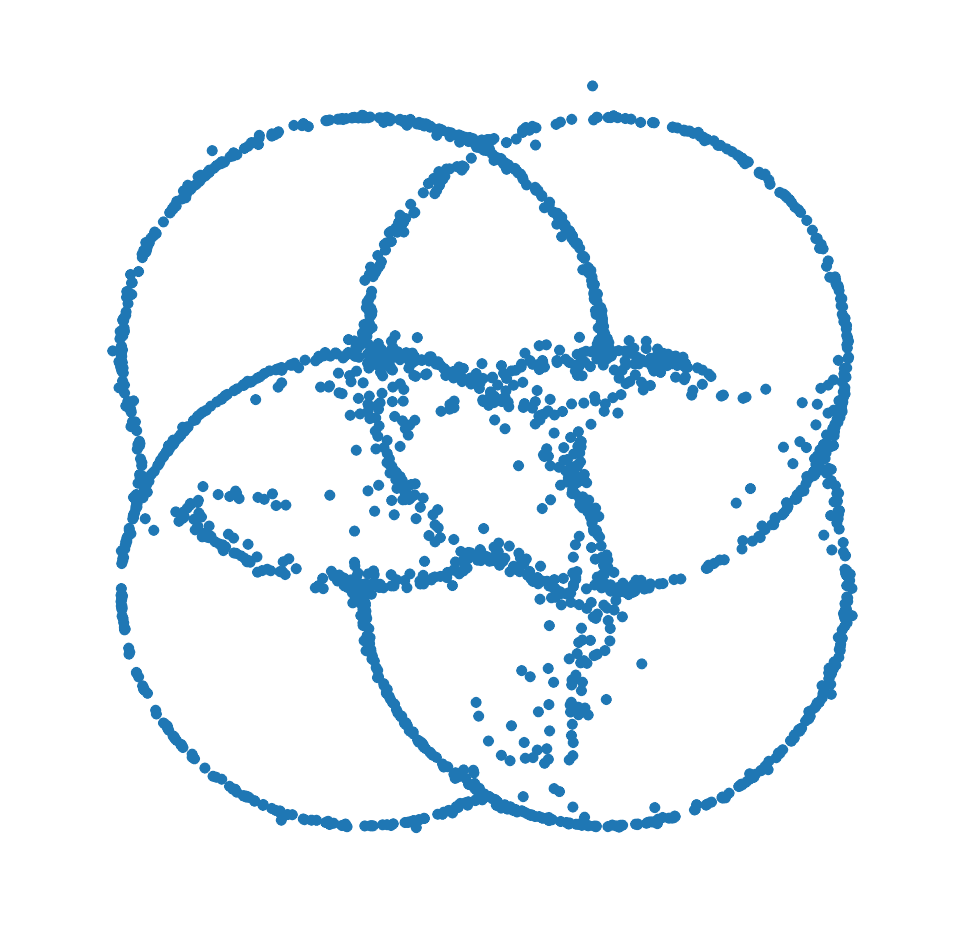} &  \includegraphics[width=15mm]{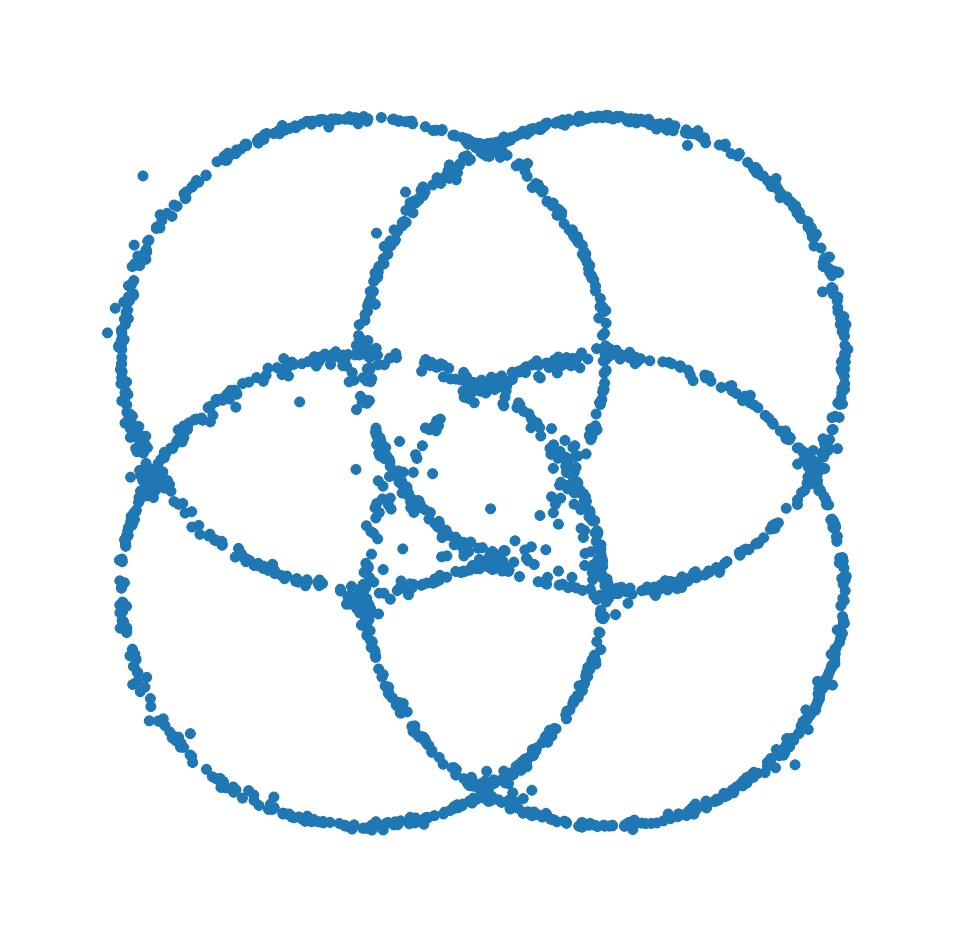} & \includegraphics[width=15mm]{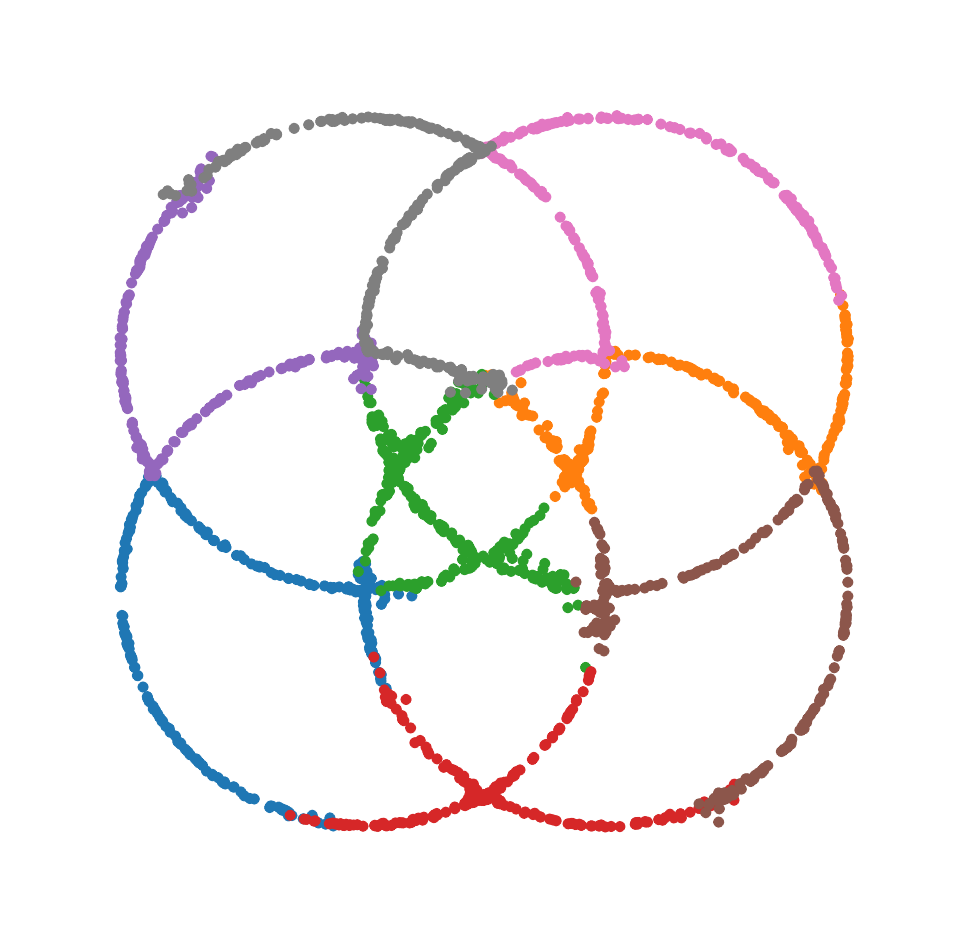} \\[-3mm]
    \rotatetitle{18mm}{\hspace*{-4mm}\footnotesize\textbf{double-moon}} & \includegraphics[width=15mm]{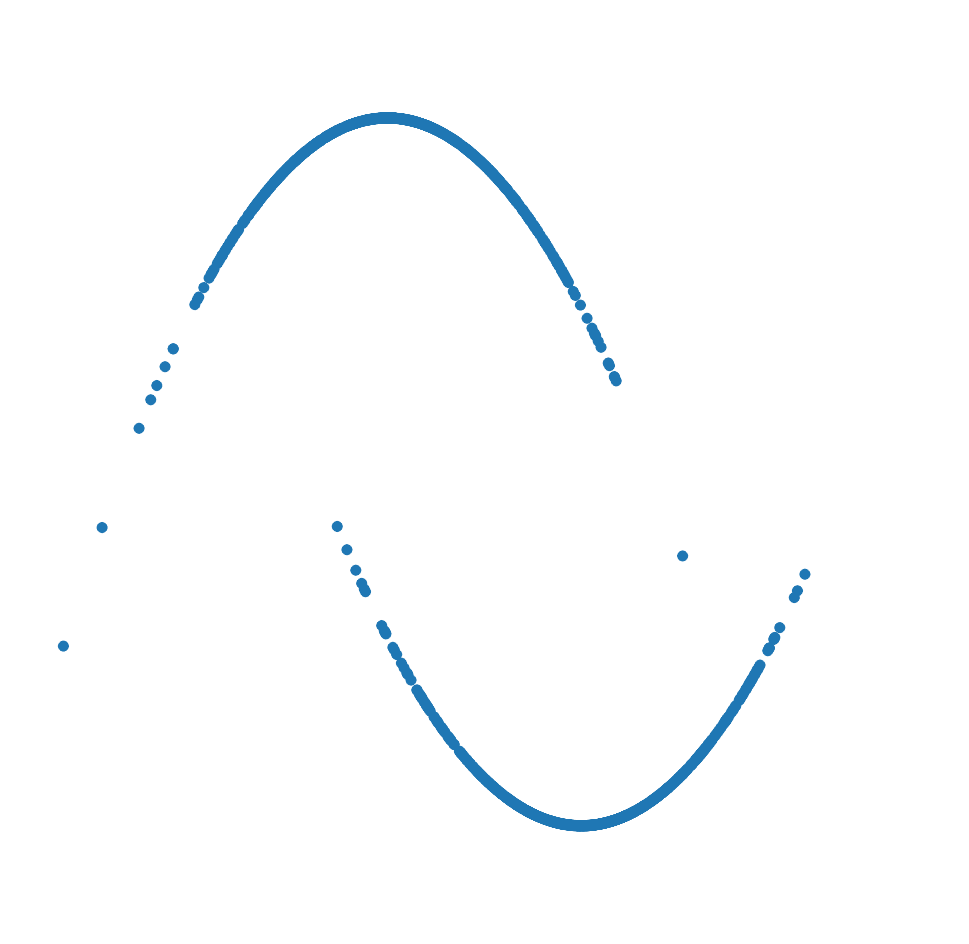} & \includegraphics[width=15mm]{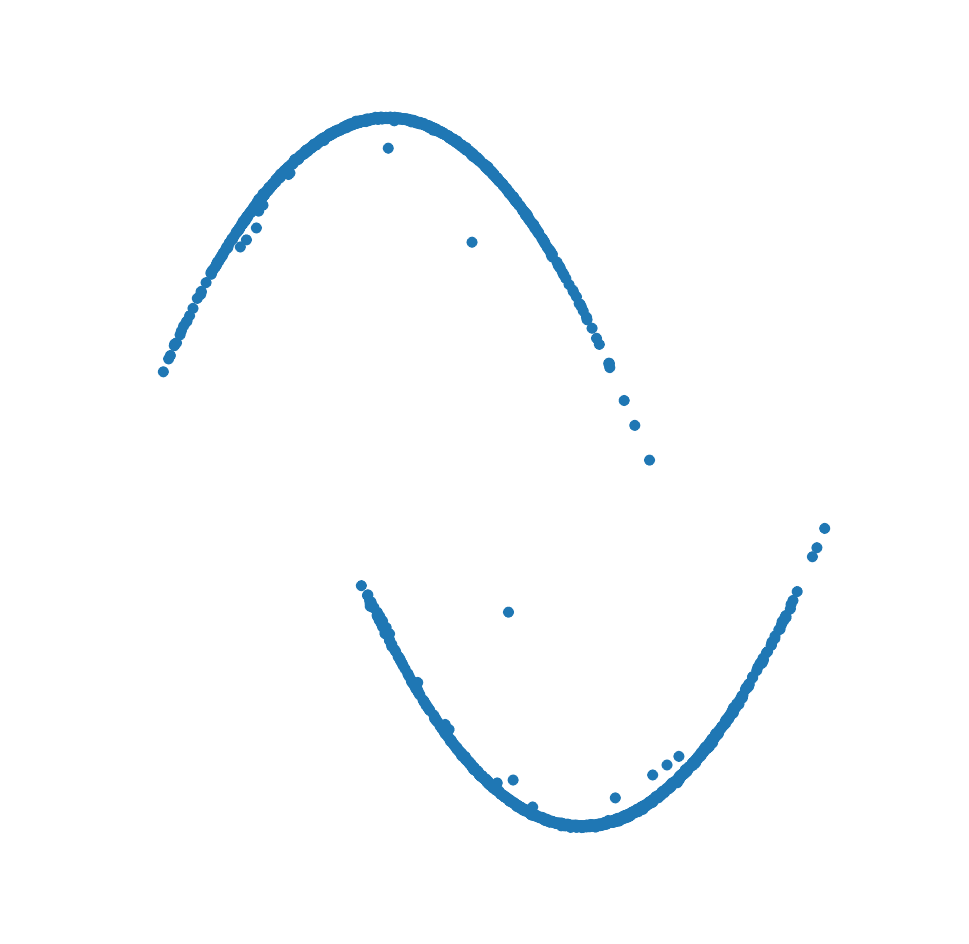} & \includegraphics[width=15mm]{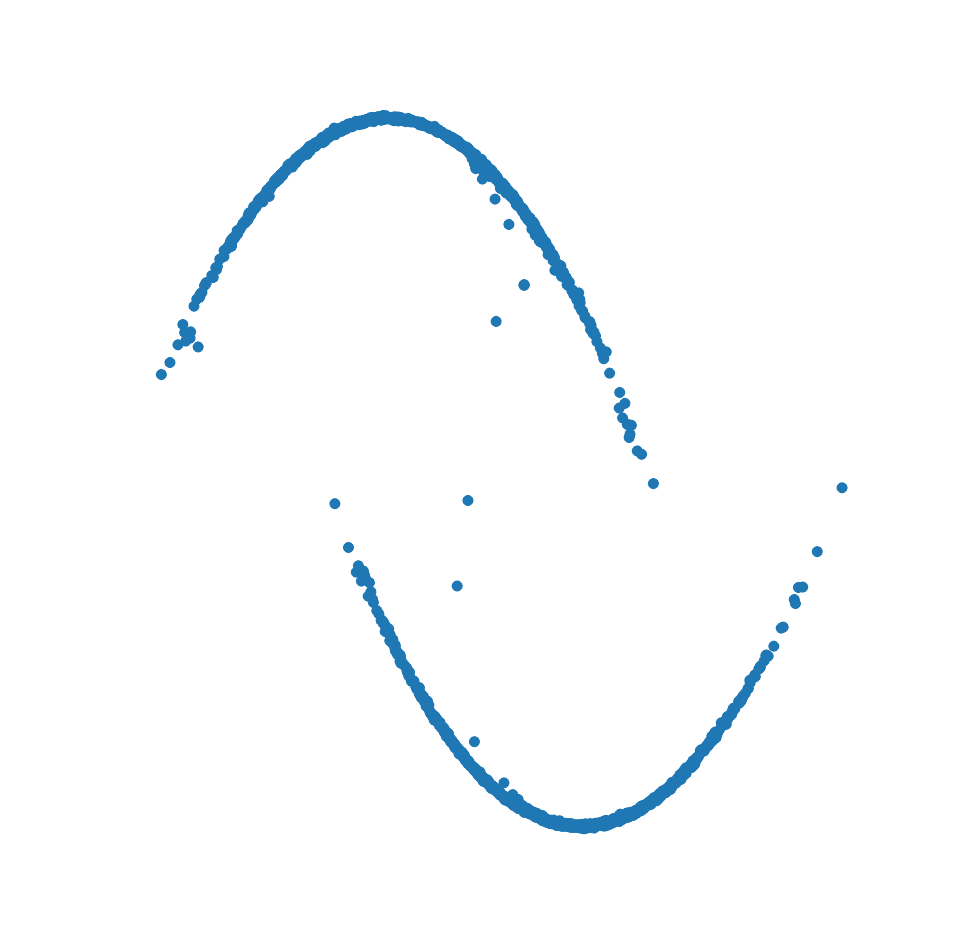} & \includegraphics[width=15mm]{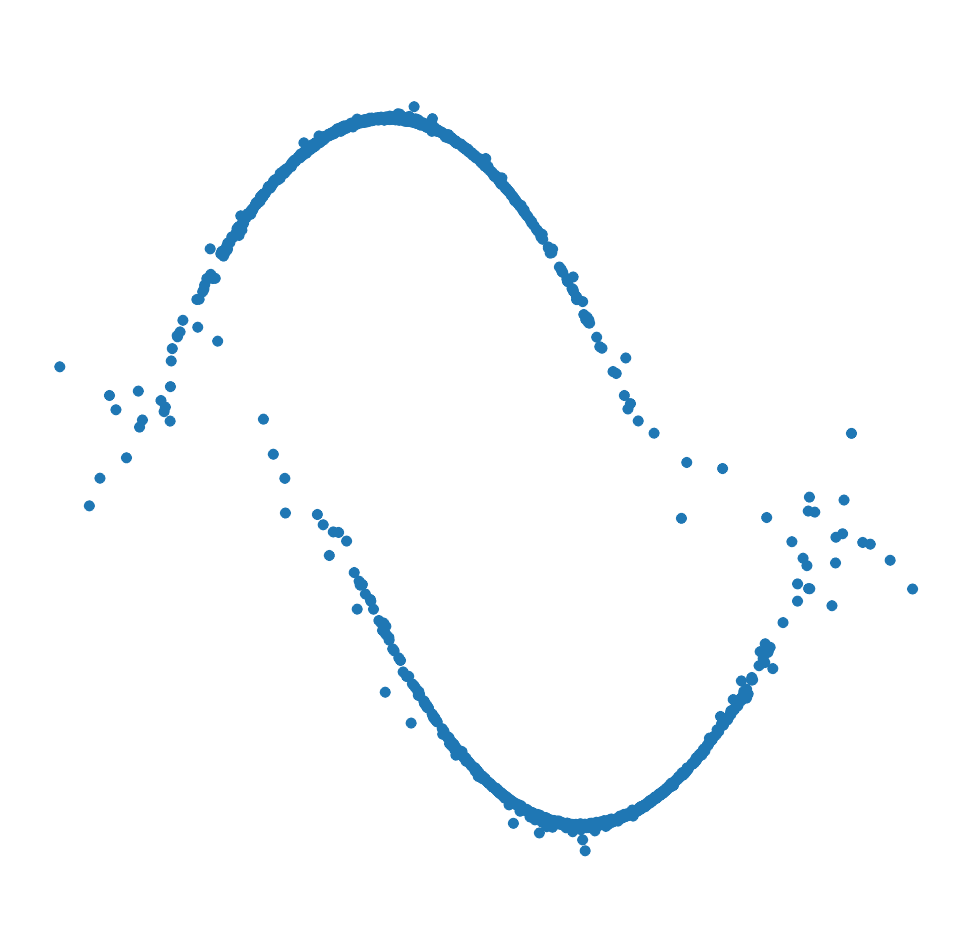} & \includegraphics[width=15mm]{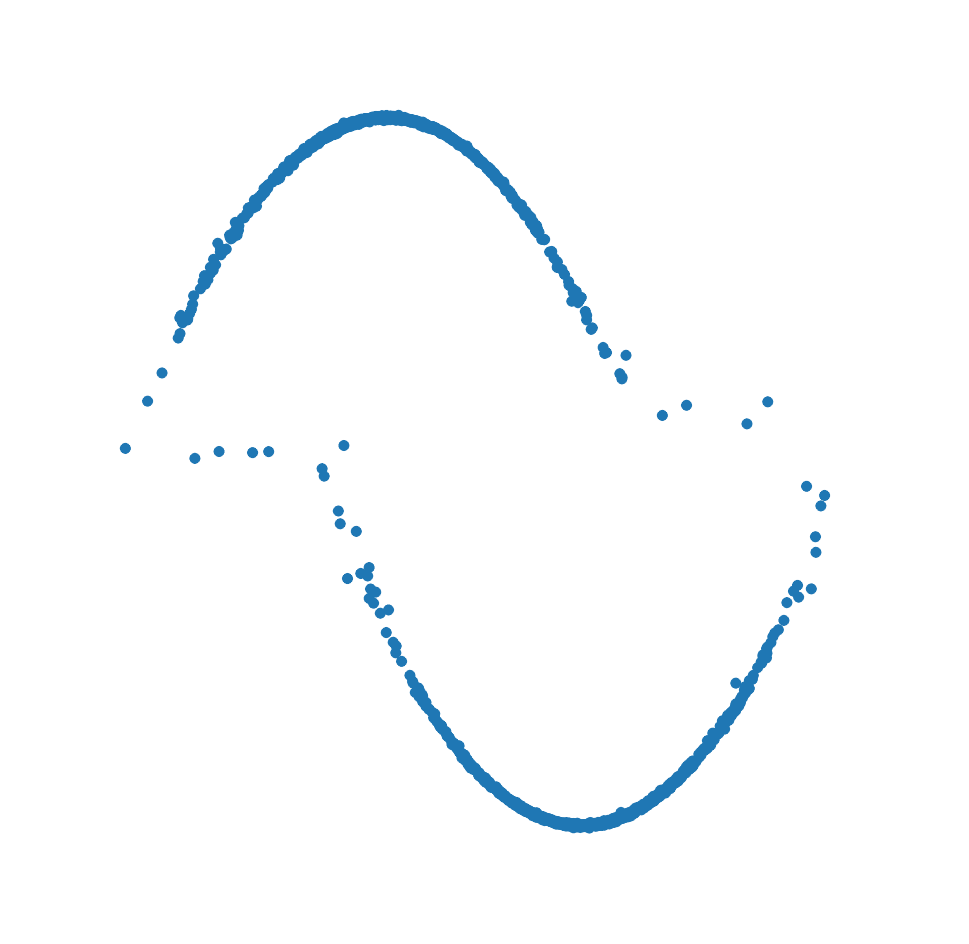} & \includegraphics[width=15mm]{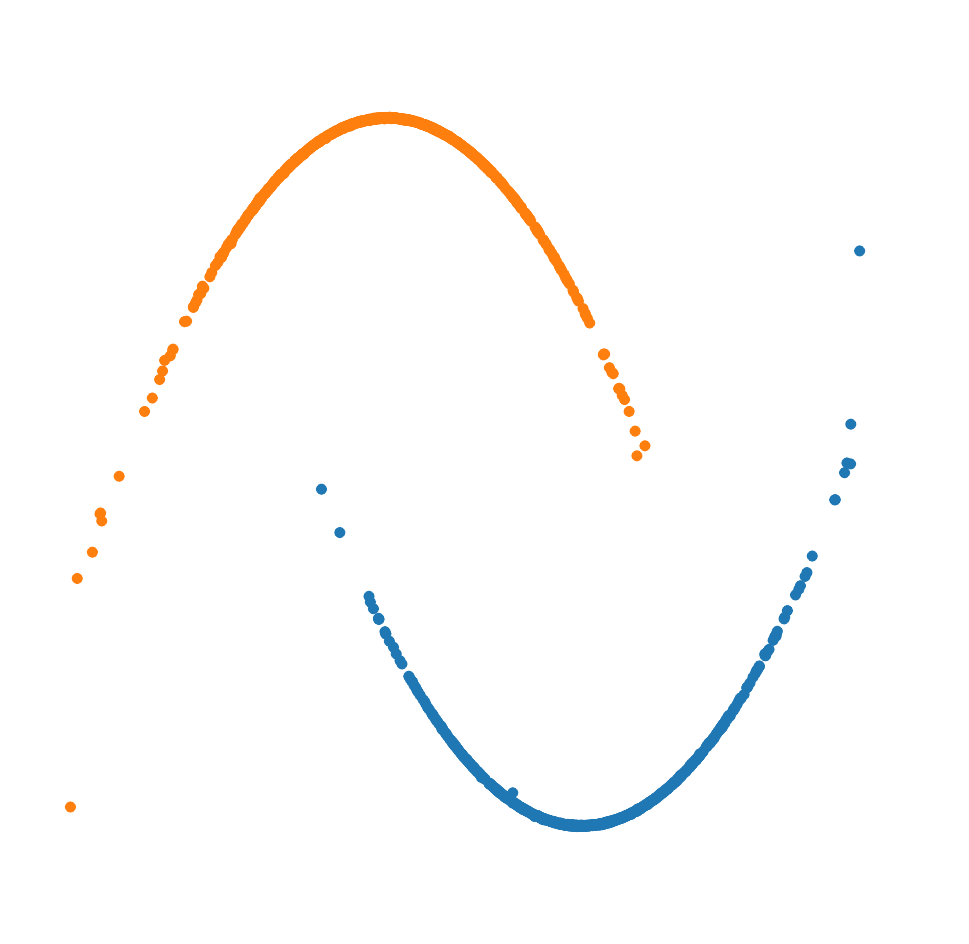} \\
  \end{tabular}
  \vspace*{-2mm}
  \caption{Point clouds generated using the proposed \emph{ChartPointFlow}, PointFlow~\cite{Yang2019}, and SoftFlow~\cite{Kim2020}.
    Color represents the chart that the point belongs to.}
  \label{fig:toydata}
  \vspace*{-4mm}
\end{figure}

\section{Background}\label{sec:background}\label{sec:background_pointflow}
To clarify the issues with existing methods, this section provides preliminary results.
We prepared synthetic datasets, namely, the circle~\cite{grathwohl2019}, 2sines~\cite{Kim2020}, four-circle~\cite{Nielsen2020}, and double-moon~\cite{Dinh2017}, each of which has only one object $X$ comprising many points $\{x_j\}$, as shown in Fig.~\ref{fig:toydata}.
The leftmost column shows the datasets.
The second and third columns show the generation results of PointFlow~\cite{Yang2019} and SoftFlow~\cite{Kim2020}, for which we employed Glow~\cite{Kingma2018} as the backbone.
The generated circles, 2sines, and four-circle show the discontinuities and blurred intersections.
The generated double-moons show the string-shaped artifacts.
Using FFJORD as the backbone, PointFlow and SoftFlow suppressed the undesired discontinuities for circle and 2sines but not for four-circle (see the fourth and fifth columns).
Moreover, they still show the string-shaped artifacts that ruin the desired disconnections.
FFJORD is a flow-based generative model inspired by a differential equation, and it learns a bijective map as a vector flow~\cite{grathwohl2019}.
Because of numerical integration, FFJORD involves significantly high computational costs.
SoftFlow did not employ FFJORD for 3D point cloud generation.

A flow-based generative model always learns a continuous deformation.
Therefore, a generated point cloud $X$ always lies on a connected manifold with no hole, as long as the latent space is a Euclidean space.
PointFlow and SoftFlow squashed 2D latent distributions to express thin structures, stretched them to trace arcs, and failed in expressing holes, intersections, and disconnections.
See Appendix~\ref{app:flow_based} for more details on flow-based models.
The same is true for methods based on GANs and AEs because a neural network is continuous in general.
This limitation is more problematic in practical tasks, as demonstrated by the results in Section~\ref{sec:reconstruction}.
These results motivate this study.


\section{Flow-Based Model with Charts}\label{sec:NF_with_charts}
Prior to \emph{ChartPointFlow}, we propose a flow-based model with charts, which is a generator of a single point cloud $X$.

\noindent\textbf{Network Structure:}\
A point generator $F$ is a flow-based generative model of a point $x\in X$ conditioned on a label $y$.
The point generator $F$ conditioned on a label $y$ is regarded as a chart, and a set of $n$ charts forms an atlas that covers the entire point cloud $X$.
The conditional log-likelihood of the point $x$ is obtained using the change of variables, as follows:
\begin{equation}
  \textstyle\log p_F(x|y) = \log p(z) + \log \left| \det \frac{\partial F^{-1}(x;y)}{\partial x} \right|,
\end{equation}
where $z$ denotes the latent variable $z=F^{-1}(x;y)$, and its prior $p(z)$ denotes the standard Gaussian distribution.
One can obtain the marginal log-likelihood $\log p_F(x)$ as the sum of all possible labels $\log p_F(x)=\log \sum_y p_F(x|y)p(y)$.
Instead, we employed a variational inference model $q_C(y|x)$, which was implemented as a neural network called a chart predictor $C$.
The evidence lower bound (ELBO) $\mathcal L_{ELBO}(F,C;x)$ is then calculated as,
\begin{equation}
  \begin{split}
    \hspace*{-1mm}\log p_F(x)
    &\textstyle=\mathbb{E}_{q_C(y|x)}\left[\log \frac{p_F(x,y)}{p_F(y|x)}\frac{q_C(y|x)}{q_C(y|x)}\right]\\
    &\textstyle\ge \mathbb{E}_{q_C(y|x)}\left[\log \frac{p_F(x|y)p(y)}{q_C(y|x)}\right]\\
    &\textstyle=\mathbb{E}_{q_C(y|x)}\left[\log p(z) \!+\! \log \left| \det \frac{\partial F^{-1}(x;y)}{\partial x} \right|\right]\hspace*{-1mm}\\
    &\textstyle\hspace*{1.3cm}-H[q_C(y|x)|p(y)]+H[q_C(y|x)]\\
    &\eqqcolon \mathcal L_{ELBO}(F,C;x),
  \end{split}
\end{equation}
where $H[q_C(y|x)|p(y)]$ and $H[q_C(y|x)]$ denote the cross-entropy and entropy, respectively.
We assume that the label prior $p(y)$ is the uniform distribution, which implies that the cross-entropy $H[q_C(y|x)|p(y)]$ has a constant value.
We emphasize that the label $y$ is inferred to maximize the ELBO in an unsupervised manner.

\vspace*{1mm}\noindent\textbf{Training:}\
The label $y$ is represented by a one-hot vector, and the ELBO $\mathcal L_{ELBO}$ is given by the weighted average over all possible labels.
This approach requires the computational cost to be proportional to the number of labels.
To avoid this issue, we employed the Gumbel-Softmax approach~\cite{Jang2017}.
Specifically,
\begin{equation}
  \begin{split}
    \tilde{y}
    & = \mathrm{softmax}((\log\pi_C(x) + \mathfrak g) / \tau),\\
    \mathfrak g & \sim \mathrm{Gumbel}(0,1),
  \end{split}\label{eq:gumbel}
\end{equation}
where $\mathfrak g$ denotes a vector, each of whose elements follows the Gumbel distribution $\mathrm{Gumbel}(0,1)$, $\tau \in (0, \infty)$ denotes the temperature of the softmax function, and $\pi_C(x)$ denotes the vector of the label posterior $q_C(y|x)$, i.e., $\left(\pi_C(x)\right)_j=q_C(y_j=1|x)$.
This approach allows us to apply the Monte Carlo sampling to the label $\tilde{y}$ in a differentiable manner.
One uses a sufficiently small temperature $\tau$, draws an almost one-hot vector $\tilde y$, substitutes it into the ELBO $\mathcal L_{ELBO}$, and trains neural networks using gradient descent algorithms.
The ELBO $\mathcal L_{ELBO}$ is approximated as,
\begin{align}
   & \!\!\mathcal L_{ELBO}(F,C;x)\simeq\tilde{\mathcal L}_{ELBO}(F,C;x)                                                                                          \\
   & \ \ \textstyle\coloneqq \log p(z)\!+\!\log \left| \det \frac{\partial F^{-1}(x;\tilde y)}{\partial x} \right|\!-\!H[q_C(y|x)|p(y)]\!+\!H[q_C(y|x)]\nonumber
\end{align}
where the vector $\tilde y$ is given by Eq.~\eqref{eq:gumbel}.
Owing to the Gumbel-Softmax approach, we emphasize that the computational cost is constant regardless of the number of charts.

When maximizing the approximated ELBO $\tilde {\mathcal L}_{ELBO}$, the entropy $H[q_C(y|x)]$ is maximized, resulting in each point belonging to all labels with the same probabilities and the charts overlapping with each other.
To assign each chart to a specific connected region of a manifold, i.e., a point cloud, we introduce a regularization term $\mathcal{L}_{MI}(x, y)$, which is based on the mutual information $I(x;y)$, as follows.
\begin{equation}
  \begin{split}
    I(y;x)
    &= H[q_C(y)] - H[q_C(y|x)]\\
    &\textstyle\simeq H\left[\frac{1}{|X|}\sum_{\tilde{x}\in X}q_C(y|\tilde{x})\right]\!-\!H[q_C(y|x)]\\
    &\eqqcolon \mathcal{L}_{MI}(C;x).
    \label{eq:mutual}
  \end{split}
\end{equation}
The maximization of the regularization term $\mathcal{L}_{MI}$ cancels out the maximization of the entropy $H[q_C(y|x)]$ in the ELBO $\tilde{\mathcal L}(F,C;X)$, and it additionally maximizes the entropy $H[q_C(y)]$.
Thus, each sample belongs to only one chart, and all charts are used with uniform probabilities.

\begin{figure}[t]
  \centering
  \includegraphics[width=19mm]{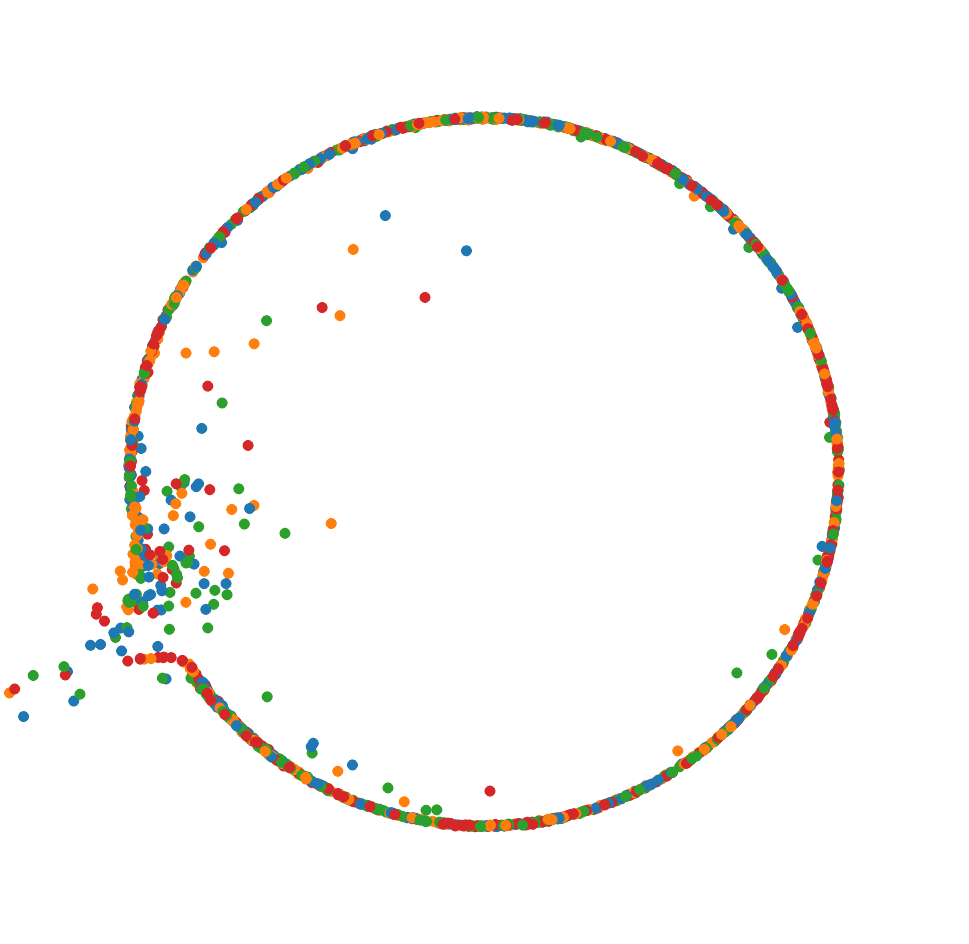}
  \hspace*{5mm}
  \includegraphics[width=19mm]{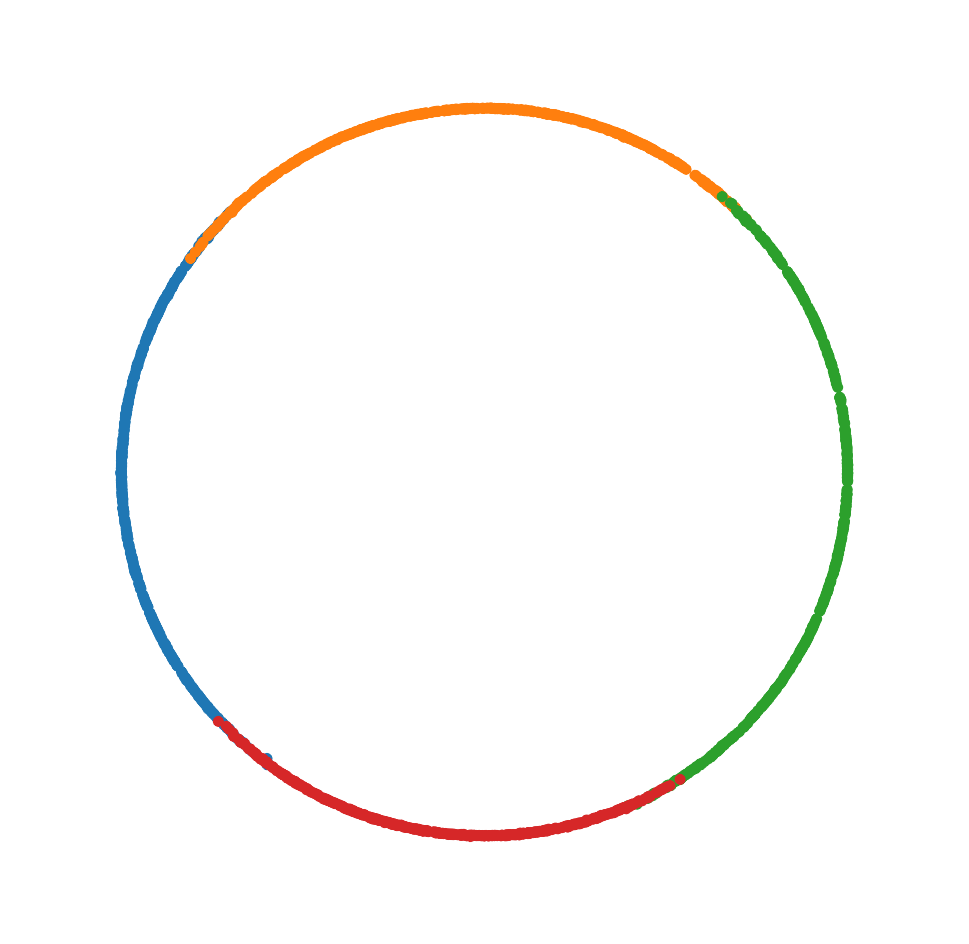}\\[-3mm]
  \caption{
    Results without (left) and with (right) the regularization term $\mathcal{L}_{MI}(x, y)$.
    Color represents the chart that the point belongs to.}
  \label{fig:regularization}
  \vspace*{-4mm}
\end{figure}

\begin{figure*}
  \includegraphics[scale=0.47]{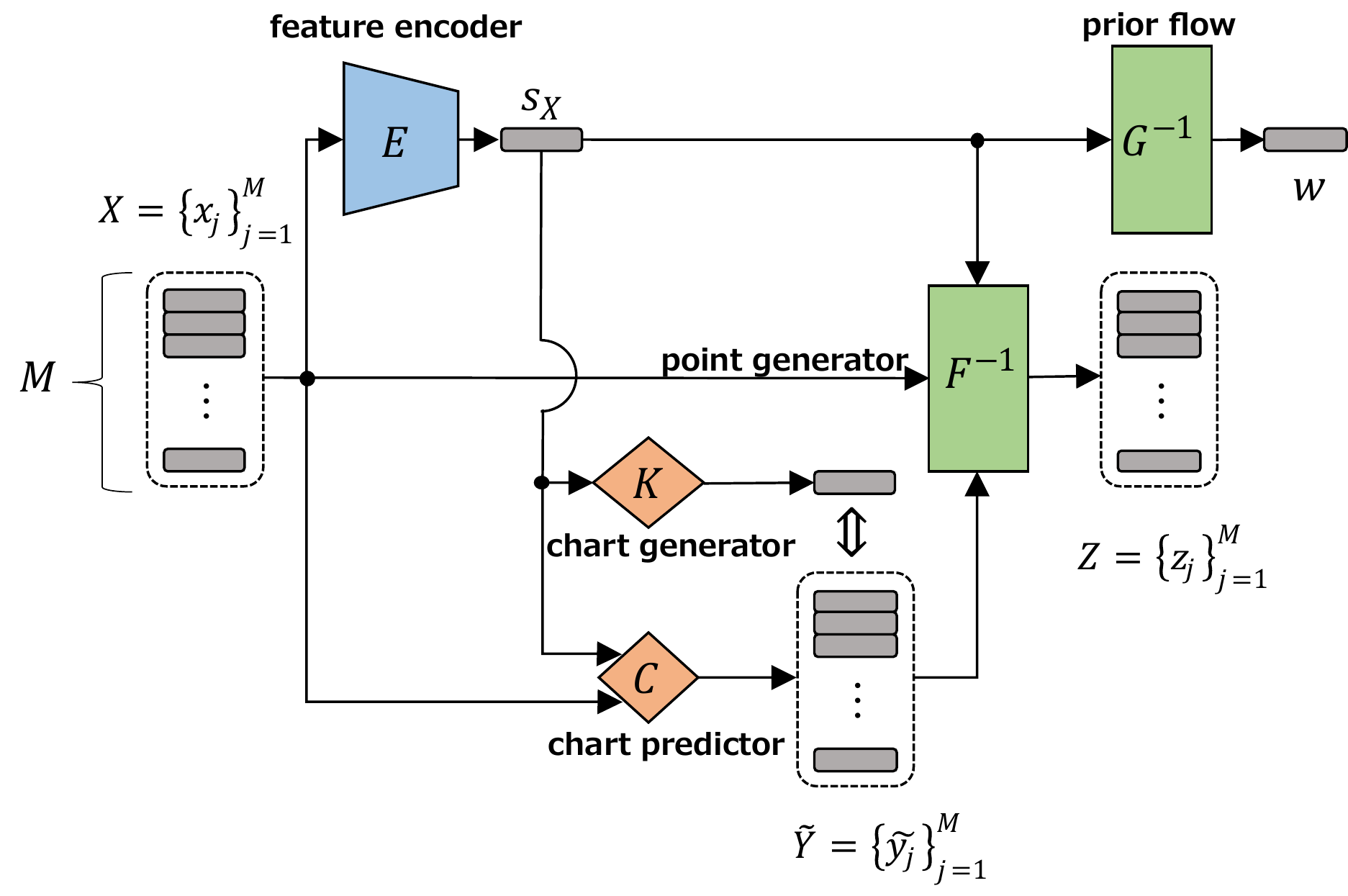}
  \hspace*{2mm}
  \includegraphics[scale=0.47]{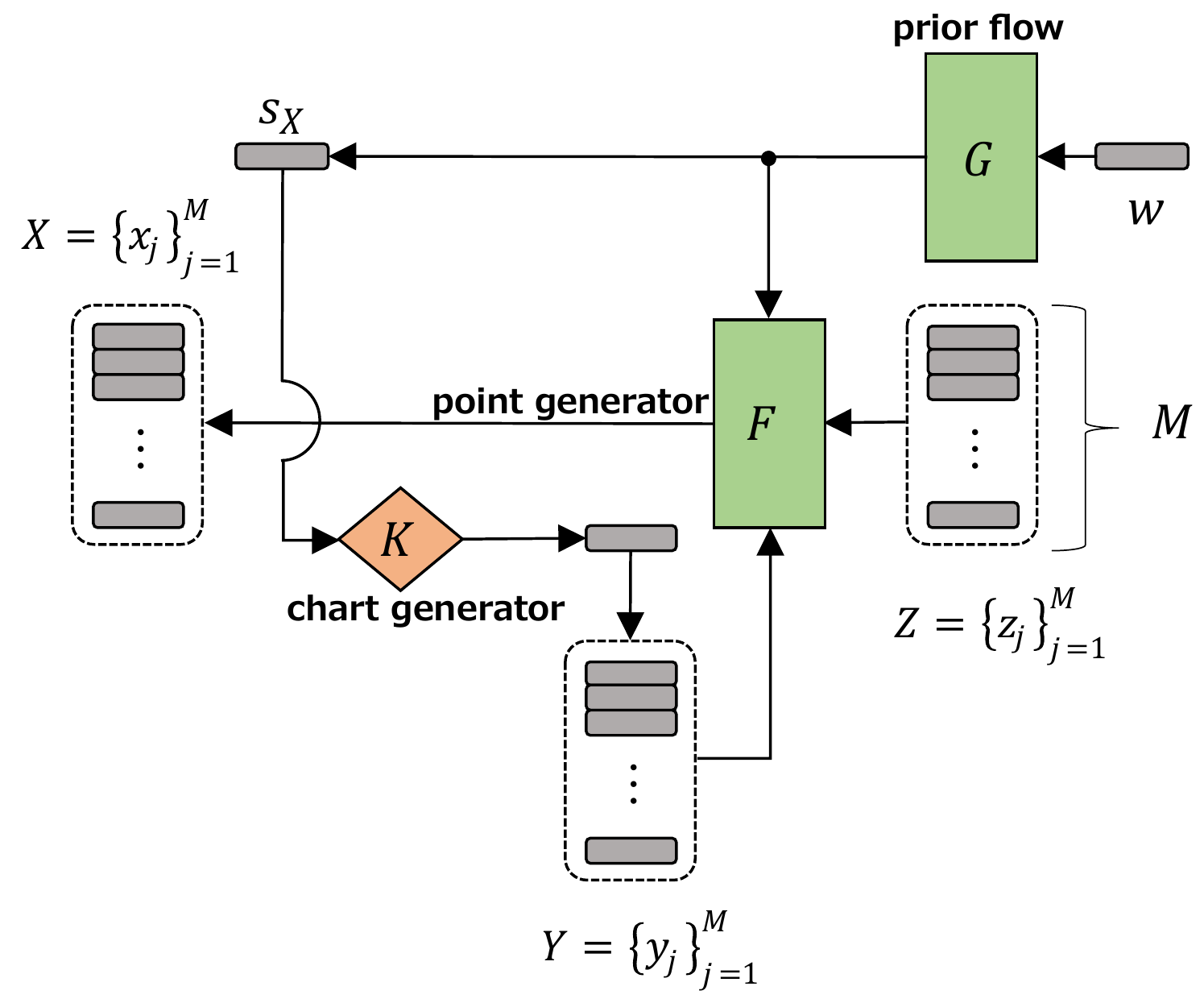}\\[-3mm]
  \caption{Architectures and data flow during the training phase (left) and the generation phase (right).
  }
  \label{fig:framework}
\end{figure*}

For the i.i.d.~assumption, the objective function to be maximized for the entire point cloud $X$ is defined using the sum over the points $x$, as follows.
\begin{align}
  \textstyle \mathcal{L}(F,C;X,\lambda) = \sum_{x\in X} \left[\tilde{\mathcal{L}}_{ELBO} +\lambda \mathcal{L}_{MI}\right],
\end{align}
where $\lambda$ adjusts the regularization term $\mathcal{L}_{MI}(x, y)$.

\vspace*{1mm}\noindent\textbf{Experiments on Synthetic Data:}\
As shown in Fig.~\ref{fig:toydata}, we conducted preliminary experiments on 2D synthetic datasets to prove the concept of the proposed method.
We employed Glow~\cite{Kingma2018} as the backbone of the point generator $F$.
We used $n=4$ charts for the circle and 2sines datasets, $n=8$ charts for the four-circle dataset, and $n=2$ charts for the double-moon dataset.
We set $\lambda$ to $1.1$ and $\tau$ to $0.1$.
For other experimental settings, we followed SoftFlow~\cite{Kim2020}, such as Adam optimizer~\cite{Kingma2015} with a batch size of $100$ for 36K iterations.
Following FFJORD~\cite{grathwohl2019}, the learning rate was set to $10^{-4}$ for Glow and to $10^{-3}$ for FFJORD.
After training, each point $x$ was drawn using the point generator $F$, as follows.
\begin{equation}
  x=F(z;y) \mbox{ for } y\sim p(y) \mbox{ and } z\sim p(z).
\end{equation}
PointFlow~\cite{Yang2019} and SoftFlow~\cite{Kim2020} were trained under the same experimental settings.
Note that the proposed method with only a single chart is the same as PointFlow.

The generated point clouds are summarized in Fig.~\ref{fig:toydata}.
PointFlow and SoftFlow generated point clouds suffering from discontinuities, blurs, and artifacts, as mentioned in Section~\ref{sec:background_pointflow}.
In contrast, the proposed method generated a circle without any discontinuity, intersections free from a severe blur, and two arcs clearly separated without any artifacts, even though the backbone was Glow.
Color represents the chart that the point belongs to.
In the circle, 2sines, and four-circle datasets, subparts are connected smoothly and form the manifold with holes.
The intersection is expressed as the intersection of the subparts.
In the double-moon, each chart is assigned to one of the arcs exclusively, and thereby expresses the disconnected manifold without artifacts.
These results imply that the proposed concept of charts works well for various topological structures, even with the same latent variable distribution $p(z)$.

The left panel of Fig.~\ref{fig:regularization} shows the results without the regularization term $\mathcal{L}_{MI}(x, y)$.
Each label is then assigned to the entire point cloud overlapping with each other, and the model generates the discontinuity.
This is because the maximization of the entropy $H[q_C(y|x)]$ results in the uniform posterior $q_C(y|x)$, and each label works similarly.


\section{ChartPointFlow}\label{sec:NF_with_ChartsPointFlow}
In this section, we extend the model proposed in Section~\ref{sec:NF_with_charts} and apply it to 3D point cloud datasets.
We name it \emph{ChartPointFlow}.
Figure~\ref{fig:framework} shows a conceptual diagram of ChartPointFlow.
We assume that a point cloud dataset $\mathcal{X}$ is composed of $N$ objects $\{X_1, X_2,\ \ldots , X_N\}$, and each object $X_i$ is represented by a cloud of $M_i$ points $\{x_1,x_2,\ldots,\\x_{M_i}\}$.

\vspace*{1mm}\noindent\textbf{Network Structure:}\
The feature encoder $E$ is the same as those used in PointFlow~\cite{Yang2019} and SoftFlow~\cite{Kim2020}.
The feature encoder $E$ is a permutation-invariant neural network that accepts a point cloud $X$ consisting of $M$ points and encodes it to a posterior $q_E(s_{X}|X)$ of a feature vector $s_{\!X}$ using the reparameterization trick~\cite{Kingma2014}.
The feature vector $s_{\!X}$ is considered a representation of the entire shape of the point cloud $X$.
With a Gaussian prior, the reparameterization trick is known to suffer from posterior collapse, where the output $s_{\!X}$ ignores the input $X$~\cite{Kingma2016,Yang2019}.
To make the prior more expressive, the feature encoder $E$ is combined with a flow-based generative model called a prior flow $G$, which maps the feature vector $s_{\!X}$ to the latent variable $w$.
The trainable prior $p_G(s_{\!X})$ of the feature vector $s_{\!X}$ is then given by,
\begin{equation}
  \textstyle\log p_G(s_{\!X}) = \log p(w) + \int_{t_0}^{t_1}\log \left | \det \frac{\partial G^{-1}(s_{\!X})}{\partial s_{\!X}} \right |,
  \label{eq:prior}
\end{equation}
where $w=G^{-1}(s_{\!X})$, and the prior $p(w)$ is set to the standard Gaussian distribution.
Thereby, the prior flow $G$ learns the distribution of point clouds.

In addition to the architectures in the previous studies, ChartPointFlow has a chart predictor $q_C(y|x,s_{\!X})$, which is introduced in Section \ref{sec:NF_with_charts}.
The chart predictor $q_C(y|x,s_{\!X})$ is conditioned on the feature vector $s_{\!X}$.
It accepts a point $x\in X$ and infers the label $y$ that corresponds to the chart that the point $x$ belongs to.
The condition on $s_{\!X}$ implies that different point clouds have different atlases, even in the same dataset.
For example, points in the same location can be part of the engine or the airframe depending on the airplane's width.
Moreover, the posterior of the label $\tilde{y}$ is $q_C(y|X,s_{\!X})=\mathbb{E}_{x\in X}\sum_j q_C(y|x,s_{\!X})$, indicating that the size of each chart depends on point cloud $X$.
A zero posterior implies that the corresponding chart is discarded.
In this way, ChartPointFlow differs significantly from AtlasNets, whose charts (patches) have the same size~\cite{Groueix2018,Deprelle2019}.

The point generator $F$ was the same as that used in SoftFlow~\cite{Kim2020} except that ours is conditioned on the label $y$, whereas that of SoftFlow is conditioned on the injected noise's intensity.
The conditional log-likelihood of a point $x$ is given by
\begin{equation}
  \textstyle \log p_F(x|y,s_{\!X}) = \log p(z) \!+\! \log \left | \det \frac{\partial F^{\!-\!1}(x;y, s_{\!X})}{\partial x} \right |\!.
  \label{eq:reconstruct}
\end{equation}

For the generation task, we further propose a neural network called the chart generator $K$, which accepts a feature vector $s_{\!X}$ and gives the posterior $p_K(y|s_X)$ of the label $y$.

\vspace*{1mm}\noindent\textbf{Objective Function:}\
Let $Y$ denote a set of labels, each of which corresponds to a point $x$ of the given point cloud $X$.
Owing to the i.i.d.~assumption, $p(Y)=\prod_j p(y_j)$, $q_C(Y|s_{\!X})=\prod_j q_C(y_j|s_{\!X})$, and $p_F(X|Y,s_{\!X})=\prod_j p_F(x_j|y_j,s_{\!X})$.
Given the above, the ELBO $\mathcal{L}_{ELBO}$ is given by,
\begin{equation}
  \begin{split}
    \hspace*{-2mm}\log p(X)
    &\textstyle\ge \mathbb E_{q_E(s_{\!X}|X)q_C(Y|X,s_{\!X})}\!\!\left[\log\frac{p_F(X|Y,s_{\!X})p(Y)p_G(s_{\!X})}{q_C(Y|X,s_{\!X})q_E(s_{\!X}|X)}\right]\hspace*{-4mm}\\
    &\textstyle=\mathbb{E}_{q_E(s_{\!X}|X)} \Bigl[
    \sum_j\Bigl\{\mathbb E_{q_C(y_j|x_j,s_{\!X})}\bigl[ \log p_F(x_j|y_j, s_{\!X})  \bigr]\\
    &\textstyle\ \ \ \ \ \ +H[q_C(y_j|x_j,s_{\!X})] - H[q_C(y_j|x_j,s_{\!X}) | p(y_j)] \Bigr\}\Bigr] \\
    &\textstyle\ \ \ \ \ \ -D_{KL}(q_E(s_{\!X}|X) \| p_G(s_{\!X}))\\
    &\eqqcolon \mathcal{L}_{ELBO}(F,C,E,G;X).
  \end{split}
\end{equation}
In practice, the expectation $\mathbb{E}_{q_C(y_j|x_j,s_{\!X})}$ over the inferred label $y_j$ is approximated using the Gumbel-Softmax approach~\cite{Jang2017} (see Eq.~\eqref{eq:gumbel}), and the expectation $\mathbb{E}_{q_E(s_{\!X}|X)}$ over the feature vector $s_{\!X}$ is approximated using Monte Carlo sampling~\cite{Kingma2014}.
The approximated ELBO is denoted by $\tilde{\mathcal{L}}_{ELBO}(F,C,E,G;X)$.

The first term of the regularization term in Eq.~\eqref{eq:mutual} forces each object to use all charts equivalently.
However, each chart may have a different size in practice.
To achieve a flexible adjustment, we introduce the coefficient terms $\mu$ and $\lambda$ as
\begin{equation}
  \mathcal{L}_{MI}(C;X,\mu, \lambda)\!\coloneqq\!\sum_{\!j}\left\{\mu H\left[\!\frac{1}{|X|}\!\sum_{\tilde x\in X}q_C(y_{\!j}|\tilde x)\right]\!-\! \lambda H[q_C(y_{\!j}|x_{\!j})]\right\}\!.\\
  \label{eq:mutual2}
\end{equation}

The objective function to be maximized is given by,
\begin{equation}
  \textstyle\mathcal{L}(F,C,E,G,K; \mathcal{X},\mu, \lambda)  = \sum_{X \in \mathcal{X}} \left[\tilde{\mathcal{L}}_{{\rm ELBO}}+ \mathcal{L}_{MI}\right].
\end{equation}
In addition, the chart generator $K$ is trained separately to estimate the label posterior $q_C(y|X)$ by maximizing the objective function
\begin{align}
  \textstyle\mathcal{L}_{CP}(K;\mathcal{X}) = -\sum_{X \in \mathcal{X}} D_{KL}(p_K(y|s_{\!X}) \| q_C(y|X)).
\end{align}

\vspace*{1mm}\noindent\textbf{Usage and Tasks:}\
For the generation tasks, one can follow the right panel of Fig.~\ref{fig:framework}.
First, draw a latent variable $w$ from the prior $p(w)$ and feed it to the prior flow $G$, obtaining a feature vector $s_{\!X}=G(w)$.
By feeding the feature vector $s_{\!X}$ to the chart generator $K$, the label posterior $p_K(y|s_{\!X})$ is obtained as a categorical distribution.
Repeat the following step $M$ times for $M$ points: draw a label $y_j$ from the posterior $p_K(y|s_{\!X})$ and a latent variable $z_j$ from the prior $p(z_j)$, feed the pair to the point generator $F$, and obtain a point $x_j=F(z_j;y_j,s_{\!X})$.
The set of the obtained points is the generated point cloud $X$.
Formally, $p(X)=\int_{s_{\!X}}p_G(s_{\!X})\prod_j \int_{y_j}p_F(x_j|y_j,s_{\!X})p_K(y_j|s_{\!X})$.

For the reconstruction or super-resolution task, feed a given point cloud $X$ to the feature encoder $E$ and obtain a feature vector $s_{\!X}$ instead of drawing the feature vector $s_{\!X}$ from the prior flow $G$.
Drawing the same number of points is called reconstruction, and adding drawn points to a given point cloud is called super-resolution.

The computational cost of ChartPointFlow is almost the same as that of the comparison methods, PointFlow~\cite{Yang2019} and SoftFlow~\cite{Kim2020}, when the same backbones are used.
Recall that the computational cost of the proposed method is constant regardless of the number of charts owing to the Gumbel-Softmax approach.
The chart predictor $C$ is used only during the training phase.
The computational cost of the chart generator $K$ is negligible because it is proportional to the number of point clouds (i.e., objects), whereas other components $E$, $F$, and $C$ require a computational cost that is proportional to the number of points in all point clouds.
In previous studies, PointFlow employed FFJORD as the backbone, but SoftFlow employed Glow.
We employed Glow as the backbone of ChartPointFlow.
Therefore, its computational cost is at the same level as that of SoftFlow and significantly smaller than that of PointFlow.


\begin{table*}[t]\centering
  \setlength{\tabcolsep}{1mm}
  \begin{tabular}{L{2mm}C{15mm}C{15mm}C{15mm}C{15mm}C{15mm}C{15mm}C{15mm}C{15mm}C{15mm}C{15mm}}
    &\multicolumn{2}{c}{\footnotesize\textbf{Result A}} & \multicolumn{2}{c}{\footnotesize\textbf{Result B}} & \multicolumn{2}{c}{\footnotesize\textbf{Result C}} & \multicolumn{2}{c}{\footnotesize\textbf{Result D}} & \multicolumn{2}{c}{\footnotesize\textbf{Result E}} \\[-1mm]
    \cmidrule(l{.75em}r{.75em}){2-3}\cmidrule(l{.75em}r{.75em}){4-5}\cmidrule(l{.75em}r{.75em}){6-7}\cmidrule(l{.75em}r{.75em}){8-9}\cmidrule(l{.75em}r{.75em}){10-11}
    \vspace*{-4mm}&
    \vspace*{-4mm}\footnotesize\textbf{Points}                                                                                            &
    \vspace*{-4mm}\footnotesize\textbf{Charts}                                                                                              &
    \vspace*{-4mm}\footnotesize\textbf{Points}                                                                                            &
    \vspace*{-4mm}\footnotesize\textbf{Charts}&
    \vspace*{-4mm}\footnotesize\textbf{Points}                                                                                            &
    \vspace*{-4mm}\footnotesize\textbf{Charts}&
    \vspace*{-4mm}\footnotesize\textbf{Points}                                                                                            &
    \vspace*{-4mm}\footnotesize\textbf{Charts}&
    \vspace*{-4mm}\footnotesize\textbf{Points}                                                                                            &
    \vspace*{-4mm}\footnotesize\textbf{Charts}                                                                                                    \\[-5.0mm]
    \rotatetitle{16mm}{\textbf{Airplane}}                                                                              &
    \includegraphics[width=16mm]{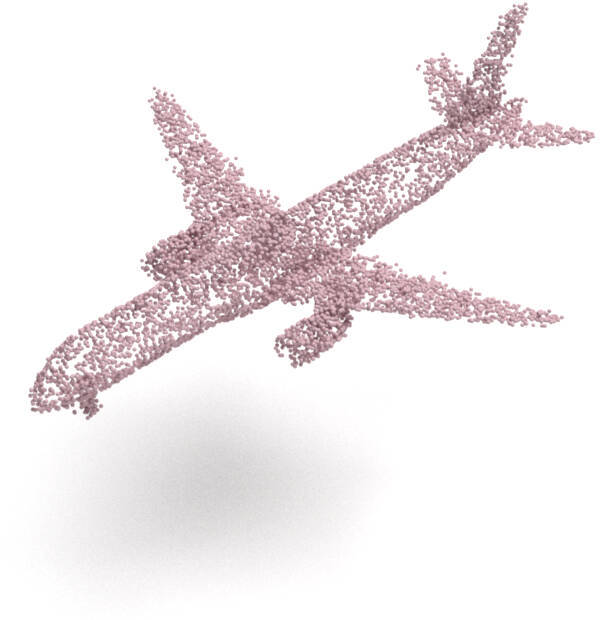}                             &
    \includegraphics[width=16mm]{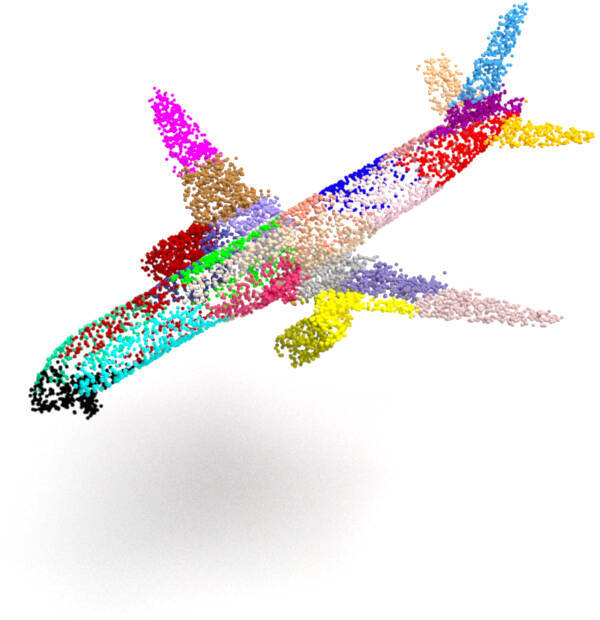}                       &
    \includegraphics[width=16mm]{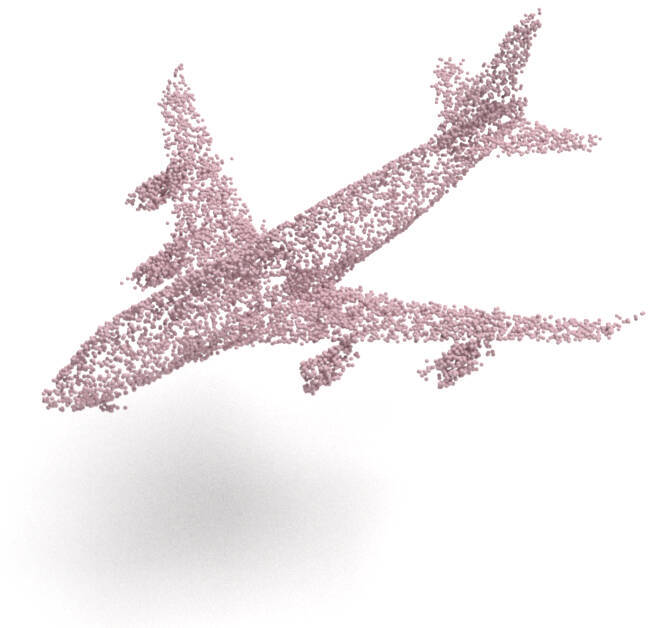}                            &
    \includegraphics[width=16mm]{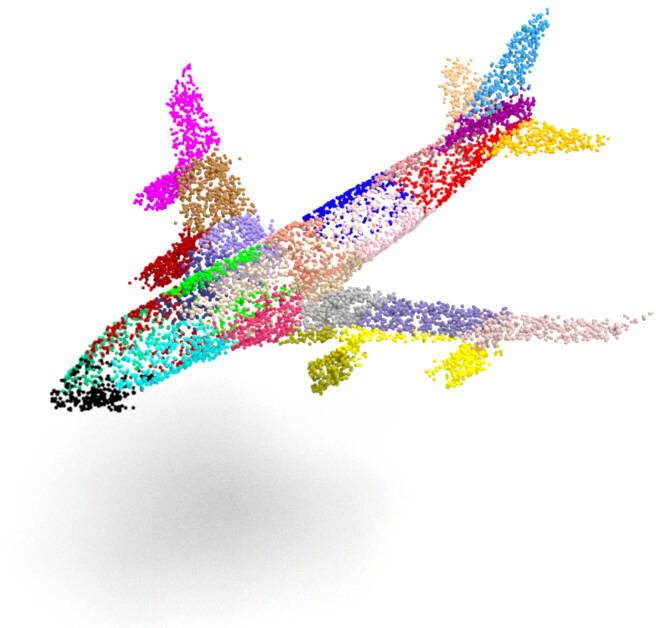}                      &
    \includegraphics[width=16mm]{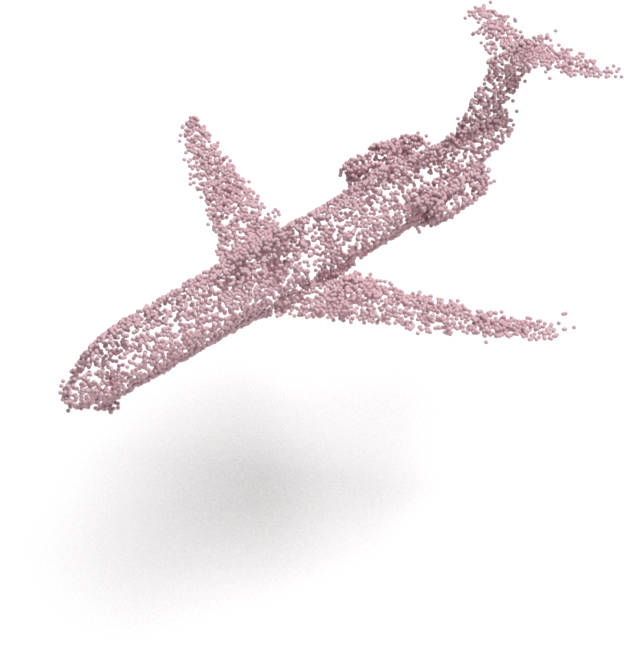}                             &
    \includegraphics[width=16mm]{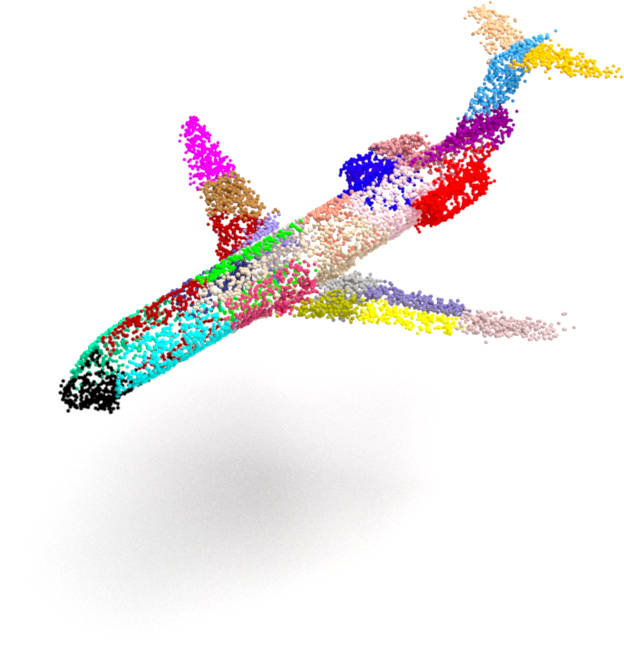}                       &
    \includegraphics[width=16mm]{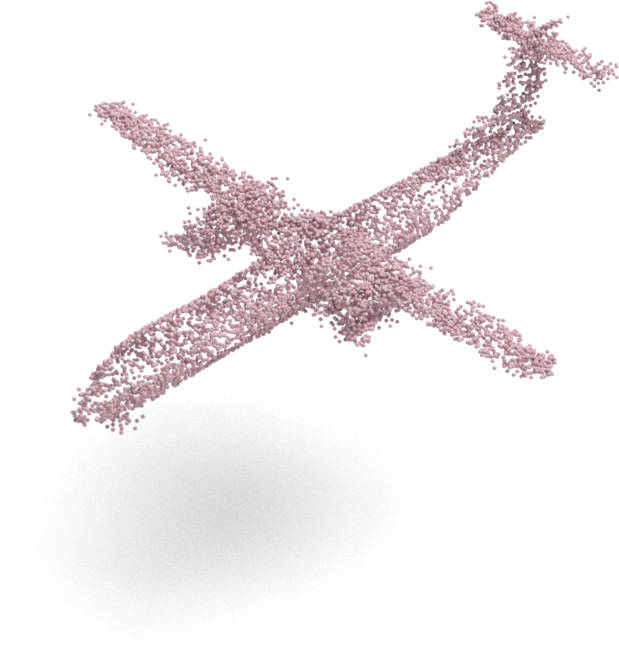}                            &
    \includegraphics[width=16mm]{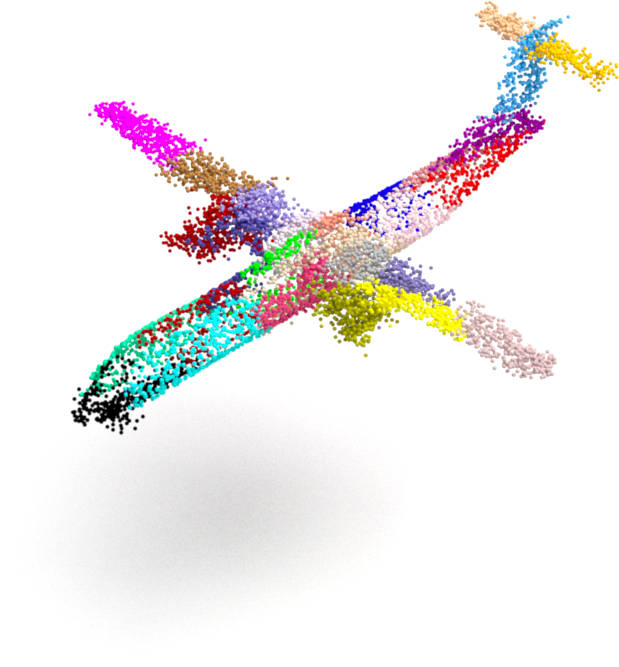}                      &
    \includegraphics[width=16mm]{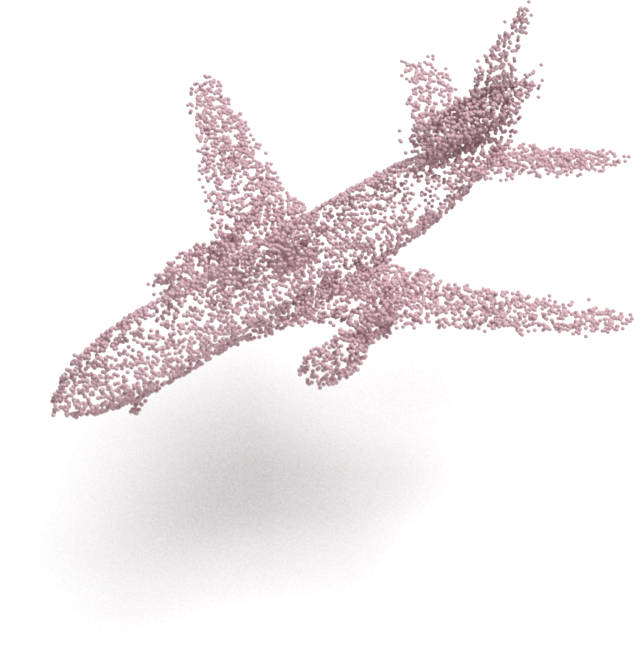}                            &
    \includegraphics[width=16mm]{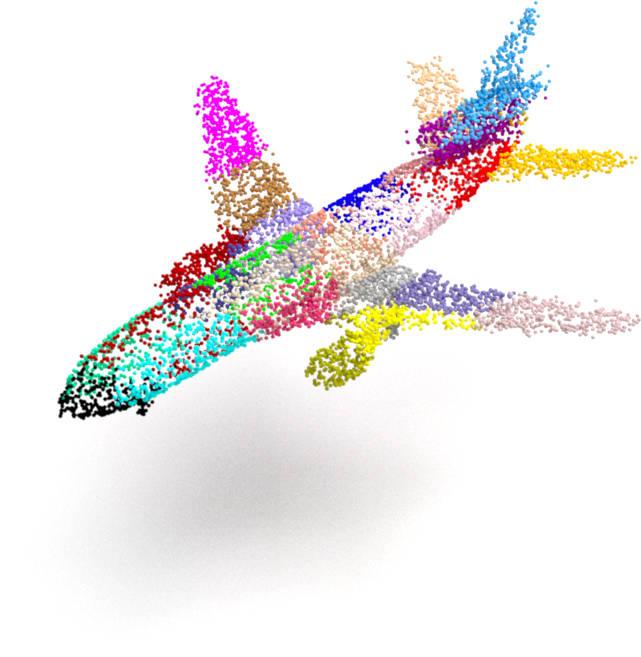}                      \\[-3.5mm]

    \rotatetitle{16mm}{\textbf{Chair}}                                                                               &
    \includegraphics[width=11mm]{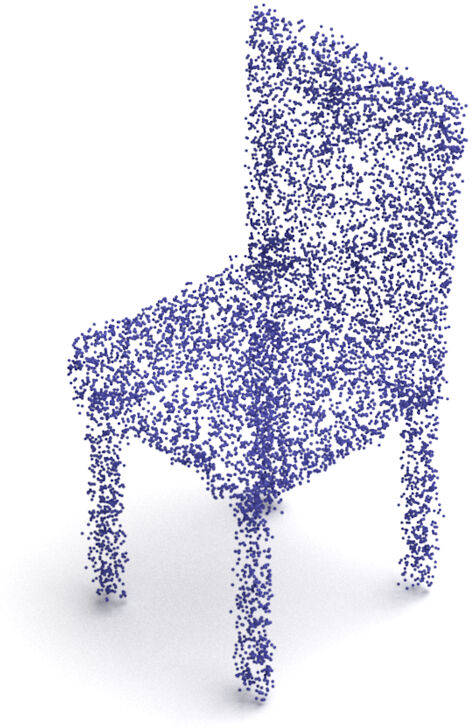}              &
    \includegraphics[width=11mm]{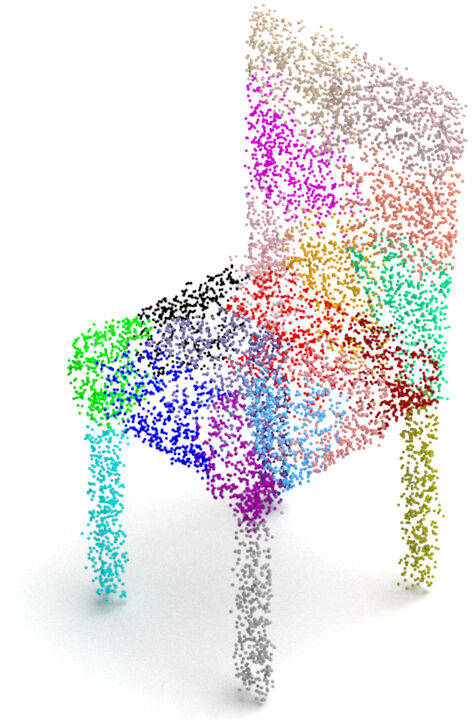}        &
    \includegraphics[width=11mm]{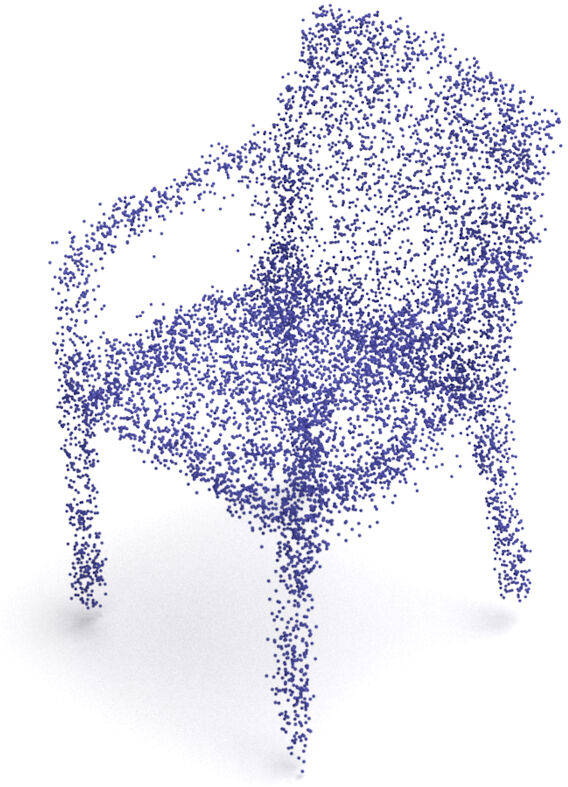}             &
    \includegraphics[width=11mm]{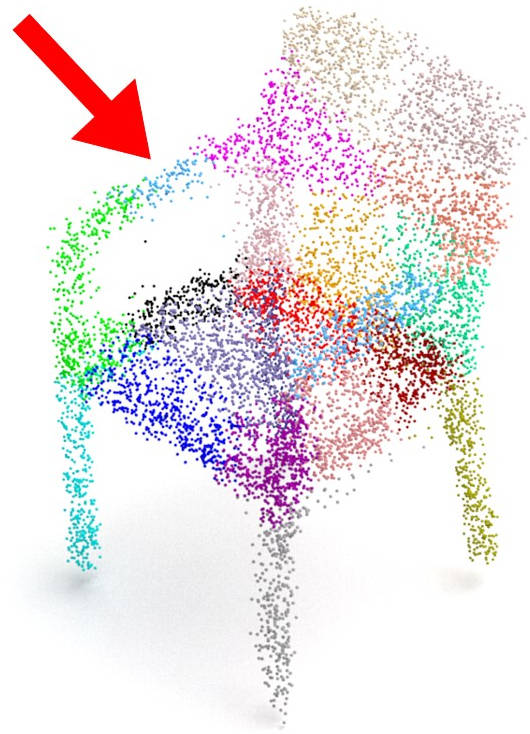} &
    \includegraphics[width=11mm]{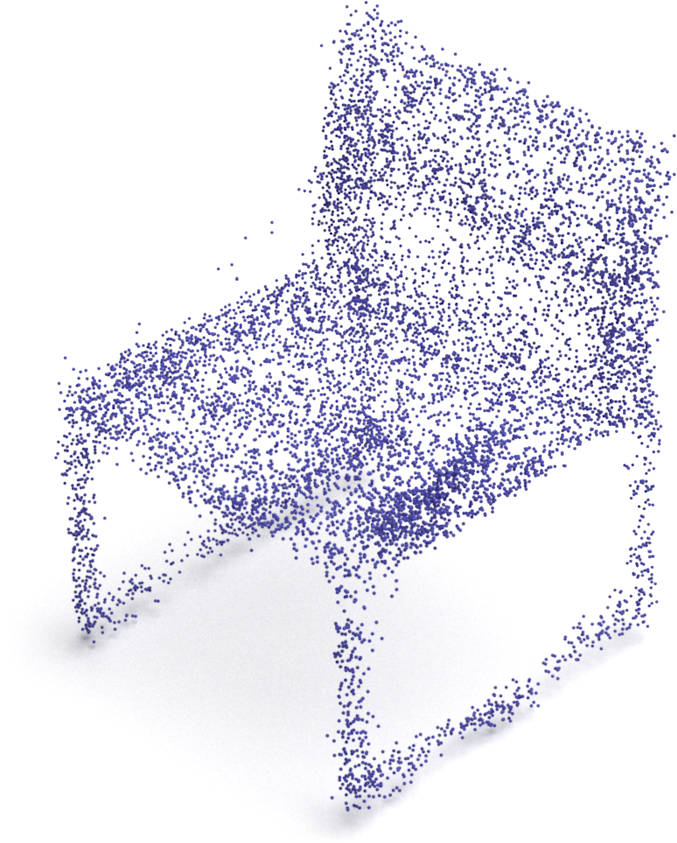}              &
    \includegraphics[width=11mm]{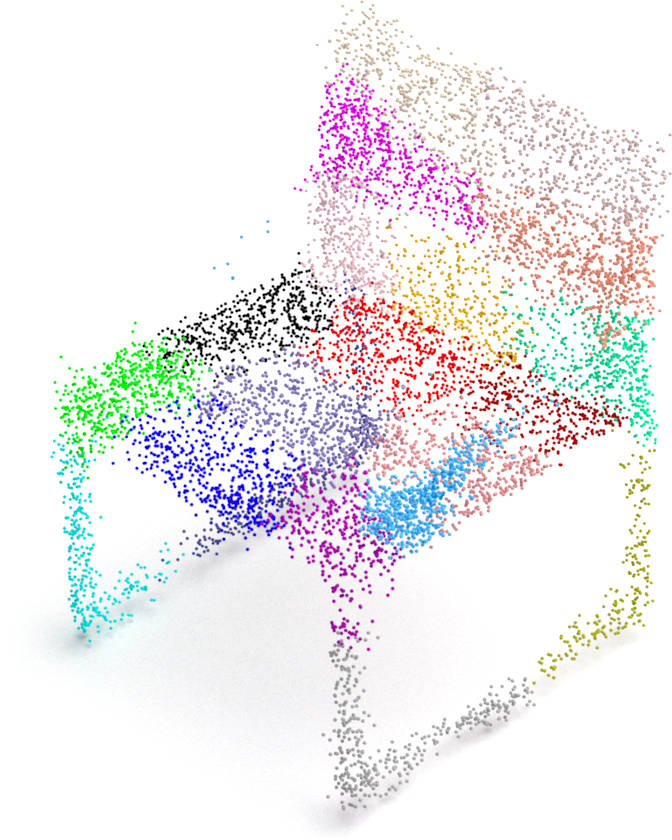}        &
    \includegraphics[width=11mm]{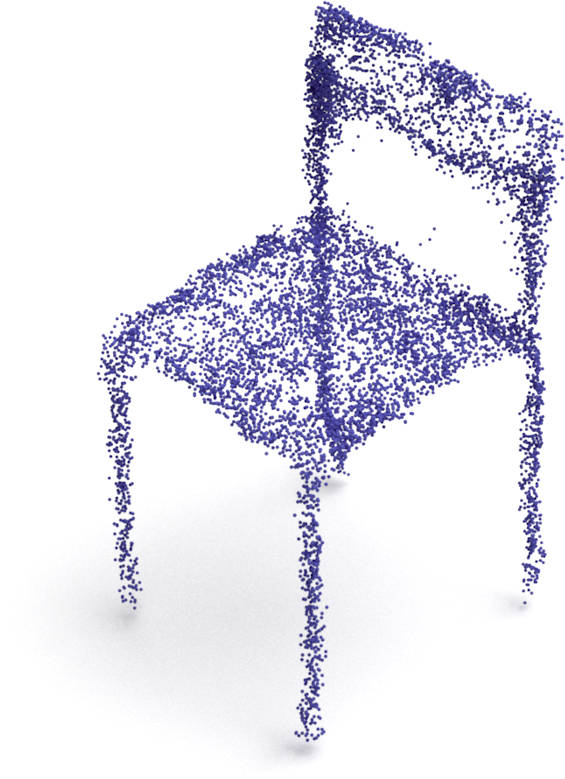}             &
    \includegraphics[width=11mm]{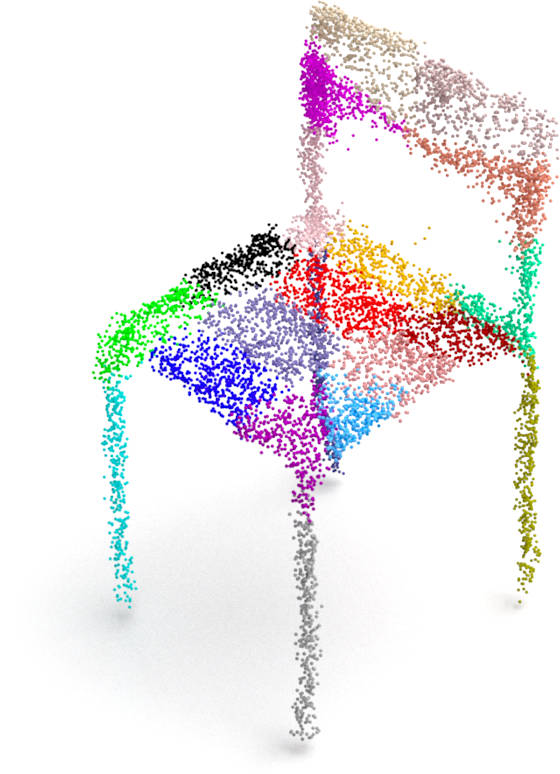} &
    \includegraphics[width=11mm]{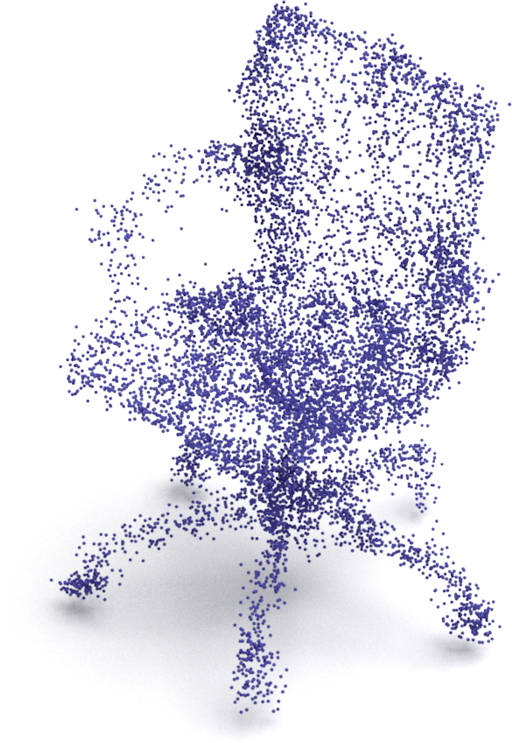}             &
    \includegraphics[width=11mm]{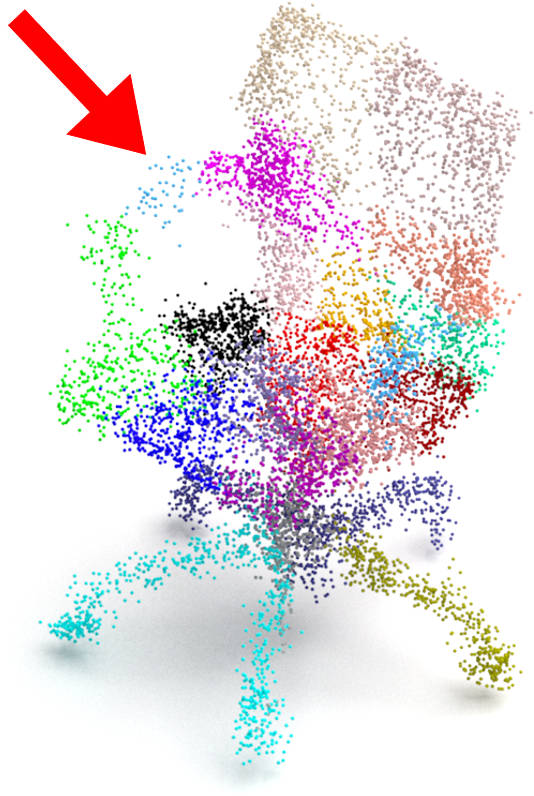} \\[-3.0mm]

    \rotatetitle{16mm}{\textbf{Car}}                                                                              &
    \includegraphics[width=16mm]{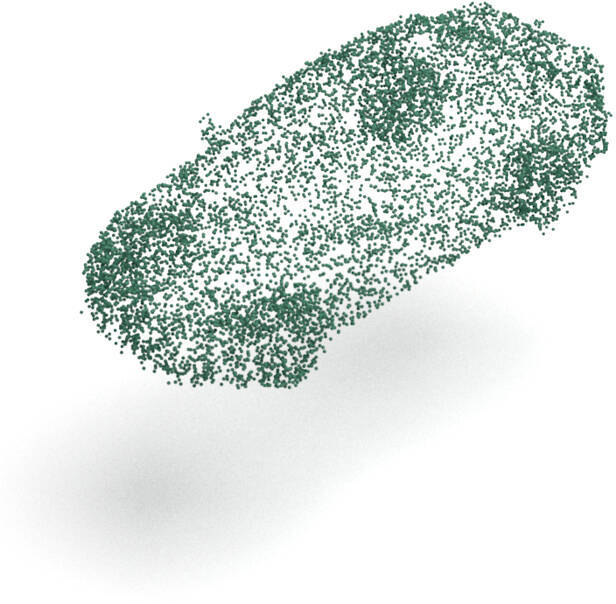}&
    \includegraphics[width=16mm]{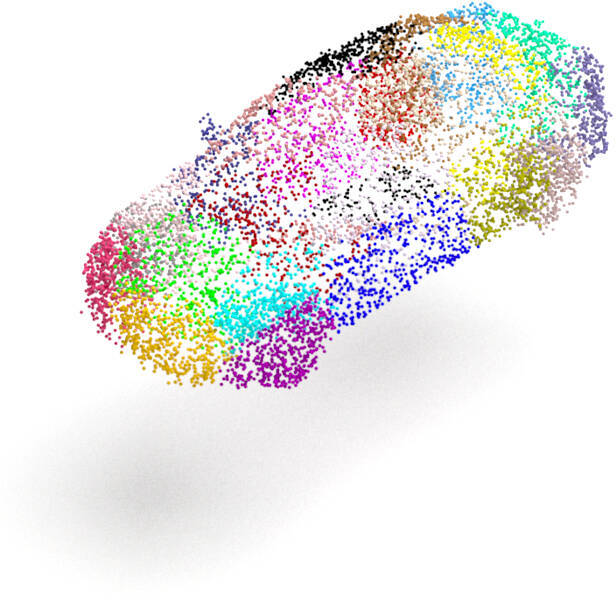}&
    \includegraphics[width=16mm]{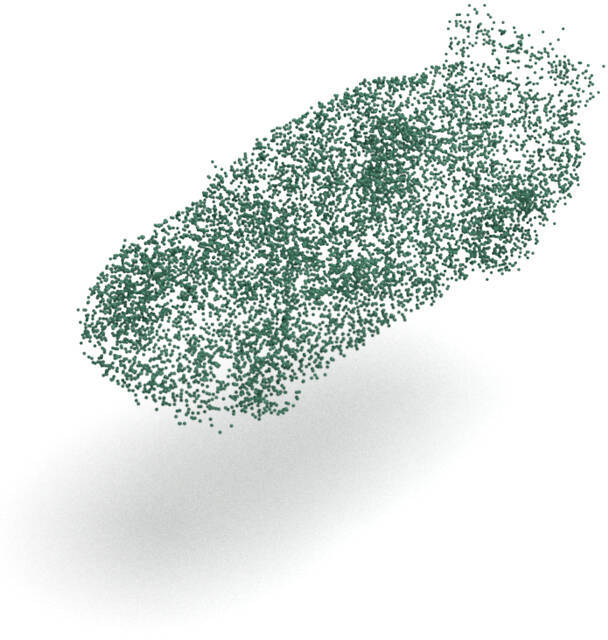}&
    \includegraphics[width=16mm]{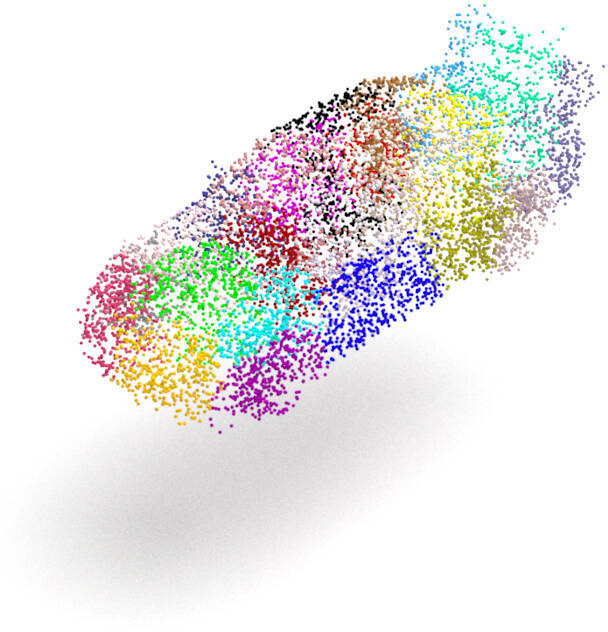}&
    \includegraphics[width=16mm]{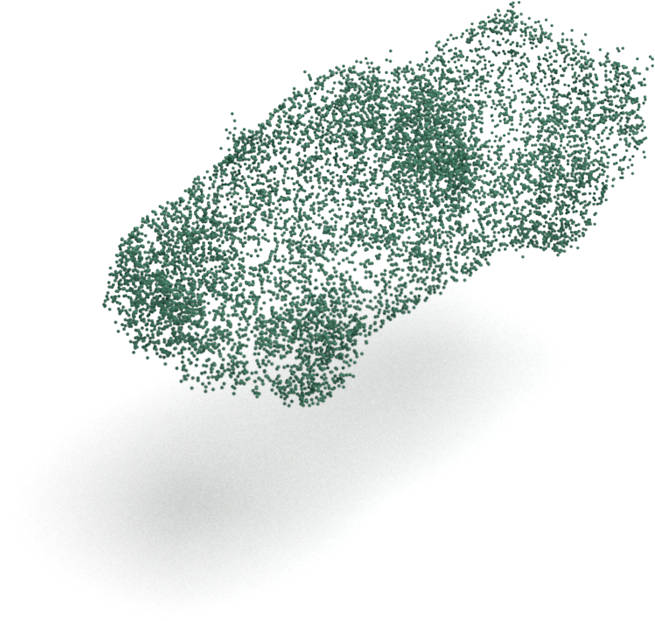}&
    \includegraphics[width=16mm]{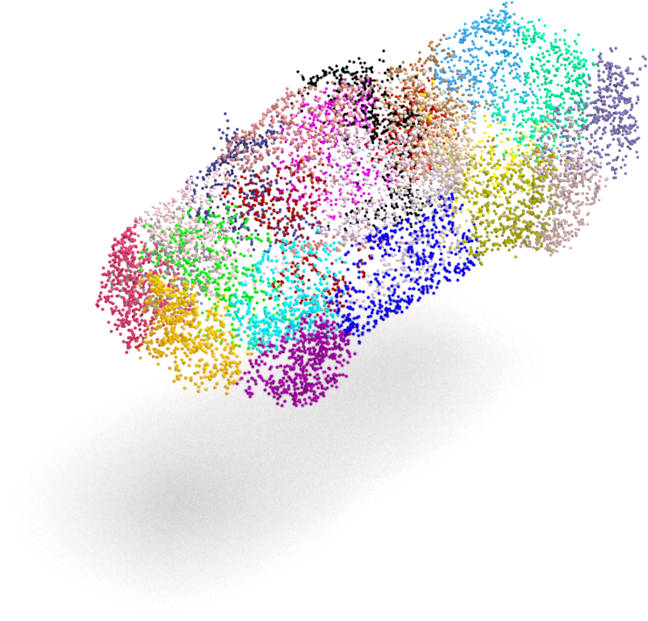}&
    \includegraphics[width=16mm]{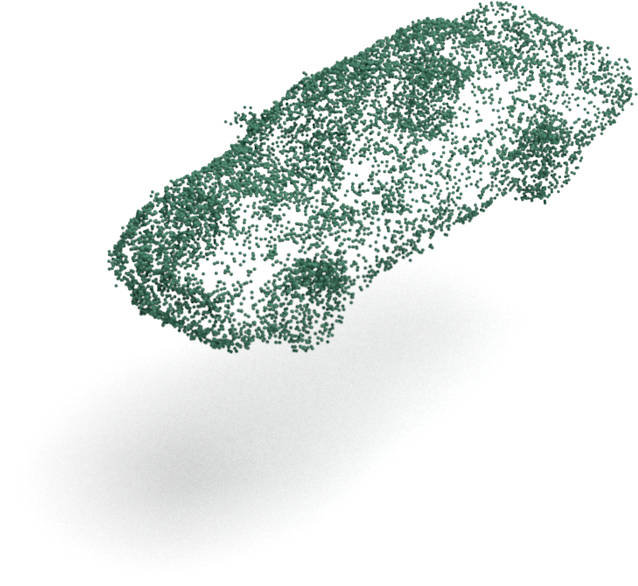}&
    \includegraphics[width=16mm]{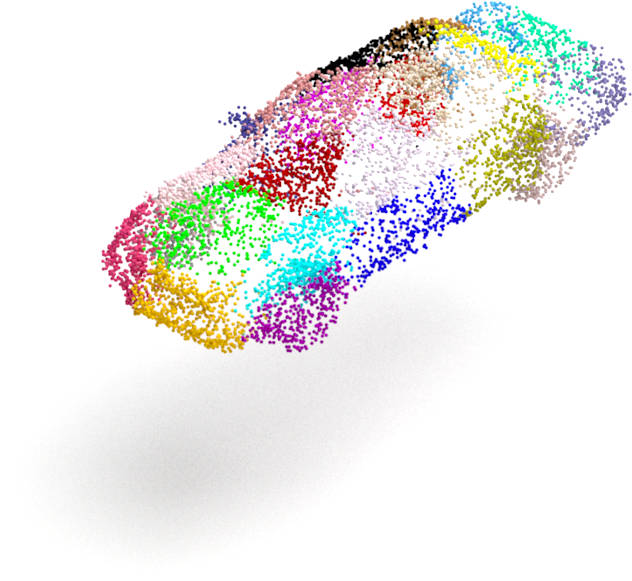}&
    \includegraphics[width=16mm]{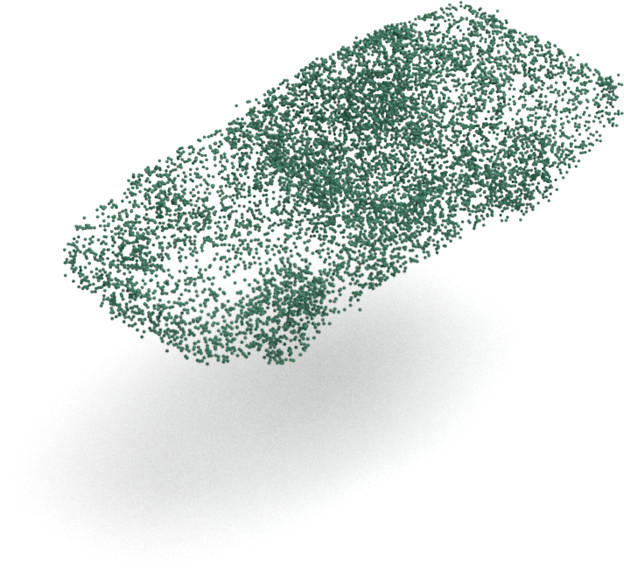}&
    \includegraphics[width=16mm]{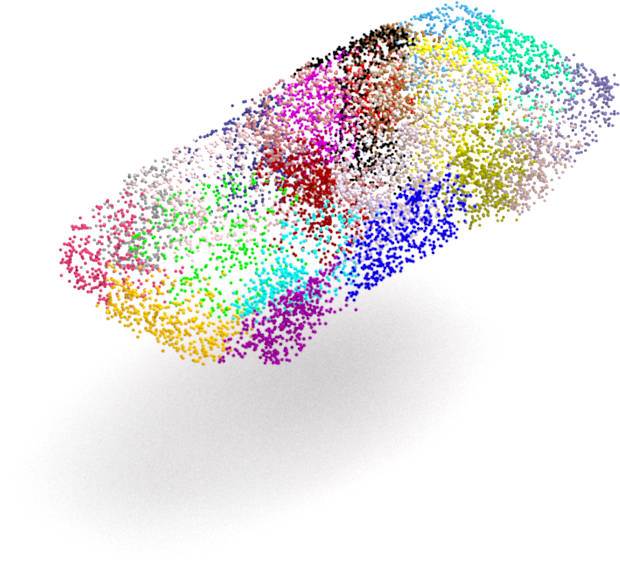}\\

  \end{tabular}
  \vspace*{-6mm}
  \captionof{figure}{Generation examples by ChartPointFlow.}
  \label{fig:generated_samples}
\end{table*}

\section{Experiments and Results}\label{sec:experiments}

\subsection{Experimental Settings}
We evaluated the performance of ChartPointFlow using the \texttt{Core.v2} of ShapeNet dataset~\cite{Chang2015}.
The dataset is composed of 513,000 unique 3D objects of 55 categories.
We selected three different categories: airplane, chair, and car, following Yang et al.~\cite{Yang2019}.

We followed the experimental settings presented in SoftFlow's release code~\cite{Kim2020}.
We trained $15$K epochs in each category using the Adam optimizer~\cite{Kingma2015} with a batch size of $128$, an initial learning rate of $2.0 \times 10^{-3}$, $\beta_1 = 0.9$, and $\beta_2 = 0.999$.
We decayed the learning rate by quarter after every $5$K epochs.
We obtained $M=2,048$ points randomly from each object $X$.

We set $\tau=0.1$ for the Gumbel-Softmax approach, set $\mu=0.05$ and $\lambda=1.0$ for the regularization term $\mathcal{L}_{MI}$, and searched the number $n$ of charts from a range of $\{4,8,12,16,20,24,28,32\}$.
The architectures of the neural networks followed those of SoftFlow~\cite{Kim2020}.
The detailed architectures are summarized in Appendix~\ref{app:architecture}.

\subsection{Evaluation Metrics}\label{sec:metrics}
To measure the distance between a pair of point clouds $X_1$ and $X_2$, we employed the earth mover's distance (EMD)~\cite{Achlioptas2018,Kim2020,Yang2019}.
The EMD is the minimum of the total travel distance of points to deform a point cloud to the other.
Specifically, the EMD is defined as
\begin{equation}
  \textstyle EMD(X_1, X_2)  = \min_{\phi:X_{1}\rightarrow X_{2}} \sum_{x\in X_{1}} \|x - \phi(x) \|_2.
\end{equation}
where both point clouds $X_1$ and $X_2$ are composed of the same number of points, $\phi$ denotes a bijective map from the point cloud $X_1$ to the other $X_2$, and $\|\cdot\|_2$ denotes the Euclidean distance on $\mathbb{R}^3$.
While Chamfer distance (CD) has also been used, recent studies have pointed out that it yields misleading results~\cite{Achlioptas2018}.
CD focuses on populated regions (e.g., a chair's seat cushion) and ignores sparsely placed points (e.g., a chair's mesh backrest).

To evaluate the similarity between a pair of sets $\mathcal{X}_1$ and $\mathcal{X}_2$ of point clouds, we employed the 1-nearest neighbor accuracy (1-NNA)~\cite{Kim2020,Lopez-Paz2017,Yang2019}, which aims to evaluate whether two distributions are identical in two-sample tests.
1-NNA is obtained as
\begin{equation}
  \hspace*{-4mm}\mathrm{1\mathchar`-NNA} (\mathcal{X}_1, \mathcal{X}_2)
  \!=\!\frac
  {\sum_{X_1 \in \mathcal{X}_1}\! \mathbbm{1} [N_{\!X_1} \!\in\!\mathcal{X}_1] \!+\! \sum_{X_2 \in \mathcal{X}_2}\! \mathbbm{1} [N_{\!X_2} \!\in\! \mathcal{X}_2]}
  {|\mathcal{X}_1| + |\mathcal{X}_2|}\!,\hspace*{-4mm}
\end{equation}
where both sets $\mathcal{X}_1$ and $\mathcal{X}_2$ are composed of the same number of point clouds, $N_{\!X_\bullet}$ denotes the nearest neighbor of $X_\bullet$ in $\mathcal{X}_1 \cup \mathcal{X}_2 - \{X_\bullet\}$, and $\mathbbm{1}[\cdot]$ denotes the indicator function.
Roughly speaking, a 1-nearest neighbor classifier classifies a given point cloud $X$ into $\mathcal{X}_1$ or $\mathcal{X}_2$ according to the nearest sample $N_X$ in terms of the EMD.
The closer to $50\%$ the accuracy of the 1-NNA is, the more similar the distributions $\mathcal{X}_1$ and $\mathcal{X}_2$ are.
Previous studies also used Jensen-Shannon divergence (JSD), minimum matching distance (MMD), and coverage (COV).
However, recent studies have revealed that they may give good scores to poor models~\cite{Kim2020,Yang2019}.
For example, JSD gives a good score to a model that generates an average shape without considering individual shapes~\cite{Yang2019}.

\begin{table}[t]\centering
  \begin{tabular}{lccc}
    \toprule
    Model                             & Airplane       & Chair          & Car            \\
    \midrule
    r-GAN~\cite{Achlioptas2018}       & 99.51          & 99.47          & 99.86          \\
    l-GAN (CD)~\cite{Achlioptas2018}  & 97.28          & 85.27          & 88.07          \\
    l-GAN (EMD)~\cite{Achlioptas2018} & 85.68          & 65.56          & 68.32          \\
    PC-GAN~\cite{Li2019}              & 92.32          & 78.37          & 90.87          \\
    ShapeGF~\cite{Cai2020}            & 81.44          & 59.60          & 60.31          \\
    PointFlow~\cite{Yang2019}         & 75.06          & 59.89          & 62.36          \\
    SoftFlow~\cite{Kim2020}           & 69.44          & 63.51          & 64.71          \\
    \midrule
    ChartPointFlow                    & \textbf{65.08} & \textbf{58.31} & \textbf{58.68} \\

    \bottomrule
  \end{tabular}
  \caption{Generation performances. Closer to 50\% is better.}
  \label{tab:generation_results}
  \vspace*{-7mm}
\end{table}

\begin{figure}[t]
  \setlength{\tabcolsep}{1mm}
  \hspace*{-5mm}
  \begin{tabular}{L{2mm}C{14mm}C{14mm}C{14mm}C{14mm}C{14mm}}
                                                                                                                              &
    \footnotesize\textbf{Reference}                                                                                                        &
    \footnotesize\textbf{ShapeGF}                                                                                                            &
    \footnotesize\textbf{PointFlow}&
    \footnotesize\textbf{SoftFlow}&
    \footnotesize\textbf{ChartPointFlow} \hspace*{3.2mm}(proposed)\\ [-4.5mm]
    \rotatetitle{18mm}{\hspace*{-21.0mm}\textbf{Airplane}}                                                                                    &
    \includegraphics[width=16mm]{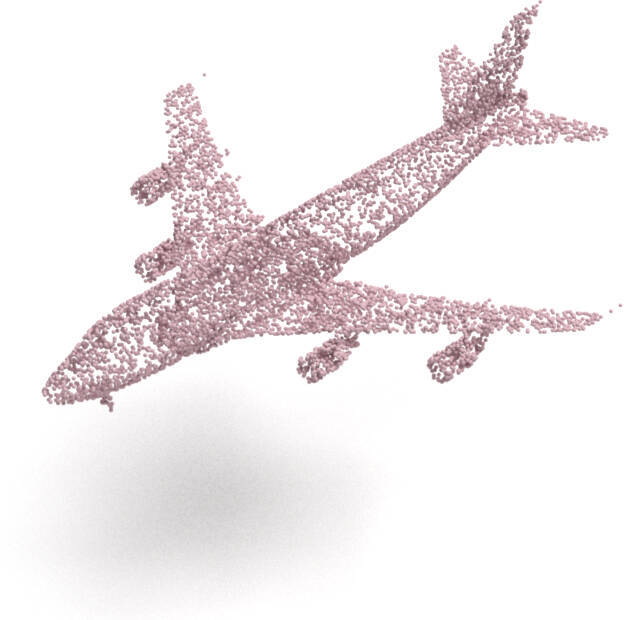}                           &
    \includegraphics[width=16mm]{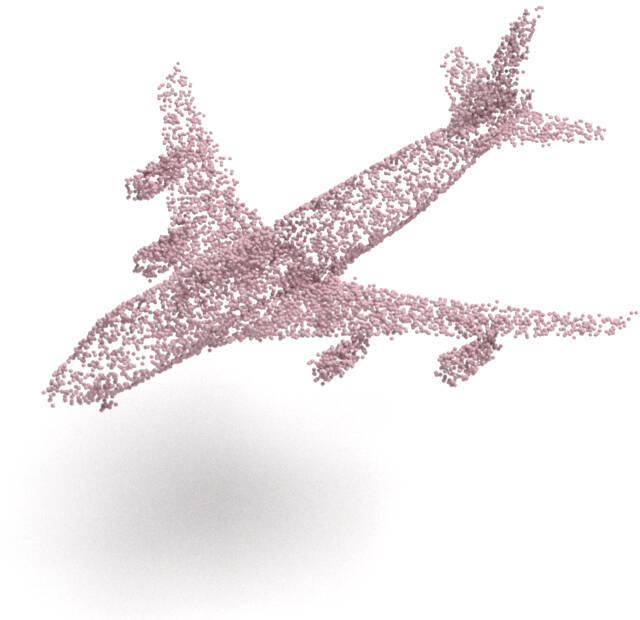}&
    \includegraphics[width=16mm]{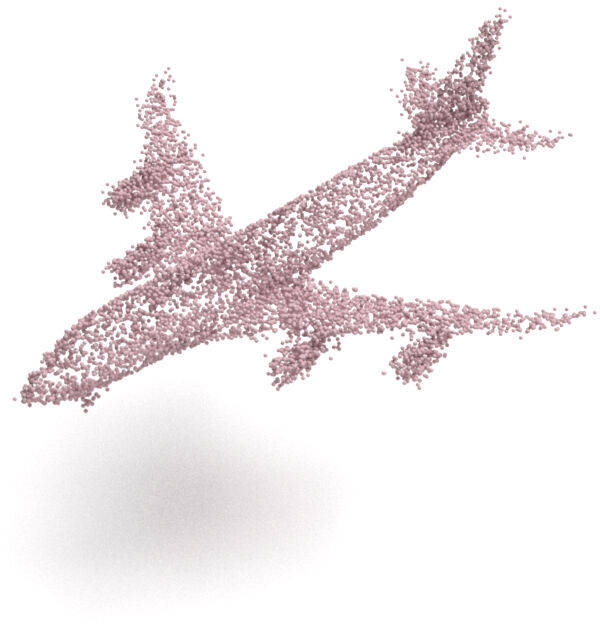}              &
    \includegraphics[width=16mm]{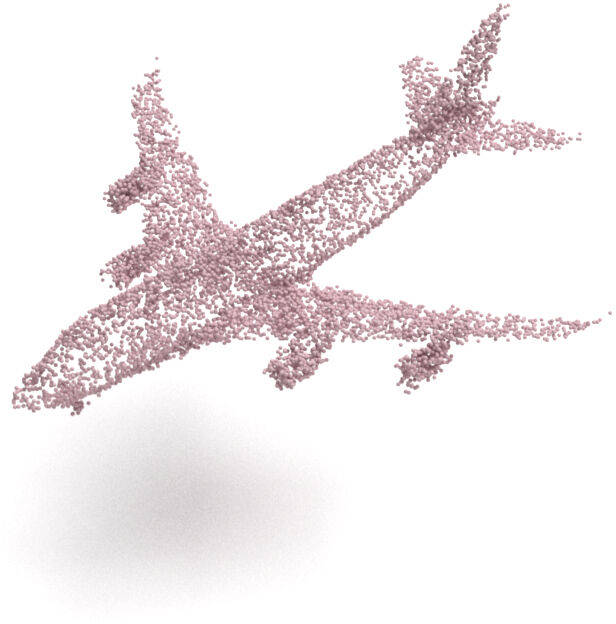}         &
    \includegraphics[width=16mm]{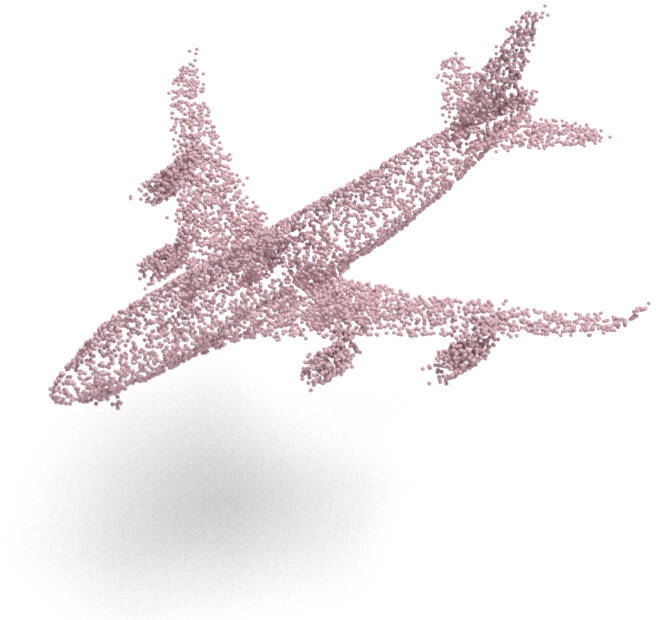}                     \\[-0.5mm]
    &
    \includegraphics[width=16mm]{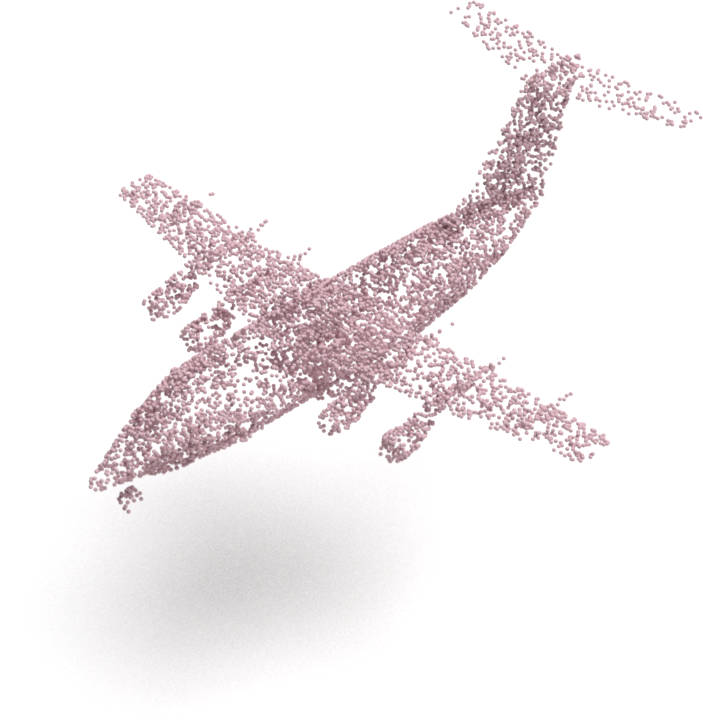}                           &
    \includegraphics[width=16mm]{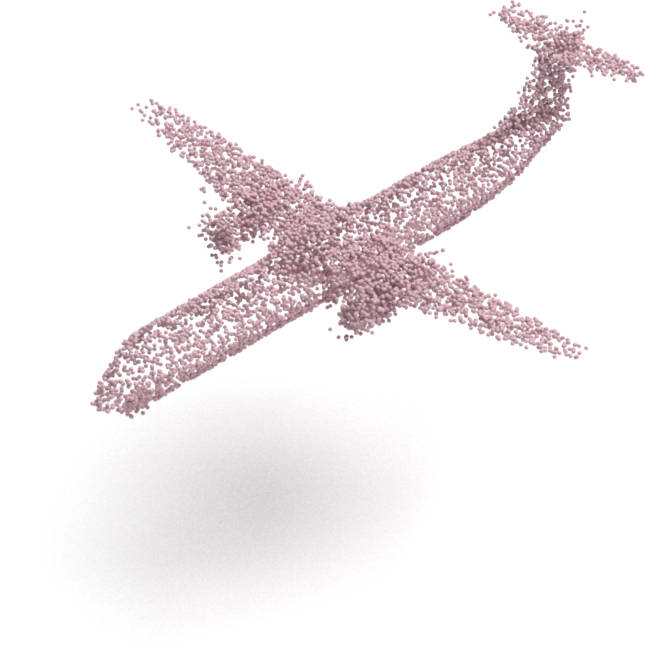}&
    \includegraphics[width=16mm]{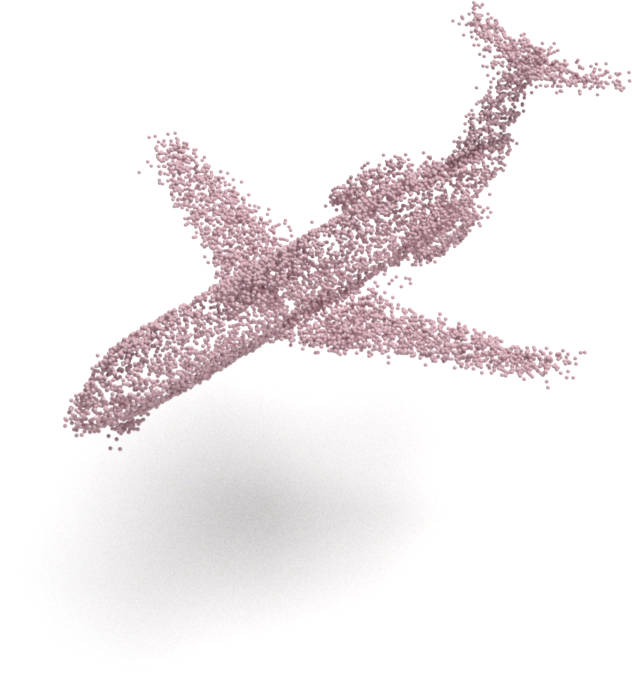}&
    \includegraphics[width=16mm]{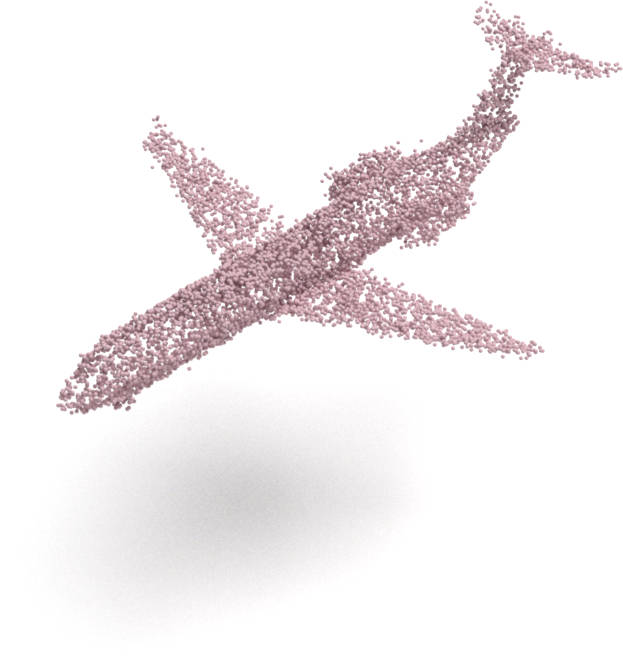}&
    \includegraphics[width=16mm]{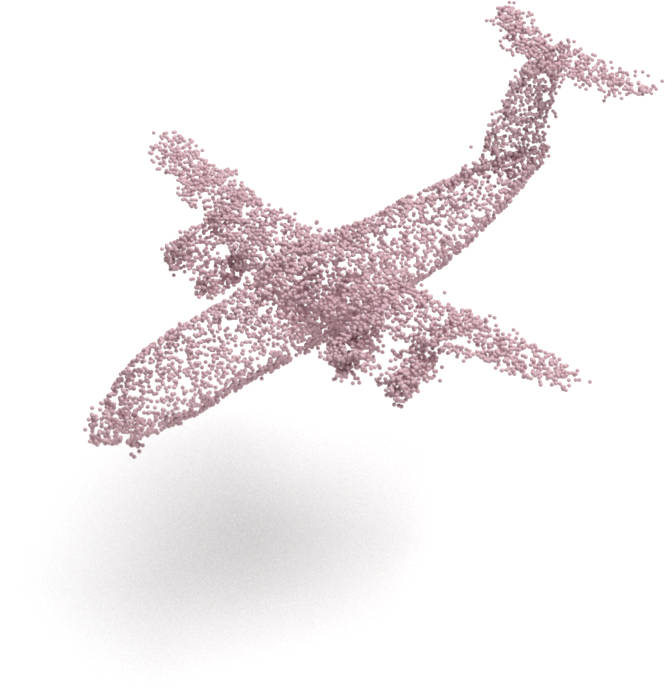}\\[-7.0mm]

    \rotatetitle{18mm}{\hspace*{-21.0mm}\textbf{Chair}}                                                                                      &
    \includegraphics[width=11mm]{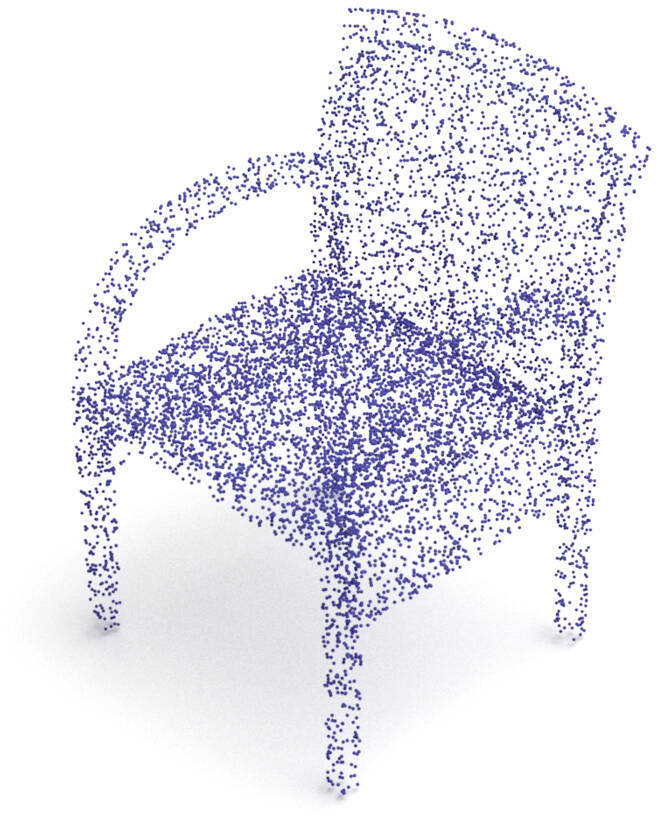}                            &
    \includegraphics[width=11mm]{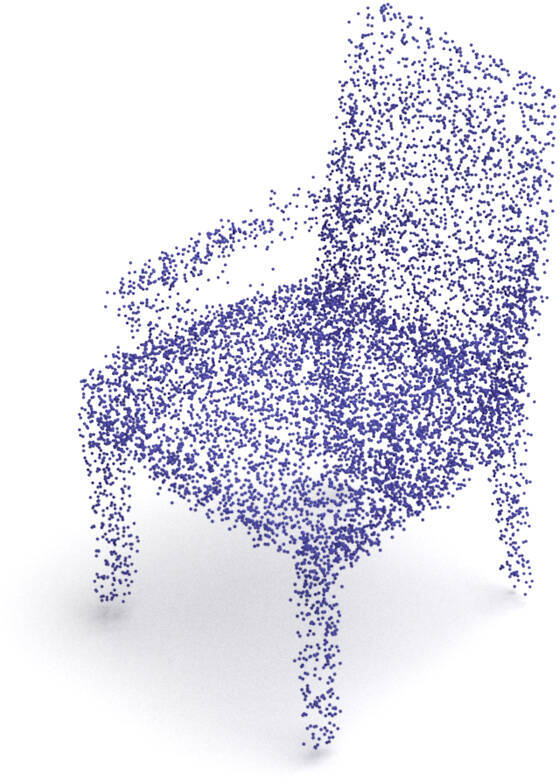}             &
    \includegraphics[width=11mm]{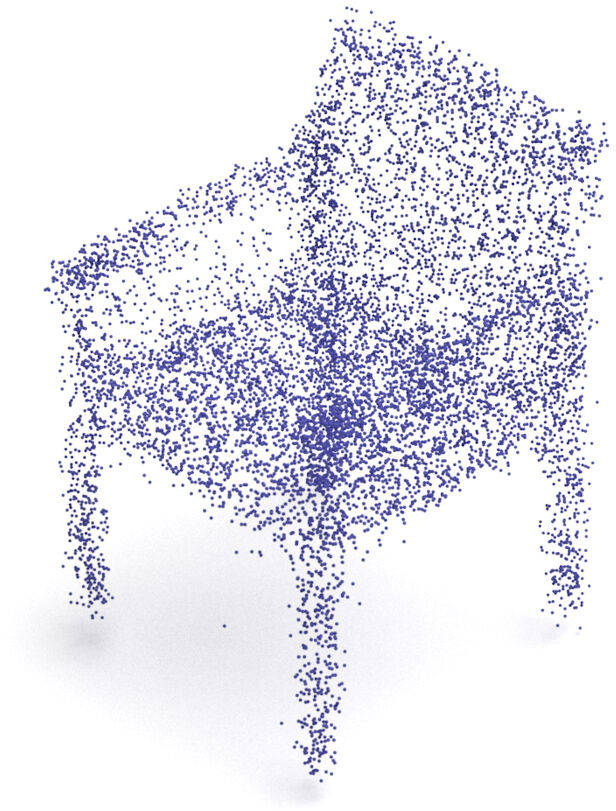}           &
    \includegraphics[width=11mm]{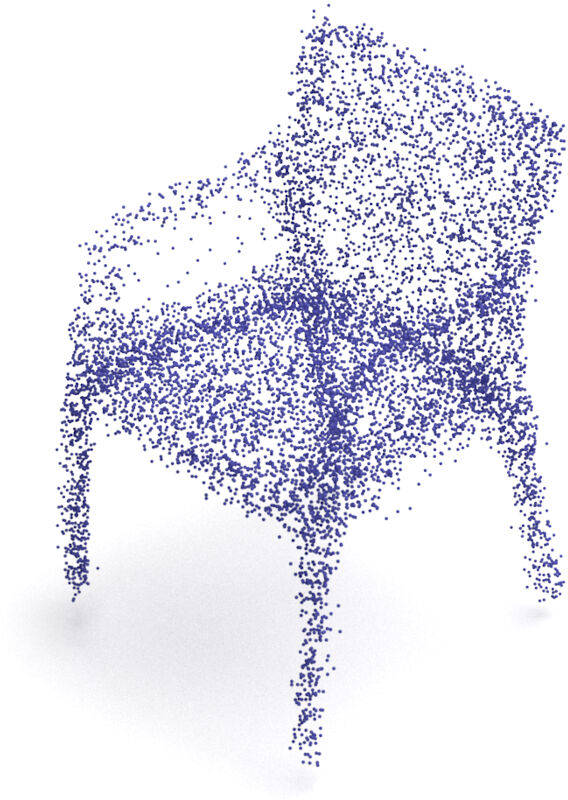}       &
    \includegraphics[width=11mm]{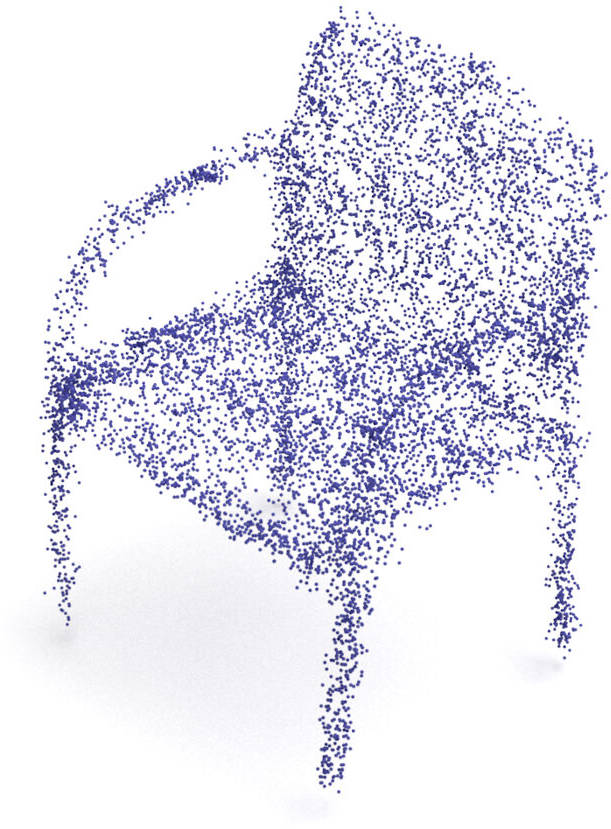} \\[-1.5mm]
    &
    \includegraphics[width=15mm]{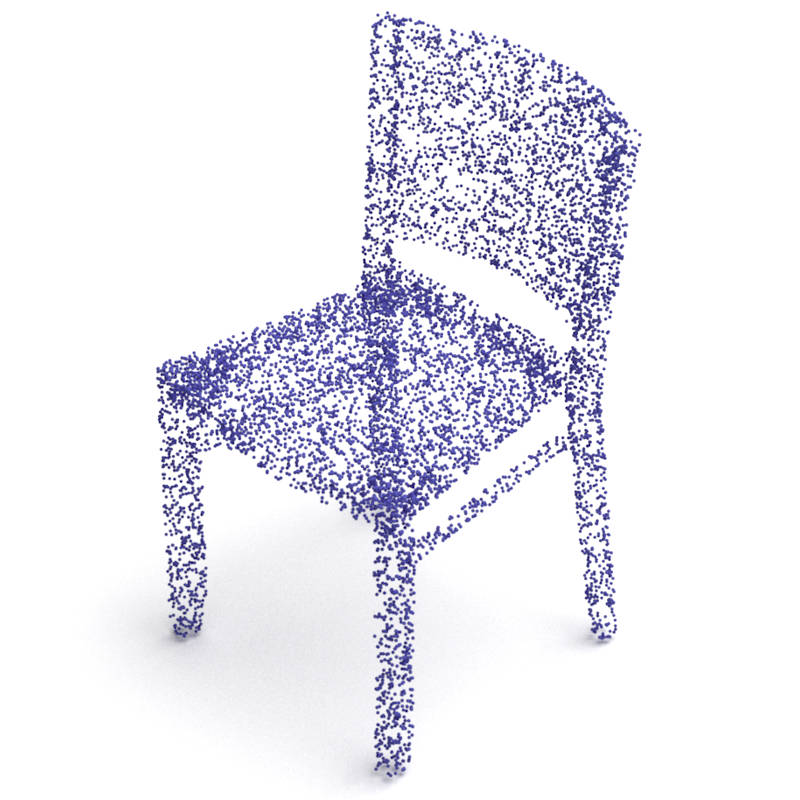}                            &
    \includegraphics[width=15mm]{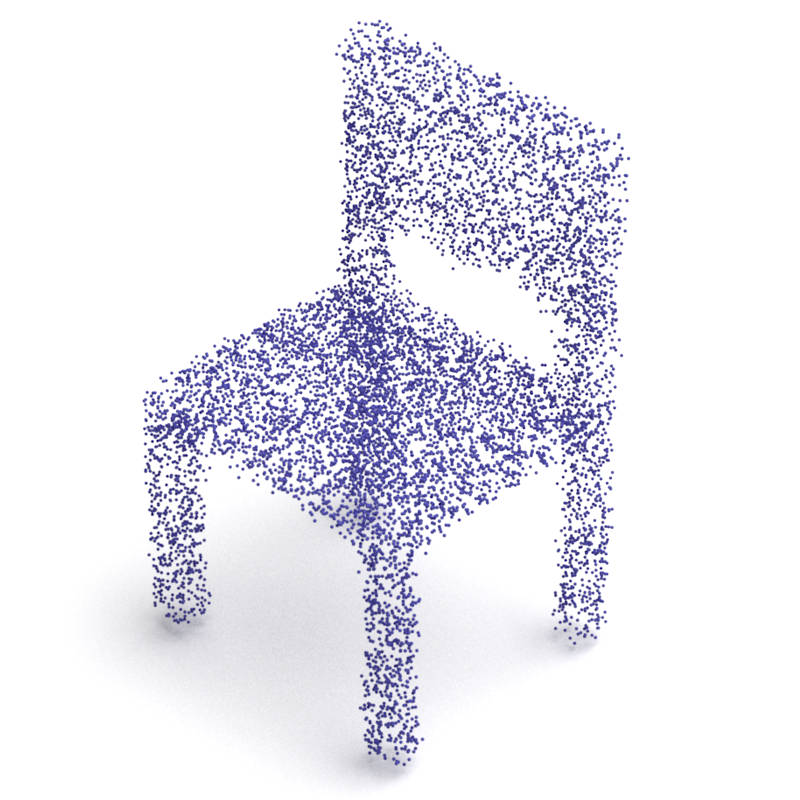}             &
    \includegraphics[width=15mm]{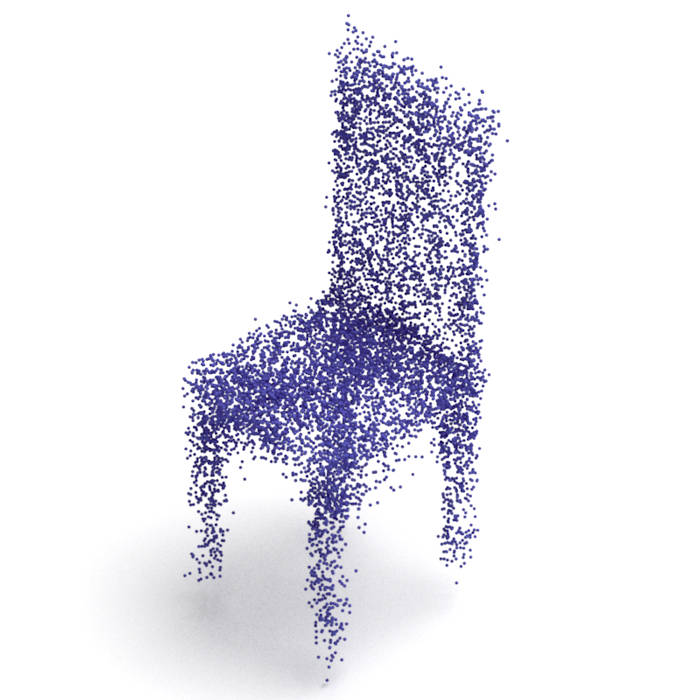}           &
    \includegraphics[width=15mm]{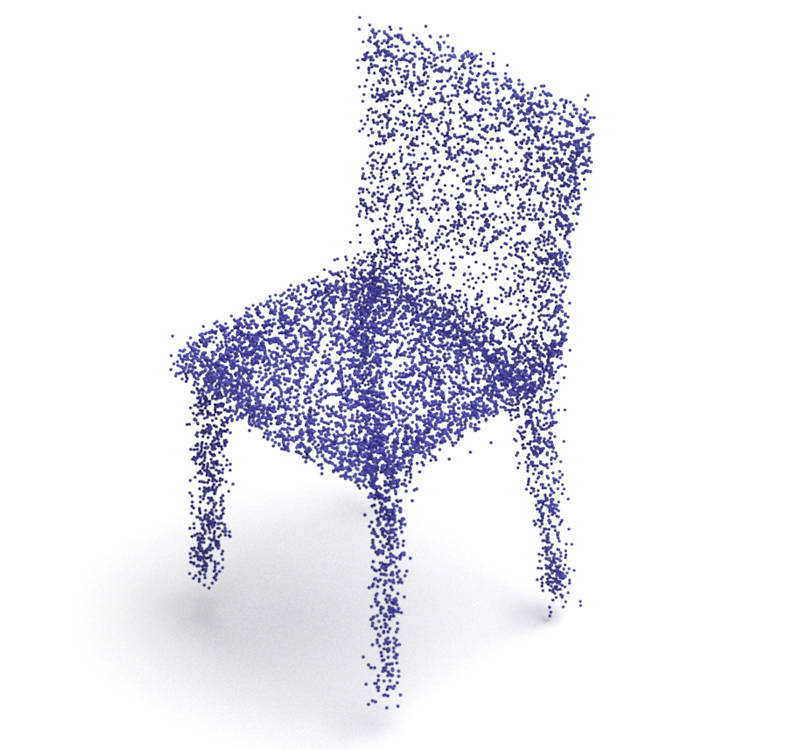}       &
    \includegraphics[width=15mm]{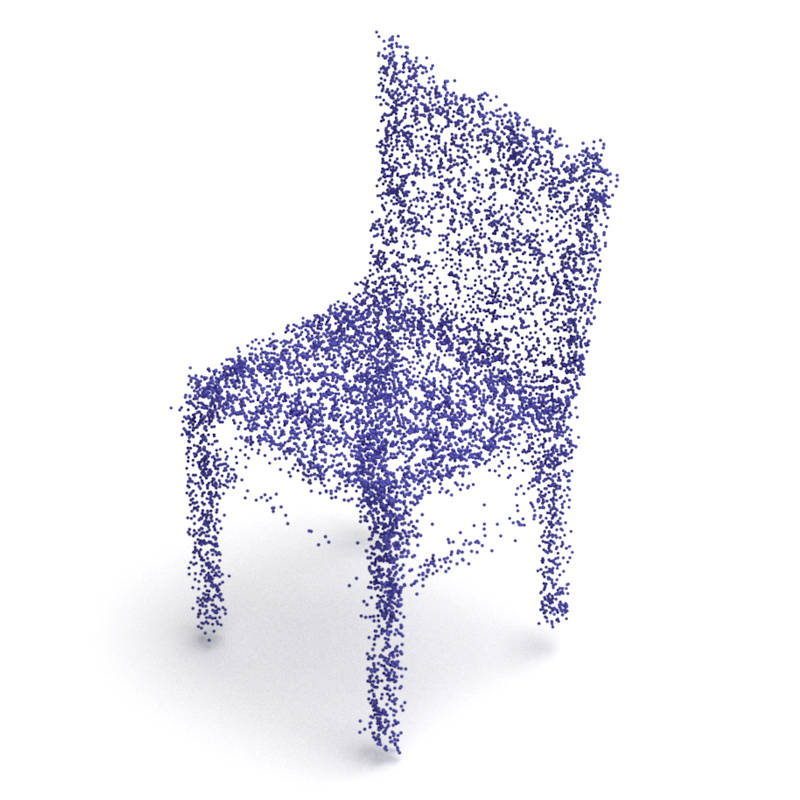} \\[-6.5mm]

    \rotatetitle{18mm}{\hspace*{-21.0mm}\textbf{Car}}                                                                                    &
    \includegraphics[width=15mm]{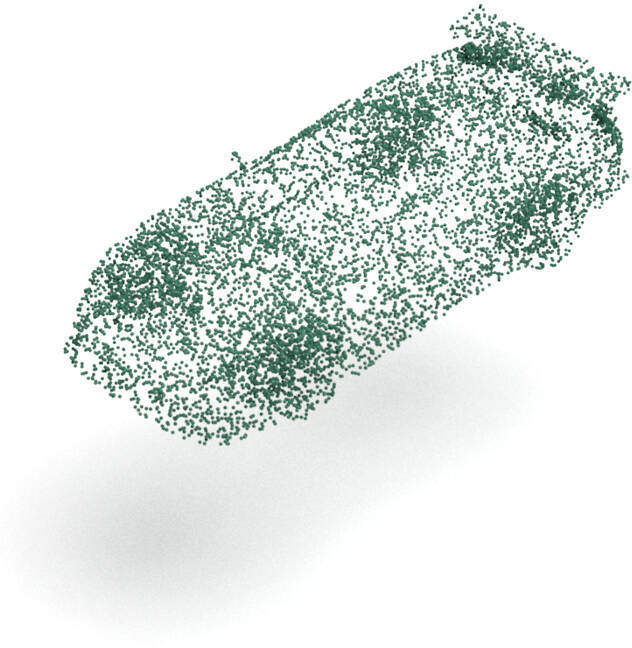} &
    \includegraphics[width=15mm]{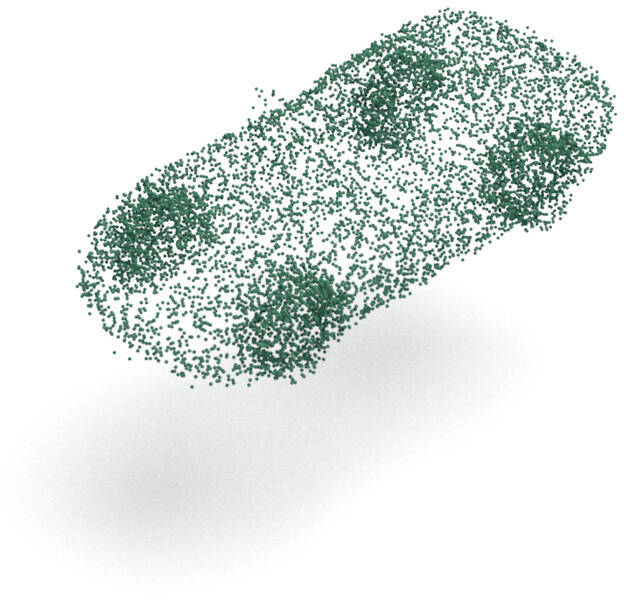}&
    \includegraphics[width=15mm]{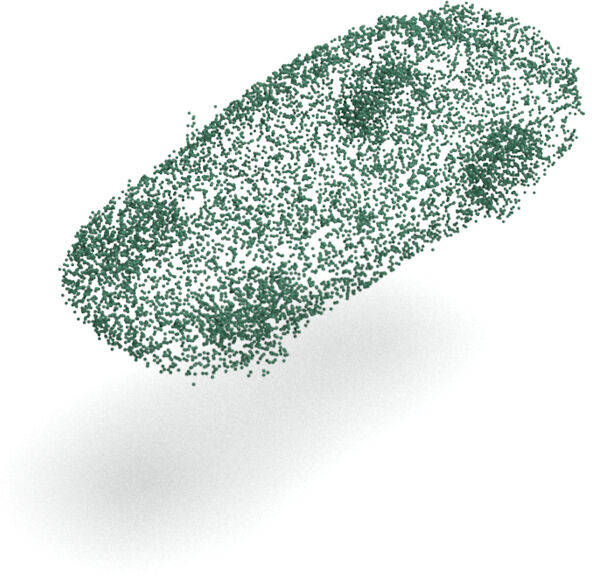} &
    \includegraphics[width=15mm]{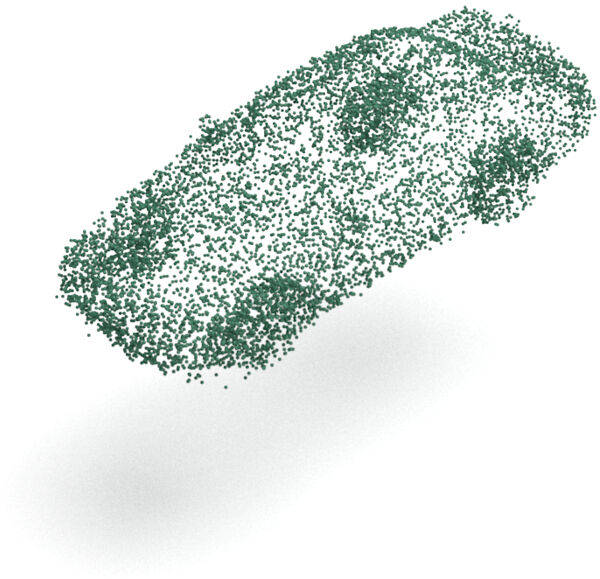}   &
    \includegraphics[width=15mm]{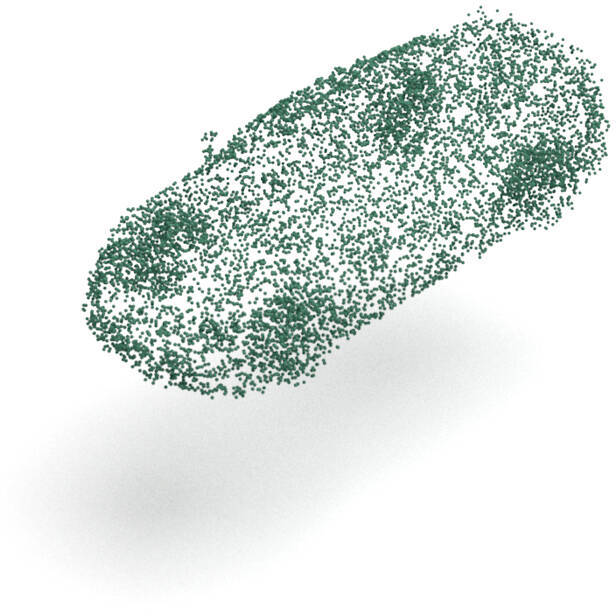}\\[-1.5mm]
    &
    \includegraphics[width=16mm]{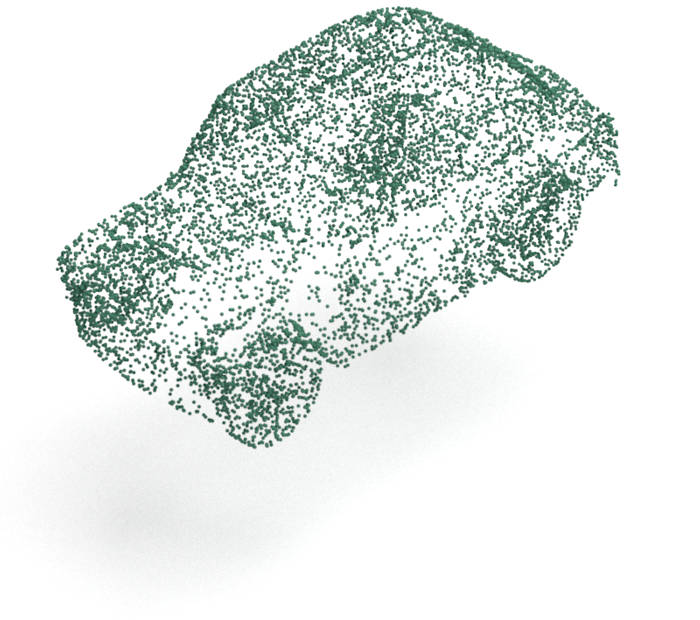} &
    \includegraphics[width=16mm]{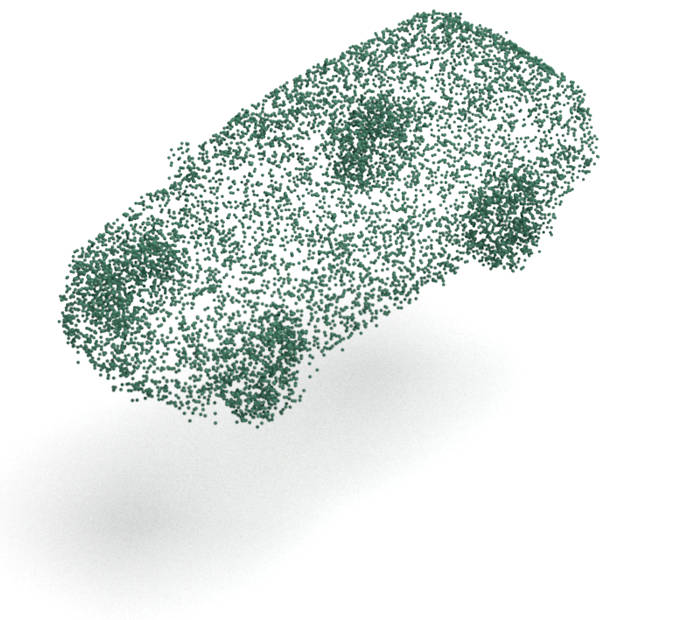}&
    \includegraphics[width=16mm]{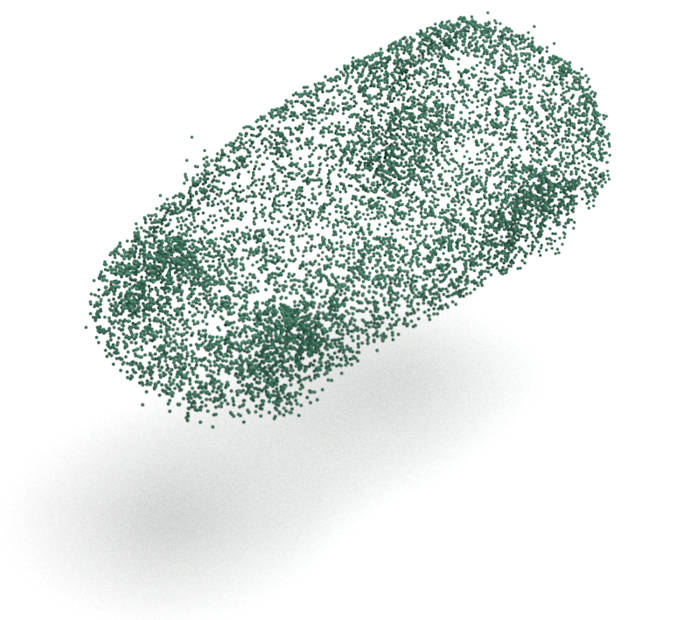} &
    \includegraphics[width=16mm]{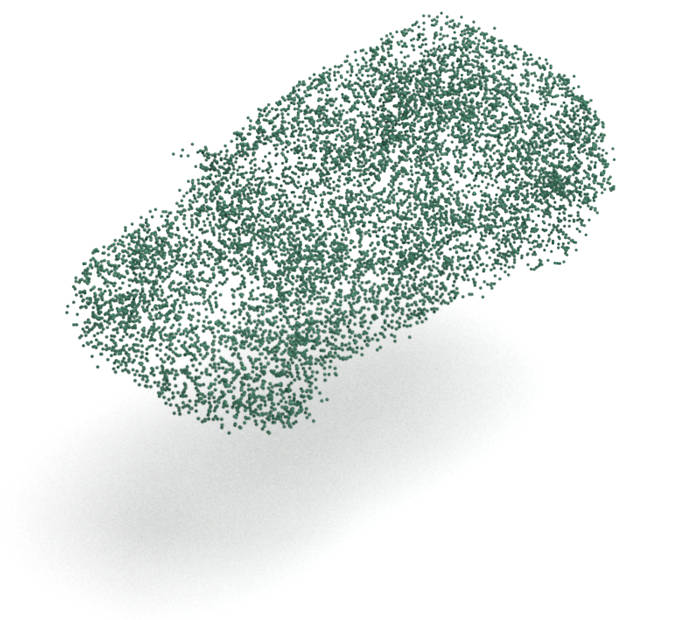}   &
    \includegraphics[width=16mm]{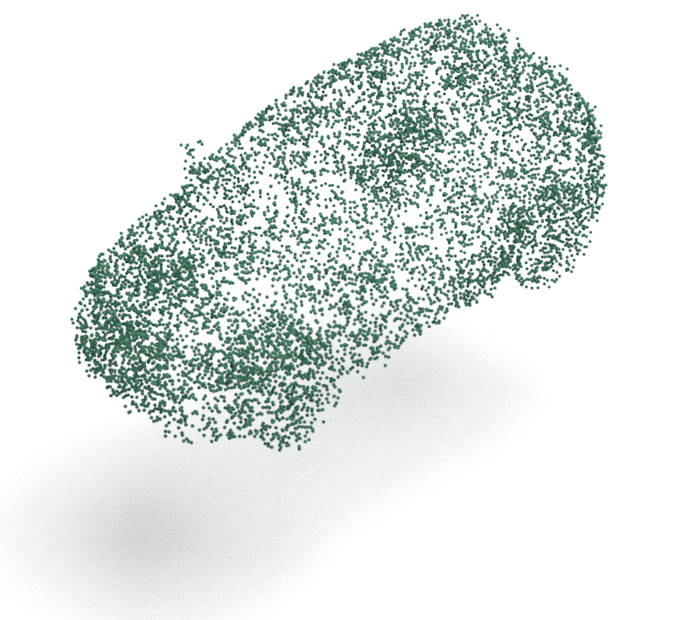}
    \\
  \end{tabular}
  \vspace*{-6mm}
  \caption{Generation examples nearest to the references taken from the datasets.}
  \label{fig:generated_samples_from_each_model}
  \vspace*{-4mm}
\end{figure}

\subsection{Generation Task}\label{sec:generation}
For the generation task, we compared ChartPointFlow with point clouds generators, namely r-GAN~\cite{Achlioptas2018}, l-GAN~\cite{Achlioptas2018}, PC-GAN~\cite{Li2019}, ShapeGF~\cite{Cai2020}, PointFlow~\cite{Yang2019}, and SoftFlow~\cite{Kim2020}.

For ChartPointFlow, we took the average results of 16 runs to suppress the variance due to the randomness in the generation, and summarized the results in Table~\ref{tab:generation_results}.
The results of ShapeGF were obtained using the official release code\footnote{\url{https://github.com/RuojinCai/ShapeGF}} under the same experimental settings, and those of the other methods for comparison were obtained from \cite{Yang2019} and \cite{Kim2020}.
The top four methods are based on GANs, ShapeGF is based on the implicit function theorem, and the others are flow-based models.
One can see that ChartPointFlow outperforms the other methods in all categories.
We provided the results of ChartPointFlow with 28, 20, and 24 charts for the airplane, chair, and car categories, respectively.
The results with different numbers of charts are summarized in Appendix~\ref{app:number_of_charts}.
ChartPointFlow achieved state-of-the-art results with 16--28 charts for all categories.
The results of the other metrics are summarized in Appendix~\ref{app:additional_metrics} just for reference.

Figure~\ref{fig:generated_samples} shows the samples generated by ChartPointFlow, each of which is composed of 10,000 points.
Each protruding subpart of an object, such as the airplane's horizontal tails, the chair's legs, and the cars' wheels, is expressed using a different chart.
The same subparts of different objects are expressed by the same charts.
The chairs in Results A, C, and D do not have armrests and do not use the charts assigned to the armrests of the chairs in Results B and E (see the red arrows).
ChartPointFlow assigned several charts to the chair's seat and armrests only when needed, thereby expressing the varying topologies.
Other results are summarized in Appendix~\ref{app:additional_images}.

For comparison, we took reference samples from the evaluation subsets and chose the nearest samples in terms of EMD from the samples generated by each model, as shown in Fig.~\ref{fig:generated_samples_from_each_model}.
We used pretrained models of PointFlow\footnote{\url{https://github.com/stevenygd/PointFlow}} and SoftFlow\footnote{\url{https://github.com/ANLGBOY/SoftFlow}} distributed by the original authors.
ChartPointFlow generated samples more similar than others, suggesting that it generated a variety of shapes.

The other GAN-based methods~\cite{Shu2019,Valsesia2019,Ramasinghe2020} used different experimental settings.
Under the same experimental settings, we confirmed that ChartPointFlow outperformed these methods (see Appendix~\ref{app:partdataset}).

\begin{figure*}[t]\centering
  \begin{tabular}{lC{16mm}C{16mm}C{16mm}C{16mm}C{16mm}C{16mm}C{16mm}}
                                           &
    \footnotesize\textbf{Input Sample} &
    \footnotesize\textbf{ShapeGF} &
    \footnotesize\textbf{AtlasNet} &
    \footnotesize\textbf{AtlasNet V2} &
    \footnotesize\textbf{PointFlow} &
    \footnotesize\textbf{SoftFlow} &
    \footnotesize\textbf{ChartPointFlow} \hspace*{2mm}(proposed)\\ [-5.5mm]
    \rotatetitle{18mm}{\hspace*{-4.0mm}\textbf{Airplane}} &
    \includegraphics[width=15mm]{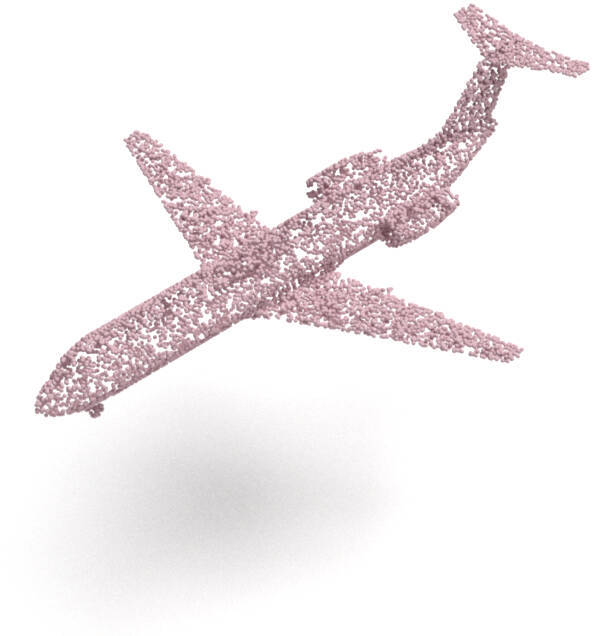} &
    \includegraphics[width=15mm]{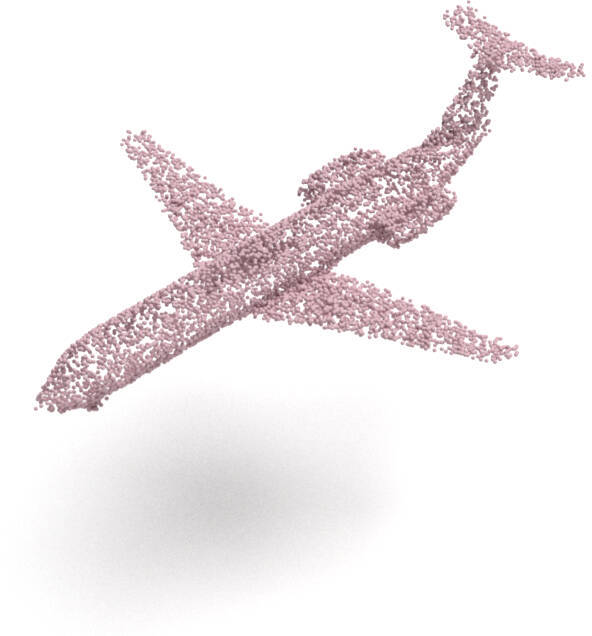}&
    \includegraphics[width=15mm]{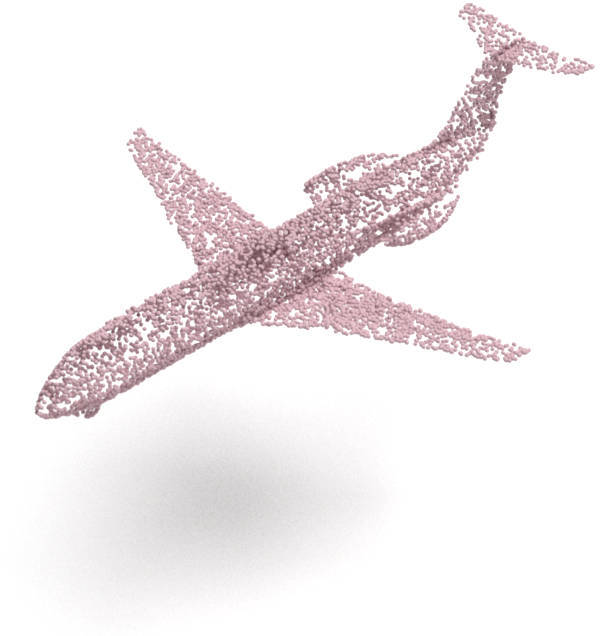}&
    \includegraphics[width=15mm]{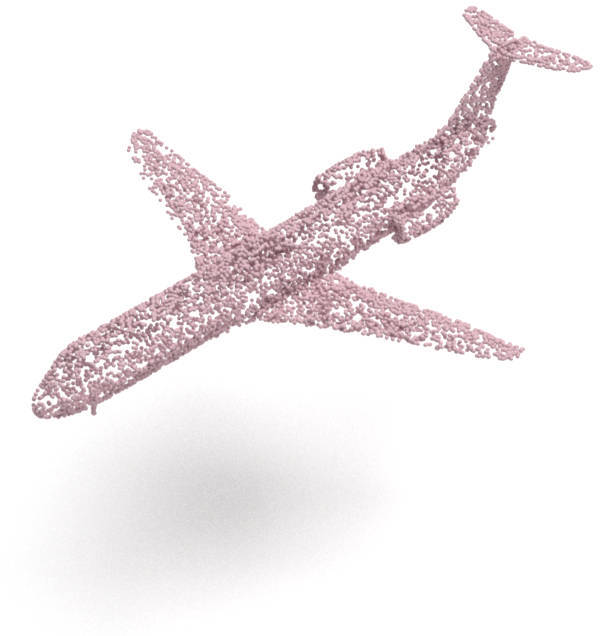}&
    \includegraphics[width=15mm]{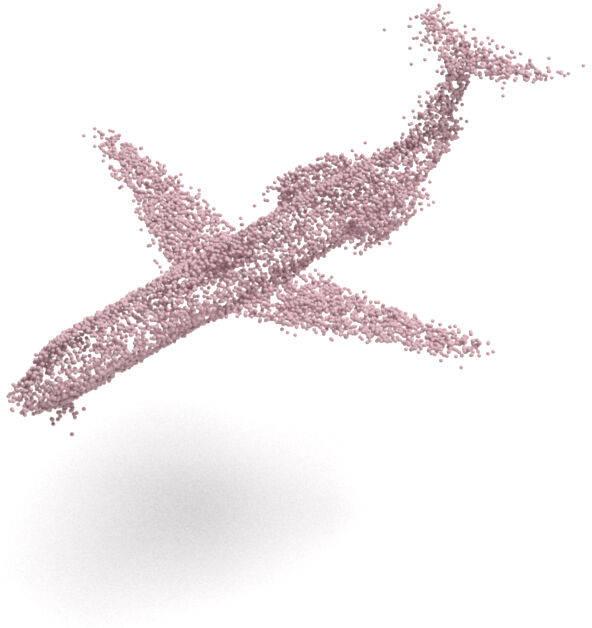}&
    \includegraphics[width=15mm]{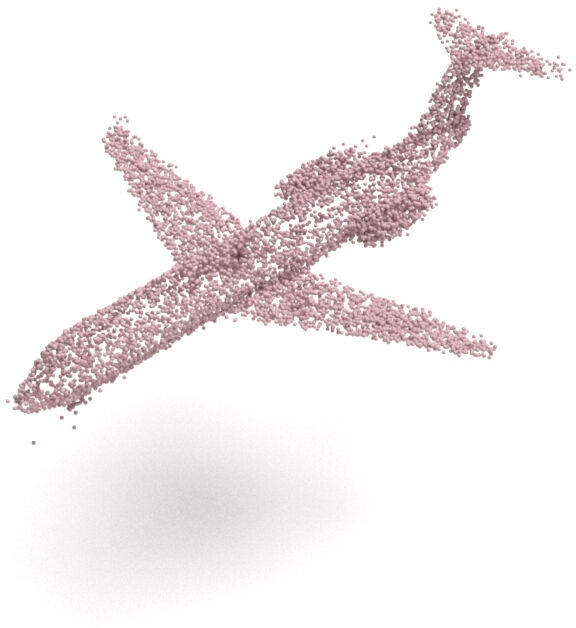}&
    \includegraphics[width=15mm]{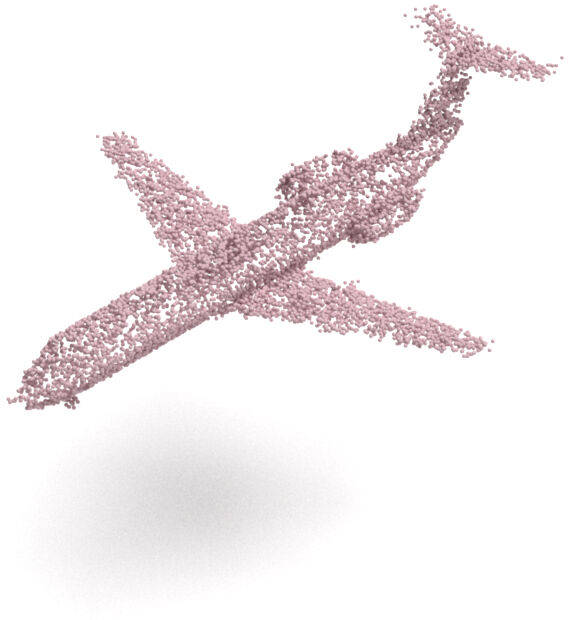}\\[-9.0mm]
    \rotatetitle{18mm}{\hspace*{-7.0mm}\textbf{Chair}} &
    \includegraphics[width=11mm]{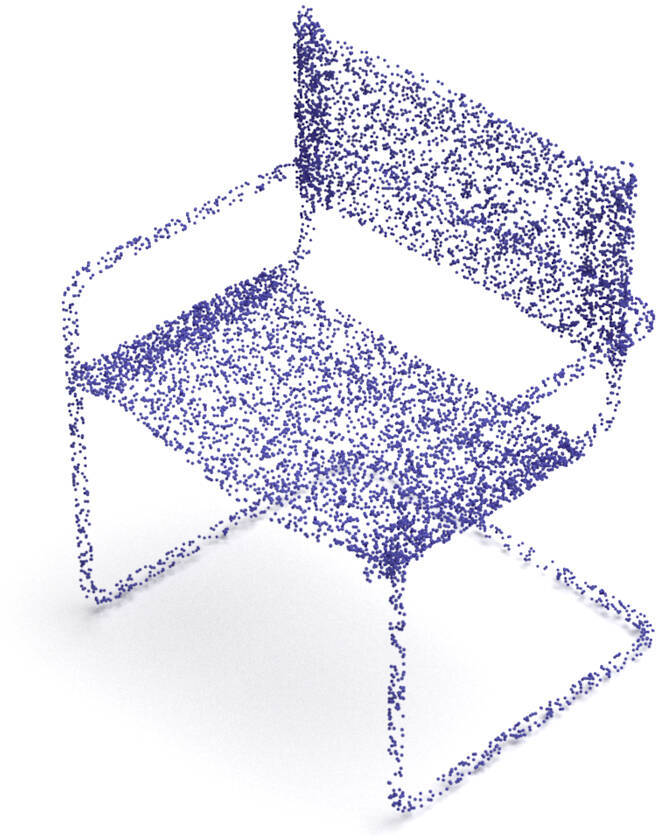}&
    \includegraphics[width=11mm]{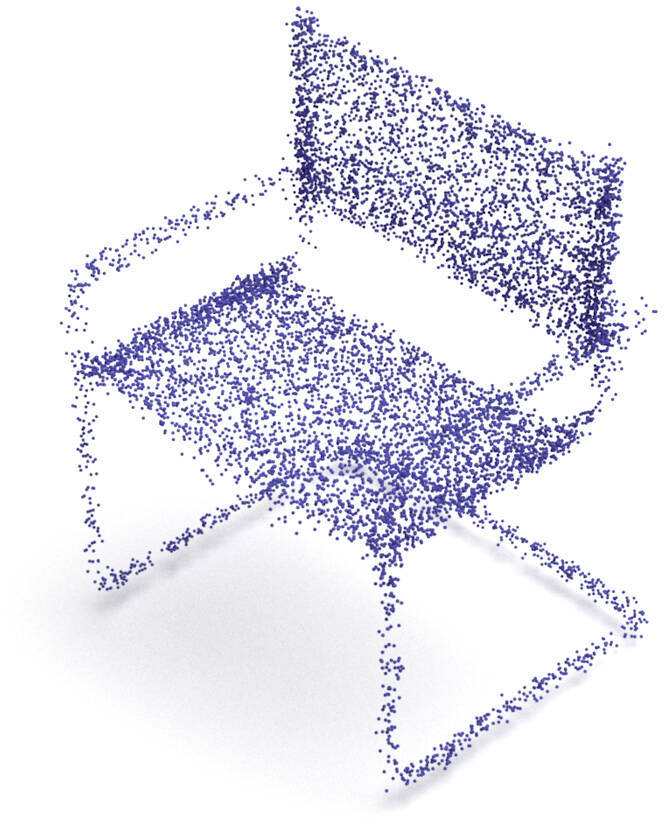}&
    \includegraphics[width=11mm]{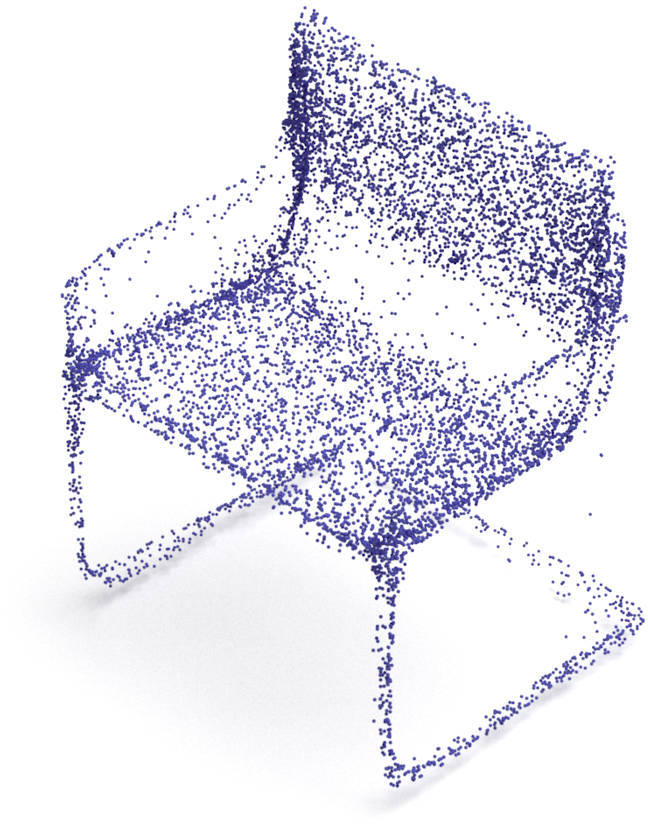}&
    \includegraphics[width=11mm]{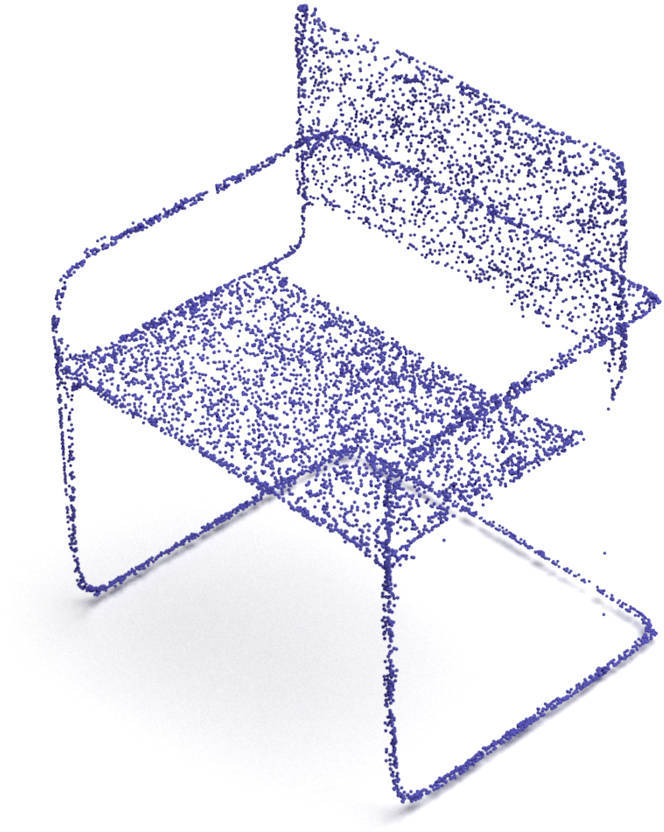}&
    \includegraphics[width=11mm]{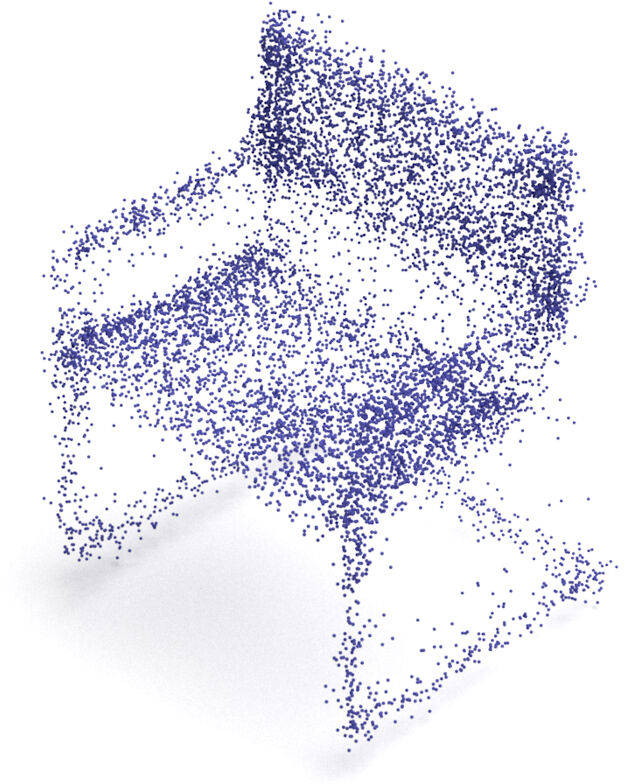}&
    \includegraphics[width=11mm]{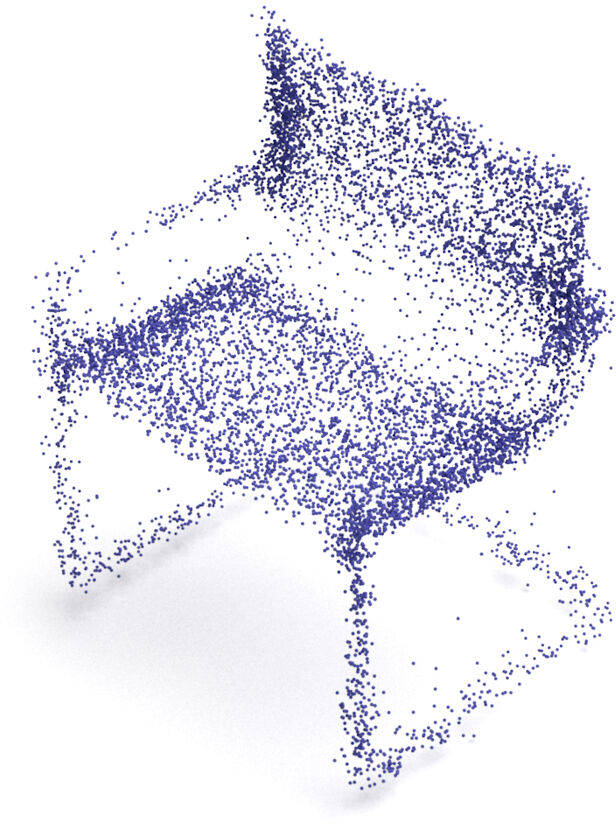}&
    \includegraphics[width=11mm]{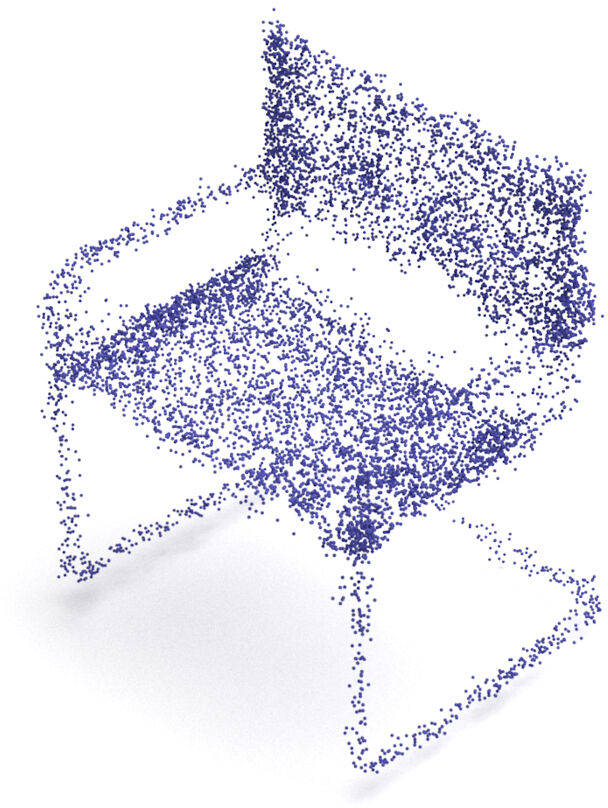}\\[-9.0mm]
    \rotatetitle{18mm}{\hspace*{-4.0mm}\textbf{Car}} &
    \includegraphics[width=15mm]{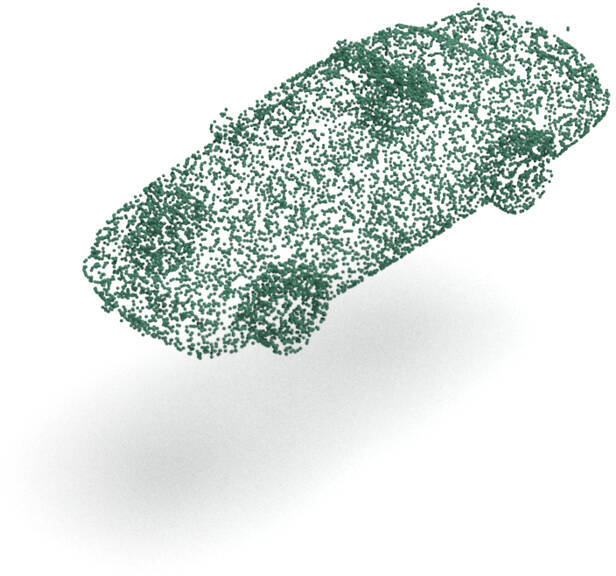}&
    \includegraphics[width=15mm]{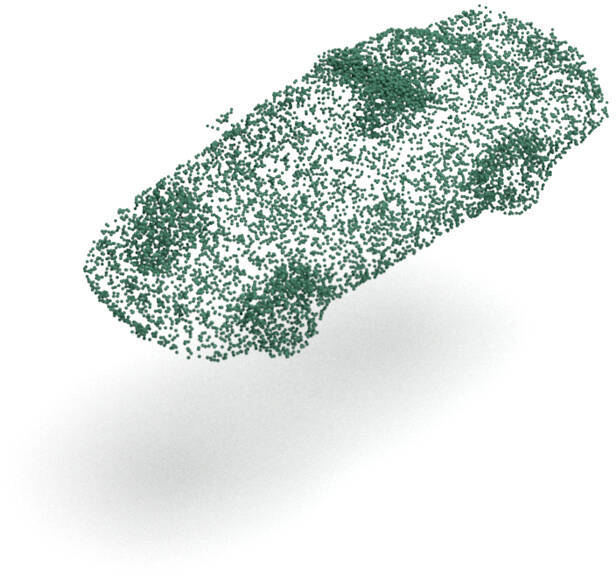}&
    \includegraphics[width=15mm]{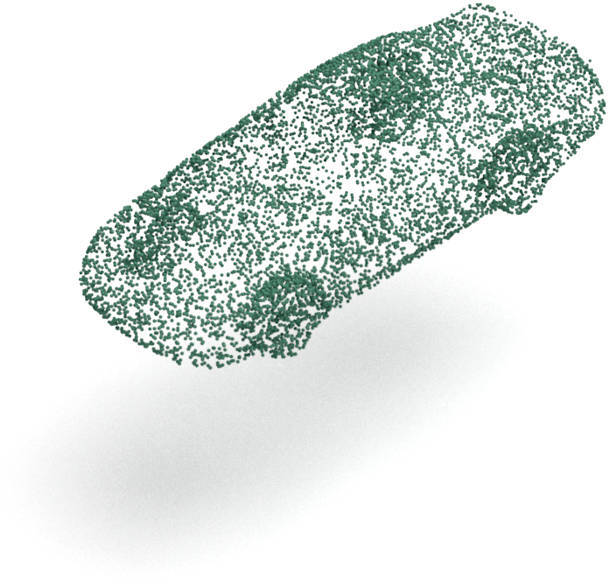}&
    \includegraphics[width=15mm]{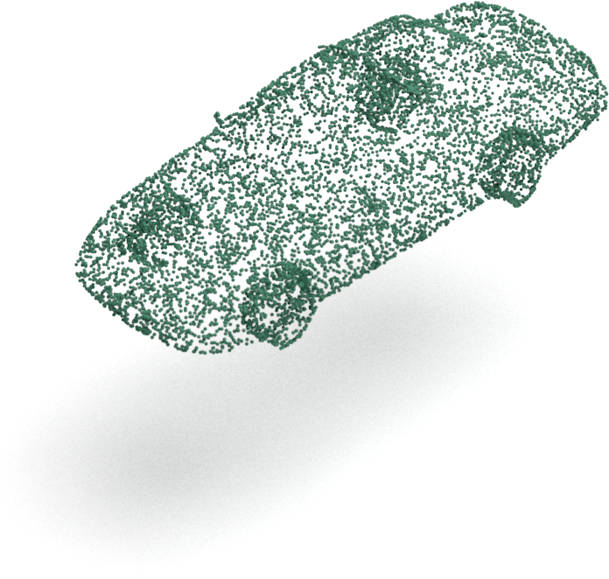}&
    \includegraphics[width=15mm]{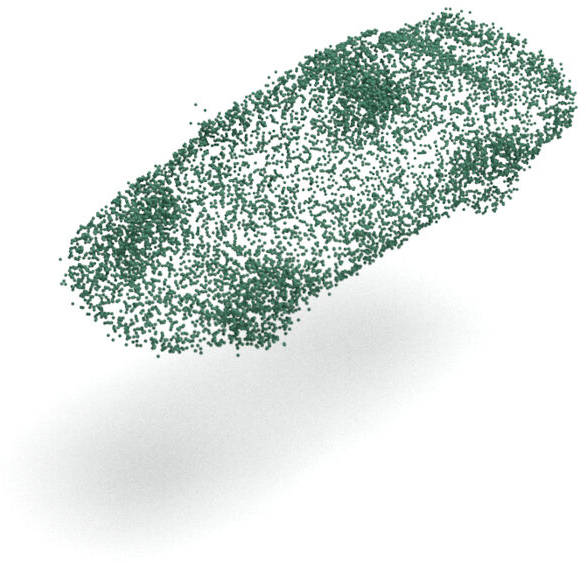}&
    \includegraphics[width=15mm]{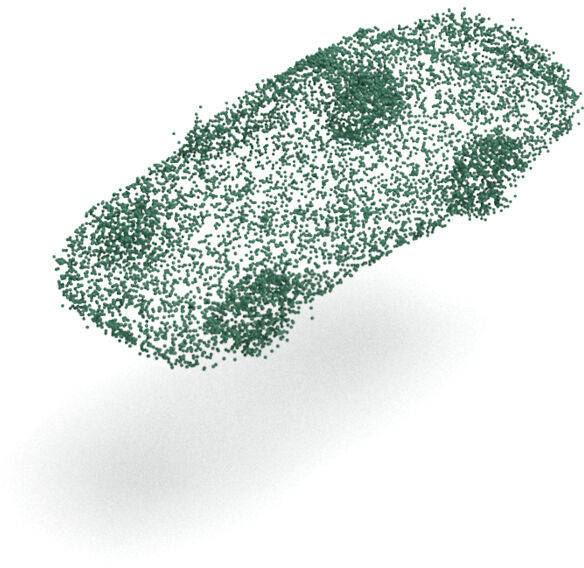}&
    \includegraphics[width=15mm]{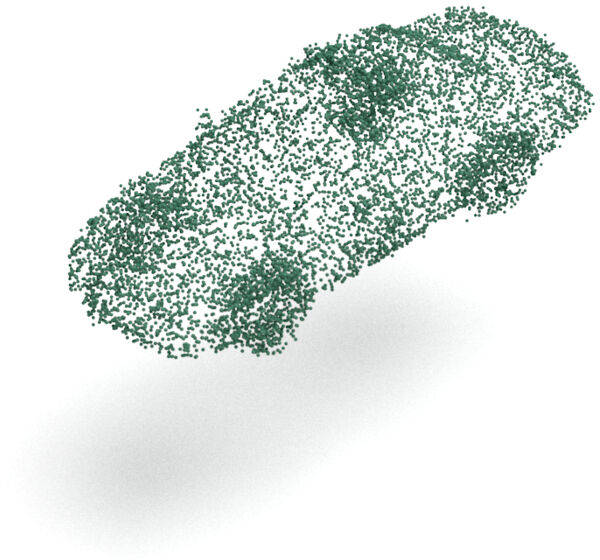}\\
  \end{tabular}
  \vspace*{-6mm}
  \captionof{figure}{Reconstruction examples.}
  \label{fig:reconstruct}
\end{figure*}

\begin{table}[t]\centering
  \begin{tabular}{lccc}
    \toprule
    Model                                & Airplane      & Chair         & Car           \\
    \midrule
    ShapeGF~\cite{Cai2020}               & 2.55          & 5.22          & 4.63          \\
    AtlasNet~\cite{Groueix2018}          & 2.95          & 6.68          & 4.75          \\
    AtlasNet V2 (PD)~\cite{Deprelle2019} & 3.28          & 5.67          & 4.51          \\
    AtlasNet V2 (PT)~\cite{Deprelle2019} & 3.57          & 5.97          & 5.13          \\
    PointFlow~\cite{Yang2019}            & 2.77          & 6.42          & 5.16          \\
    SoftFlow~\cite{Kim2020}              & 2.60          & 6.60          & 5.08          \\
    \midrule
    ChartPointFlow                       & \textbf{2.23} & \textbf{4.62} & \textbf{3.96} \\
    \bottomrule
  \end{tabular}
  \caption{Reconstruction errors. Smaller is better.}
  \label{tab:reconstruction_results}
  \vspace*{-7mm}
\end{table}

\subsection{Reconstruction Task}\label{sec:reconstruction}
For the reconstruction (or super-resolution) task, we measured the EMD between a reference point cloud and a reconstructed one, and summarized the results averaged over five trials (see Table \ref{tab:reconstruction_results}).
In this section, we used the pretrained model of PointFlow, which was trained only on the reconstruction task, whereas ChartPointFlow and SoftFlow were trained only on the generation task.
We also evaluated AtlasNet~\cite{Groueix2018} and AtlasNet V2 (patch deformation (PD) and point translation (PT))~\cite{Deprelle2019} with 25 patches (P25), which are specialized for the reconstruction task.
Using the original codes\footnote{\url{https://github.com/ThibaultGROUEIX/AtlasNet}}$^,$\footnote{\url{https://github.com/TheoDEPRELLE/AtlasNetV2}}, we trained AtlasNets ourselves under the same experimental settings.
We also evaluated ShapeGF~\cite{Cai2020}.

ChartPointFlow outperformed all the comparison methods in all categories.
The improvement from the performances of PointFlow and SoftFlow is the most significant for the chair category.
This may be because the chair category shows the varying shapes of armrests and legs and the varying number of holes in the backrest, i.e., the varying topologies.
Figure~\ref{fig:reconstruct} shows that ChartPointFlow reconstructed such shapes clearly.\footnote{AtlasNet V2 (PT) deals with a fixed number of points; thus, it is unavailable for reconstruction of a point cloud more dense than that used at the training phase.}
Because of the same reason, AtlasNet V2 outperformed PointFlow and SoftFlow in the chair category, but not in other categories.
Moreover, ChartPointFlow reconstructed even the airplane's front wheel and the car's mirrors.
PointFlow and SoftFlow generate shapes with different topologies only for simple target domains (e.g., 2D synthetic datasets, as shown in Fig.~\ref{fig:toydata}), and they suffer from blurs and artifacts in practice.
AtlasNet V2 reconstructed objects that are sharper than input objects; in other words, they have difficulty in expressing small subparts with accurate densities.
This is because AtlasNet V2 deformed the fixed number of fixed-size 2D patches.

See Appendices~\ref{app:number_of_charts} and~\ref{app:additional_images} for more results.

\subsection{Unsupervised Segmentation}
ChartPointFlow and AtlasNets assign each point to one of the charts (or patches~\cite{Groueix2018,Deprelle2019}).
This process can be regarded as clustering or unsupervised segmentation.
We evaluated the performances of ChartPointFlow and AtlasNets on the unsupervised part segmentation task.
The \texttt{PartDataset} of ShapeNet dataset contains labels corresponding to semantic parts for part segmentation, such as wings of an airplane~\cite{Yi2016}.
In particular, each of the three used categories is divided into four parts.

\begin{table}[t]\centering
  \begin{tabular}{lccc}
    \toprule
    Model                                & Airplane                      & Chair                         & Car                           \\
    \midrule
    AtlasNet~\cite{Groueix2018}          & 0.22 / 0.76                   & 0.23 / 0.74                   & 0.11 / 0.71                   \\
    AtlasNet V2 (PD)~\cite{Deprelle2019} & 0.25 / 0.79                   & 0.24 / 0.75                   & 0.13 / 0.72                   \\
    AtlasNet V2 (PT)~\cite{Deprelle2019} & 0.27 / \textbf{0.80}          & 0.24 / 0.74                   & 0.17 / 0.73                   \\
    \midrule
    ChartPointFlow                       & \textbf{0.30} / \textbf{0.80} & \textbf{0.35} / \textbf{0.86} & \textbf{0.19} / \textbf{0.79} \\
    \bottomrule
  \end{tabular}
  \caption{Segmentation performances (NMI/purity) with 25 clusters. Larger is better.}
  \label{tab:part_segmentation}
  \vspace*{-7mm}
  \setlength{\tabcolsep}{1mm}
\end{table}

After training, we fed all the unseen objects to a model to assign points to charts, and we obtained the purity (PUR) and normalized mutual information (NMI).
They are defined as,
\begin{equation}
  \hspace*{-5mm}\begin{split}
    \mathrm{PUR}(Y,\hat Y)&\textstyle=\frac{1}{|Y|}\sum_l \max_k \#\{y_j|y_j=k \mbox{ for } j \mbox{ such that }\hat y_j=l\},\\
    \mathrm{NMI}(Y,\hat Y)&\textstyle=\frac{2\ I(Y,\hat Y)}{H(Y)+H(\hat Y)},
  \end{split}\hspace*{-5mm}
\end{equation}
where $y_j\in Y$ and $\hat y_j \in \hat Y$ denote the ground truth label and the estimated chart of a point $x_j\in X$, respectively.
$l$ denotes the $l$-th cluster estimated by a model, and $k$ denotes the $k$-th ground truth label.

We evaluated ChartPointFlow and AtlasNets with 25 clusters (called charts or patches), which is the default number for AtlasNets.
AtlasNets do not have a chart predictor.
Instead, we performed a reconstruction task and a 1-nearest neighbor classification.
Specifically, we assigned a given point to the chart that the nearest reconstructed point belongs to.
ChartPointFlow outperformed AtlasNets for both criteria in all categories, except for the purity for airplane, as summarized in Table~\ref{tab:part_segmentation}.
See Appendix~\ref{app:number_of_charts} for the results with different numbers of charts.
We obtained the results of the part segmentation by assigning a label to each cluster so as to maximize the purity, as shown in Fig.~\ref{fig:segment}.
ChartPointFlow segmented the tail wing of an airplane, the legs of a chair, and the wheels of a car more clearly than AtlasNets, which contaminated the leg part of a chair with the seat part and the backrest part.
Because AtlasNets employed fixed-size patches, a patch used for a leg of a chair was used for different parts of other chairs when their legs were much smaller.

\begin{table}[t]\centering
  \setlength{\tabcolsep}{1mm}
  \hspace*{-4mm}\begin{tabular}{L{2mm}C{14mm}C{14mm}C{14mm}C{14mm}C{14mm}}
    &
    \footnotesize\textbf{Ground Truth}&
    \footnotesize\textbf{AtlasNet}&
    \footnotesize\textbf{AtlasNet V2} (PD)&
    \footnotesize\textbf{AtlasNet V2} (PT)&
    \footnotesize\textbf{ChartPointFlow} \hspace*{3.2mm}(proposed)\\ [-4.5mm]
    \rotatetitle{18mm}{\hspace*{-21.0mm}\textbf{Airplane}} &
    \includegraphics[width=17mm]{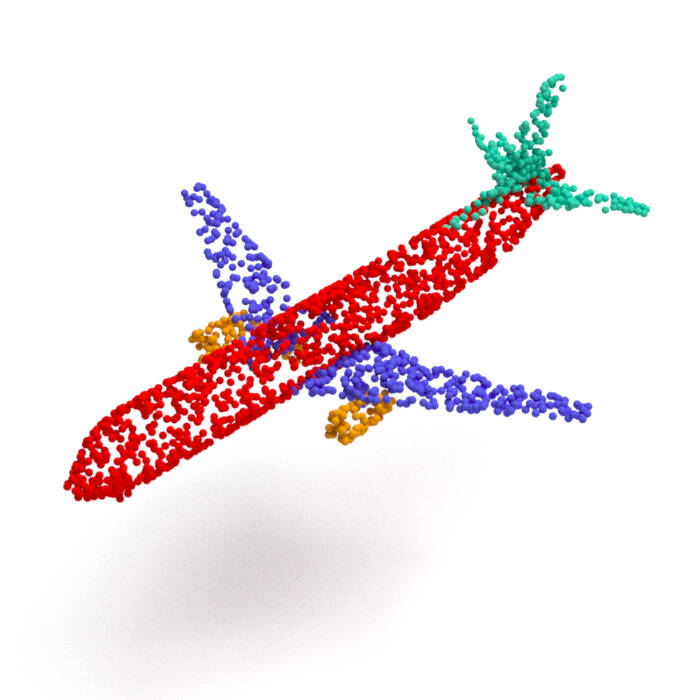}&
    \includegraphics[width=17mm]{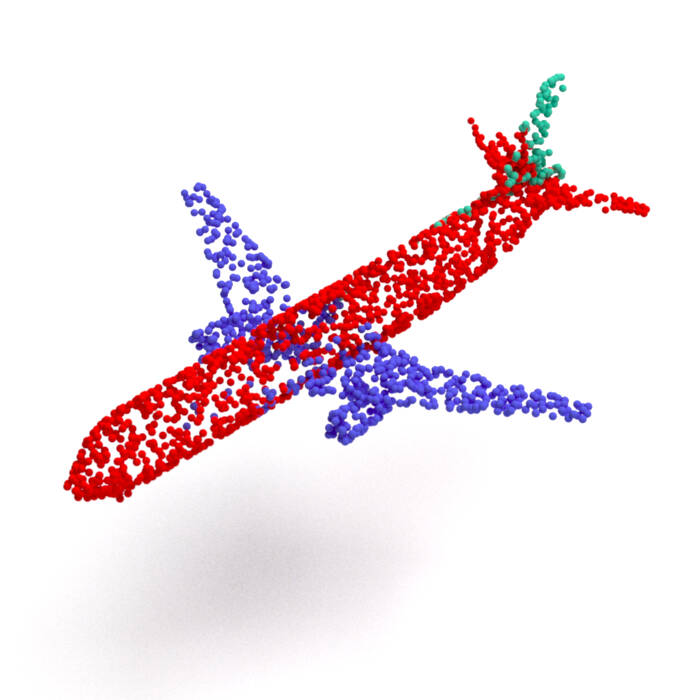}        &
    \includegraphics[width=17mm]{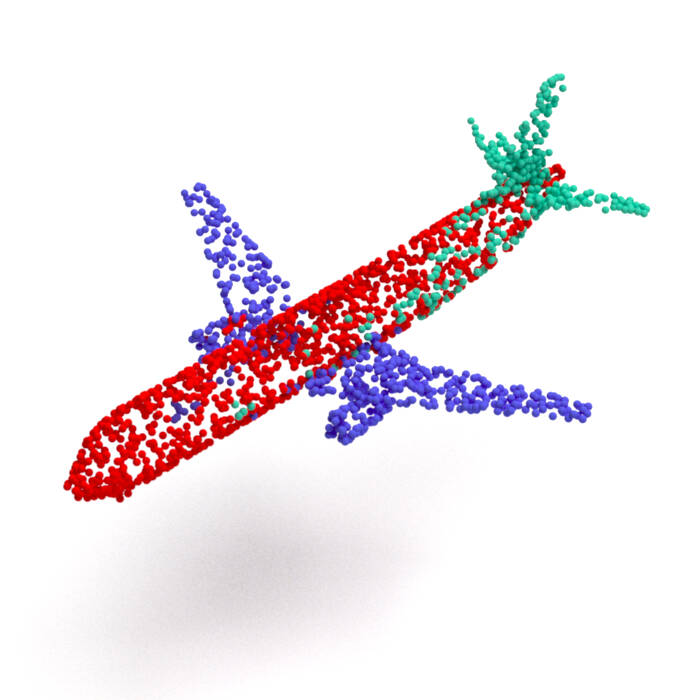} &
    \includegraphics[width=17mm]{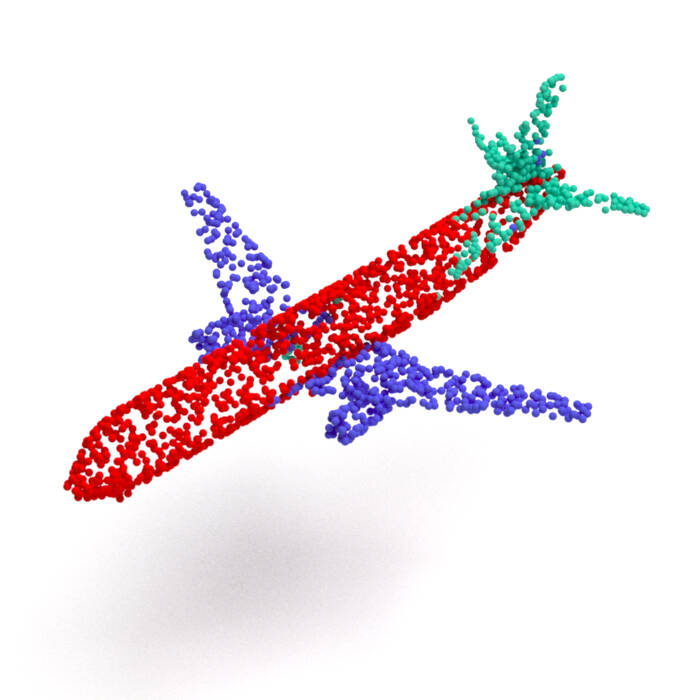} &
    \includegraphics[width=17mm]{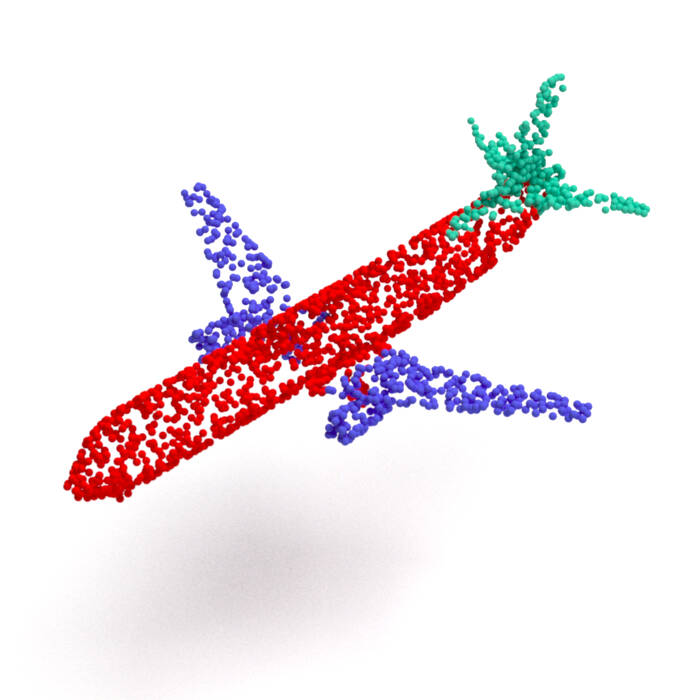}       \\
    &
    \includegraphics[width=17mm]{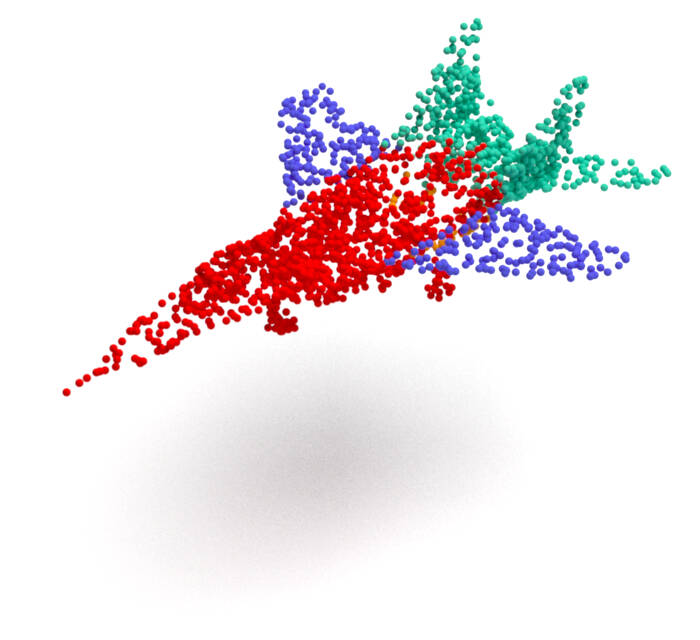}&
    \includegraphics[width=17mm]{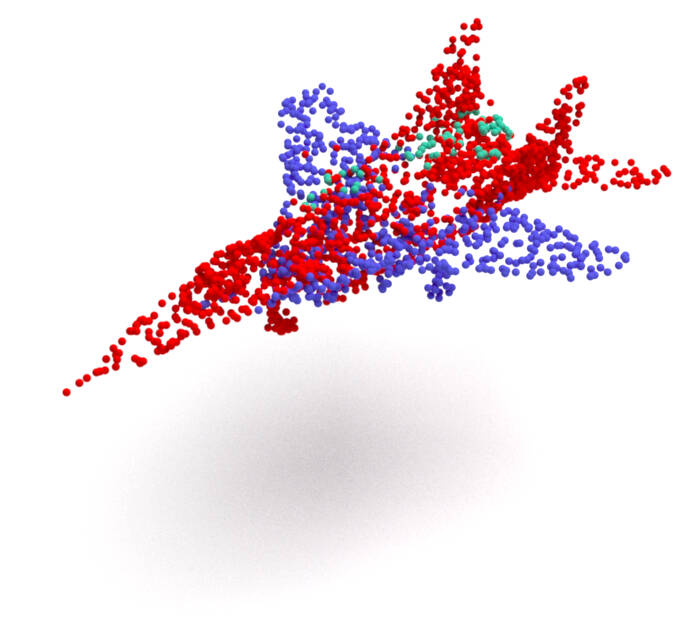}        &
    \includegraphics[width=17mm]{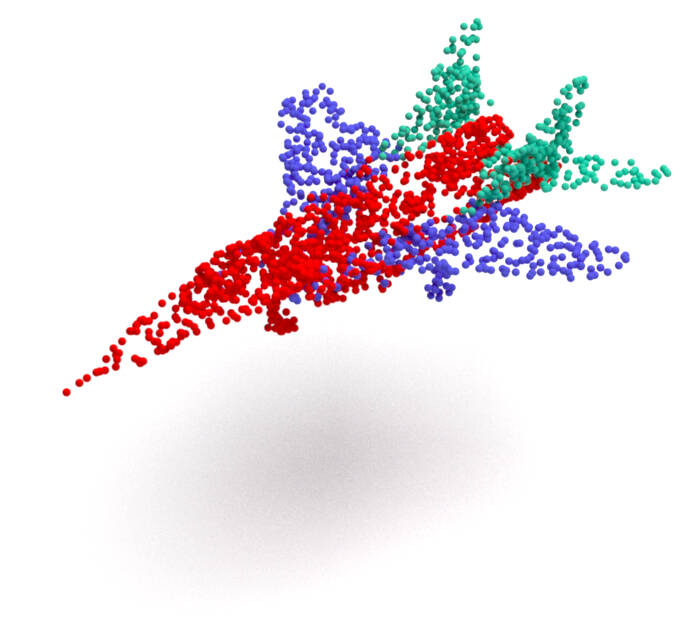} &
    \includegraphics[width=17mm]{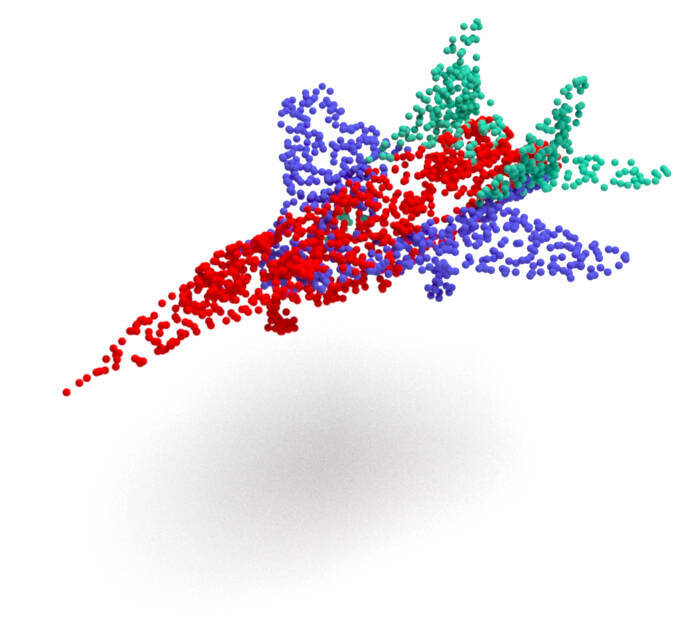} &
    \includegraphics[width=17mm]{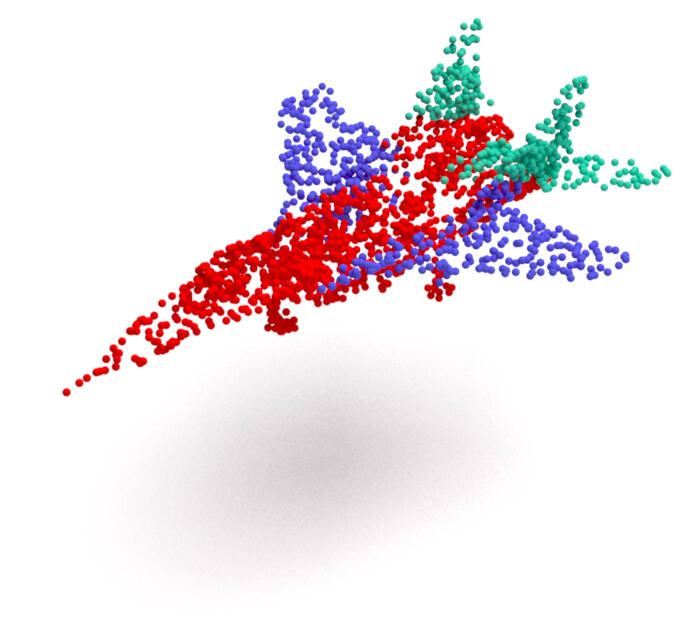}       \\[-5.5mm]
    \rotatetitle{18mm}{\hspace*{-21.0mm}\textbf{Chair}} &
    \includegraphics[width=17mm]{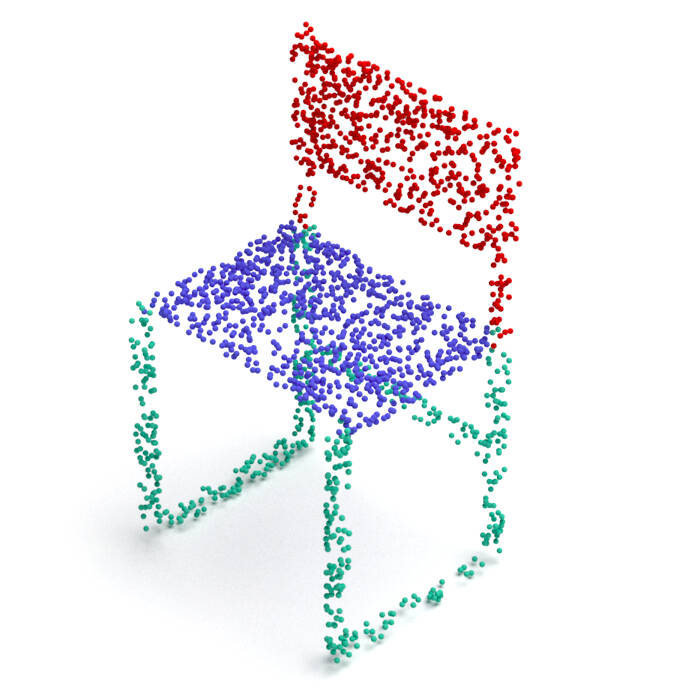}&
    \includegraphics[width=17mm]{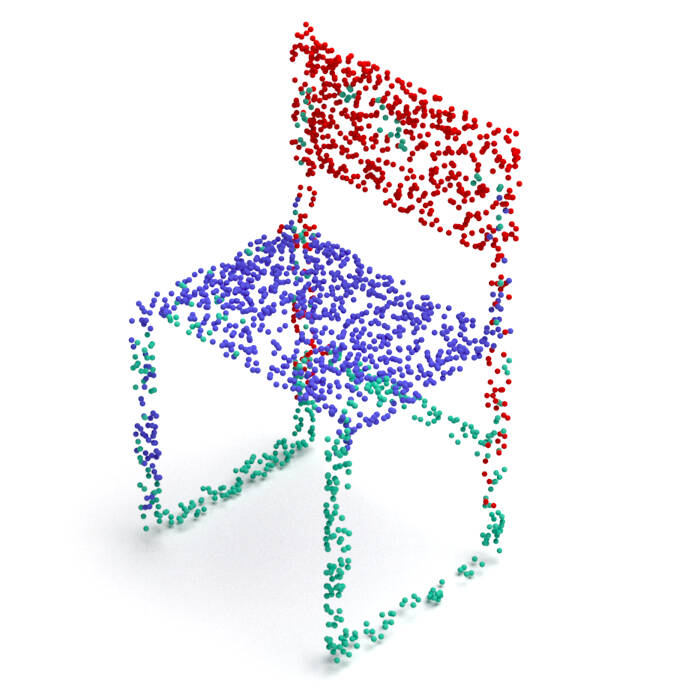}&
    \includegraphics[width=17mm]{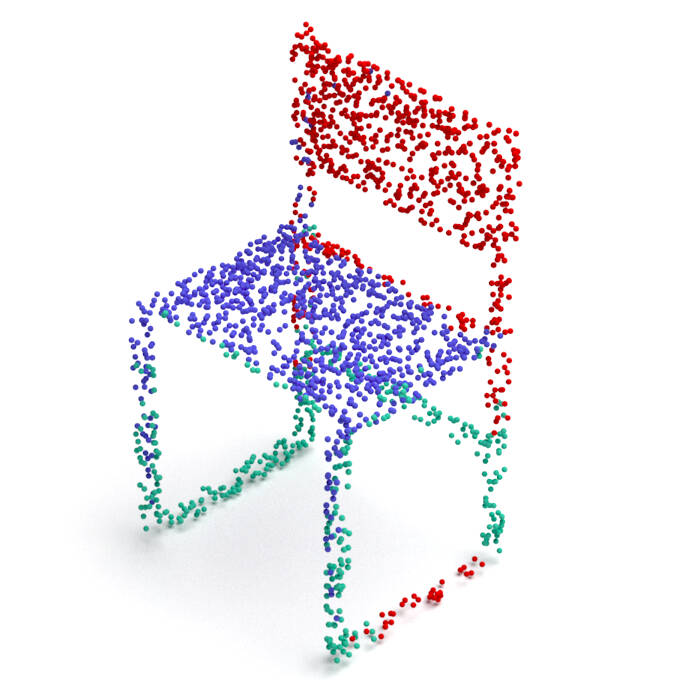}&
    \includegraphics[width=17mm]{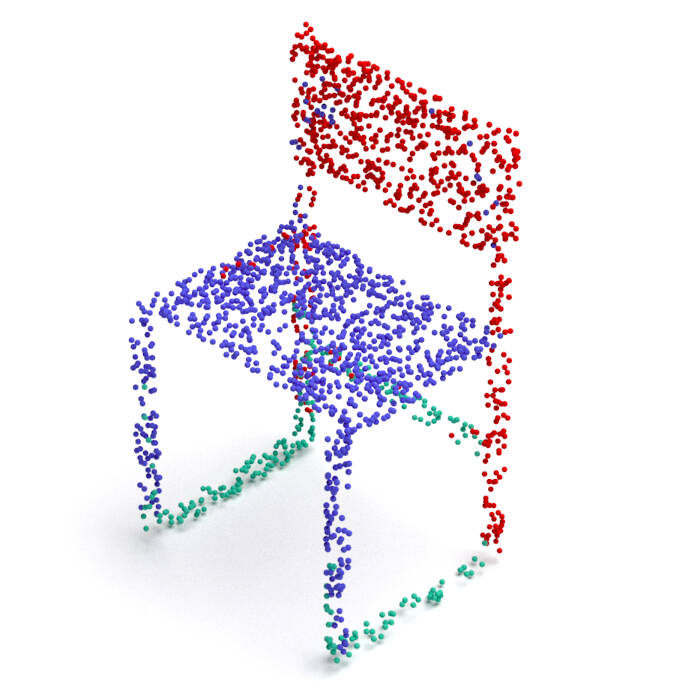}&
    \includegraphics[width=17mm]{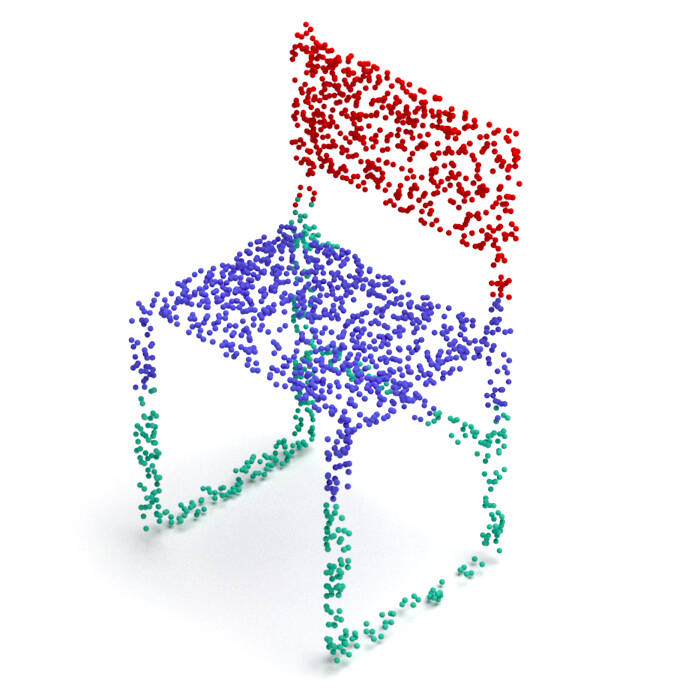}\\[-1.5mm]
    &
    \includegraphics[width=17mm]{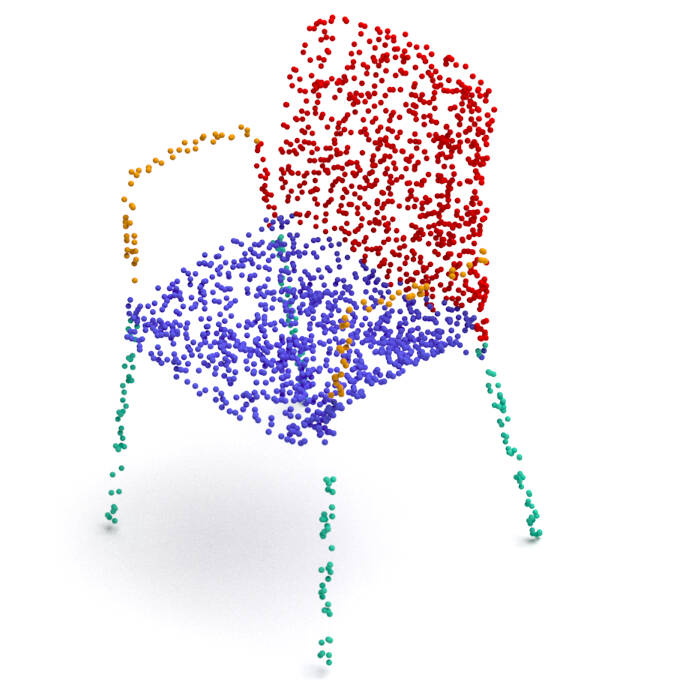}&
    \includegraphics[width=17mm]{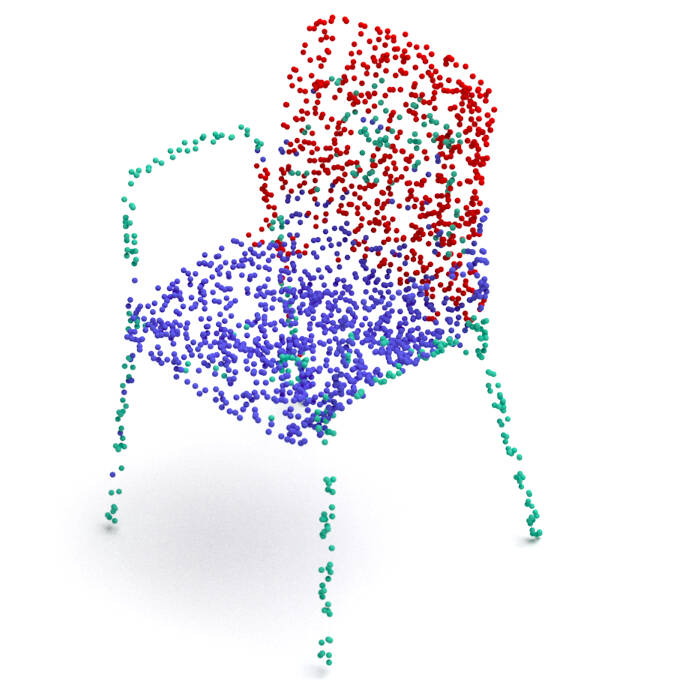}&
    \includegraphics[width=17mm]{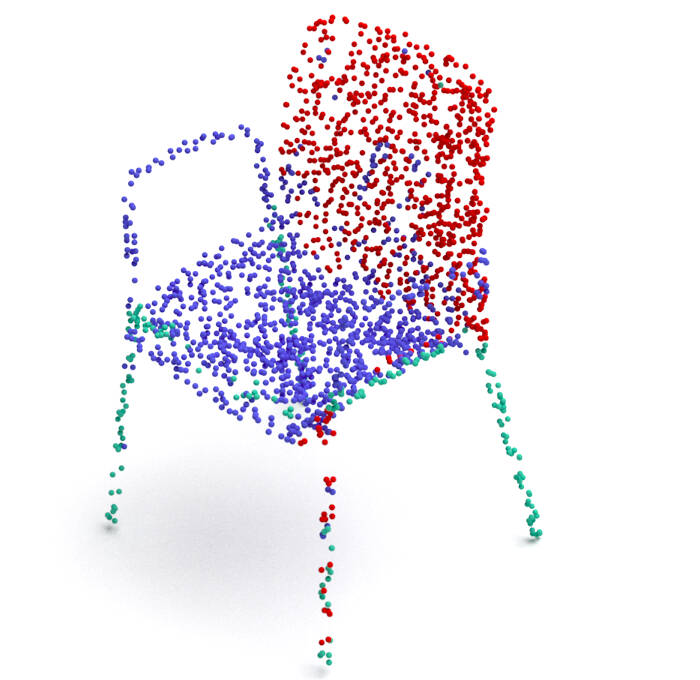}&
    \includegraphics[width=17mm]{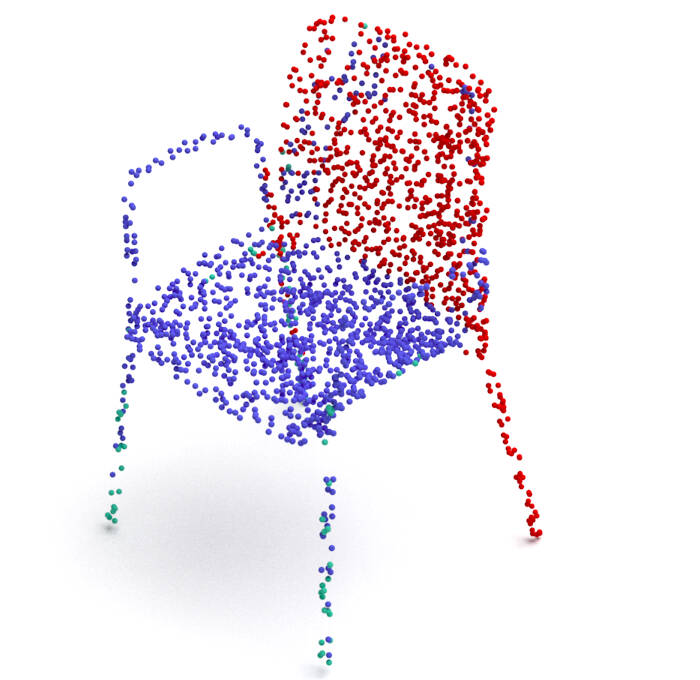}&
    \includegraphics[width=17mm]{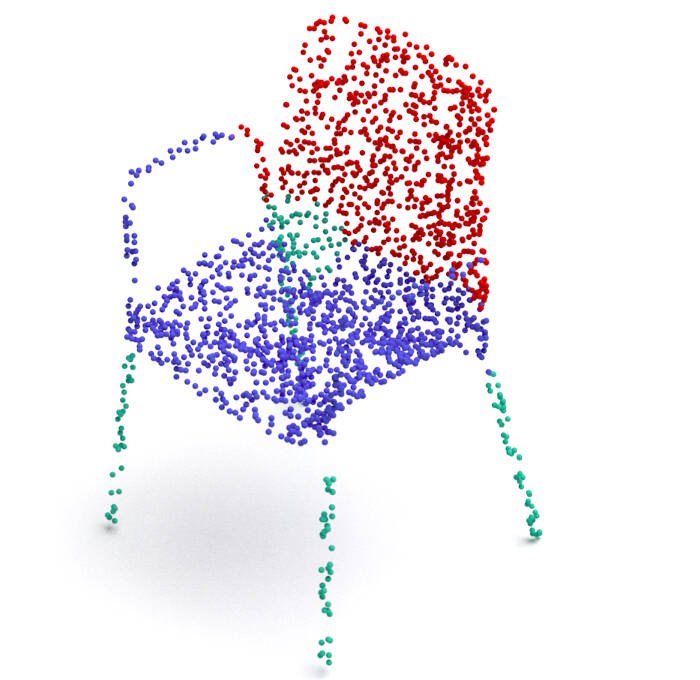}\\[-6.5mm]
    \rotatetitle{18mm}{\hspace*{-21.0mm}\textbf{Car}} &
    \includegraphics[width=17mm]{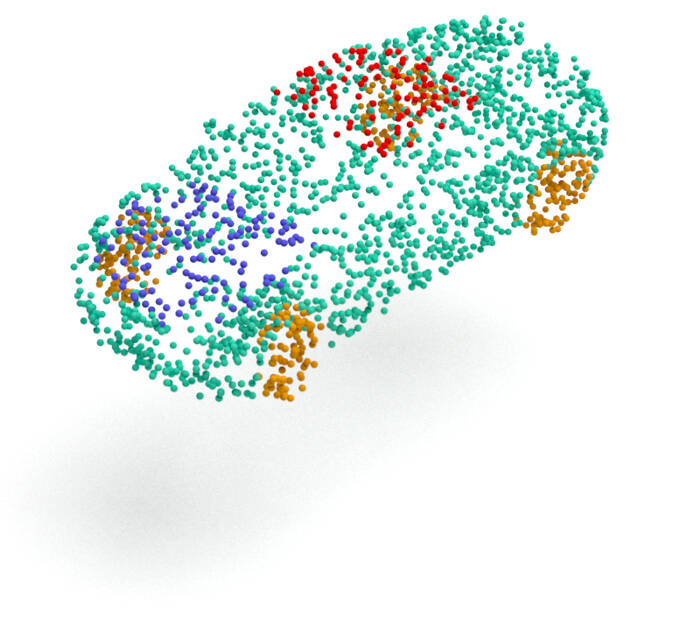}&
    \includegraphics[width=17mm]{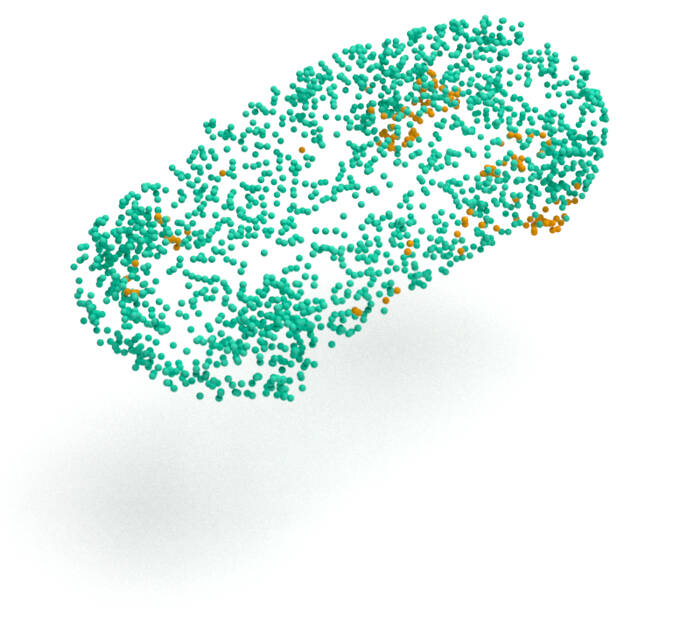}&
    \includegraphics[width=17mm]{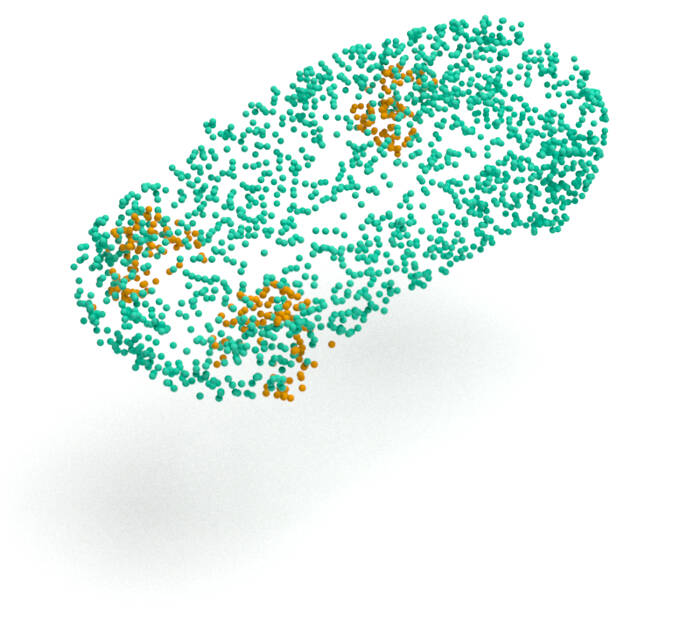}&
    \includegraphics[width=17mm]{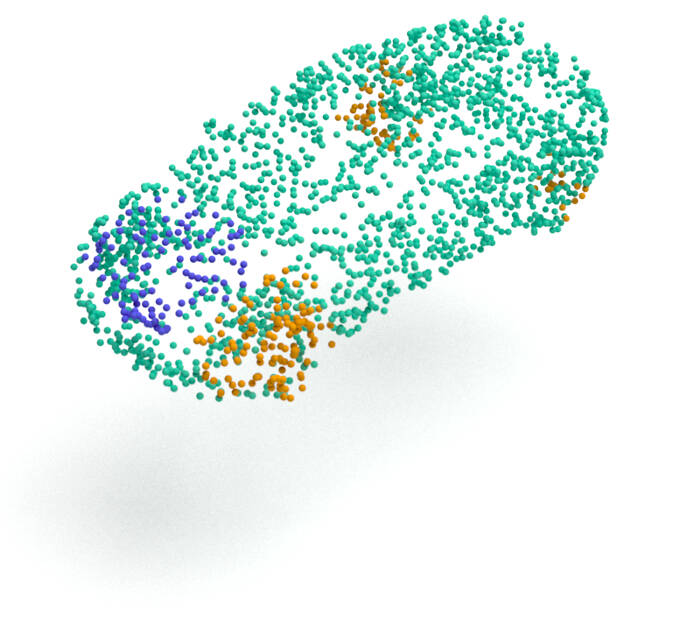}&
    \includegraphics[width=17mm]{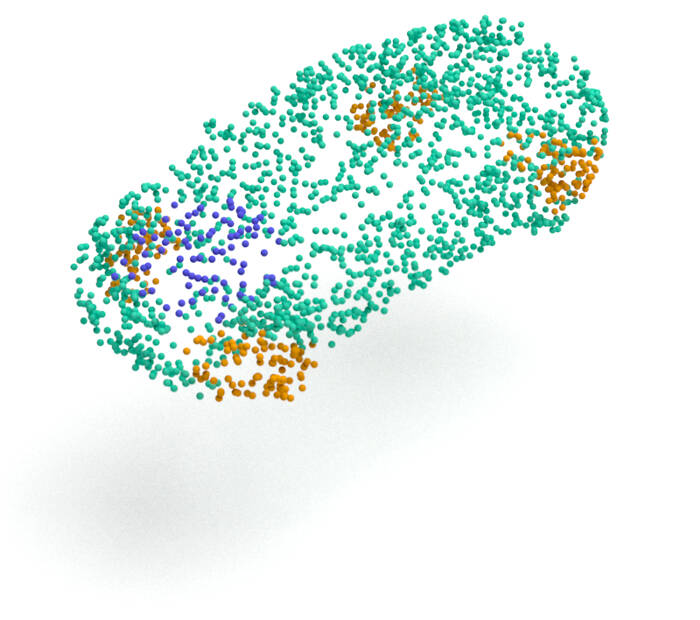}
    \\[-1.8mm]
    &
    \includegraphics[width=17mm]{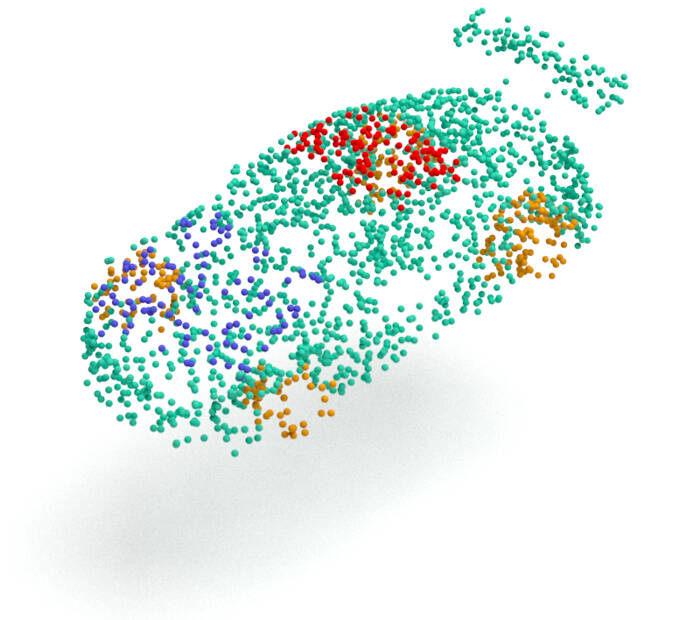}&
    \includegraphics[width=17mm]{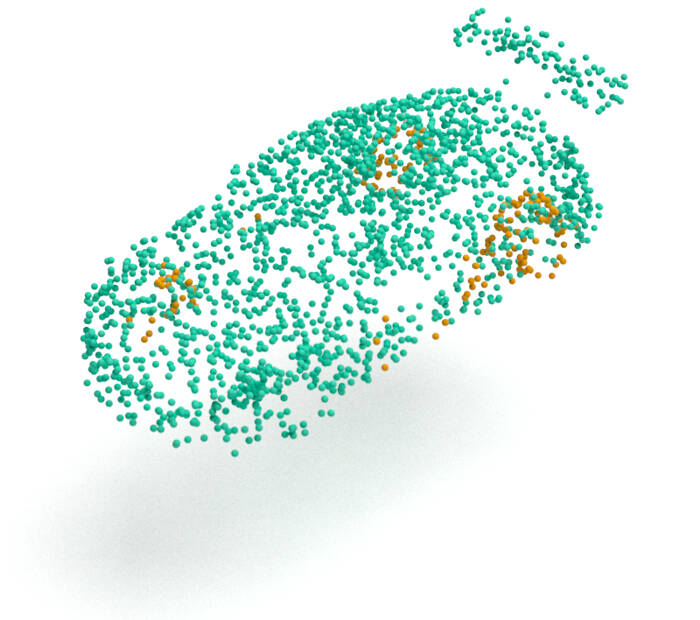}&
    \includegraphics[width=17mm]{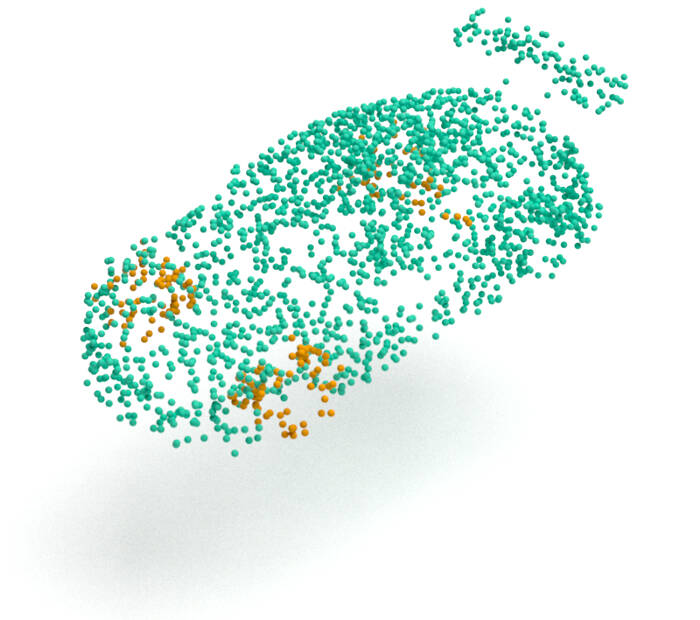}&
    \includegraphics[width=17mm]{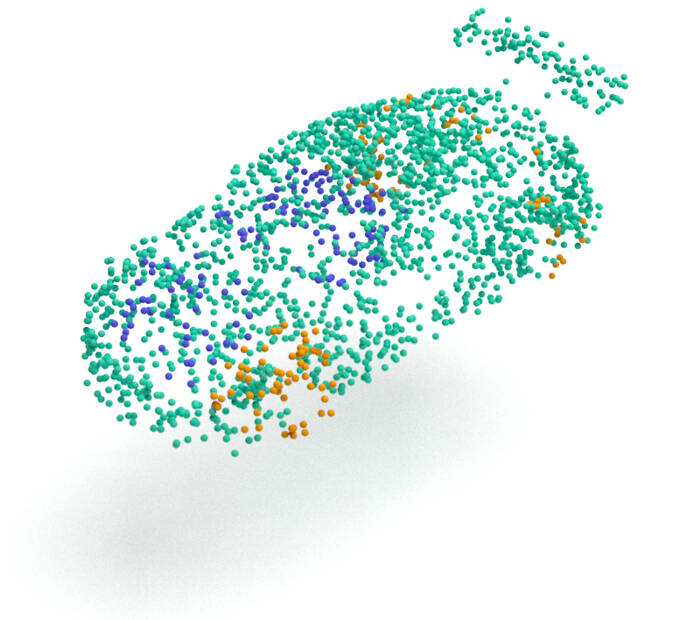}&
    \includegraphics[width=17mm]{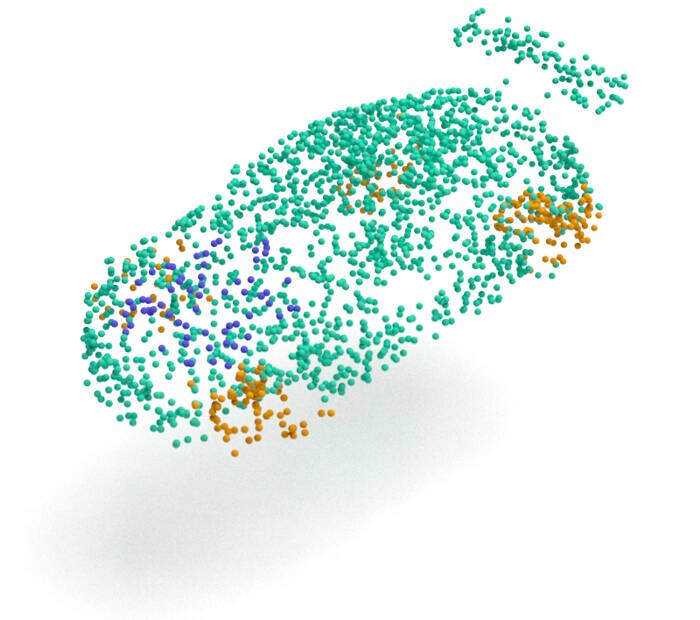} \\
  \end{tabular}
  \vspace*{-4mm}
  \captionof{figure}{Results of unsupervised part segmentation.}
  \label{fig:segment}
  \vspace*{-1mm}
\end{table}

\section{Conclusion}
In this study, we proposed ChartPointFlow, which is a flow-based generative model of point clouds that employs multiple charts.
Each chart is assigned to a semantic subpart of a point cloud, thereby expressing a variety of shapes with different topologies.
Owing to Monte Carlo sampling, the computational cost is of the same order as that of the case without charts.
The performance was evaluated using four 2D synthetic datasets and three 3D practical datasets, and the results demonstrated that ChartPointFlow generates various point clouds of various shapes with better accuracies than the comparison methods.

\begin{acks}
  This work was partially supported by the MIC/SCOPE \#172107101, JST-CREST (JPMJCR1914), and JSPS KAKENHI (19H04172, 19K20344).
\end{acks}

\clearpage
\newpage


\begin{thebibliography}{51}

  
  \ifx \showCODEN    \undefined \def \showCODEN     #1{\unskip}     \fi
  \ifx \showDOI      \undefined \def \showDOI       #1{#1}\fi
  \ifx \showISBNx    \undefined \def \showISBNx     #1{\unskip}     \fi
  \ifx \showISBNxiii \undefined \def \showISBNxiii  #1{\unskip}     \fi
  \ifx \showISSN     \undefined \def \showISSN      #1{\unskip}     \fi
  \ifx \showLCCN     \undefined \def \showLCCN      #1{\unskip}     \fi
  \ifx \shownote     \undefined \def \shownote      #1{#1}          \fi
  \ifx \showarticletitle \undefined \def \showarticletitle #1{#1}   \fi
  \ifx \showURL      \undefined \def \showURL       {\relax}        \fi
  \providecommand\bibfield[2]{#2}
  \providecommand\bibinfo[2]{#2}
  \providecommand\natexlab[1]{#1}
  \providecommand\showeprint[2][]{arXiv:#2}
  
  \bibitem[\protect\citeauthoryear{Achlioptas, Diamanti, Mitliagkas, and
    Guibas}{Achlioptas et~al\mbox{.}}{2018}]%
          {Achlioptas2018}
  \bibfield{author}{\bibinfo{person}{Panos Achlioptas}, \bibinfo{person}{Olga
    Diamanti}, \bibinfo{person}{Ioannis Mitliagkas}, {and}
    \bibinfo{person}{Leonidas Guibas}.} \bibinfo{year}{2018}\natexlab{}.
  \newblock \showarticletitle{{Learning representations and generative models for
    3d point clouds}}. In \bibinfo{booktitle}{\emph{International Conference on
    Machine Learning (ICML)}}.
  \newblock
  
  
  \bibitem[\protect\citeauthoryear{Arshad and Beksi}{Arshad and Beksi}{2020}]%
          {Arshad2020}
  \bibfield{author}{\bibinfo{person}{M. Arshad} {and} \bibinfo{person}{William~J.
    Beksi}.} \bibinfo{year}{2020}\natexlab{}.
  \newblock \showarticletitle{A Progressive Conditional Generative Adversarial
    Network for Generating Dense and Colored 3D Point Clouds}. In
    \bibinfo{booktitle}{\emph{International Conference on 3D Vision (3DV)}}.
  \newblock
  
  
  \bibitem[\protect\citeauthoryear{Cai, Yang, Averbuch-Elor, Hao, Belongie,
    Snavely, and Hariharan}{Cai et~al\mbox{.}}{2020}]%
          {Cai2020}
  \bibfield{author}{\bibinfo{person}{Ruojin Cai}, \bibinfo{person}{Guandao Yang},
    \bibinfo{person}{Hadar Averbuch-Elor}, \bibinfo{person}{Zekun Hao},
    \bibinfo{person}{Serge Belongie}, \bibinfo{person}{Noah Snavely}, {and}
    \bibinfo{person}{Bharath Hariharan}.} \bibinfo{year}{2020}\natexlab{}.
  \newblock \showarticletitle{{Learning Gradient Fields for Shape Generation}}.
    In \bibinfo{booktitle}{\emph{European Conference on Computer Vision (ECCV)}}.
  \newblock
  
  
  \bibitem[\protect\citeauthoryear{Chang, Funkhouser, Guibas, Hanrahan, Huang,
    Li, Savarese, Savva, Song, Su, Xiao, Yi, and Yu}{Chang et~al\mbox{.}}{2015}]%
          {Chang2015}
  \bibfield{author}{\bibinfo{person}{Angel~X. Chang}, \bibinfo{person}{Thomas
    Funkhouser}, \bibinfo{person}{Leonidas Guibas}, \bibinfo{person}{Pat
    Hanrahan}, \bibinfo{person}{Qixing Huang}, \bibinfo{person}{Zimo Li},
    \bibinfo{person}{Silvio Savarese}, \bibinfo{person}{Manolis Savva},
    \bibinfo{person}{Shuran Song}, \bibinfo{person}{Hao Su},
    \bibinfo{person}{Jianxiong Xiao}, \bibinfo{person}{Li Yi}, {and}
    \bibinfo{person}{Fisher Yu}.} \bibinfo{year}{2015}\natexlab{}.
  \newblock \showarticletitle{{ShapeNet: An information-rich 3d model
    repository}}.
  \newblock \bibinfo{journal}{\emph{arXiv preprint arXiv:1512.03012}}
    (\bibinfo{year}{2015}).
  \newblock
  
  
  \bibitem[\protect\citeauthoryear{Deprelle, Groueix, Fisher, Kim, Russell, and
    Aubry}{Deprelle et~al\mbox{.}}{2019}]%
          {Deprelle2019}
  \bibfield{author}{\bibinfo{person}{Theo Deprelle}, \bibinfo{person}{Thibault
    Groueix}, \bibinfo{person}{Matthew Fisher}, \bibinfo{person}{Vladimir~G.
    Kim}, \bibinfo{person}{Bryan~C. Russell}, {and} \bibinfo{person}{Mathieu
    Aubry}.} \bibinfo{year}{2019}\natexlab{}.
  \newblock \showarticletitle{{Learning elementary structures for 3D shape
    generation and matching}}. In \bibinfo{booktitle}{\emph{Advances in Neural
    Information Processing Systems (NeurIPS)}}.
  \newblock
  
  
  \bibitem[\protect\citeauthoryear{Dinh, Sohl-Dickstein, and Bengio}{Dinh
    et~al\mbox{.}}{2017}]%
          {Dinh2017}
  \bibfield{author}{\bibinfo{person}{Laurent Dinh}, \bibinfo{person}{Jascha
    Sohl-Dickstein}, {and} \bibinfo{person}{Samy Bengio}.}
    \bibinfo{year}{2017}\natexlab{}.
  \newblock \showarticletitle{{Density estimation using real NVP}}. In
    \bibinfo{booktitle}{\emph{International Conference on Learning
    Representations (ICLR)}}.
  \newblock
  
  
  \bibitem[\protect\citeauthoryear{Edwards and Storkey}{Edwards and
    Storkey}{2016}]%
          {Edwards2016}
  \bibfield{author}{\bibinfo{person}{Harrison Edwards} {and}
    \bibinfo{person}{Amos Storkey}.} \bibinfo{year}{2016}\natexlab{}.
  \newblock \showarticletitle{{Towards a Neural Statistician}}. In
    \bibinfo{booktitle}{\emph{International Conference on Learning
    Representations (ICLR)}}.
  \newblock
  
  
  \bibitem[\protect\citeauthoryear{Gadelha, Wang, and Maji}{Gadelha
    et~al\mbox{.}}{2018}]%
          {Gadelha2018}
  \bibfield{author}{\bibinfo{person}{Matheus Gadelha}, \bibinfo{person}{Rui
    Wang}, {and} \bibinfo{person}{Subhransu Maji}.}
    \bibinfo{year}{2018}\natexlab{}.
  \newblock \showarticletitle{Multiresolution Tree Networks for 3D Point Cloud
    Processing}. In \bibinfo{booktitle}{\emph{European Conference on Computer
    Vision (ECCV)}}.
  \newblock
  
  
  \bibitem[\protect\citeauthoryear{Goodfellow, Pouget-abadie, Mirza, Xu, and
    Warde-farley}{Goodfellow et~al\mbox{.}}{2014}]%
          {Goodfellow2014}
  \bibfield{author}{\bibinfo{person}{Ian~J Goodfellow}, \bibinfo{person}{Jean
    Pouget-abadie}, \bibinfo{person}{Mehdi Mirza}, \bibinfo{person}{Bing Xu},
    {and} \bibinfo{person}{David Warde-farley}.} \bibinfo{year}{2014}\natexlab{}.
  \newblock \showarticletitle{{Generative Adversarial Nets}}. In
    \bibinfo{booktitle}{\emph{Advances in Neural Information Processing Systems
    (NIPS)}}.
  \newblock
  
  
  \bibitem[\protect\citeauthoryear{Grathwohl, Chen, Bettencourt, Sutskever, and
    Duvenaud}{Grathwohl et~al\mbox{.}}{2019}]%
          {grathwohl2019}
  \bibfield{author}{\bibinfo{person}{Will Grathwohl}, \bibinfo{person}{Ricky
    T.~Q. Chen}, \bibinfo{person}{Jesse Bettencourt}, \bibinfo{person}{Ilya
    Sutskever}, {and} \bibinfo{person}{David Duvenaud}.}
    \bibinfo{year}{2019}\natexlab{}.
  \newblock \showarticletitle{FFJORD: Free-form Continuous Dynamics for Scalable
    Reversible Generative Models}. In \bibinfo{booktitle}{\emph{International
    Conference on Learning Representations (ICLR)}}.
  \newblock
  
  
  \bibitem[\protect\citeauthoryear{Groueix, Fisher, Kim, Russell, and
    Aubry}{Groueix et~al\mbox{.}}{2018}]%
          {Groueix2018}
  \bibfield{author}{\bibinfo{person}{Thibault Groueix}, \bibinfo{person}{Matthew
    Fisher}, \bibinfo{person}{Vladimir~G. Kim}, \bibinfo{person}{Bryan~C.
    Russell}, {and} \bibinfo{person}{Mathieu Aubry}.}
    \bibinfo{year}{2018}\natexlab{}.
  \newblock \showarticletitle{{AtlasNet: A Papier-M{\^{a}}ch{\'{e}} Approach to
    Learning 3D Surface Generation}}. In \bibinfo{booktitle}{\emph{Computer
    Vision and Pattern Recognition (CVPR)}}.
  \newblock
  
  
  \bibitem[\protect\citeauthoryear{Guo, Wang, Hu, Liu, Liu, and Bennamoun}{Guo
    et~al\mbox{.}}{2019}]%
          {Guo2019a}
  \bibfield{author}{\bibinfo{person}{Yulan Guo}, \bibinfo{person}{Hanyun Wang},
    \bibinfo{person}{Qingyong Hu}, \bibinfo{person}{Hao Liu}, \bibinfo{person}{Li
    Liu}, {and} \bibinfo{person}{Mohammed Bennamoun}.}
    \bibinfo{year}{2019}\natexlab{}.
  \newblock \showarticletitle{{Deep learning for 3D point clouds: A survey}}.
  \newblock \bibinfo{journal}{\emph{IEEE Transactions on Pattern Analysis and
    Machine Intelligence}} (\bibinfo{year}{2019}).
  \newblock
  
  
  \bibitem[\protect\citeauthoryear{Hui, Xu, Xie, Qian, and Yang}{Hui
    et~al\mbox{.}}{2020}]%
          {Hui2020}
  \bibfield{author}{\bibinfo{person}{Le Hui}, \bibinfo{person}{Rui Xu},
    \bibinfo{person}{Jin Xie}, \bibinfo{person}{Jianjun Qian}, {and}
    \bibinfo{person}{Jian Yang}.} \bibinfo{year}{2020}\natexlab{}.
  \newblock \showarticletitle{{Progressive Point Cloud Deconvolution Generation
    Network}}. In \bibinfo{booktitle}{\emph{European Conference on Computer
    Vision (ECCV)}}.
  \newblock
  
  
  \bibitem[\protect\citeauthoryear{Hutchinson}{Hutchinson}{1989}]%
          {Hutchinson1989}
  \bibfield{author}{\bibinfo{person}{M.F. Hutchinson}.}
    \bibinfo{year}{1989}\natexlab{}.
  \newblock \showarticletitle{A Stochastic Estimator of the Trace of the
    Influence Matrix for Laplacian Smoothing Splines}.
  \newblock \bibinfo{journal}{\emph{Communications in Statistics - Simulation and
    Computation}}  \bibinfo{volume}{18} (\bibinfo{year}{1989}),
    \bibinfo{pages}{1059--1076}.
  \newblock
  
  
  \bibitem[\protect\citeauthoryear{Ioffe and Szegedy}{Ioffe and Szegedy}{2015}]%
          {Ioffe2015}
  \bibfield{author}{\bibinfo{person}{Sergey Ioffe} {and}
    \bibinfo{person}{Christian Szegedy}.} \bibinfo{year}{2015}\natexlab{}.
  \newblock \showarticletitle{{Batch Normalization: Accelerating Deep Network
    Training by Reducing Internal Covariate Shift}}. In
    \bibinfo{booktitle}{\emph{International Conference on Machine Learning
    (ICML)}}.
  \newblock
  
  
  \bibitem[\protect\citeauthoryear{Jacobsen, Smeulders, and Oyallon}{Jacobsen
    et~al\mbox{.}}{2018}]%
          {Smeulders2018}
  \bibfield{author}{\bibinfo{person}{J{\"{o}}rn-Henrik Jacobsen},
    \bibinfo{person}{Arnold Smeulders}, {and} \bibinfo{person}{Edouard Oyallon}.}
    \bibinfo{year}{2018}\natexlab{}.
  \newblock \showarticletitle{{i-RevNet: Deep Invertible Networks}}. In
    \bibinfo{booktitle}{\emph{International Conference on Learning
    Representations (ICLR)}}.
  \newblock
  
  
  \bibitem[\protect\citeauthoryear{Jang, Gu, and Poole}{Jang
    et~al\mbox{.}}{2017}]%
          {Jang2017}
  \bibfield{author}{\bibinfo{person}{Eric Jang}, \bibinfo{person}{Shixiang Gu},
    {and} \bibinfo{person}{Ben Poole}.} \bibinfo{year}{2017}\natexlab{}.
  \newblock \showarticletitle{{Categorical reparameterization with
    gumbel-softmax}}. In \bibinfo{booktitle}{\emph{International Conference on
    Learning Representations (ICLR)}}.
  \newblock
  
  
  \bibitem[\protect\citeauthoryear{Kim, Lee, Kang, Lee, and Kim}{Kim
    et~al\mbox{.}}{2020}]%
          {Kim2020}
  \bibfield{author}{\bibinfo{person}{Hyeongju Kim}, \bibinfo{person}{Hyeonseung
    Lee}, \bibinfo{person}{Woo~Hyun Kang}, \bibinfo{person}{Joun~Yeop Lee}, {and}
    \bibinfo{person}{Nam~Soo Kim}.} \bibinfo{year}{2020}\natexlab{}.
  \newblock \showarticletitle{{SoftFlow: Probabilistic Framework for Normalizing
    Flow on Manifolds}}. In \bibinfo{booktitle}{\emph{Advances in Neural
    Information Processing Systems (NeurIPS)}}.
  \newblock
  
  
  \bibitem[\protect\citeauthoryear{Kingma and Ba}{Kingma and Ba}{2015}]%
          {Kingma2015}
  \bibfield{author}{\bibinfo{person}{Diederik~P. Kingma} {and}
    \bibinfo{person}{Jimmy Ba}.} \bibinfo{year}{2015}\natexlab{}.
  \newblock \showarticletitle{{Adam: A Method for Stochastic Optimization}}. In
    \bibinfo{booktitle}{\emph{International Conference on Learning
    Representations (ICLR)}}.
  \newblock
  
  
  \bibitem[\protect\citeauthoryear{Kingma and Dhariwal}{Kingma and
    Dhariwal}{2018}]%
          {Kingma2018}
  \bibfield{author}{\bibinfo{person}{Diederik~P. Kingma} {and}
    \bibinfo{person}{Prafulla Dhariwal}.} \bibinfo{year}{2018}\natexlab{}.
  \newblock \showarticletitle{{Glow: Generative flow with invertible 1x1
    convolutions}}. In \bibinfo{booktitle}{\emph{Advances in Neural Information
    Processing Systems (NeurIPS)}}.
  \newblock
  
  
  \bibitem[\protect\citeauthoryear{Kingma, Rezende, and Welling}{Kingma
    et~al\mbox{.}}{2014}]%
          {Kingma2014a}
  \bibfield{author}{\bibinfo{person}{Diederik~P. Kingma},
    \bibinfo{person}{Danilo~J. Rezende}, {and} \bibinfo{person}{Max Welling}.}
    \bibinfo{year}{2014}\natexlab{}.
  \newblock \showarticletitle{{Semi-supervised Learning with Deep Generative
    Models}}. In \bibinfo{booktitle}{\emph{Advances in Neural Information
    Processing Systems (NIPS)}}.
  \newblock
  
  
  \bibitem[\protect\citeauthoryear{Kingma, Salimans, Jozefowicz, Chen, Sutskever,
    and Welling}{Kingma et~al\mbox{.}}{2016}]%
          {Kingma2016}
  \bibfield{author}{\bibinfo{person}{Diederik~P. Kingma}, \bibinfo{person}{Tim
    Salimans}, \bibinfo{person}{Rafal Jozefowicz}, \bibinfo{person}{Xi Chen},
    \bibinfo{person}{Ilya Sutskever}, {and} \bibinfo{person}{Max Welling}.}
    \bibinfo{year}{2016}\natexlab{}.
  \newblock \showarticletitle{{Improved variational inference with inverse
    autoregressive flow}}. In \bibinfo{booktitle}{\emph{Advances in Neural
    Information Processing Systems (NIPS)}}.
  \newblock
  
  
  \bibitem[\protect\citeauthoryear{Kingma and Welling}{Kingma and
    Welling}{2014}]%
          {Kingma2014}
  \bibfield{author}{\bibinfo{person}{Diederik~P. Kingma} {and}
    \bibinfo{person}{Max Welling}.} \bibinfo{year}{2014}\natexlab{}.
  \newblock \showarticletitle{{Auto-encoding variational bayes}}. In
    \bibinfo{booktitle}{\emph{International Conference on Learning
    Representations (ICLR)}}.
  \newblock
  
  
  \bibitem[\protect\citeauthoryear{Klokov, Boyer, and Verbeek}{Klokov
    et~al\mbox{.}}{2020}]%
          {Klokov2020}
  \bibfield{author}{\bibinfo{person}{Roman Klokov}, \bibinfo{person}{Edmond
    Boyer}, {and} \bibinfo{person}{Jakob Verbeek}.}
    \bibinfo{year}{2020}\natexlab{}.
  \newblock \showarticletitle{{Discrete Point Flow Networks for Efficient Point
    Cloud Generation}}. In \bibinfo{booktitle}{\emph{European Conference on
    Computer Vision (ECCV)}}.
  \newblock
  
  
  \bibitem[\protect\citeauthoryear{Klokov and Lempitsky}{Klokov and
    Lempitsky}{2017}]%
          {Klokov2017}
  \bibfield{author}{\bibinfo{person}{Roman Klokov} {and} \bibinfo{person}{Victor
    Lempitsky}.} \bibinfo{year}{2017}\natexlab{}.
  \newblock \showarticletitle{{Escape from Cells: Deep Kd-Networks for the
    Recognition of 3D Point Cloud Models}}. In
    \bibinfo{booktitle}{\emph{International Conference on Conputer Vision
    (ICCV)}}.
  \newblock
  
  
  \bibitem[\protect\citeauthoryear{Li, Zaheer, Zhang, P{\'{o}}czos, and
    Salakhutdinov}{Li et~al\mbox{.}}{2019}]%
          {Li2019}
  \bibfield{author}{\bibinfo{person}{Chun~Liang Li}, \bibinfo{person}{Manzil
    Zaheer}, \bibinfo{person}{Yang Zhang}, \bibinfo{person}{Barnab{\'{a}}s
    P{\'{o}}czos}, {and} \bibinfo{person}{Ruslan Salakhutdinov}.}
    \bibinfo{year}{2019}\natexlab{}.
  \newblock \showarticletitle{{Point cloud gan}}. In
    \bibinfo{booktitle}{\emph{Deep Generative Models for Highly Structured Data,
    International Conference on Learning Representations (ICLR) Workshop}}.
  \newblock
  
  
  \bibitem[\protect\citeauthoryear{Liu, Han, Wen, Liu, and Zwicker}{Liu
    et~al\mbox{.}}{2019}]%
          {Liu2019}
  \bibfield{author}{\bibinfo{person}{Xinhai Liu}, \bibinfo{person}{Zhizhong Han},
    \bibinfo{person}{Xin Wen}, \bibinfo{person}{Yu-Shen Liu}, {and}
    \bibinfo{person}{Matthias Zwicker}.} \bibinfo{year}{2019}\natexlab{}.
  \newblock \showarticletitle{L2G Auto-Encoder: Understanding Point Clouds by
    Local-to-Global Reconstruction with Hierarchical Self-Attention}. In
    \bibinfo{booktitle}{\emph{ACM International Conference on Multimedia (MM)}}.
  \newblock
  
  
  \bibitem[\protect\citeauthoryear{Lopez-Paz and Oquab}{Lopez-Paz and
    Oquab}{2017}]%
          {Lopez-Paz2017}
  \bibfield{author}{\bibinfo{person}{David Lopez-Paz} {and}
    \bibinfo{person}{Maxime Oquab}.} \bibinfo{year}{2017}\natexlab{}.
  \newblock \showarticletitle{{Revisiting classifier two-sample tests}}. In
    \bibinfo{booktitle}{\emph{International Conference on Learning
    Representations (ICLR)}}.
  \newblock
  
  
  \bibitem[\protect\citeauthoryear{Lou, Lim, Katsman, Huang, Jiang, Lim, and {De
    Sa}}{Lou et~al\mbox{.}}{2020}]%
          {Lou2020}
  \bibfield{author}{\bibinfo{person}{Aaron Lou}, \bibinfo{person}{Derek Lim},
    \bibinfo{person}{Isay Katsman}, \bibinfo{person}{Leo Huang},
    \bibinfo{person}{Qingxuan Jiang}, \bibinfo{person}{Ser-Nam Lim}, {and}
    \bibinfo{person}{Christopher {De Sa}}.} \bibinfo{year}{2020}\natexlab{}.
  \newblock \showarticletitle{{Neural Manifold Ordinary Differential Equations}}.
    In \bibinfo{booktitle}{\emph{Advances in Neural Information Processing
    Systems (NeurIPS)}}.
  \newblock
  
  
  \bibitem[\protect\citeauthoryear{Luo and Hu}{Luo and Hu}{2020}]%
          {Luo2020}
  \bibfield{author}{\bibinfo{person}{Shitong Luo} {and} \bibinfo{person}{Wei
    Hu}.} \bibinfo{year}{2020}\natexlab{}.
  \newblock \showarticletitle{Differentiable Manifold Reconstruction for Point
    Cloud Denoising}. In \bibinfo{booktitle}{\emph{ACM International Conference
    on Multimedia (MM)}}.
  \newblock
  
  
  \bibitem[\protect\citeauthoryear{Mirza and Osindero}{Mirza and
    Osindero}{2014}]%
          {Mirza2014}
  \bibfield{author}{\bibinfo{person}{Mehdi Mirza} {and} \bibinfo{person}{Simon
    Osindero}.} \bibinfo{year}{2014}\natexlab{}.
  \newblock \showarticletitle{{Conditional Generative Adversarial Nets}}.
  \newblock \bibinfo{journal}{\emph{arXiv preprint arXiv:1411.1784}}
    (\bibinfo{year}{2014}).
  \newblock
  
  
  \bibitem[\protect\citeauthoryear{Nielsen, Jaini, Hoogeboom, Winther, and
    Welling}{Nielsen et~al\mbox{.}}{2020}]%
          {Nielsen2020}
  \bibfield{author}{\bibinfo{person}{Didrik Nielsen}, \bibinfo{person}{Priyank
    Jaini}, \bibinfo{person}{Emiel Hoogeboom}, \bibinfo{person}{Ole Winther},
    {and} \bibinfo{person}{Max Welling}.} \bibinfo{year}{2020}\natexlab{}.
  \newblock \showarticletitle{{SurVAE Flows: Surjections to Bridge the Gap
    between VAEs and Flows}}. In \bibinfo{booktitle}{\emph{Advances in Neural
    Information Processing Systems (NeurIPS)}}.
  \newblock
  
  
  \bibitem[\protect\citeauthoryear{Papamakarios, Pavlakou, and
    Murray}{Papamakarios et~al\mbox{.}}{2017}]%
          {Papamakarios2017}
  \bibfield{author}{\bibinfo{person}{George Papamakarios}, \bibinfo{person}{Theo
    Pavlakou}, {and} \bibinfo{person}{Iain Murray}.}
    \bibinfo{year}{2017}\natexlab{}.
  \newblock \showarticletitle{{Masked Autoregressive Flow for Density
    Estimation}}. In \bibinfo{booktitle}{\emph{Advances in Neural Information
    Processing Systems (NIPS)}}.
  \newblock
  
  
  \bibitem[\protect\citeauthoryear{Qi, Su, Mo, and Guibas}{Qi
    et~al\mbox{.}}{2017a}]%
          {Qi2017}
  \bibfield{author}{\bibinfo{person}{Charles~R. Qi}, \bibinfo{person}{Hao Su},
    \bibinfo{person}{Kaichun Mo}, {and} \bibinfo{person}{Leonidas~J. Guibas}.}
    \bibinfo{year}{2017}\natexlab{a}.
  \newblock \showarticletitle{{PointNet: Deep learning on point sets for 3D
    classification and segmentation}}. In \bibinfo{booktitle}{\emph{Computer
    Vision and Pattern Recognition (CVPR)}}.
  \newblock
  
  
  \bibitem[\protect\citeauthoryear{Qi, Yi, Su, and Guibas}{Qi
    et~al\mbox{.}}{2017b}]%
          {Qi2017a}
  \bibfield{author}{\bibinfo{person}{Charles~R. Qi}, \bibinfo{person}{Li Yi},
    \bibinfo{person}{Hao Su}, {and} \bibinfo{person}{Leonidas~J. Guibas}.}
    \bibinfo{year}{2017}\natexlab{b}.
  \newblock \showarticletitle{{PointNet++: Deep hierarchical feature learning on
    point sets in a metric space}}. In \bibinfo{booktitle}{\emph{Advances in
    Neural Information Processing Systems (NeurIPS)}}.
  \newblock
  
  
  \bibitem[\protect\citeauthoryear{Ramasinghe, Khan, Barnes, and
    Gould}{Ramasinghe et~al\mbox{.}}{2020}]%
          {Ramasinghe2020}
  \bibfield{author}{\bibinfo{person}{Sameera Ramasinghe}, \bibinfo{person}{Salman
    Khan}, \bibinfo{person}{Nick Barnes}, {and} \bibinfo{person}{Stephen Gould}.}
    \bibinfo{year}{2020}\natexlab{}.
  \newblock \showarticletitle{{Spectral-GANs for High-Resolution 3D Point-cloud
    Generation}}.
  \newblock \bibinfo{journal}{\emph{arXiv preprint arXiv:1912.01800}}
    (\bibinfo{year}{2020}).
  \newblock
  
  
  \bibitem[\protect\citeauthoryear{Rezende and Mohamed}{Rezende and
    Mohamed}{2015}]%
          {Rezende2015}
  \bibfield{author}{\bibinfo{person}{Danilo~Jimenez Rezende} {and}
    \bibinfo{person}{Shakir Mohamed}.} \bibinfo{year}{2015}\natexlab{}.
  \newblock \showarticletitle{{Variational inference with normalizing flows}}. In
    \bibinfo{booktitle}{\emph{International Conference on Machine Learning
    (ICML)}}.
  \newblock
  
  
  \bibitem[\protect\citeauthoryear{Rezende, Papamakarios, Racani{\`{e}}re,
    Albergo, Kanwar, Shanahan, and Cranmer}{Rezende et~al\mbox{.}}{2020}]%
          {Rezende2020}
  \bibfield{author}{\bibinfo{person}{Danilo~Jimenez Rezende},
    \bibinfo{person}{George Papamakarios}, \bibinfo{person}{S{\'{e}}bastien
    Racani{\`{e}}re}, \bibinfo{person}{Michael~S. Albergo},
    \bibinfo{person}{Gurtej Kanwar}, \bibinfo{person}{Phiala~E. Shanahan}, {and}
    \bibinfo{person}{Kyle Cranmer}.} \bibinfo{year}{2020}\natexlab{}.
  \newblock \showarticletitle{{Normalizing Flows on Tori and Spheres}}. In
    \bibinfo{booktitle}{\emph{International Conference on Machine Learning
    (ICML)}}.
  \newblock
  
  
  \bibitem[\protect\citeauthoryear{Shu, Park, and Kwon}{Shu
    et~al\mbox{.}}{2019}]%
          {Shu2019}
  \bibfield{author}{\bibinfo{person}{Dongwook Shu}, \bibinfo{person}{Sung~Woo
    Park}, {and} \bibinfo{person}{Junseok Kwon}.}
    \bibinfo{year}{2019}\natexlab{}.
  \newblock \showarticletitle{{3D point cloud generative adversarial network
    based on tree structured graph convolutions}}. In
    \bibinfo{booktitle}{\emph{International Conference on Conputer Vision
    (ICCV)}}.
  \newblock
  
  
  \bibitem[\protect\citeauthoryear{Su, Jampani, Sun, Maji, Kalogerakis, Yang, and
    Kautz}{Su et~al\mbox{.}}{2018}]%
          {Su2018}
  \bibfield{author}{\bibinfo{person}{Hang Su}, \bibinfo{person}{Varun Jampani},
    \bibinfo{person}{Deqing Sun}, \bibinfo{person}{Subhransu Maji},
    \bibinfo{person}{Evangelos Kalogerakis}, \bibinfo{person}{Ming~Hsuan Yang},
    {and} \bibinfo{person}{Jan Kautz}.} \bibinfo{year}{2018}\natexlab{}.
  \newblock \showarticletitle{{SPLATNet: Sparse Lattice Networks for Point Cloud
    Processing}}. In \bibinfo{booktitle}{\emph{Computer Vision and Pattern
    Recognition (CVPR)}}.
  \newblock
  
  
  \bibitem[\protect\citeauthoryear{Sun, Lian, and Xiao}{Sun
    et~al\mbox{.}}{2019}]%
          {Sun2019}
  \bibfield{author}{\bibinfo{person}{Xiao Sun}, \bibinfo{person}{Zhouhui Lian},
    {and} \bibinfo{person}{Jianguo Xiao}.} \bibinfo{year}{2019}\natexlab{}.
  \newblock \showarticletitle{SRINet: Learning Strictly Rotation-Invariant
    Representations for Point Cloud Classification and Segmentation}. In
    \bibinfo{booktitle}{\emph{ACM International Conference on Multimedia (MM)}}.
  \newblock
  
  
  \bibitem[\protect\citeauthoryear{Teshima, Ishikawa, Tojo, Oono, Ikeda, and
    Sugiyama}{Teshima et~al\mbox{.}}{2020}]%
          {Teshima2020}
  \bibfield{author}{\bibinfo{person}{Takeshi Teshima}, \bibinfo{person}{Isao
    Ishikawa}, \bibinfo{person}{Koichi Tojo}, \bibinfo{person}{Kenta Oono},
    \bibinfo{person}{Masahiro Ikeda}, {and} \bibinfo{person}{Masashi Sugiyama}.}
    \bibinfo{year}{2020}\natexlab{}.
  \newblock \showarticletitle{{Coupling-based Invertible Neural Networks Are
    Universal Diffeomorphism Approximators}}. In
    \bibinfo{booktitle}{\emph{Advances in Neural Information Processing Systems
    (NeurIPS)}}.
  \newblock
  
  
  \bibitem[\protect\citeauthoryear{Valsesia, Fracastoro, and Magli}{Valsesia
    et~al\mbox{.}}{2019}]%
          {Valsesia2019}
  \bibfield{author}{\bibinfo{person}{Diego Valsesia}, \bibinfo{person}{Giulia
    Fracastoro}, {and} \bibinfo{person}{Enrico Magli}.}
    \bibinfo{year}{2019}\natexlab{}.
  \newblock \showarticletitle{{Learning localized generative models for 3D point
    clouds via graph convolution}}. In \bibinfo{booktitle}{\emph{International
    Conference on Learning Representations (ICLR)}}.
  \newblock
  
  
  \bibitem[\protect\citeauthoryear{Wang, Huang, Hou, Zhang, and Shan}{Wang
    et~al\mbox{.}}{2019}]%
          {Wang2019}
  \bibfield{author}{\bibinfo{person}{Lei Wang}, \bibinfo{person}{Yuchun Huang},
    \bibinfo{person}{Yaolin Hou}, \bibinfo{person}{Shenman Zhang}, {and}
    \bibinfo{person}{Jie Shan}.} \bibinfo{year}{2019}\natexlab{}.
  \newblock \showarticletitle{{Graph Attention Convolution for Point Cloud
    Semantic Segmentation}}. In \bibinfo{booktitle}{\emph{Computer Vision and
    Pattern Recognition (CVPR)}}.
  \newblock
  
  
  \bibitem[\protect\citeauthoryear{Wang and Neumann}{Wang and Neumann}{2020}]%
          {Wang2020}
  \bibfield{author}{\bibinfo{person}{Panqu Wang} {and} \bibinfo{person}{Ulrich
    Neumann}.} \bibinfo{year}{2020}\natexlab{}.
  \newblock \showarticletitle{{Grid-GCN for Fast and Scalable Point Cloud
    Learning}}. In \bibinfo{booktitle}{\emph{Computer Vision and Pattern
    Recognition (CVPR)}}.
  \newblock
  
  
  \bibitem[\protect\citeauthoryear{Yan, Zheng, Li, Wang, and Cui}{Yan
    et~al\mbox{.}}{2020}]%
          {Yan2020}
  \bibfield{author}{\bibinfo{person}{Xu Yan}, \bibinfo{person}{Chaoda Zheng},
    \bibinfo{person}{Zhen Li}, \bibinfo{person}{Sheng Wang}, {and}
    \bibinfo{person}{Shuguang Cui}.} \bibinfo{year}{2020}\natexlab{}.
  \newblock \showarticletitle{{PointASNL: Robust Point Clouds Processing Using
    Nonlocal Neural Networks With Adaptive Sampling}}. In
    \bibinfo{booktitle}{\emph{Computer Vision and Pattern Recognition (CVPR)}}.
  \newblock
  
  
  \bibitem[\protect\citeauthoryear{Yang, Huang, Hao, Liu, Belongie, and
    Hariharan}{Yang et~al\mbox{.}}{2019}]%
          {Yang2019}
  \bibfield{author}{\bibinfo{person}{Guandao Yang}, \bibinfo{person}{Xun Huang},
    \bibinfo{person}{Zekun Hao}, \bibinfo{person}{Ming~Yu Liu},
    \bibinfo{person}{Serge Belongie}, {and} \bibinfo{person}{Bharath Hariharan}.}
    \bibinfo{year}{2019}\natexlab{}.
  \newblock \showarticletitle{{Pointflow: 3D point cloud generation with
    continuous normalizing flows}}. In \bibinfo{booktitle}{\emph{International
    Conference on Conputer Vision (ICCV)}}.
  \newblock
  
  
  \bibitem[\protect\citeauthoryear{Yi, Kim, Ceylan, Shen, Yan, Su, Lu, Huang,
    Sheffer, and Guibas}{Yi et~al\mbox{.}}{2016}]%
          {Yi2016}
  \bibfield{author}{\bibinfo{person}{Li Yi}, \bibinfo{person}{Vladimir~G. Kim},
    \bibinfo{person}{Duygu Ceylan}, \bibinfo{person}{I-Chao Shen},
    \bibinfo{person}{Mengyan Yan}, \bibinfo{person}{Hao Su},
    \bibinfo{person}{Cewu Lu}, \bibinfo{person}{Qixing Huang},
    \bibinfo{person}{Alla Sheffer}, {and} \bibinfo{person}{Leonidas Guibas}.}
    \bibinfo{year}{2016}\natexlab{}.
  \newblock \showarticletitle{A Scalable Active Framework for Region Annotation
    in 3D Shape Collections}.
  \newblock \bibinfo{journal}{\emph{SIGGRAPH Asia}}.
  \newblock
  
  
  \bibitem[\protect\citeauthoryear{Zaheer, Kottur, Ravanbhakhsh, P{\'{o}}czos,
    Salakhutdinov, and Smola}{Zaheer et~al\mbox{.}}{2017}]%
          {Zaheer2017}
  \bibfield{author}{\bibinfo{person}{Manzil Zaheer}, \bibinfo{person}{Satwik
    Kottur}, \bibinfo{person}{Siamak Ravanbhakhsh},
    \bibinfo{person}{Barnab{\'{a}}s P{\'{o}}czos}, \bibinfo{person}{Ruslan
    Salakhutdinov}, {and} \bibinfo{person}{Alexander~J. Smola}.}
    \bibinfo{year}{2017}\natexlab{}.
  \newblock \showarticletitle{{Deep sets}}. In \bibinfo{booktitle}{\emph{Advances
    in Neural Information Processing Systems (NeurIPS)}}.
  \newblock
  
  
  \bibitem[\protect\citeauthoryear{Zamorski, Zi{\c{e}}ba, Klukowski, Nowak,
    Kurach, Stokowiec, and Trzci{\'{n}}ski}{Zamorski et~al\mbox{.}}{2020}]%
          {Zamorski2020}
  \bibfield{author}{\bibinfo{person}{Maciej Zamorski}, \bibinfo{person}{Maciej
    Zi{\c{e}}ba}, \bibinfo{person}{Piotr Klukowski}, \bibinfo{person}{Rafa{\l}
    Nowak}, \bibinfo{person}{Karol Kurach}, \bibinfo{person}{Wojciech Stokowiec},
    {and} \bibinfo{person}{Tomasz Trzci{\'{n}}ski}.}
    \bibinfo{year}{2020}\natexlab{}.
  \newblock \showarticletitle{{Adversarial autoencoders for compact
    representations of 3D point clouds}}.
  \newblock \bibinfo{journal}{\emph{Computer Vision and Image Understanding
    (CVIU)}}  \bibinfo{volume}{193} (\bibinfo{year}{2020}).
  \newblock
  
  
  \bibitem[\protect\citeauthoryear{Zhao}{Zhao}{2018}]%
          {Zhao2018}
  \bibfield{author}{\bibinfo{person}{Na Zhao}.} \bibinfo{year}{2018}\natexlab{}.
  \newblock \showarticletitle{End2End Semantic Segmentation for 3D Indoor
    Scenes}. In \bibinfo{booktitle}{\emph{ACM International Conference on
    Multimedia (MM)}}.
  \newblock
  
  
  \end{thebibliography}


\clearpage
\newpage
\renewcommand\thetable{A\arabic{table}}
\setcounter{table}{0}
\renewcommand\thefigure{A\arabic{figure}}
\setcounter{figure}{0}
\appendix
{\Huge Appendix}
\section{Flow-based Generative Model}\label{app:flow_based}
A flow-based generative model (or a normalizing flow) $f$ is a neural network composed of a sequence of $L$ invertible transformations $g_{0}, \dots, g_{L-1}$, i.e., $f=g_{L-1}\circ\dots\circ g_{0}$~\cite{Dinh2017,Kingma2018}.
The model $f$ maps a latent variable $z$ to a sample $x$ in the data space, i.e., $x=f(z)$).
Specifically,
\begin{align}
  z=h_0
  \underset{g^{-1}_{0}}{\overset{g_{0}}{\rightleftarrows}}
  h_{1}
  \underset{g^{-1}_{1}}{\overset{g_{1}}{\rightleftarrows}}
  h_{2}
  \cdots
  \underset{g^{-1}_{L-1}}{\overset{g_{L-1}}{\rightleftarrows}}
  h_L=x.
\end{align}
Given the map $f$, the log-likelihood of a sample $x$ is obtained using the change of variables, which is expressed as
\begin{equation}
  \begin{split}
    \log p(x)
    &\textstyle= \log p(z) - \log \left| \det \frac{\partial f}{\partial z} \right|\\
    &\textstyle= \log p(z) - \sum_{i=0}^{L-1}\log \left| \det \frac{\partial g_{i}}{\partial h_{i}} \right|\\
    &\textstyle= \log p(z) + \log \left| \det \frac{\partial f^{-1}}{\partial x} \right|\\
    &\textstyle= \log p(z) + \sum_{i=0}^{L-1}\log \left| \det \frac{\partial g^{-1}_{i}}{\partial h_{i+1}} \right|,
  \end{split}\label{eq:log_likelihood_NF}
\end{equation}
where $p(z)$ denotes a prior, and $\log|\det \partial g_{i} / \partial h_{i}|$ denotes the log-absolute-determinant of the Jacobian matrix $\partial g_{i} / \partial h_{i}$.
The prior $p(z)$ of the latent variable $z$ is often set to a simple distribution, such as the standard Gaussian distribution.

Because the calculation of the log-determinant is computationally expensive, each map $g_{i}$ is often given by a neural network with a specially designed architecture.
A coupling-based network is composed of two sub-networks, each of which is applied to the other alternatively~\cite{Dinh2017,Kingma2018}.
Then, each Jacobian matrix is triangular, and its determinant is easily obtained.
A coupling-based network has been proven to approximate arbitrary diffeomorphisms~\cite{Teshima2020}.
An autoregressive flow generates an element of the output one-by-one using the remaining elements, also leading to triangular Jacobian matrices~\cite{Kingma2016,Papamakarios2017}.
Some other architectures have also been proposed~\cite{Smeulders2018}.

In contrast, a continuous normalizing flow, namely, FFJORD~\cite{grathwohl2019}, defines the map $f$ as the integral of an ordinary differential equation (ODE) $\mathrm{d}h/\mathrm{d}t=g(h,t)$ and allows general architectures, where $z=h\left(t_{0}\right)$ and $x=h\left(t_{1}\right)$.
Given either a sample $x$ or a latent variable $z$, one can obtain the other by solving the initial value problem using an ODE solver (e.g., the Dormand-Prince method), as follows.
\begin{equation}
  \textstyle
  x=z+\int_{t_{0}}^{t_{1}} g(h(\xi), \xi) \mathrm{d}\xi,\mbox{ or }
  z=x+\int_{t_{1}}^{t_{0}} g(h(\xi), \xi) \mathrm{d}\xi.
\end{equation}
The log-likelihood is given by
\begin{equation}
  \begin{split}
    \textstyle \log p\left(x\right)
    &\textstyle=\log p\left(z\right)-\int_{t_{0}}^{t_{1}} \operatorname{Tr}\left(\frac{\partial g(h;t)}{\partial h(t)}\right) \mathrm{d}t.
  \end{split}\label{eq:log_likelihood_CNF}
\end{equation}
The log-absolute-determinant is obtained using Hutchinson's trace estimator~\cite{Hutchinson1989}.
Therefore, the number of function evaluations, and hence the computational cost, are much larger than those of the discrete counterpart.

\section{Model Architecture}\label{app:architecture}
ChartPointFlow is adaptable to any network architecture.
In the experiments in Section~\ref{sec:experiments}, we employed architectures similar to those of PointFlow~\cite{Li2019} and SoftFlow~\cite{Kim2020}, as summarized below.

For the feature encoder $E$, we employed the same architecture as that used in PointFlow~\cite{Yang2019}.
In particular, the former part was implemented as four 1D convolutional layers with $128$-$128$-$256$-$512$ channels and a kernel size of 1.
This architecture is equivalent to fully-connected layers applied to each point independently.
The latter part was composed of a max-pooling over the points, followed by three fully-connected layers with $256$-$128$-$128$ units.
Applied to a set of points, the feature encoder $E$ obtains a permutation-invariant joint representation $s_{\!X}$~\cite{Edwards2016}.
Each hidden layer was followed by a batch normalization~\cite{Ioffe2015} and the ReLU function.
Using the reparameterization trick, the output was regarded as the posterior $q_E(s_{X}|X)$ of the feature vector $s_{\!X}$.

The prior flow $G$ was also the same as that used in PointFlow~\cite{Yang2019}.
It was composed of three concatsquash layers of $256$-$256$-$128$ units sandwiched by moving batch normalizations.
A concatsquash layer is implemented in FFJORD's release code~\cite{grathwohl2019}, and it is expressed by
\begin{equation}
  CS(\chi, \xi) = (W_\chi \chi + b_\chi) \sigma(W_\xi \xi + b_\xi) + W_b \xi + b_b,
\end{equation}
where $W_\chi$, $b_\chi$, $W_\xi$, $b_\xi$, $W_b$, and $b_b$ are trainable parameters, and $\sigma$ denotes the sigmoid function.
$\chi$ and $\xi$ denote the input and condition, respectively, which were a point $x$ and the time $t$ in the prior flow $G$.
The first two concatsquash layers were followed by tanh functions as the activation function.

The point generator $F$ was the same as that used in SoftFlow~\cite{Kim2020} except that ours accepts the label $y$ as a condition, whereas SoftFlow accepts the injected noise's intensity as a condition.
It was composed of nine blocks, each of which was composed of an actnorm, invertible 1x1 convolution~\cite{Kingma2018}, and autoregressive layer~\cite{Kim2020}.
An autoregressive layer was composed of three concatsquash layers with 256 units, followed by a tanh function.
The input $\chi$ is a point $x$ and the condition $\xi$ is the feature vector $s_{\!X}$.
Preliminary experiments suggest that the point generator $F$ of PointFlow, which is based on a continuous normalizing flow, potentially improves the performance, and that it requires too much computational cost for our equipment.

The chart predictor $C$ was composed of three concatsquash layers with $256$-$256$-$n$ units, where $n$ is the number of charts.
The chart generator $K$ was composed of five fully-connected layers with $256$-$512$-$256$-$128$-$n$ units.
Each hidden layer was followed by the ReLU function.

\section{Additional Results}\label{app:additional_results}

\subsection{Number of Charts}\label{app:number_of_charts}
We also provide the results of the generation and reconstruction tasks with the varying number of charts in Tables~\ref{tab:generation_results_chart}, \ref{tab:reconstruction_results_chart}, and \ref{tab:segmentation_results_chart}.
Recall that the computational cost of the proposed method is constant regardless of the number $n$ of charts owing to the Gumbel-Softmax approach~\cite{Jang2017}.

\subsection{Additional Metrics}\label{app:additional_metrics}
Chamfer distance (CD) has been used as a distance between two point clouds $X_1$ and $X_2$.
Jensen-Shannon divergence (JSD), minimum matching distance (MMD), and coverage (COV) have been used to measure the similarity between two point cloud sets $\mathcal{X}_1$ and $\mathcal{X}_2$.
However, previous studies pointed out that these measures may give good scores to poor models~\cite{Achlioptas2018,Kim2020,Yang2019}.

CD is defined as the sum of the squared distance of each point to the nearest point among the points obtained from the other point cloud.
Specifically,
\begin{equation}
  CD(X_1, X_2)  = \sum_{x\in X_{1}} \min_{\xi \in X_2} \|x-\xi \|^{2}_{2} + \sum_{x \in X_{2}} \min_{\xi \in X_1} \| x-\xi\|^{2}_{2}.
\end{equation}

JSD measures the distance between two empirical distributions $P_1$ and $P_2$.
For JSD, a canonical voxel grid was introduced, the number of points lying in each voxel was counted, and then an empirical probability distribution was obtained for each of the reference and generated sets.
MMD is the distance between a point cloud in the reference set and its nearest neighbor in the generated set.
COV measures the fraction of point clouds in the reference set that can be matched with at least one point cloud in the generated set.

We summarized the results evaluated using these measures in Tables~\ref{tab:generation_results_chart}--\ref{tab:reconstruction_results_all_metrics} just for reference.
ChartPointFlow achieved the best scores in most criteria for the generation task, as shown in Table~\ref{tab:generation_results_all_metrics}.

\subsection{Additional Images}\label{app:additional_images}
We also provide additional results for the qualitative assessment.

\begin{figure*}[t]
  \centering
  \setlength{\tabcolsep}{0mm}
  \begin{tabular}{cccccccc}
    \includegraphics[width=21mm]{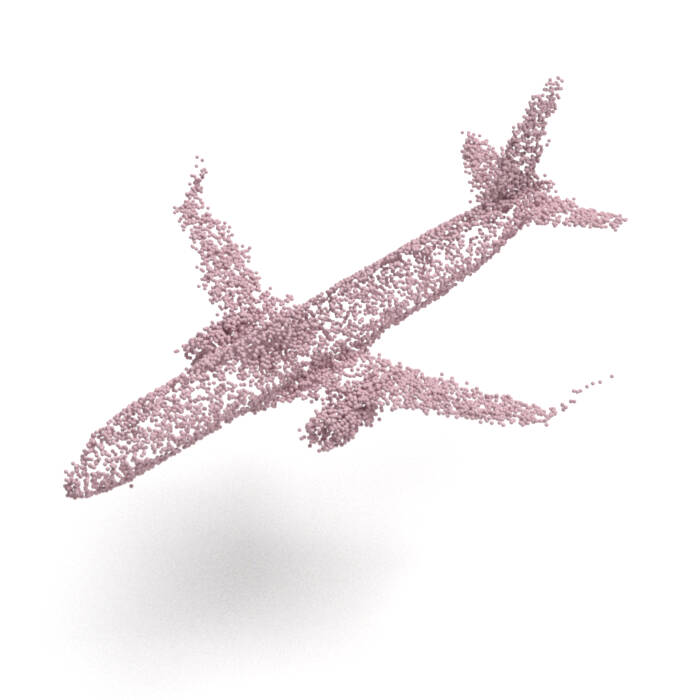}         &
    \includegraphics[width=21mm]{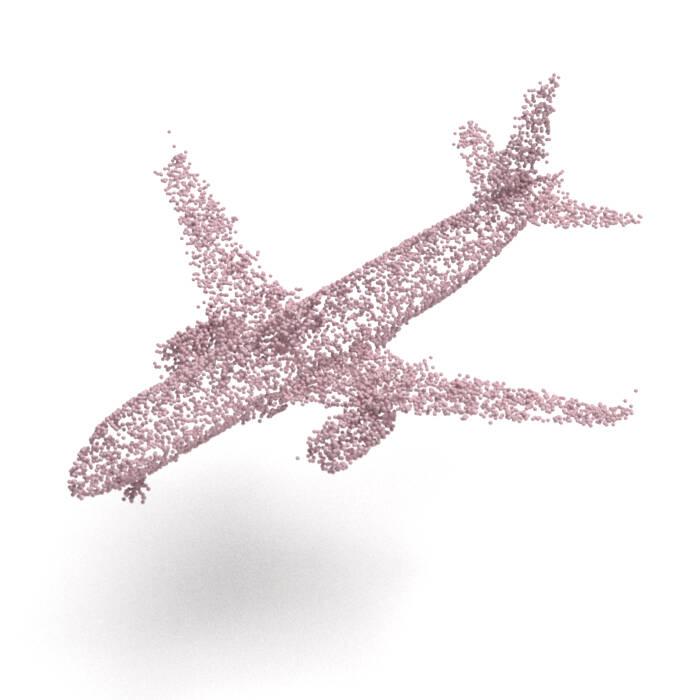}         &
    \includegraphics[width=21mm]{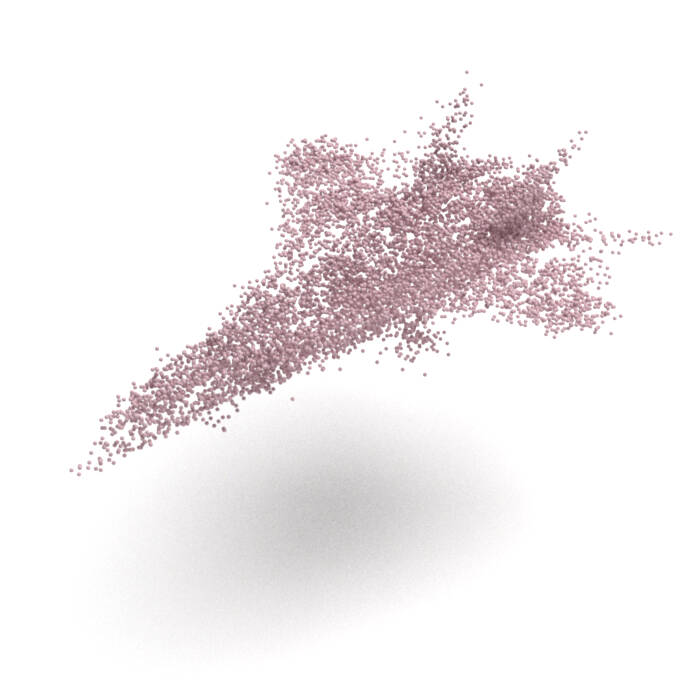}        &
    \includegraphics[width=21mm]{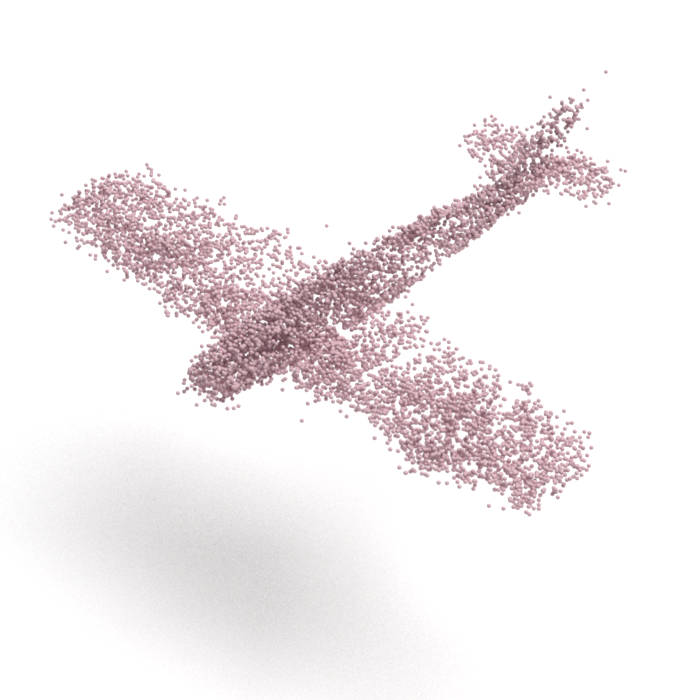}        &
    \includegraphics[width=21mm]{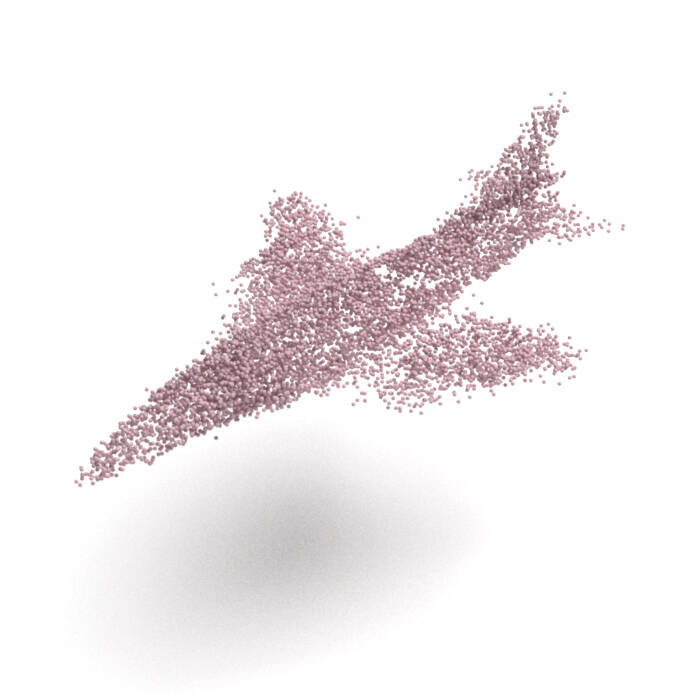}        &
    \includegraphics[width=21mm]{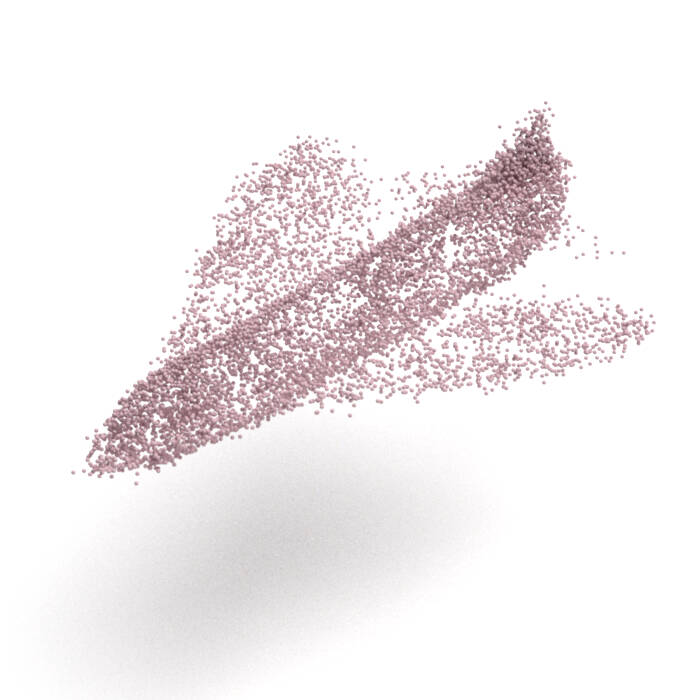}        &
    \includegraphics[width=21mm]{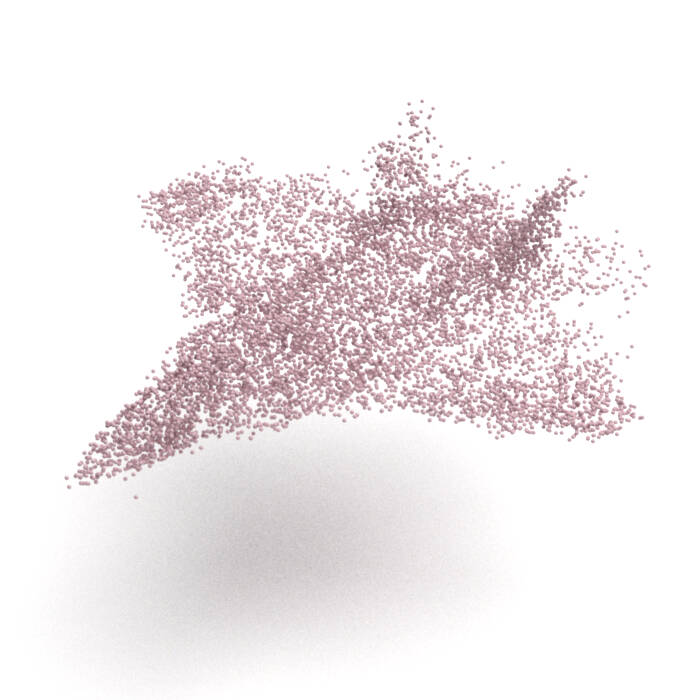}        &
    \includegraphics[width=21mm]{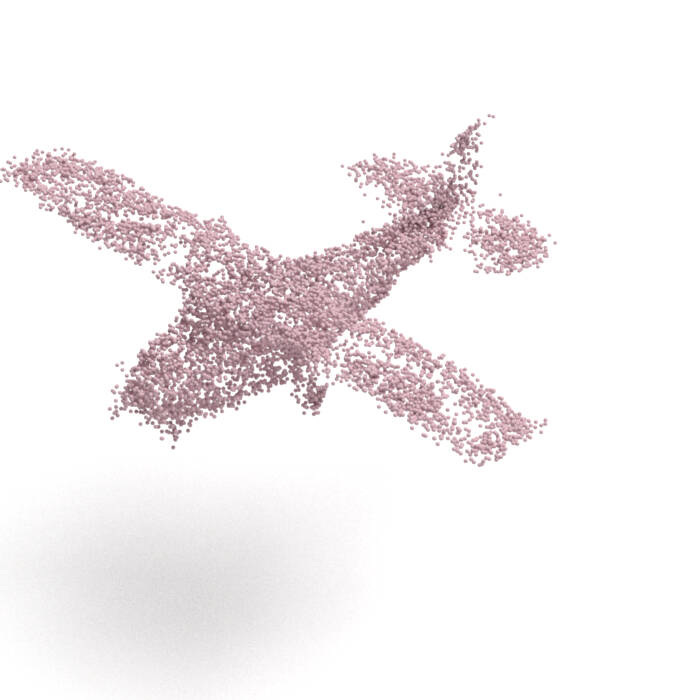}          \\
    \includegraphics[width=21mm]{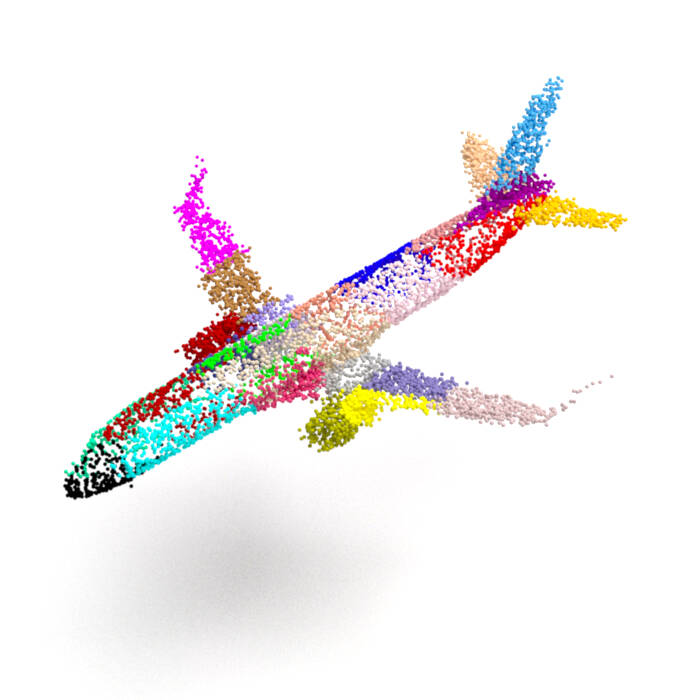}   &
    \includegraphics[width=21mm]{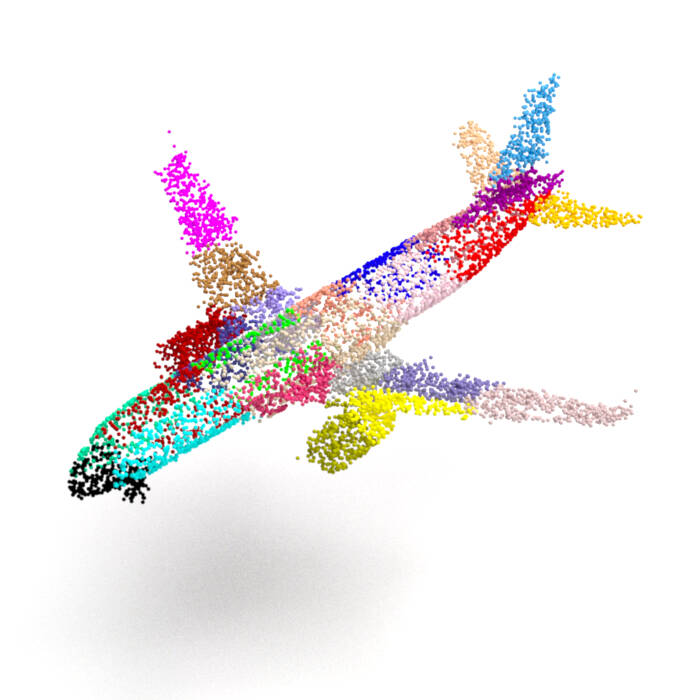}   &
    \includegraphics[width=21mm]{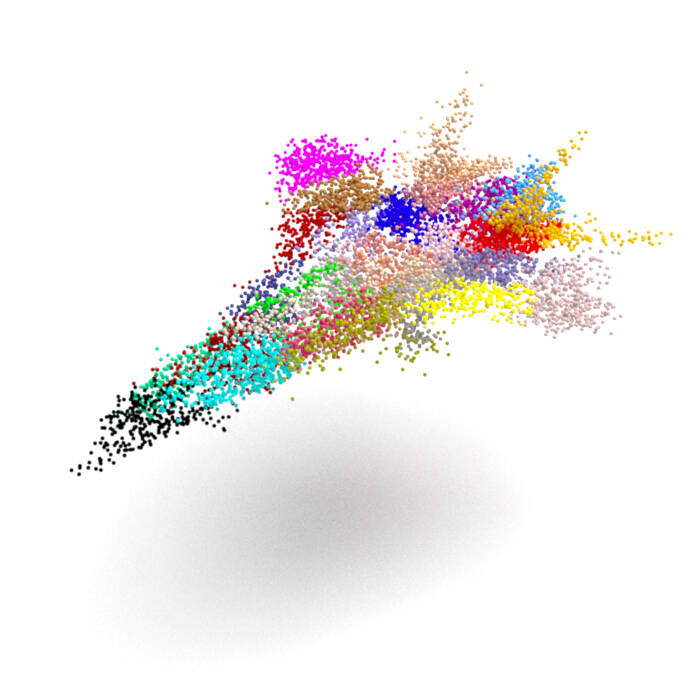}  &
    \includegraphics[width=21mm]{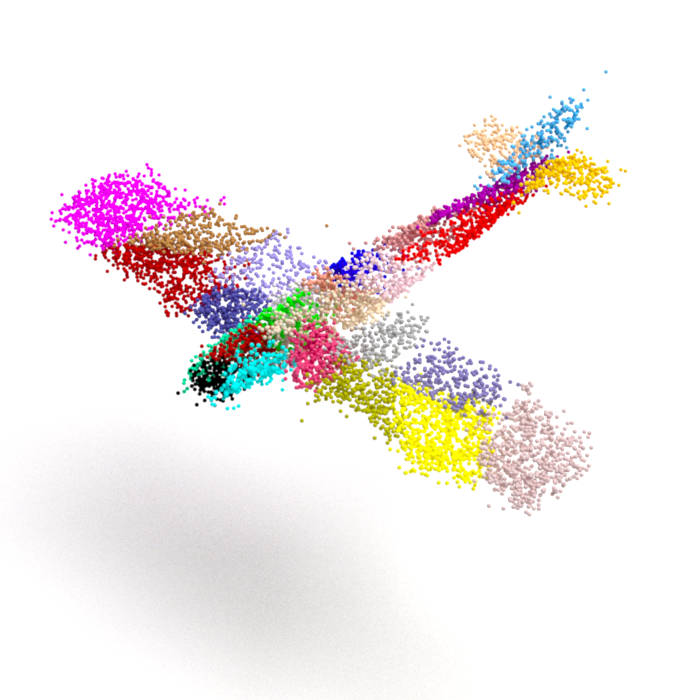}  &
    \includegraphics[width=21mm]{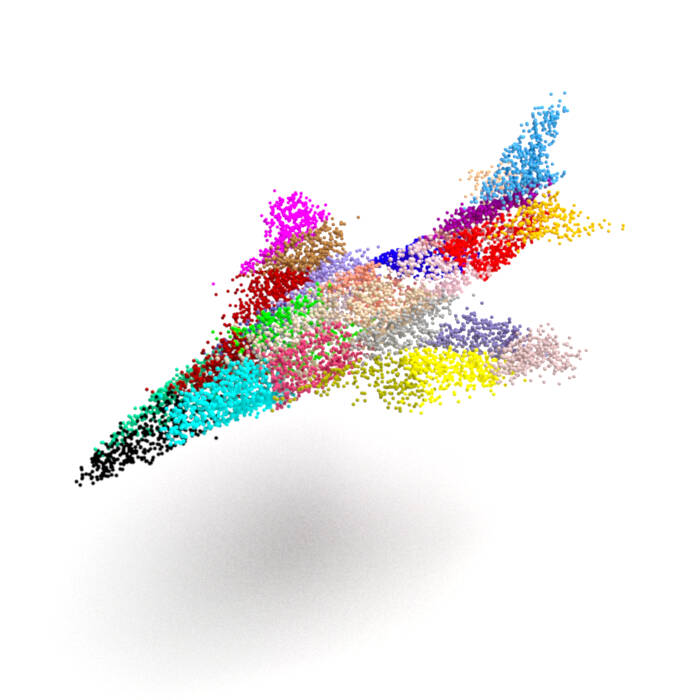}  &
    \includegraphics[width=21mm]{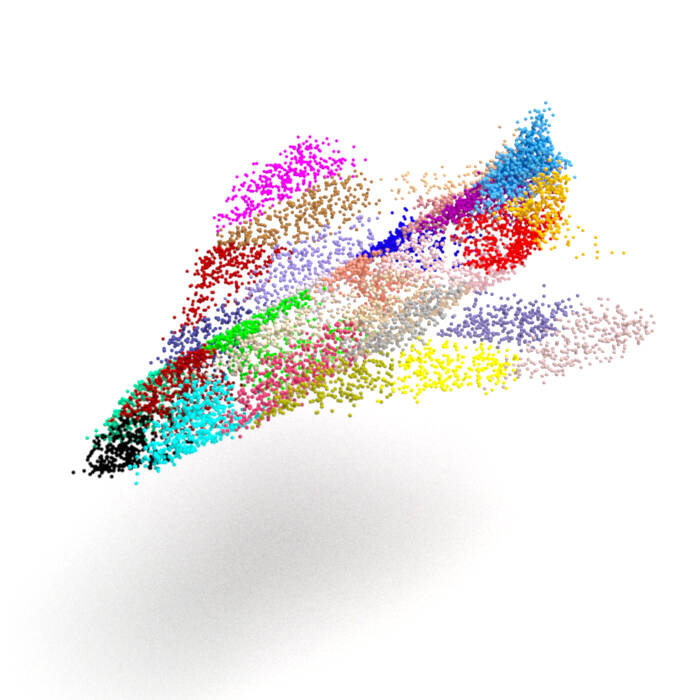}  &
    \includegraphics[width=21mm]{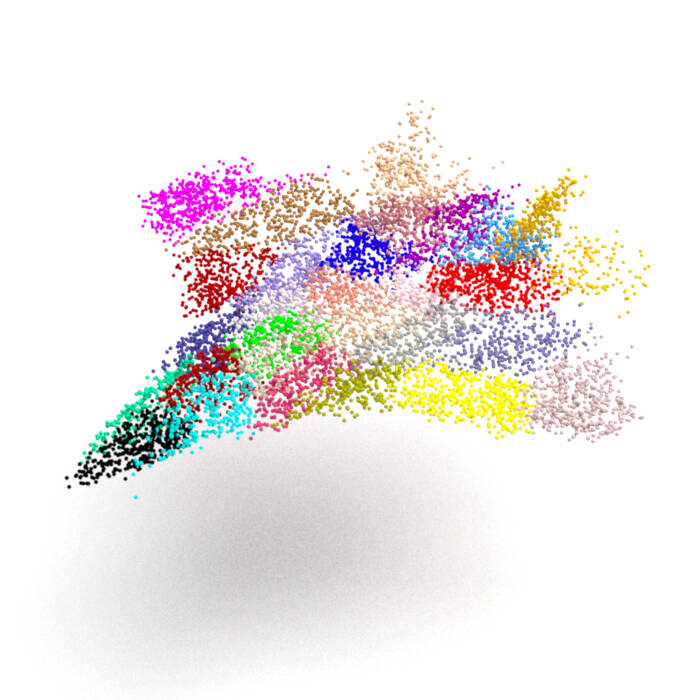}  &
    \includegraphics[width=21mm]{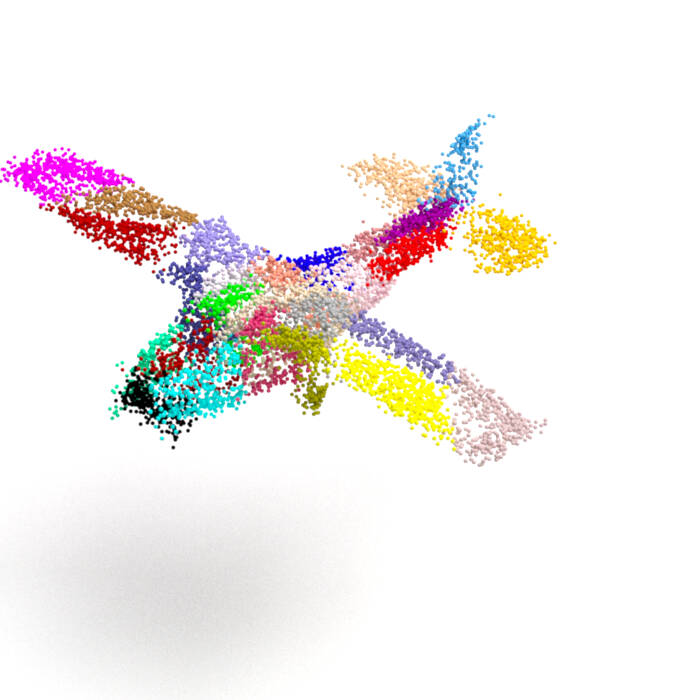}    \\
\end{tabular}
  \caption{Generation examples of airplane by ChartPointFlow.}
  \label{fig:appendix_generated_airplane}
\end{figure*}

\begin{figure*}[t]
  \centering
  \setlength{\tabcolsep}{0mm}
  \begin{tabular}{cccccccc}
    \includegraphics[width=21mm]{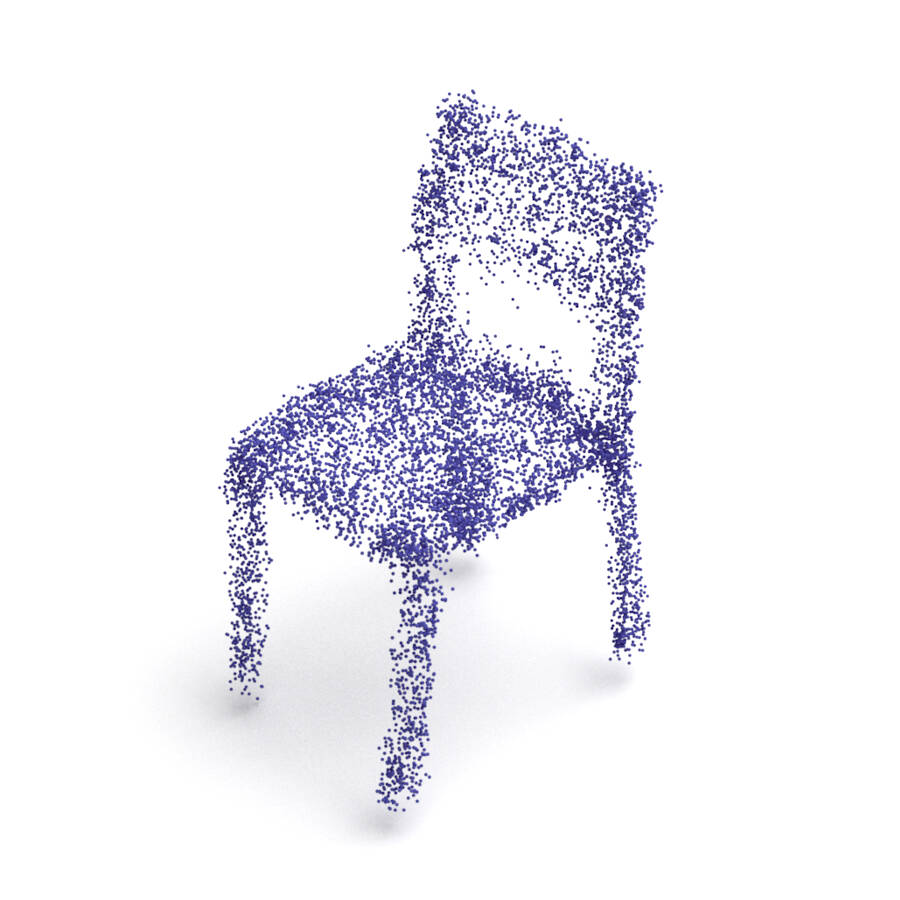}         &
    \includegraphics[width=21mm]{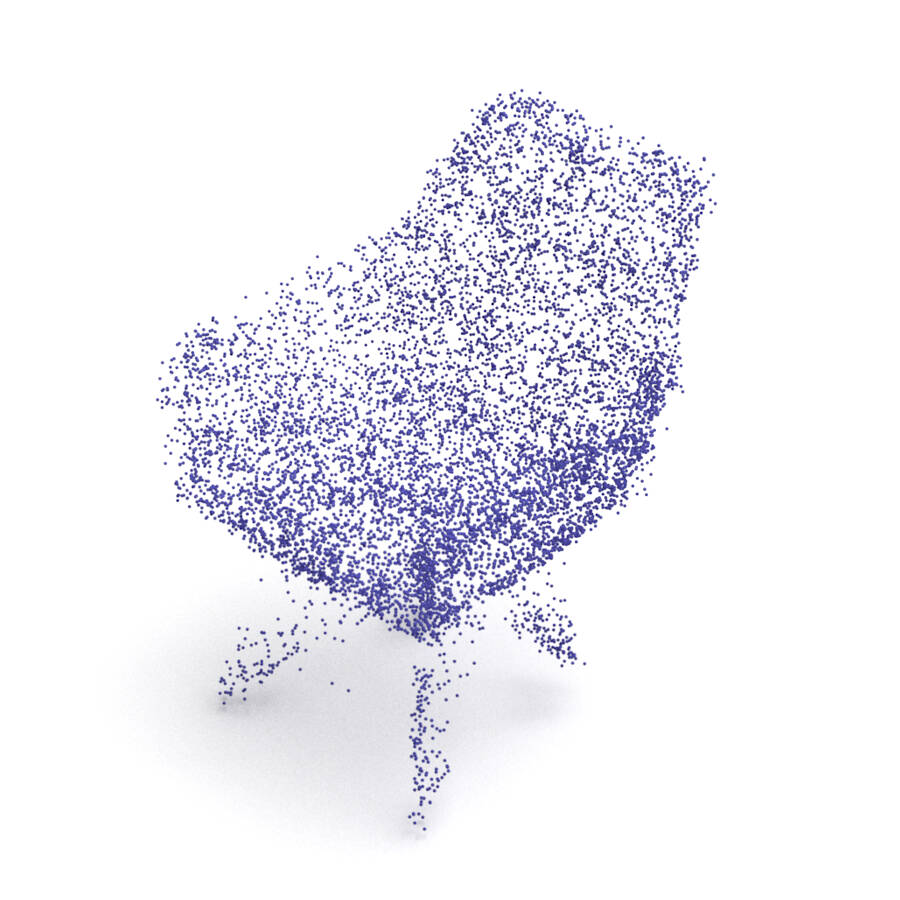}         &
    \includegraphics[width=21mm]{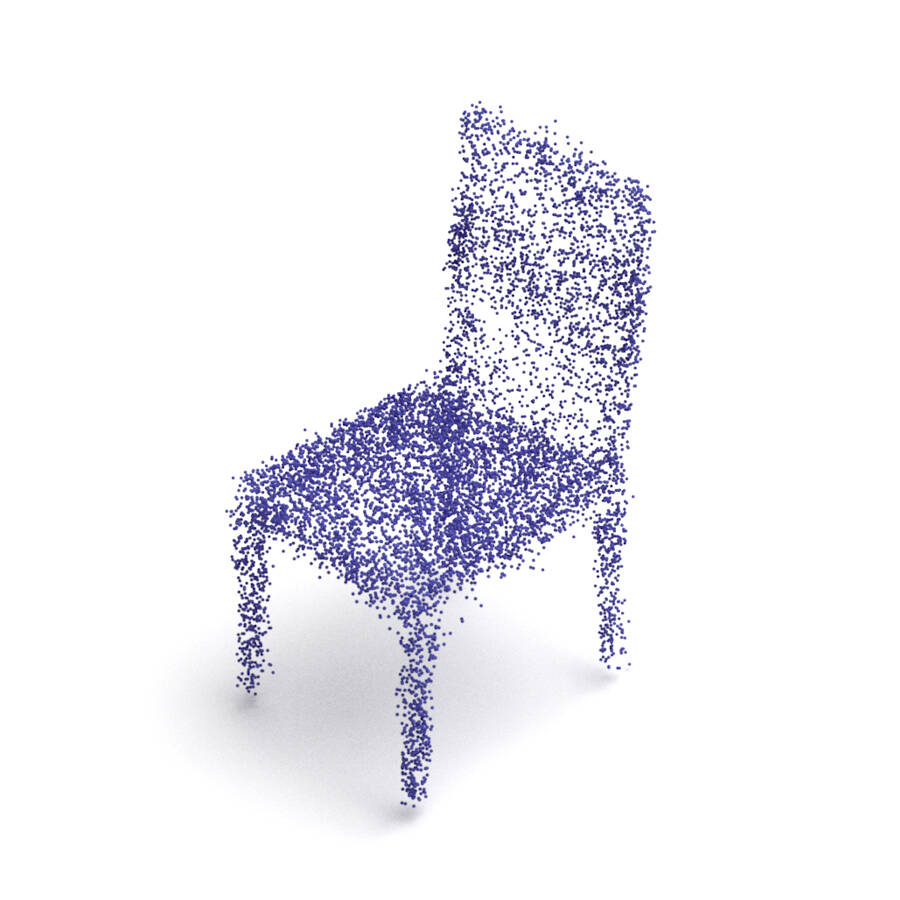}         &
    \includegraphics[width=21mm]{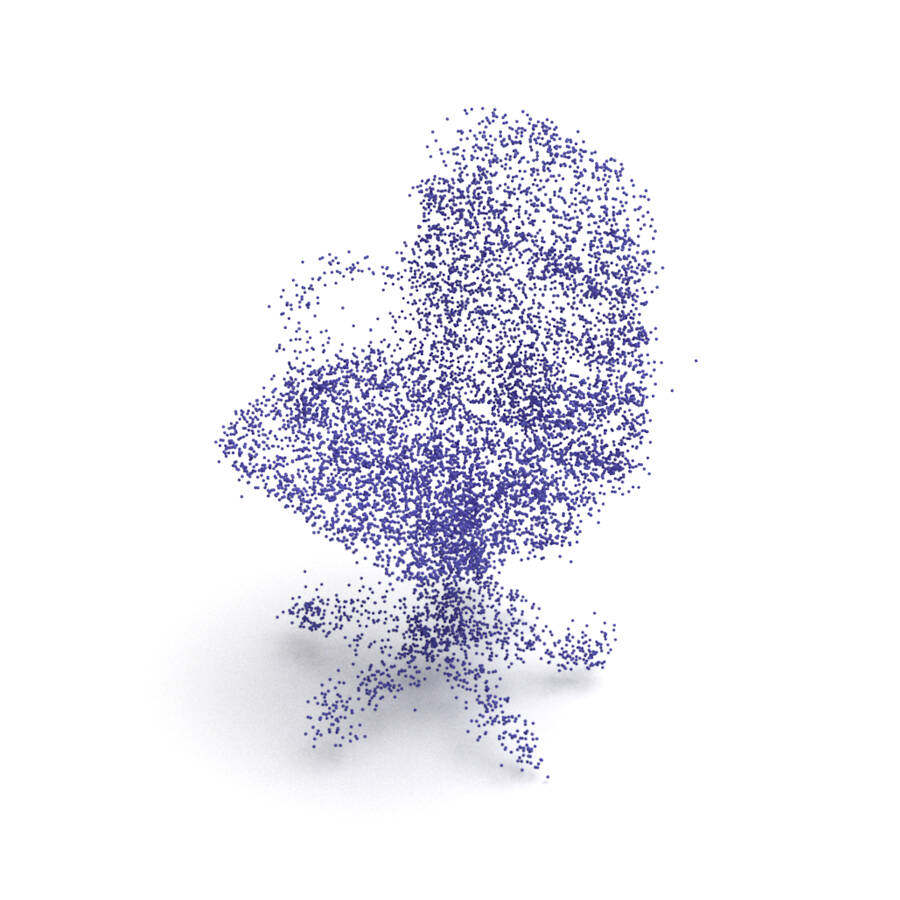}        &
    \includegraphics[width=21mm]{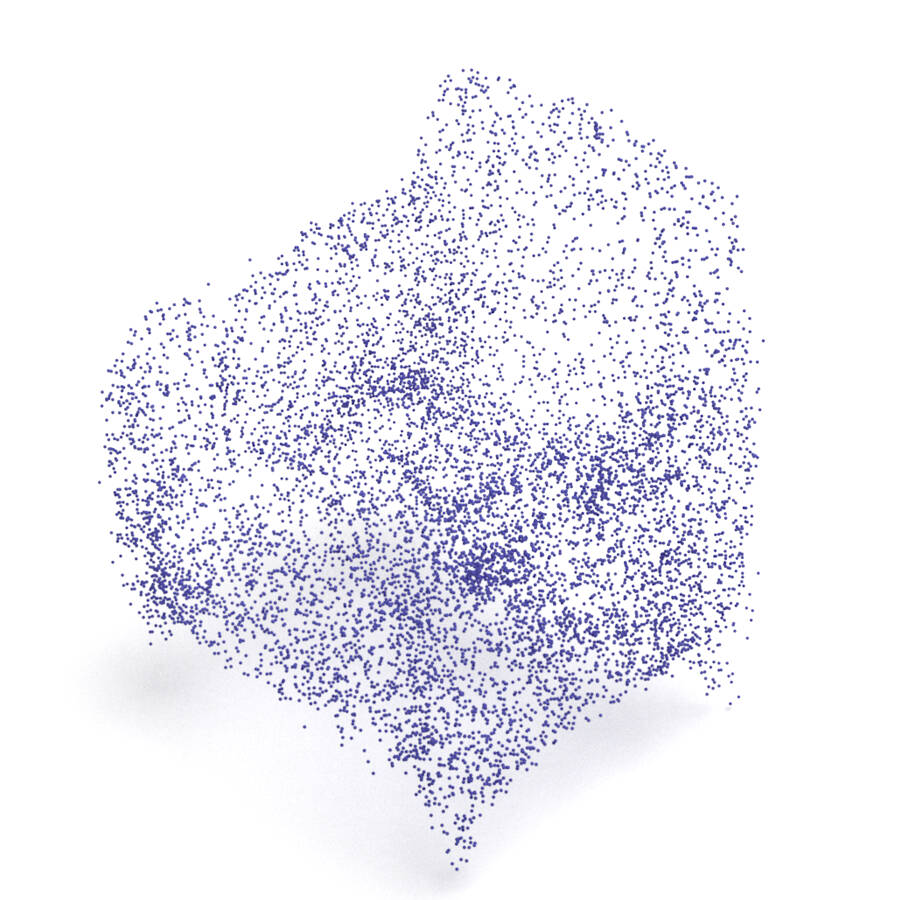}        &
    \includegraphics[width=21mm]{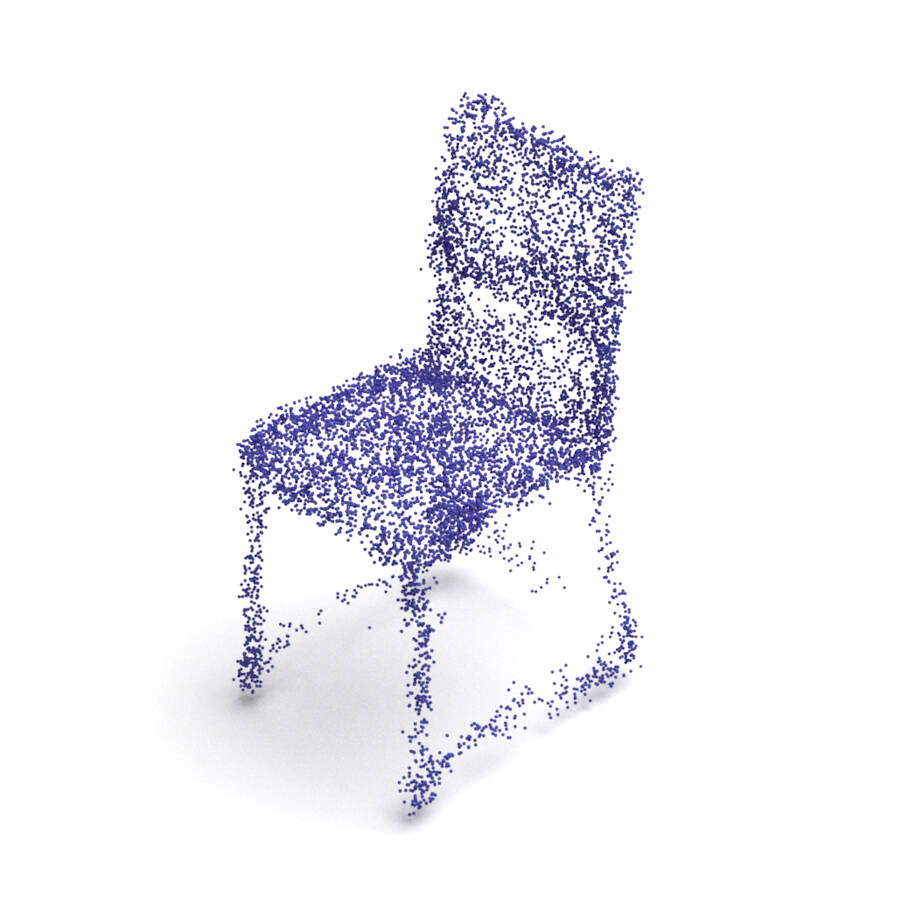}        &
    \includegraphics[width=21mm]{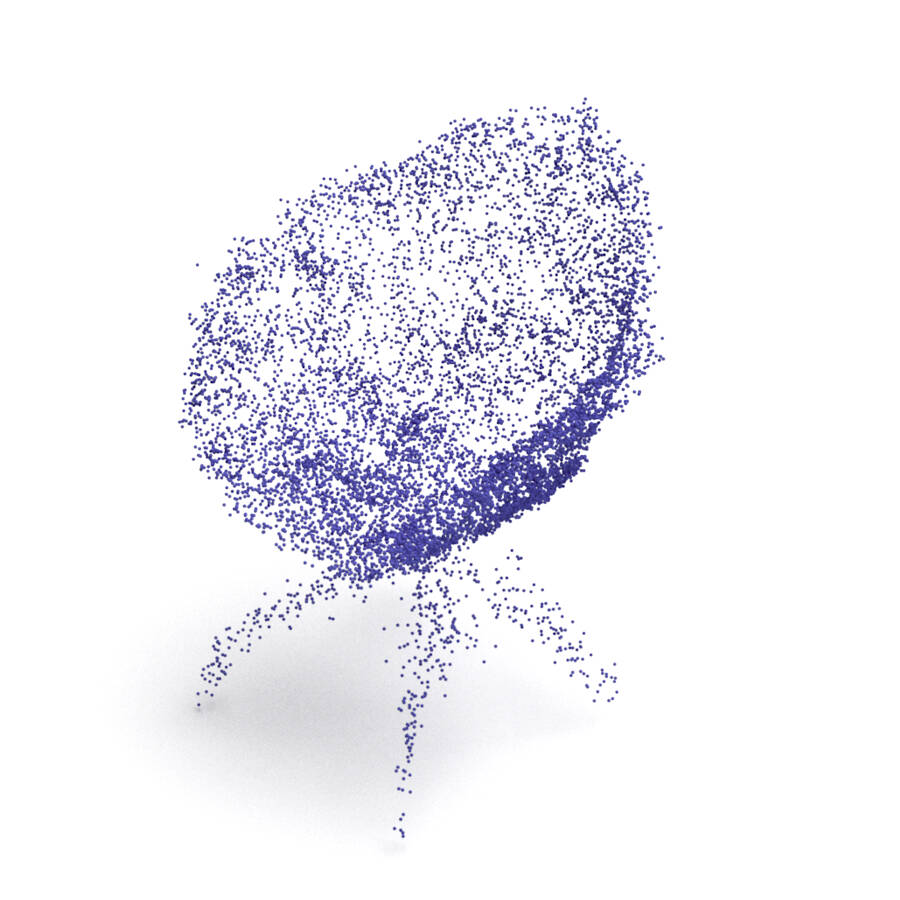}        &
    \includegraphics[width=21mm]{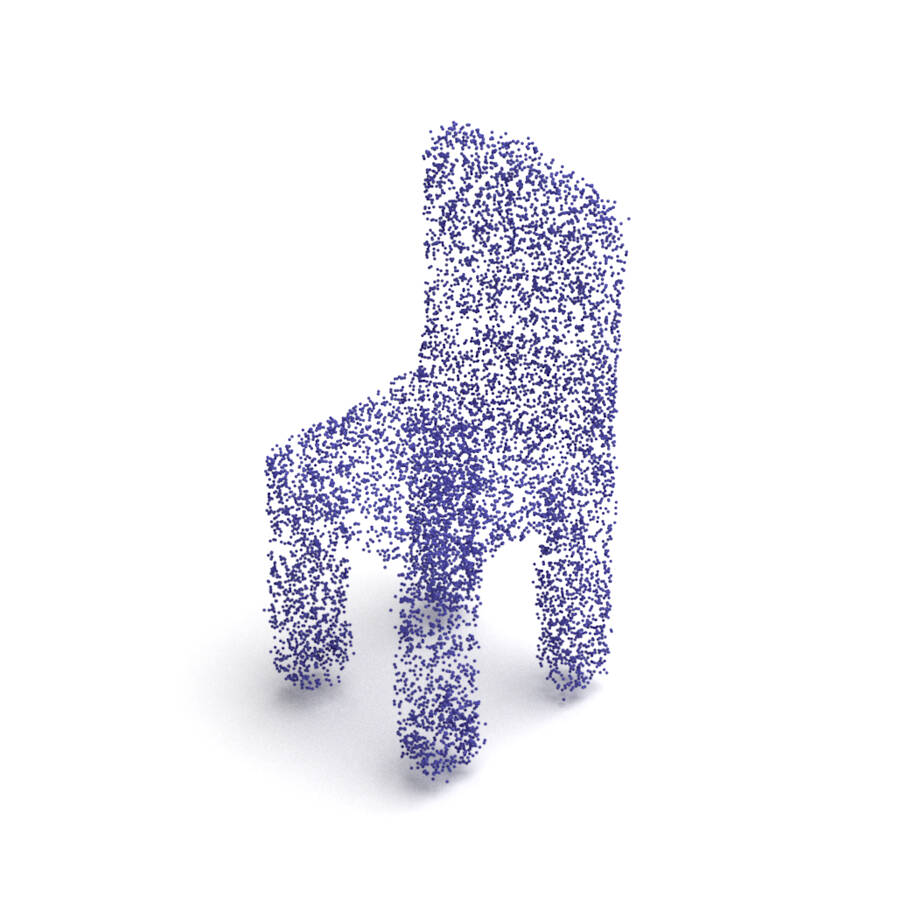}          \\
    \includegraphics[width=21mm]{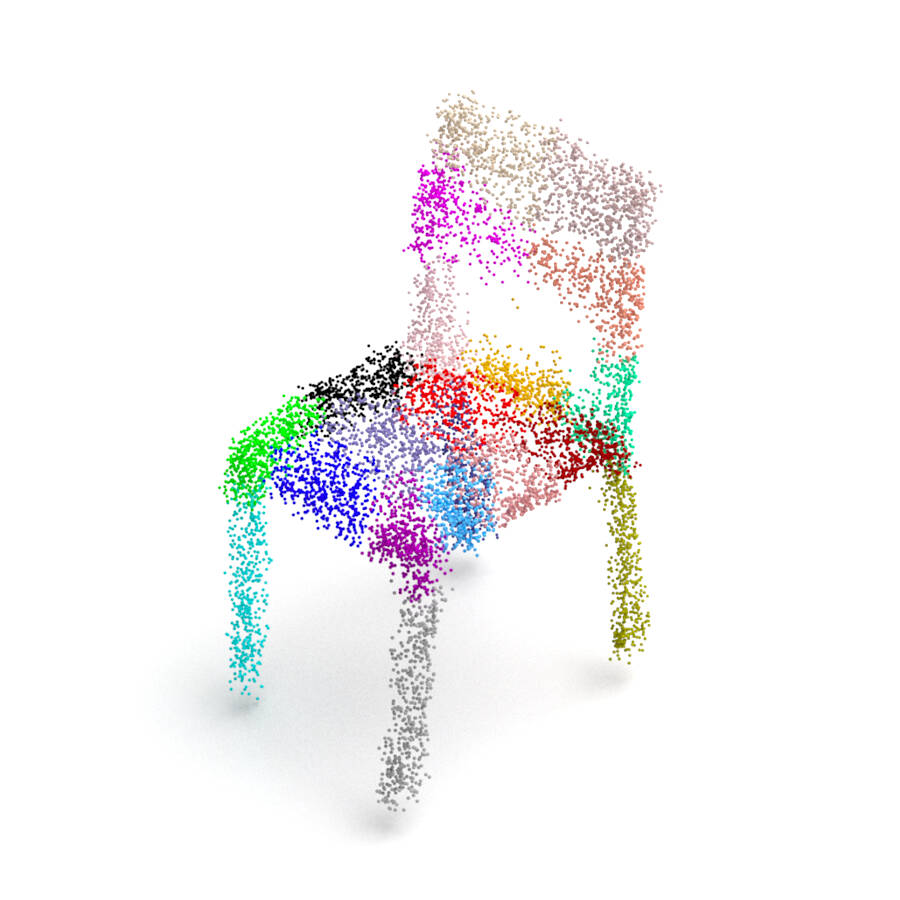}   &
    \includegraphics[width=21mm]{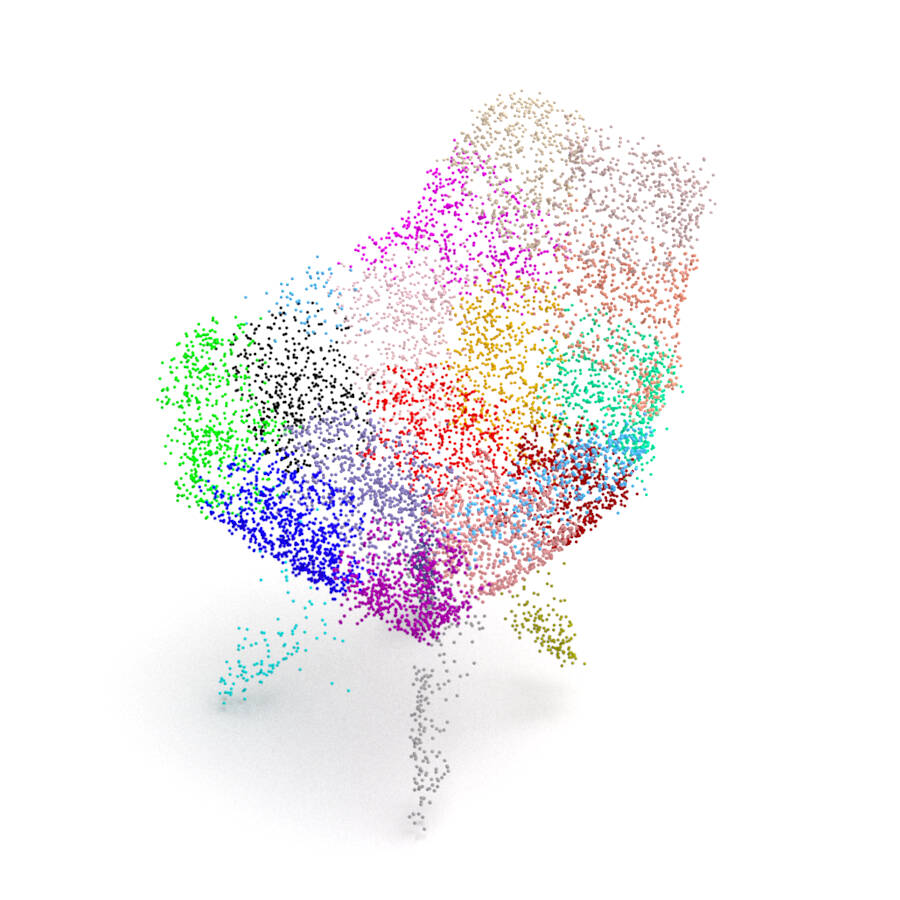}   &
    \includegraphics[width=21mm]{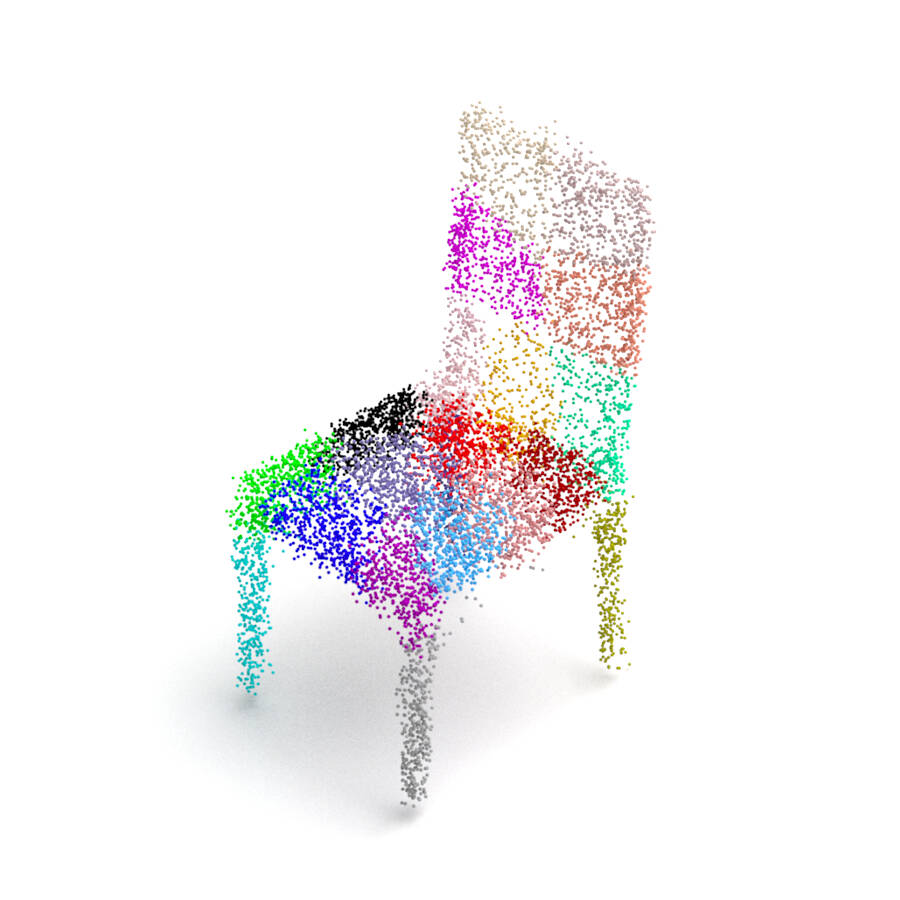}   &
    \includegraphics[width=21mm]{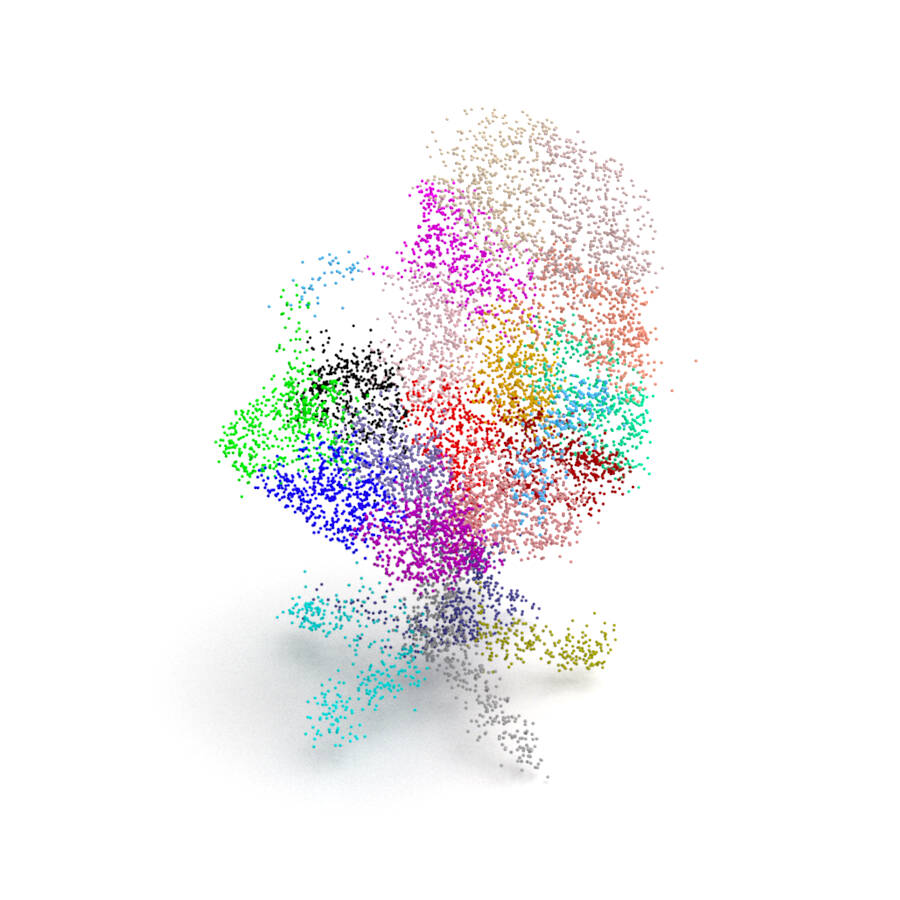}  &
    \includegraphics[width=21mm]{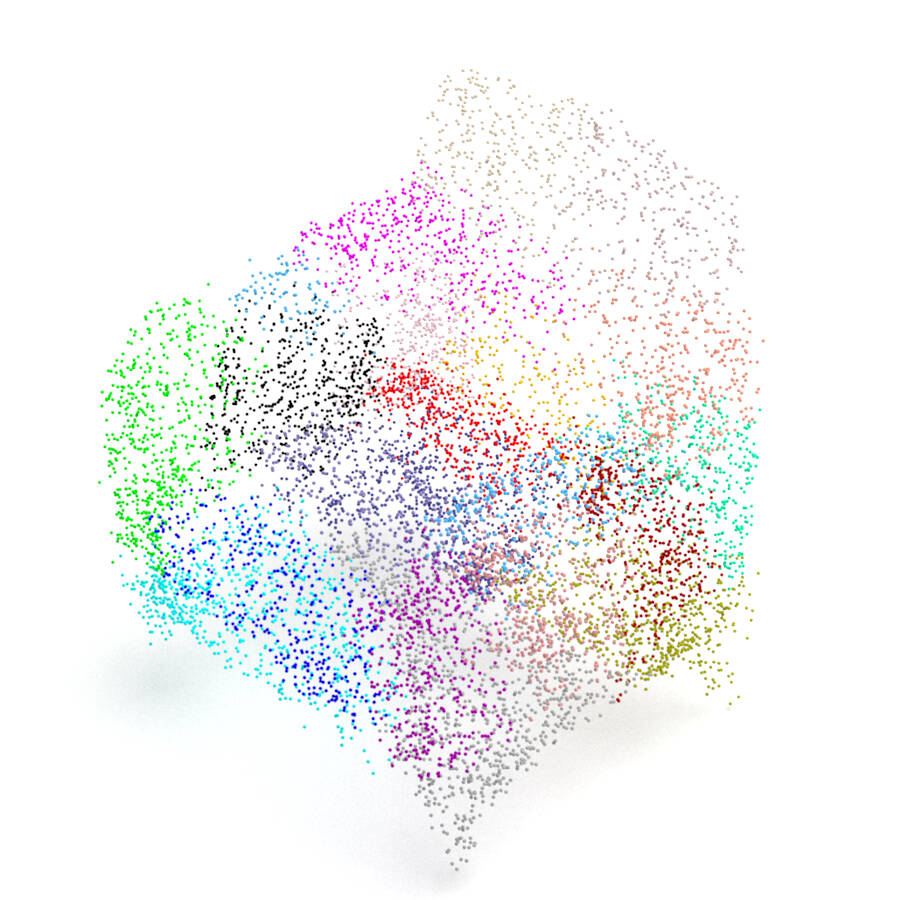}  &
    \includegraphics[width=21mm]{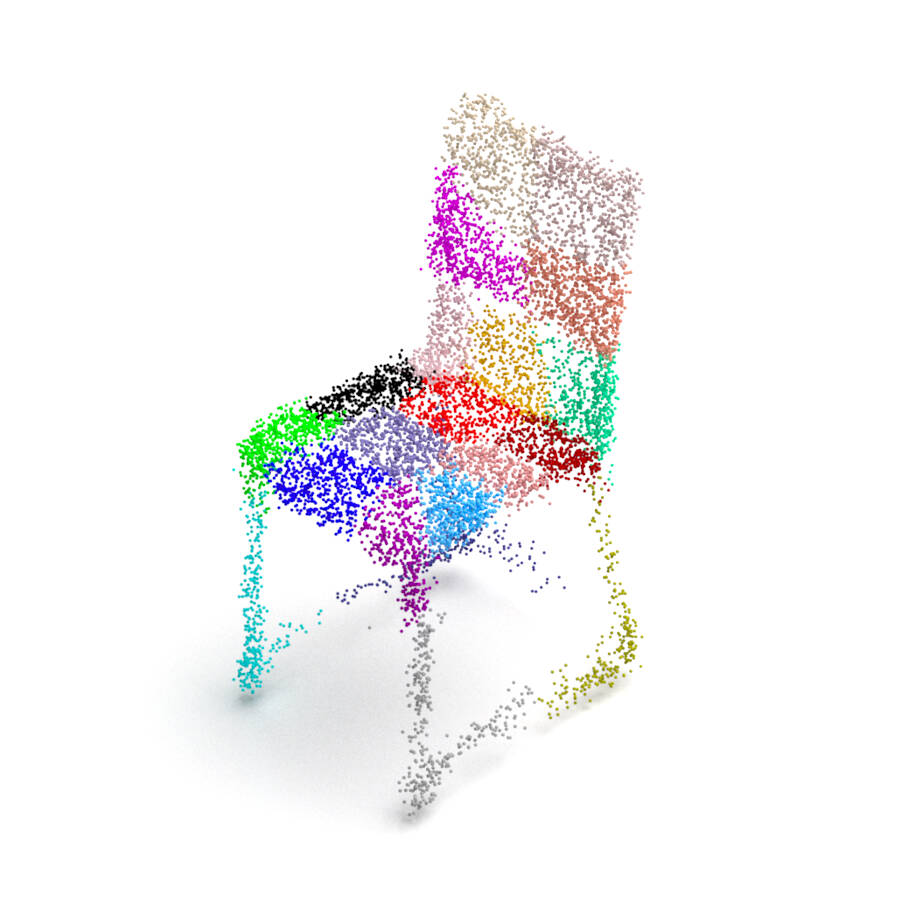}  &
    \includegraphics[width=21mm]{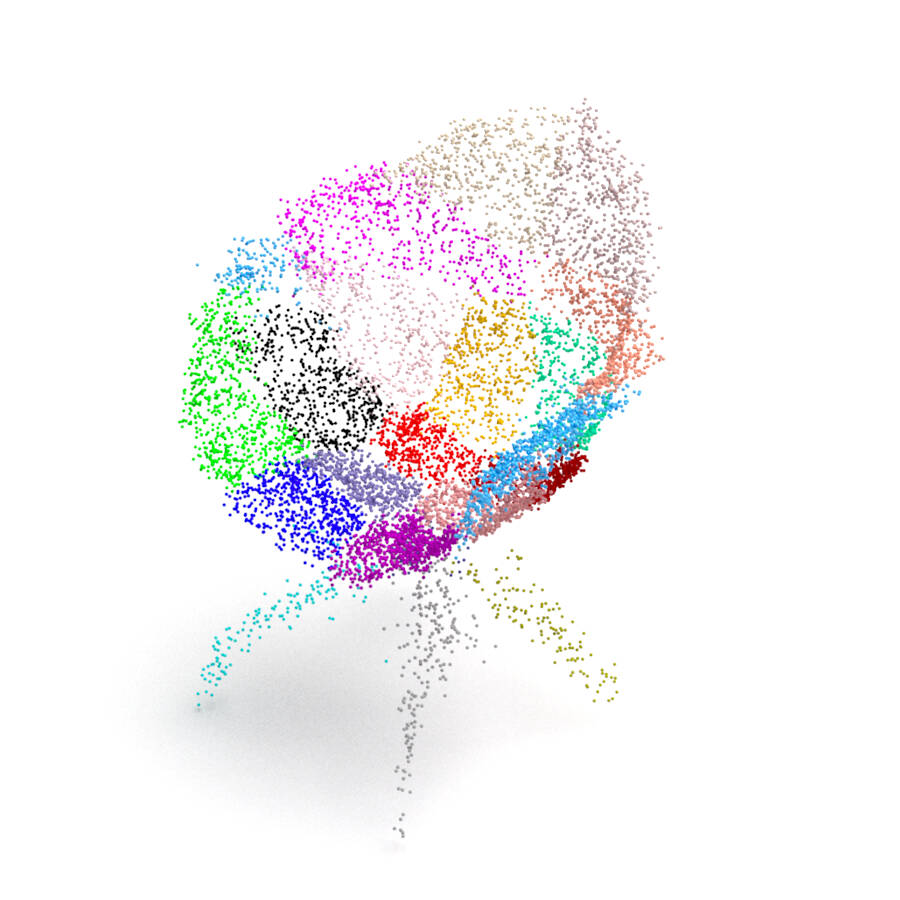}  &
    \includegraphics[width=21mm]{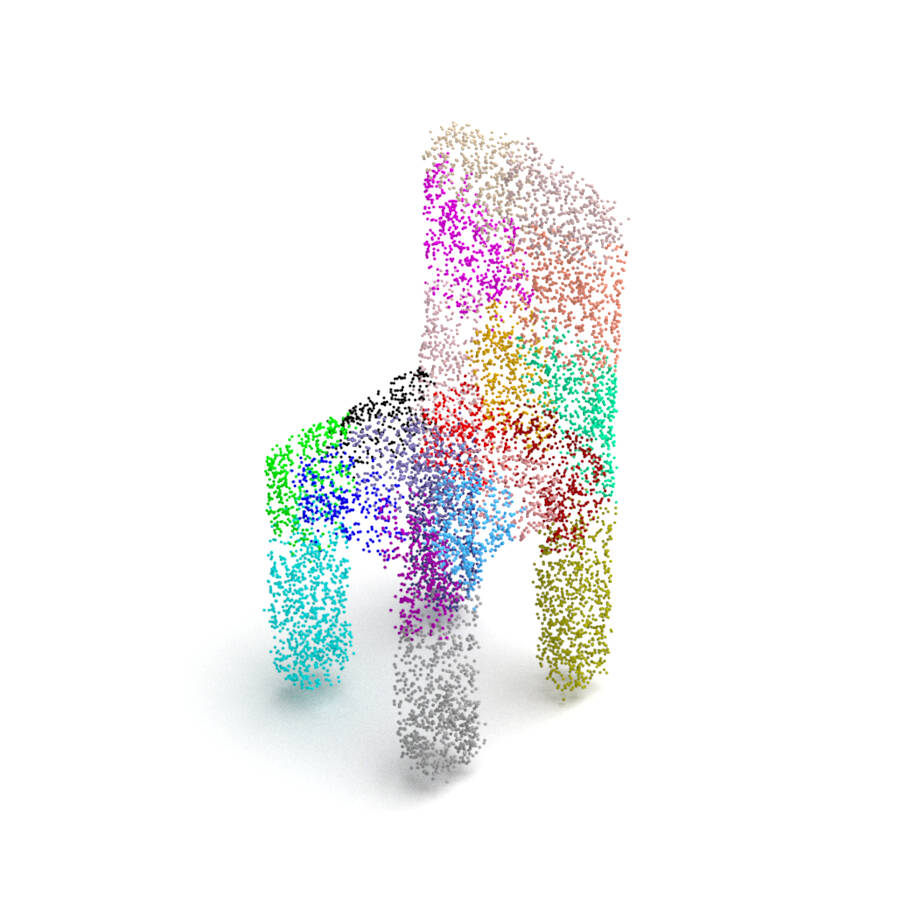}    \\
\end{tabular}
  \caption{Generation examples of chair by ChartPointFlow.}
  \label{fig:appendix_generated_chair}
\end{figure*}

\begin{figure*}[t]
  \centering
  \setlength{\tabcolsep}{0mm}
  \begin{tabular}{cccccccc}
    \includegraphics[width=21mm]{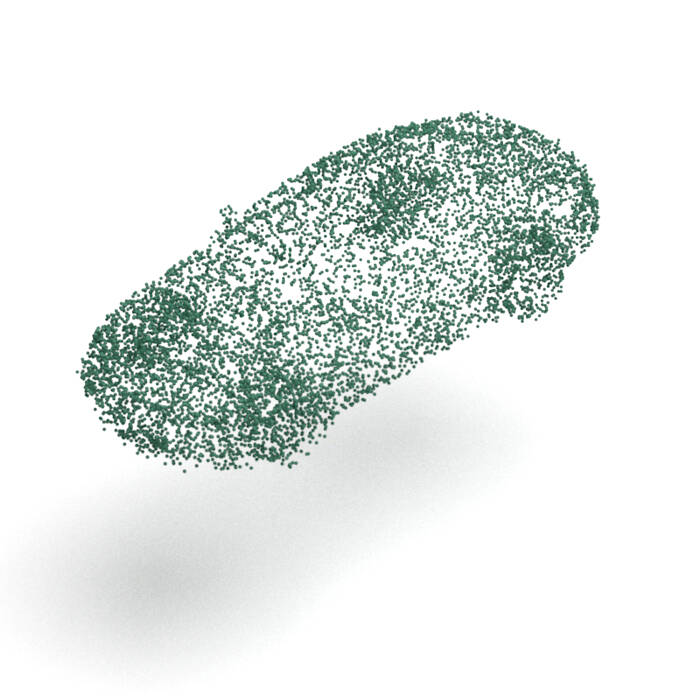}         &
    \includegraphics[width=21mm]{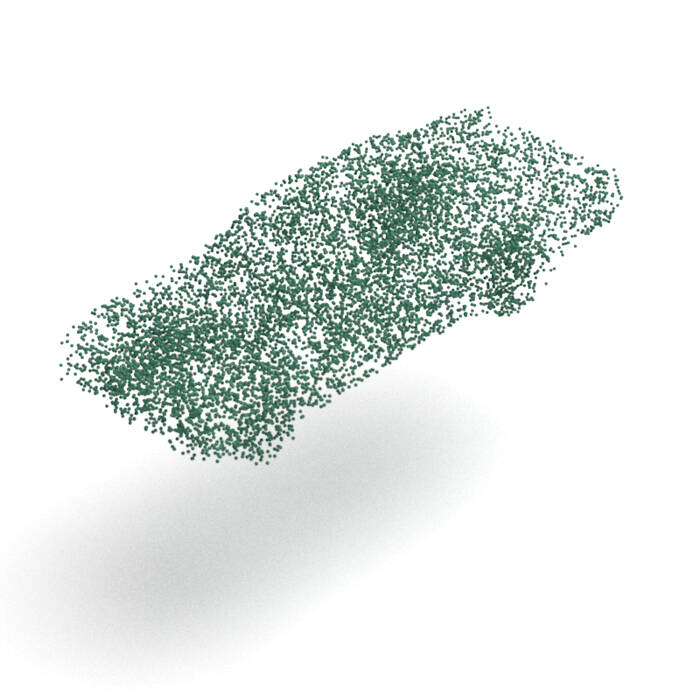}         &
    \includegraphics[width=21mm]{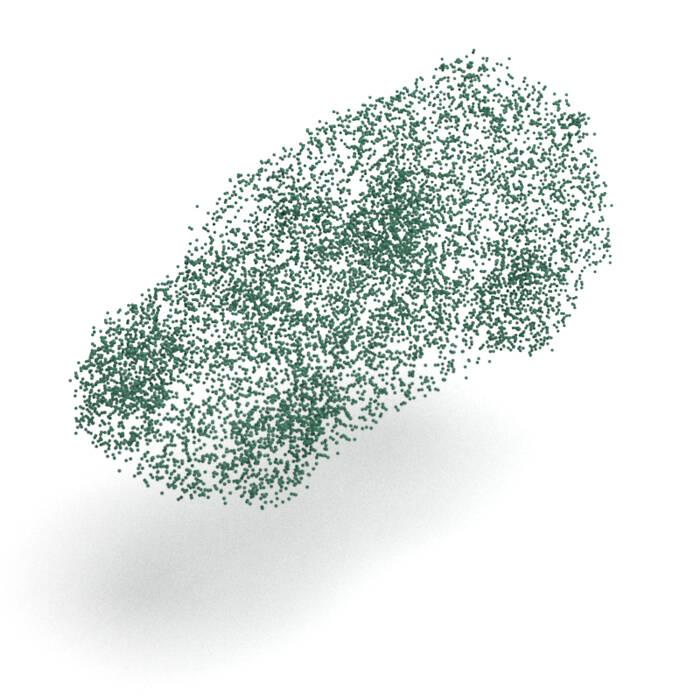}         &
    \includegraphics[width=21mm]{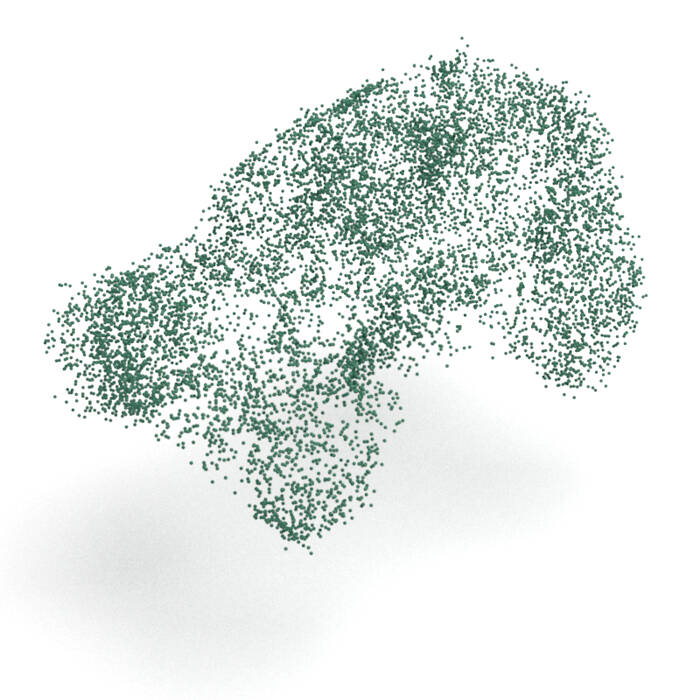}        &
    \includegraphics[width=21mm]{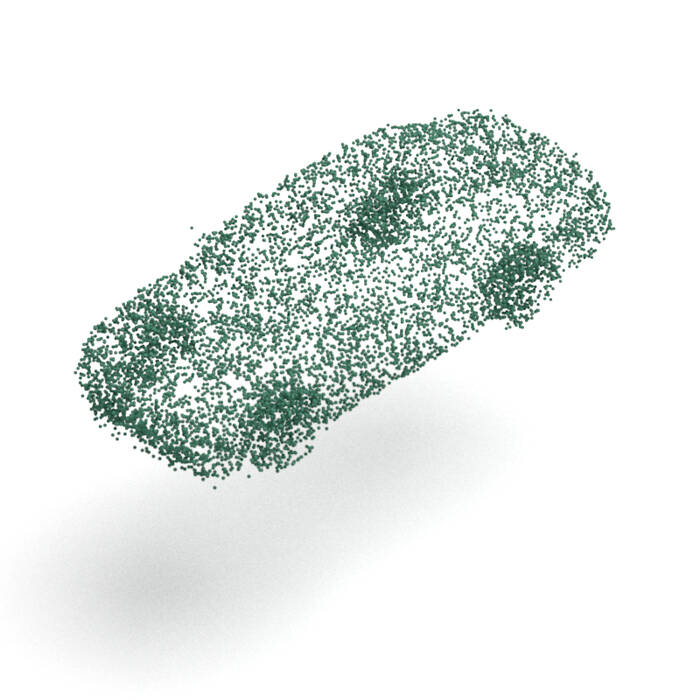}        &
    \includegraphics[width=21mm]{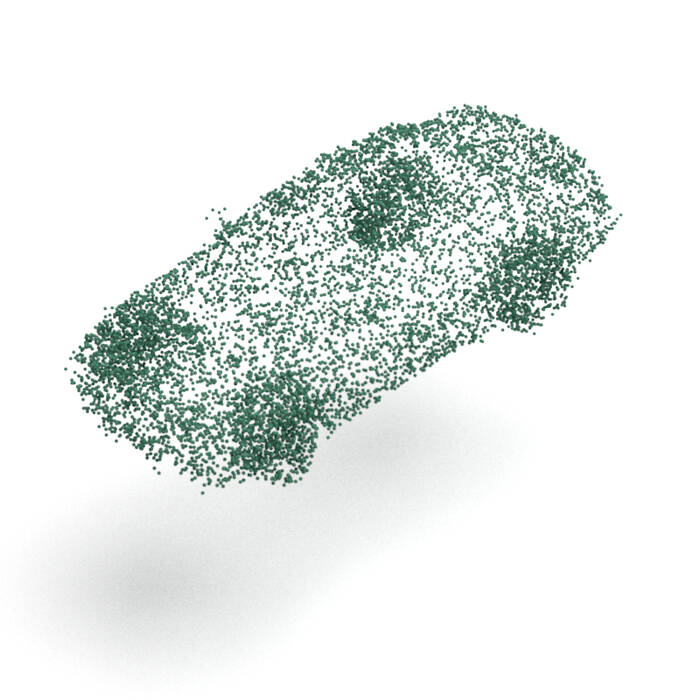}       &
    \includegraphics[width=21mm]{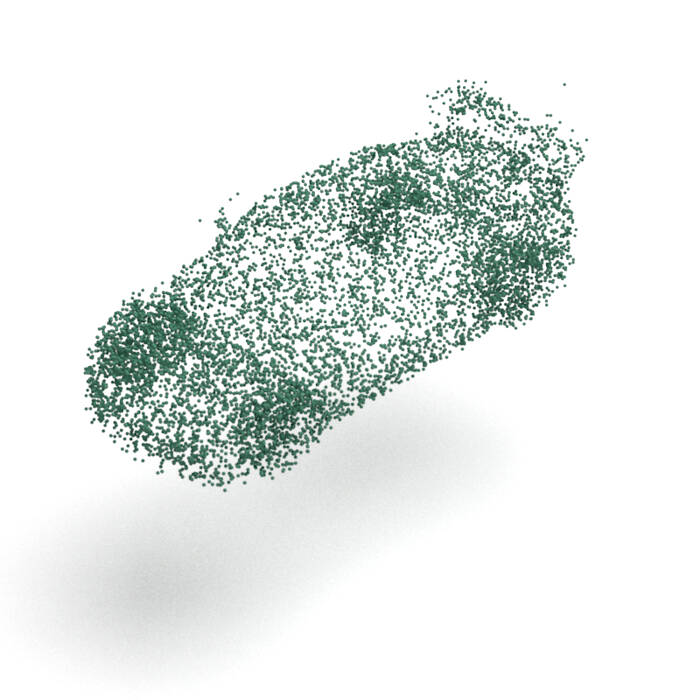}       &
    \includegraphics[width=21mm]{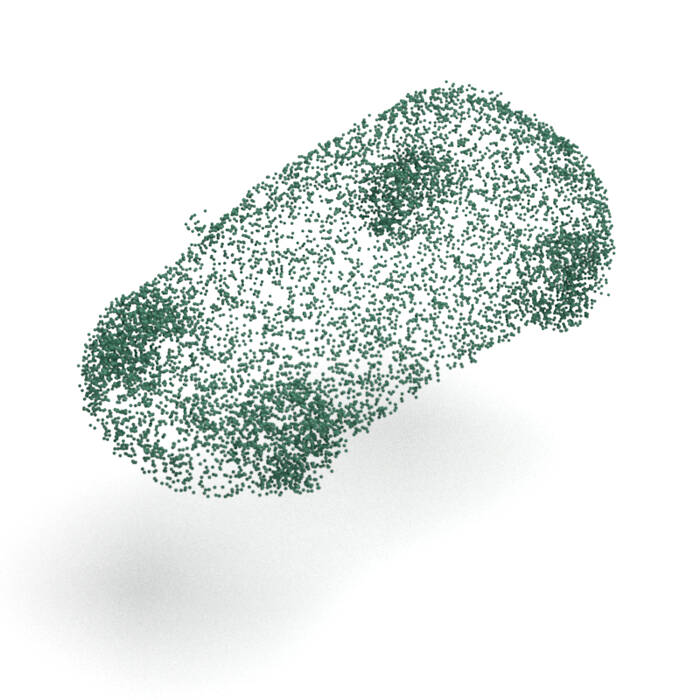}         \\
    \includegraphics[width=21mm]{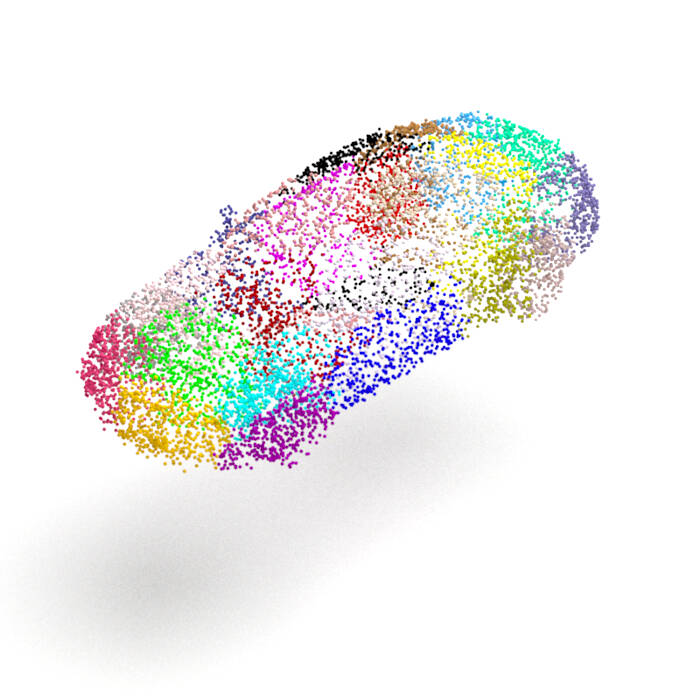}   &
    \includegraphics[width=21mm]{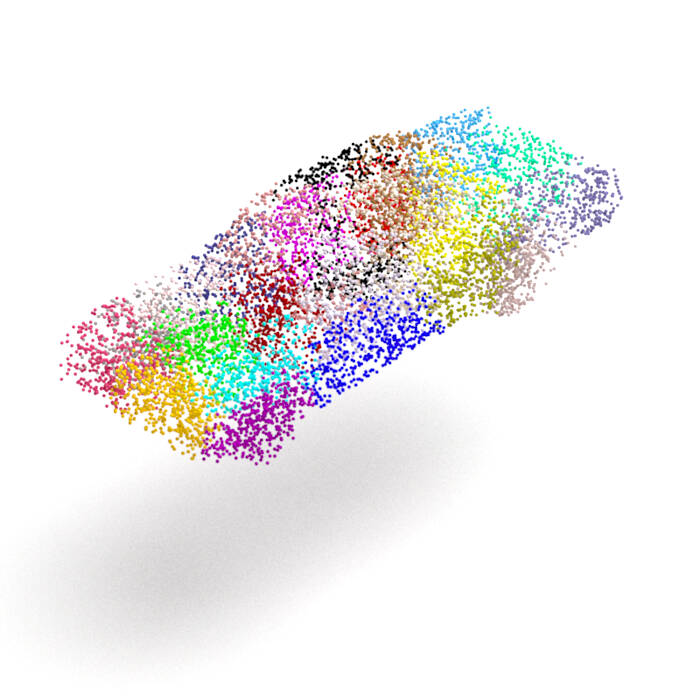}   &
    \includegraphics[width=21mm]{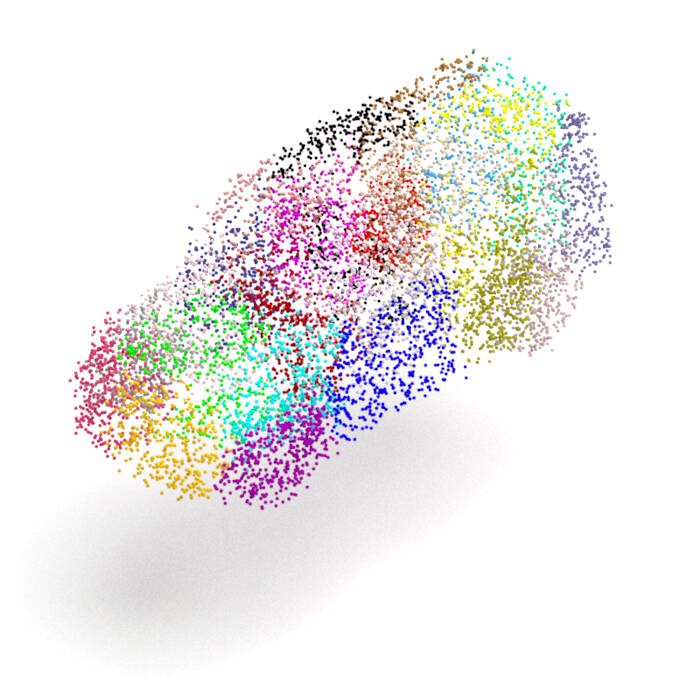}   &
    \includegraphics[width=21mm]{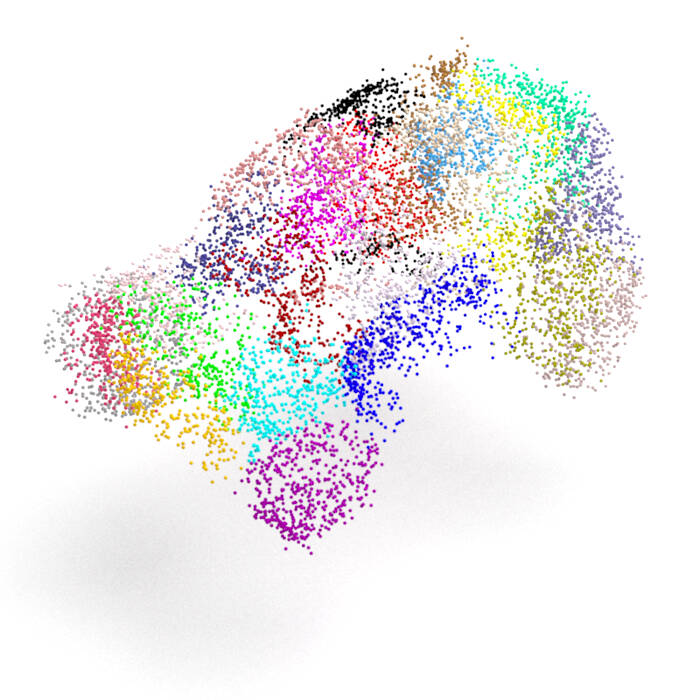}  &
    \includegraphics[width=21mm]{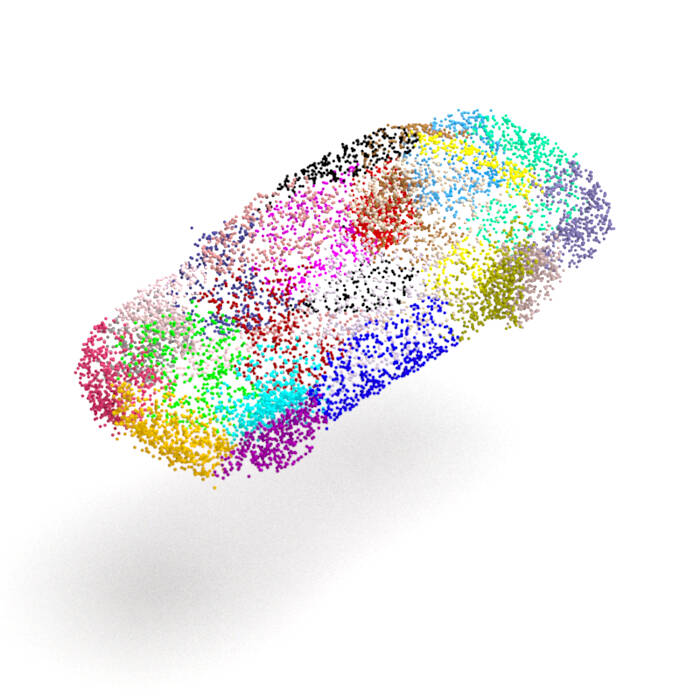}  &
    \includegraphics[width=21mm]{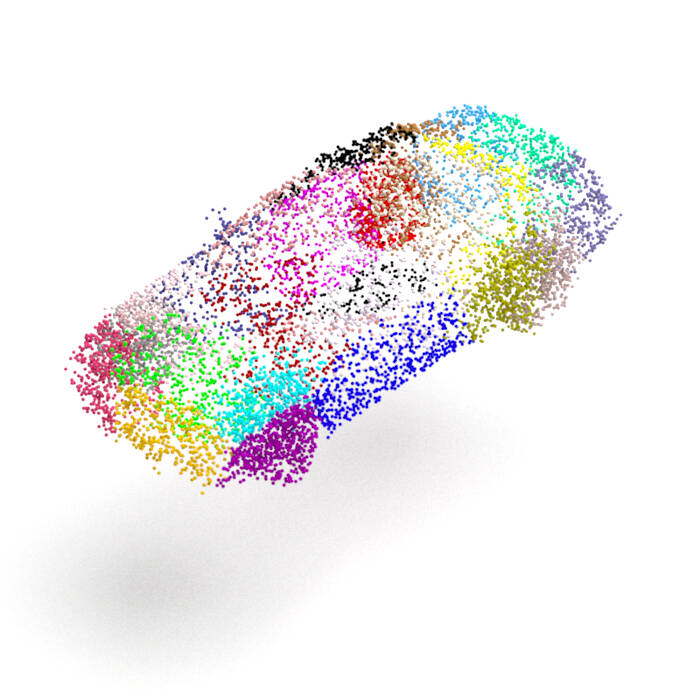} &
    \includegraphics[width=21mm]{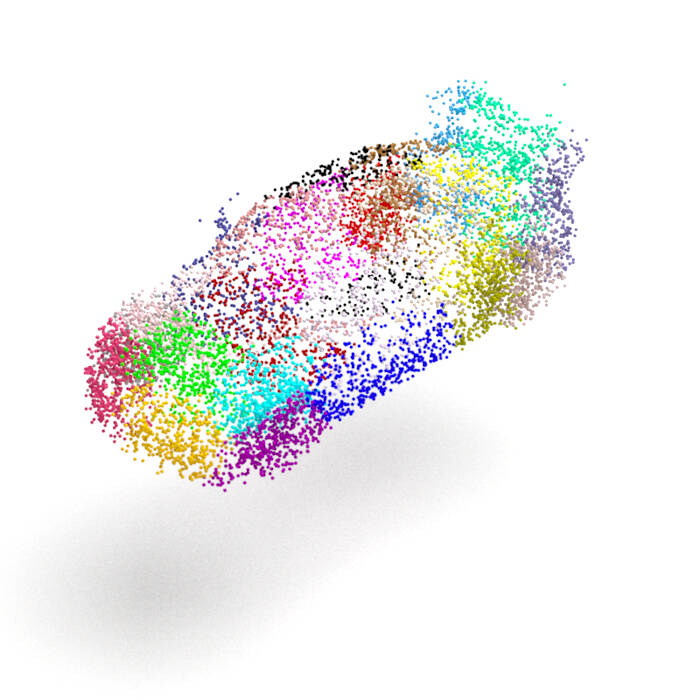} &
    \includegraphics[width=21mm]{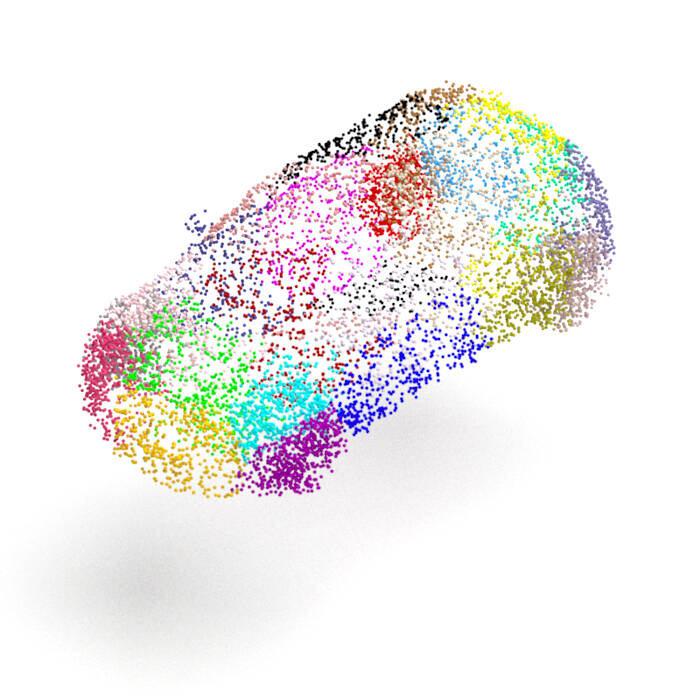}   \\
\end{tabular}
  \caption{Generation examples of car by ChartPointFlow.}
  \label{fig:appendix_generated_car}
\end{figure*}

Figures~\ref{fig:appendix_generated_airplane}--\ref{fig:appendix_generated_car} summarize samples generated by ChartPointFlow.
One can see that a wide variety of objects are generated, and the same chart is assigned to the same subpart across objects, such as the airplane wings, chair legs, and car wheels.
For example, in Fig.~\ref{fig:appendix_generated_airplane}, the charts denoted by yellow, purple, and pink colors cover the front half, rear half, and wing tip of the left wing of an airplane, respectively.
The assignment is independent of the absolute position or the shape of the left wing.
This is true even for a stealth aircraft, whose left wing is not separated from the main body.
Therefore, we conclude that ChartPointFlow learned the fine-grained semantic information.

\begin{figure*}[t]
  \centering
  \setlength{\tabcolsep}{1mm}
  \begin{tabular}{ccccccc}
    \includegraphics[width=21mm]{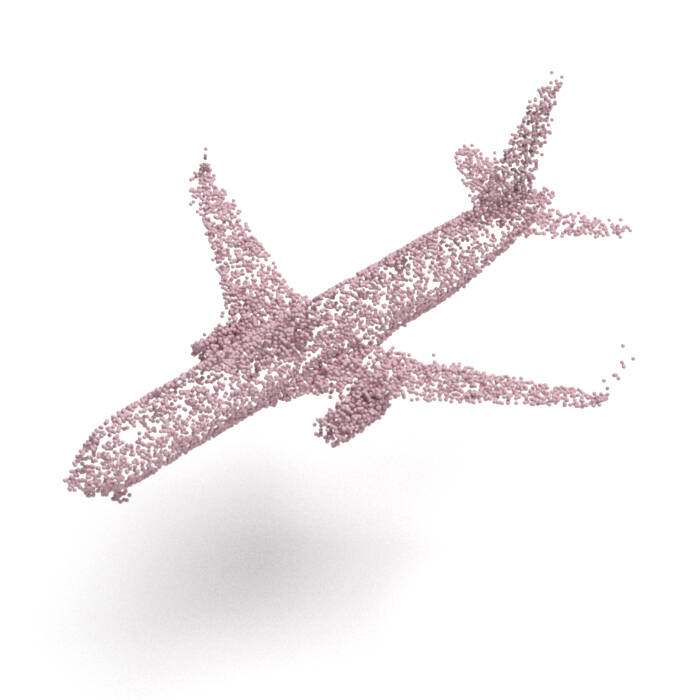}           &
    \includegraphics[width=21mm]{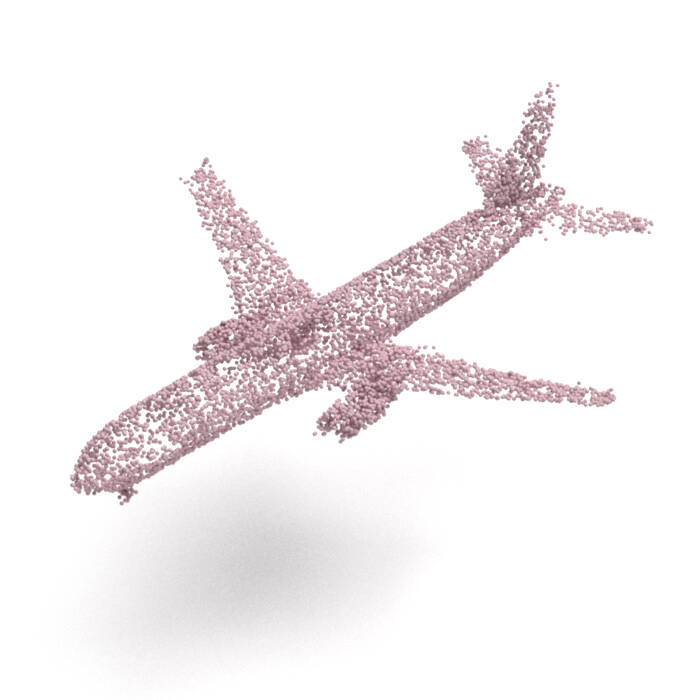}           &
    \includegraphics[width=21mm]{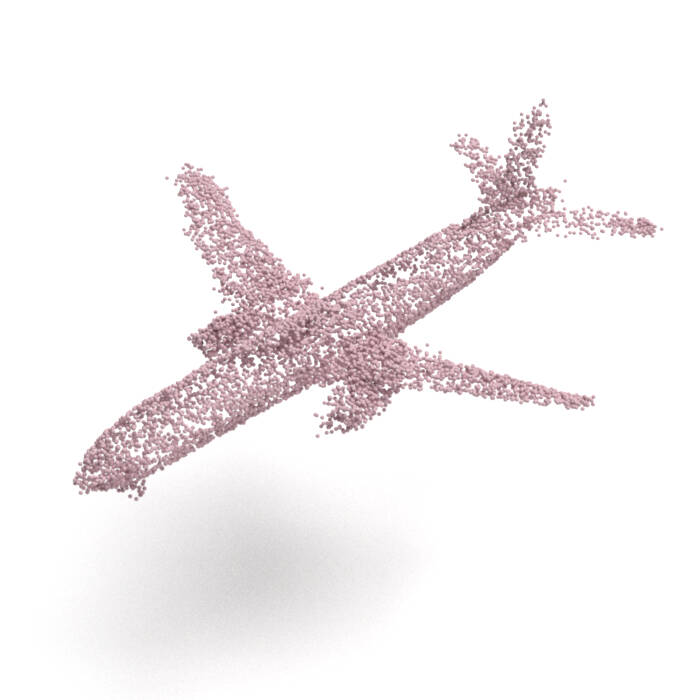}           &
    \includegraphics[width=21mm]{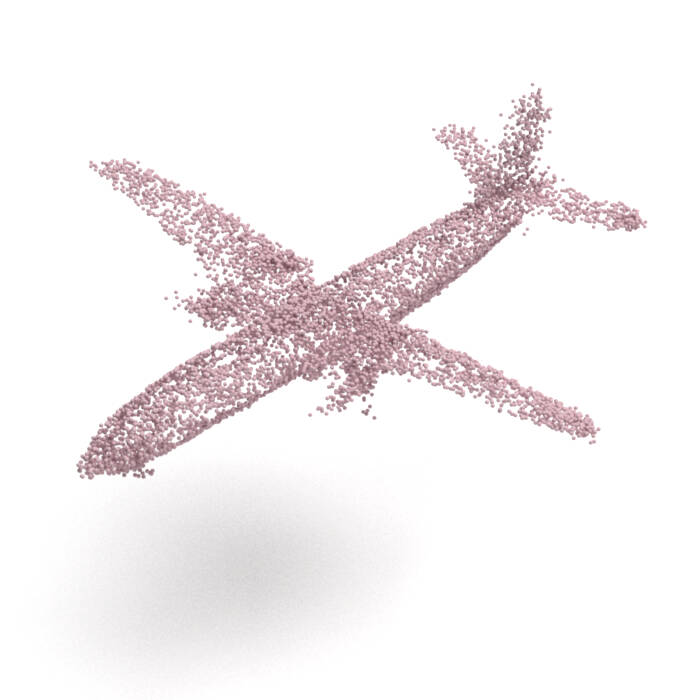}           &
    \includegraphics[width=21mm]{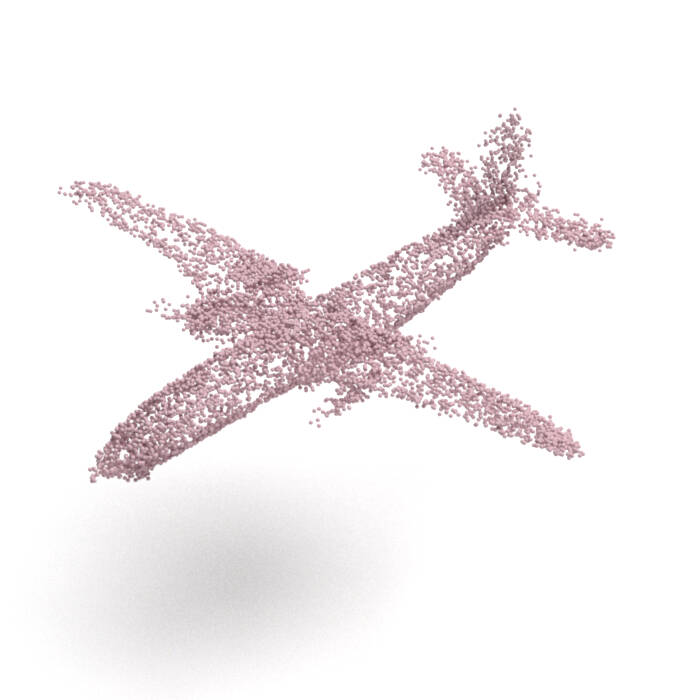}           &
    \includegraphics[width=21mm]{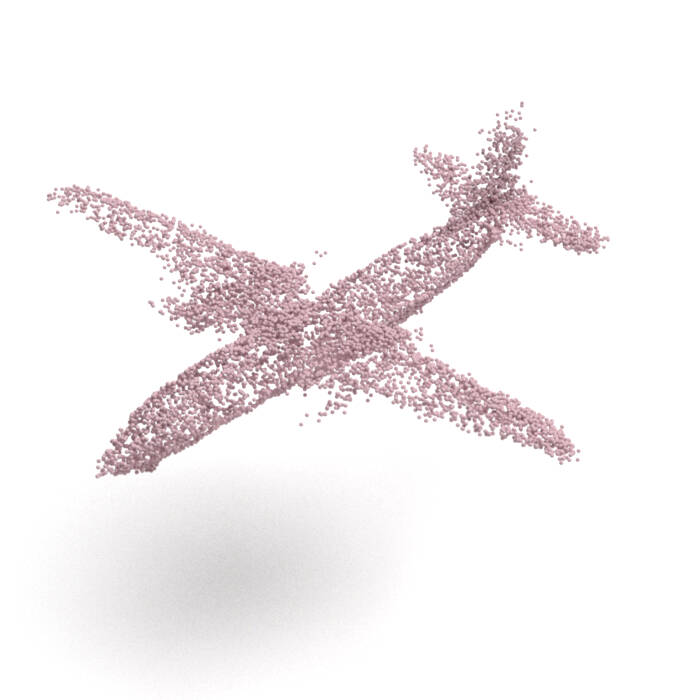}           &
    \includegraphics[width=21mm]{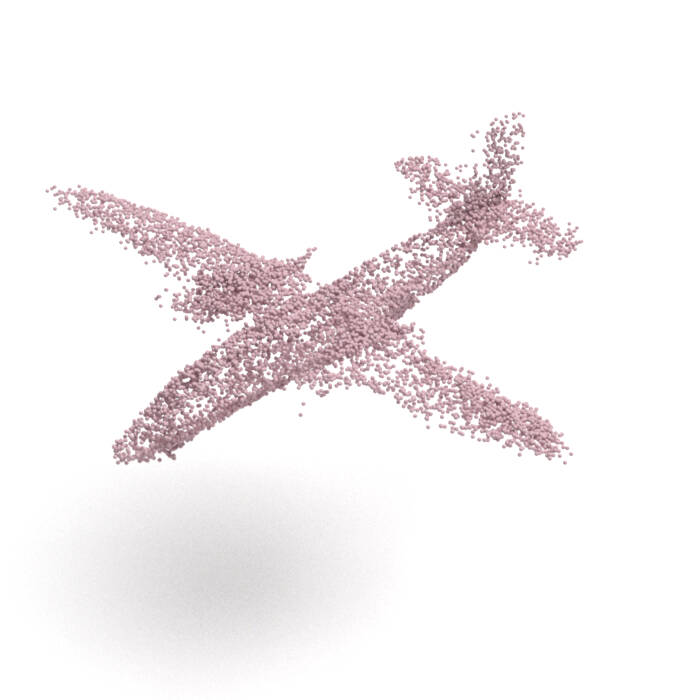}            \\
    \includegraphics[width=21mm]{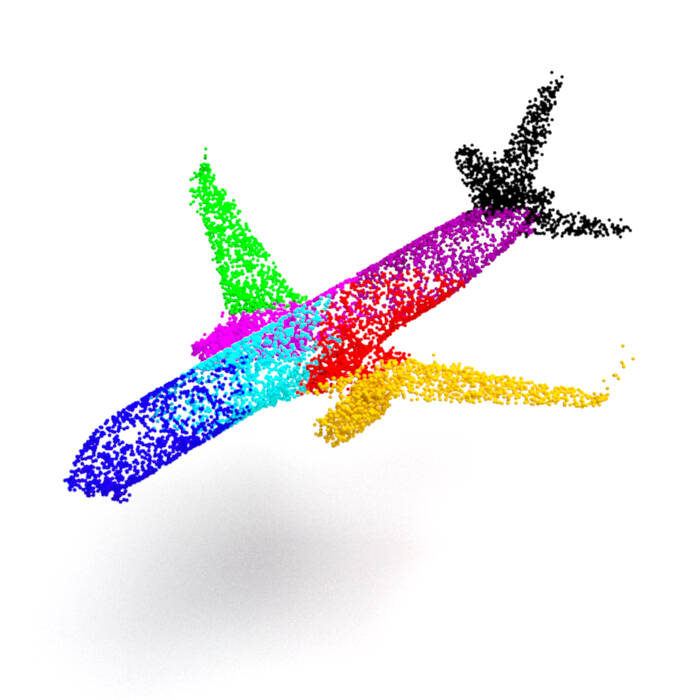}     &
    \includegraphics[width=21mm]{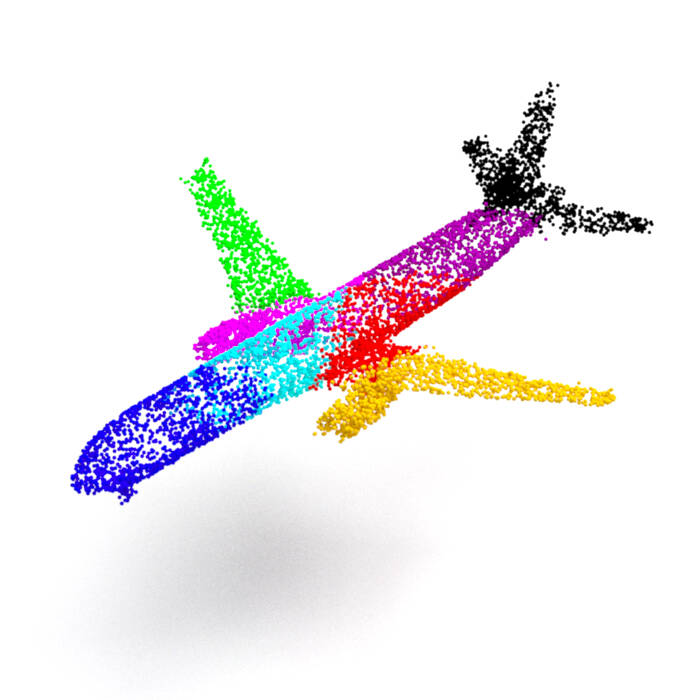}     &
    \includegraphics[width=21mm]{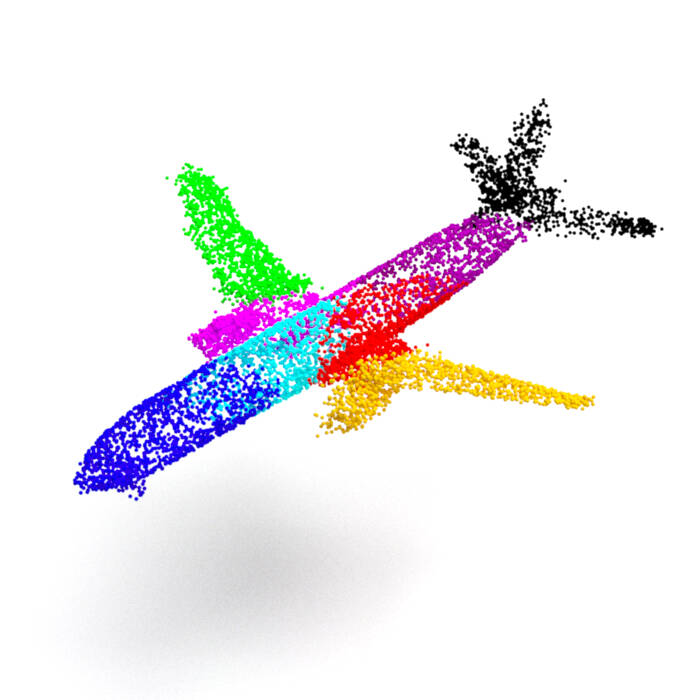}     &
    \includegraphics[width=21mm]{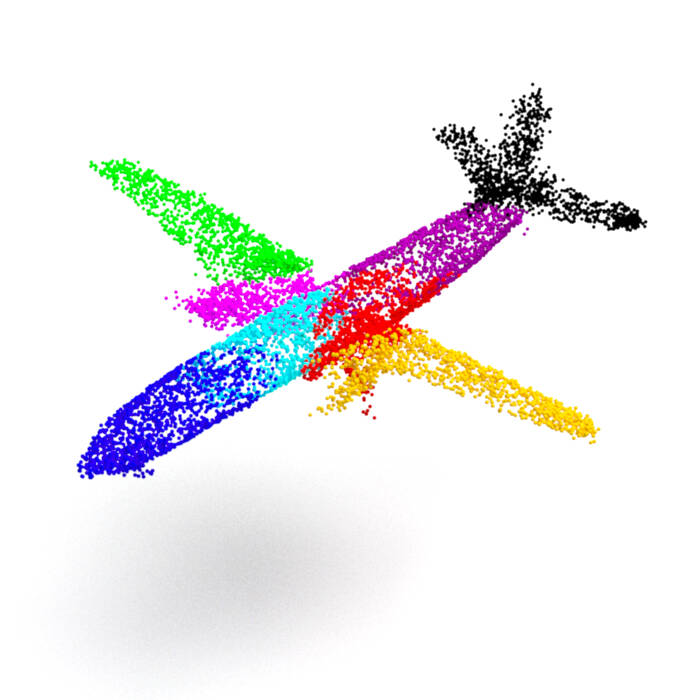}     &
    \includegraphics[width=21mm]{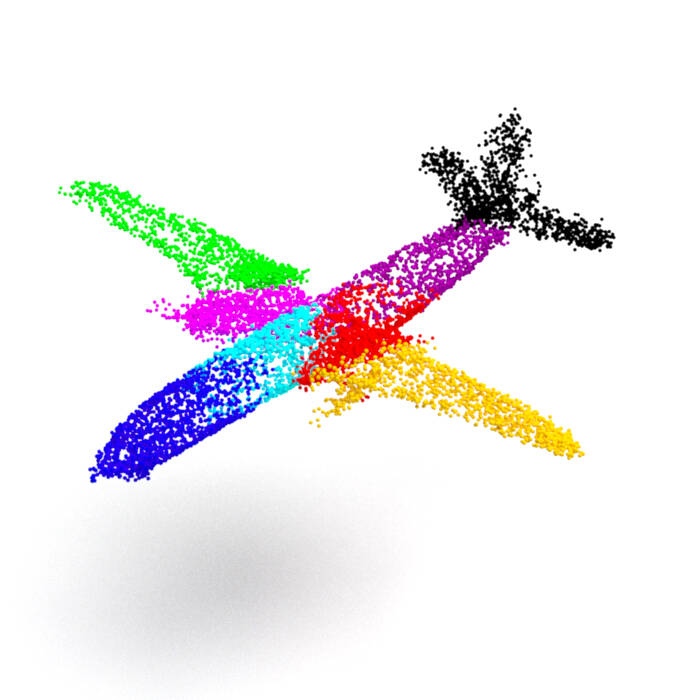}     &
    \includegraphics[width=21mm]{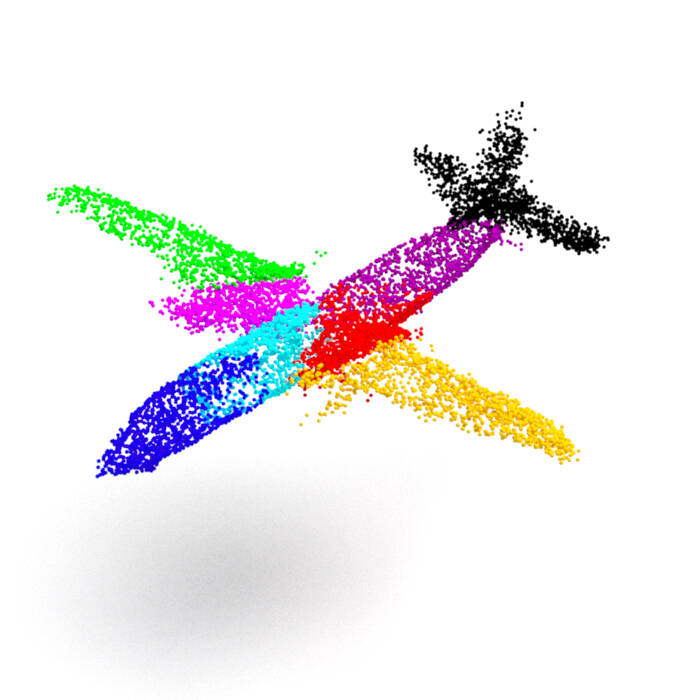}     &
    \includegraphics[width=21mm]{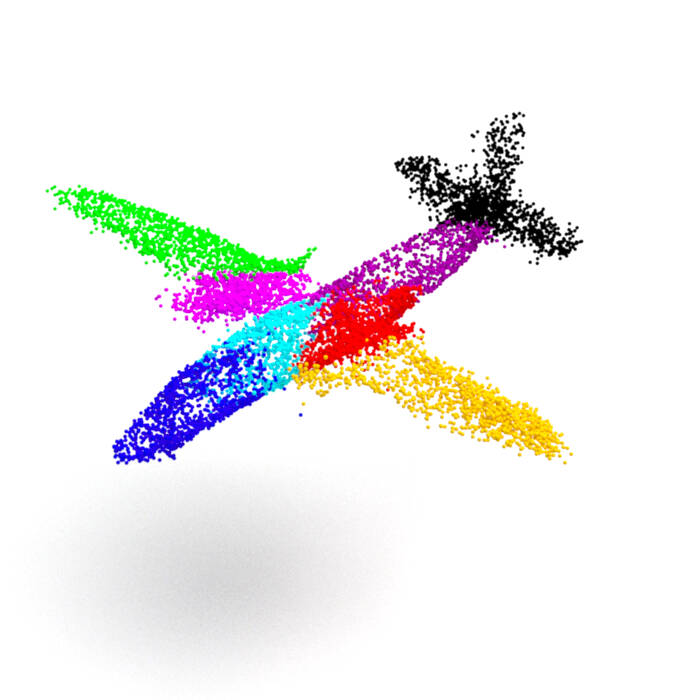}      \\
\end{tabular}
  \caption{Linear interpolation of the feature vector $s_{\!X}$ between two point clouds.}
  \label{fig:appendix_interpolation}
\end{figure*}

Figure~\ref{fig:appendix_interpolation} shows the point clouds obtained through the linear interpolation of the feature vector $s_{\!X}$ between two point clouds.
To improve the visibility, we set the number $n$ of charts to $8$.
At the leftmost column in the chair category, each of the four legs is covered by a different chart.
With the changing feature vector $s_{\!X}$, the two legs on each side come close to each other and collide, forming a different structure.
In this way, ChartPointFlow expresses a variety of shapes through a continuous deformation.

\begin{figure*}[t]
  \centering
  \setlength{\tabcolsep}{1mm}
  \begin{tabular}{lccccccc}
    \rotatetitle{18mm}{\textbf{Data}}                                                      &
    \includegraphics[width=21mm]{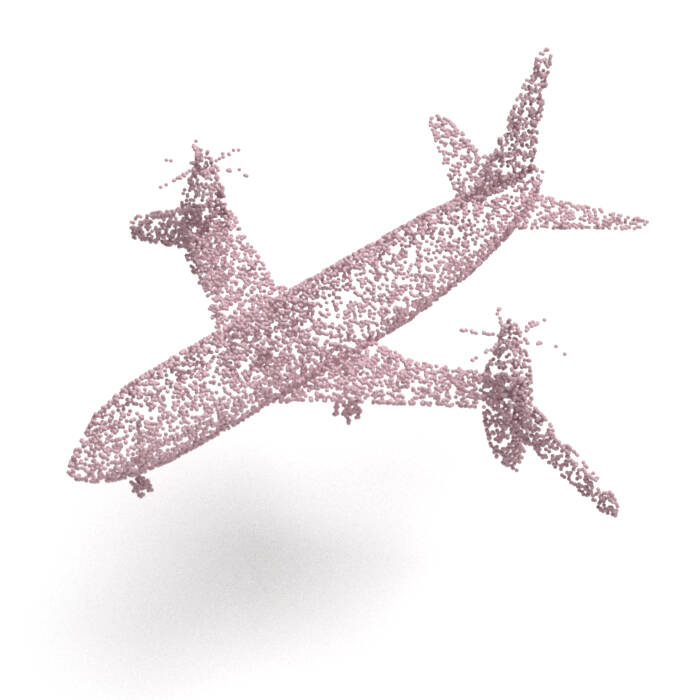}      &
    \includegraphics[width=21mm]{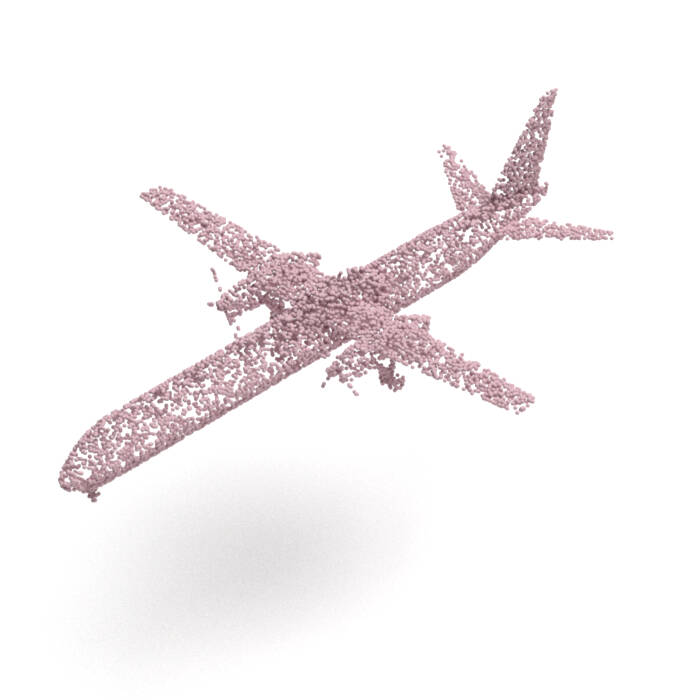}     &
    \includegraphics[width=21mm]{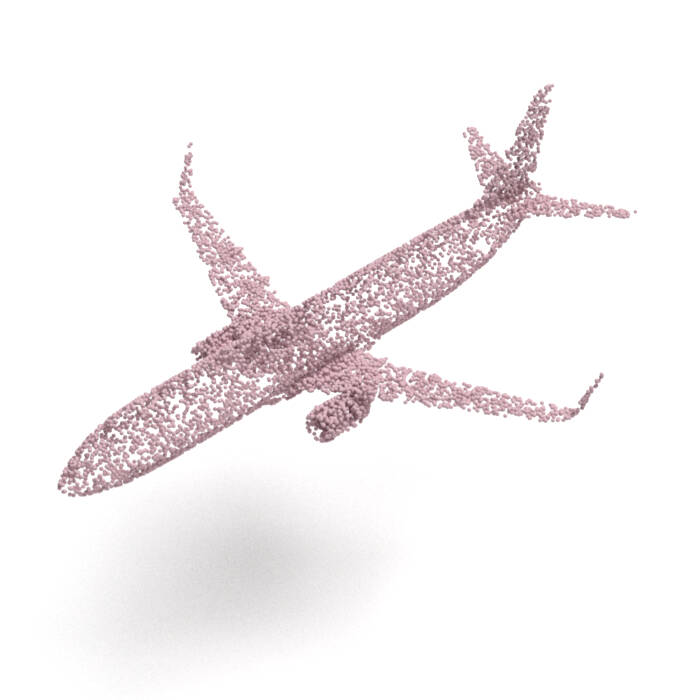}     &
    \includegraphics[width=21mm]{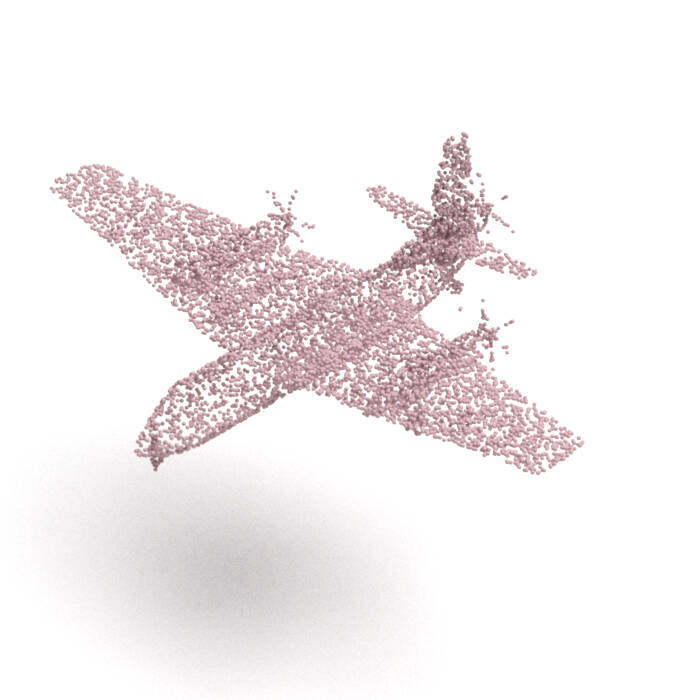}     &
    \includegraphics[width=21mm]{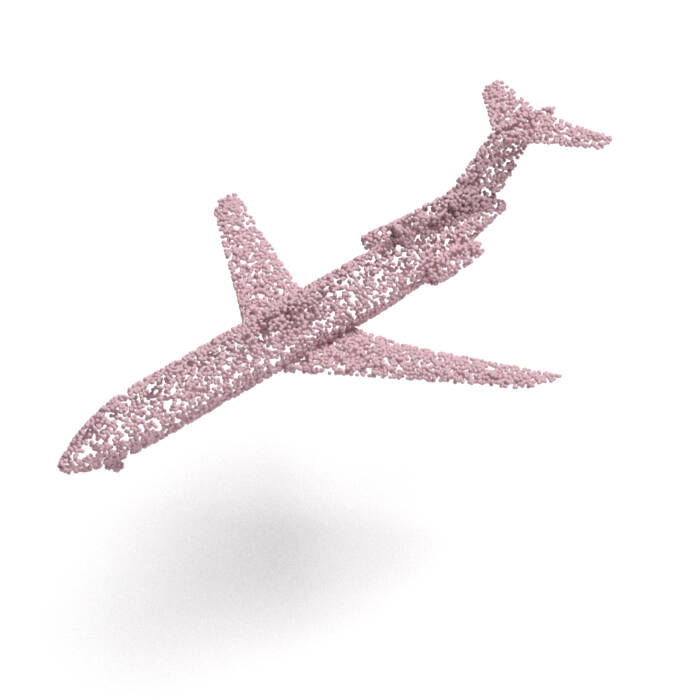}     &
    \includegraphics[width=21mm]{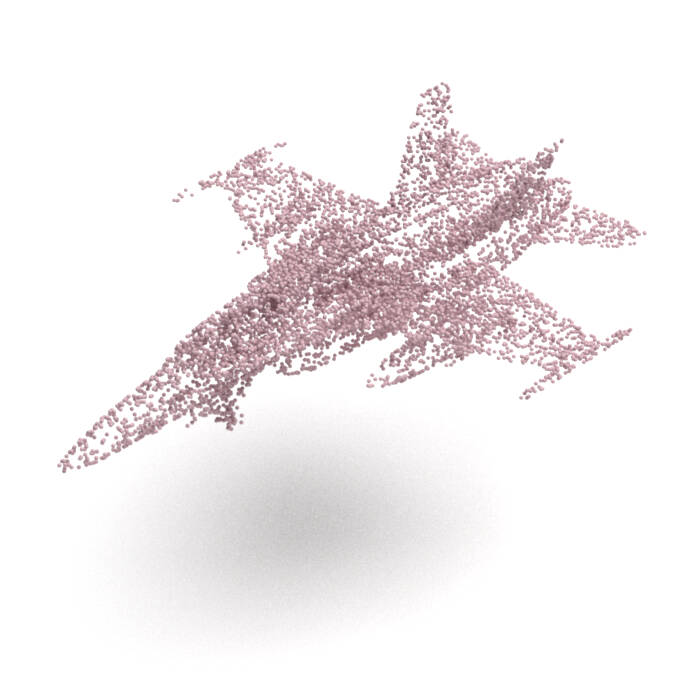}     &
    \includegraphics[width=21mm]{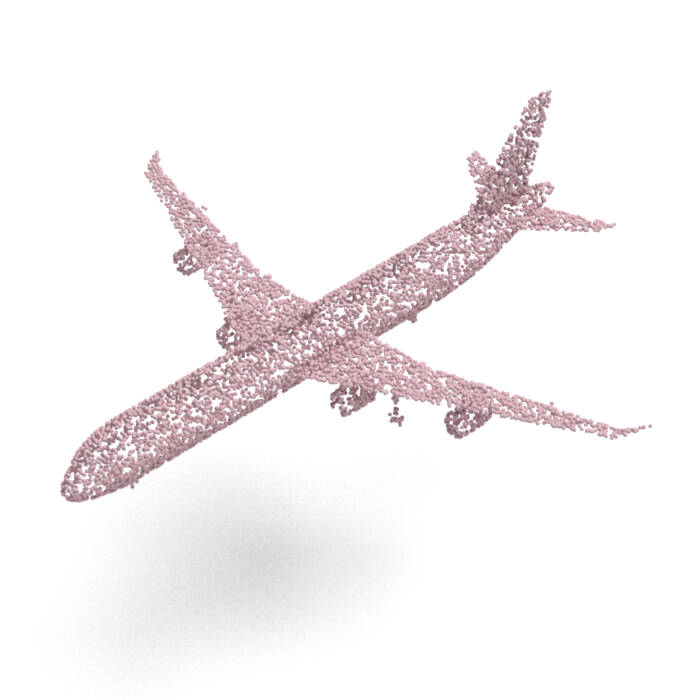}       \\
    \rotatetitle{18mm}{\textbf{ShapeGF}}                                                   &
    \includegraphics[width=21mm]{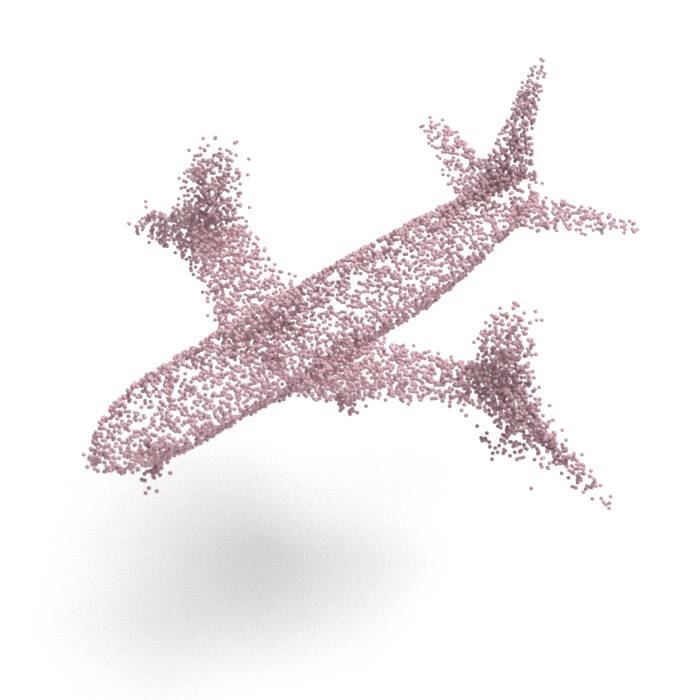} &
    \includegraphics[width=21mm]{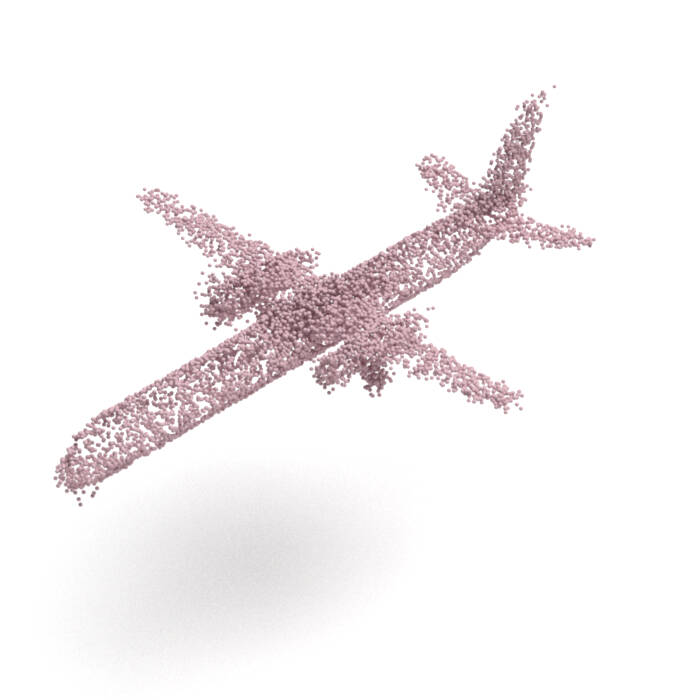} &
    \includegraphics[width=21mm]{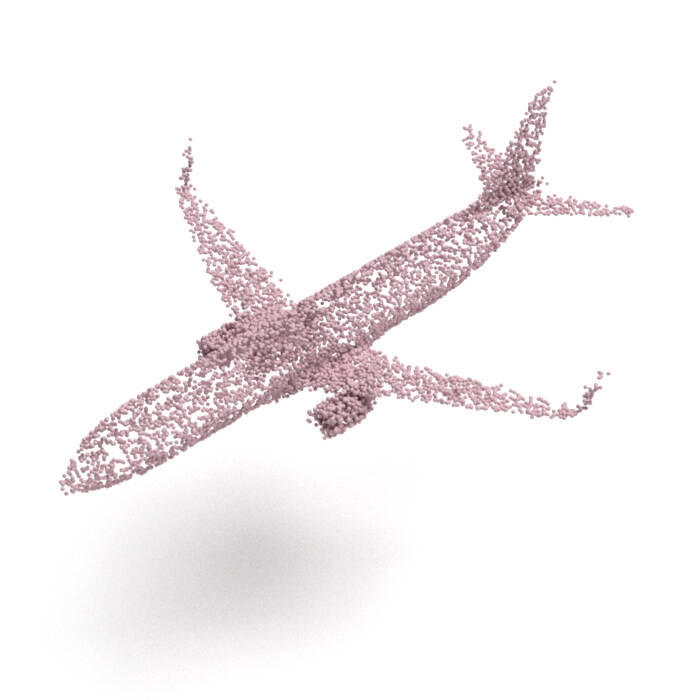} &
    \includegraphics[width=21mm]{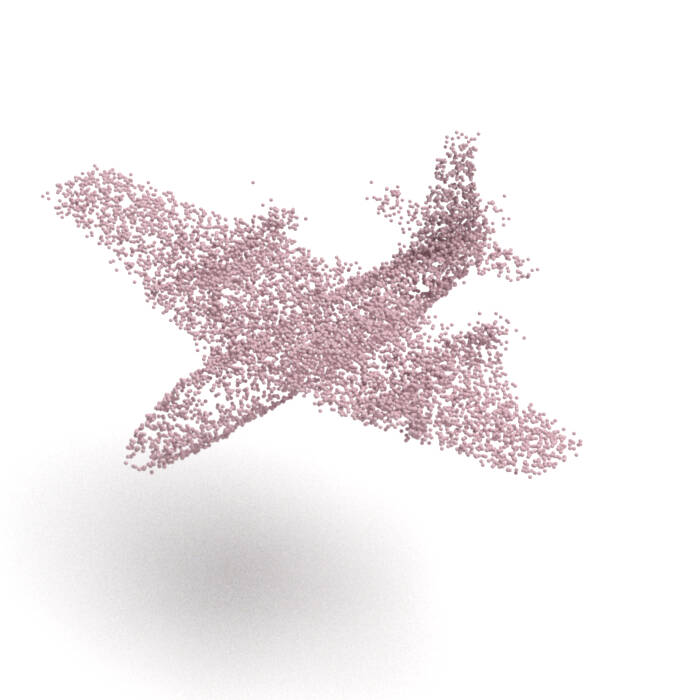} &
    \includegraphics[width=21mm]{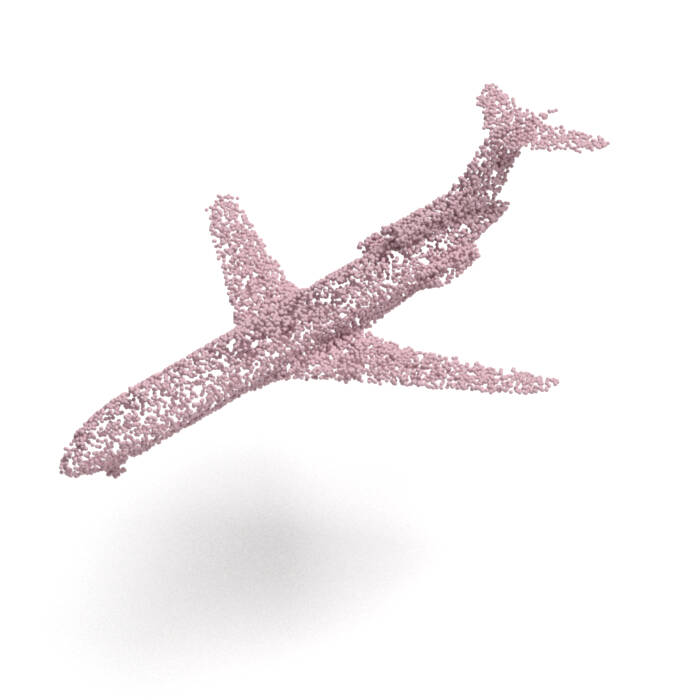} &
    \includegraphics[width=21mm]{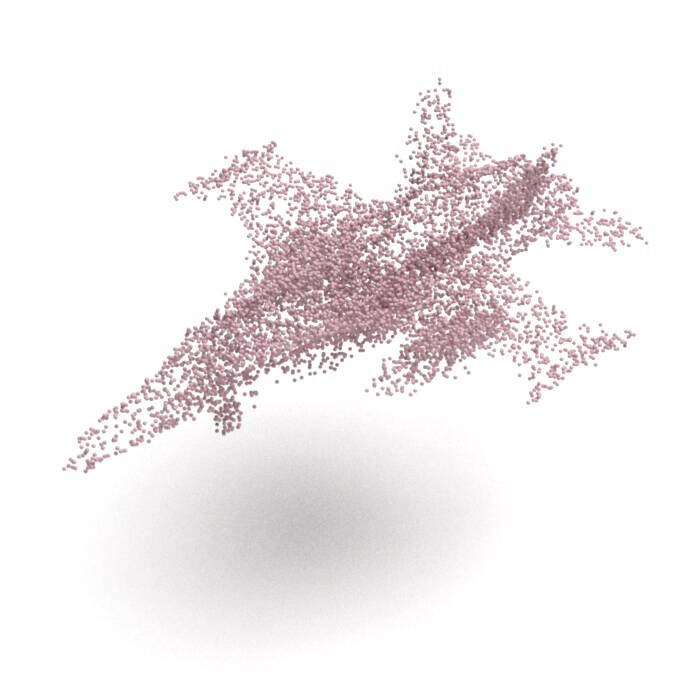} &
    \includegraphics[width=21mm]{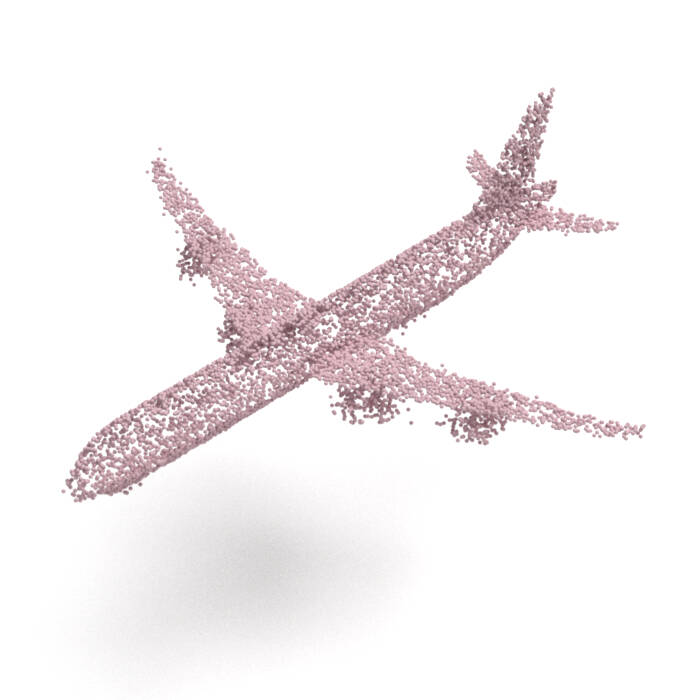}   \\
\end{tabular}
  \caption{Reconstruction examples of seen airplanes (i.e., super-resolution).}
  \label{fig:appendix_reconstruct_seen_airplane}
\end{figure*}

\begin{figure*}[t]
  \centering
  \setlength{\tabcolsep}{1mm}
  \begin{tabular}{lccccccc}
    \rotatetitle{18mm}{\textbf{Data}}                                                        &
    \includegraphics[width=21mm]{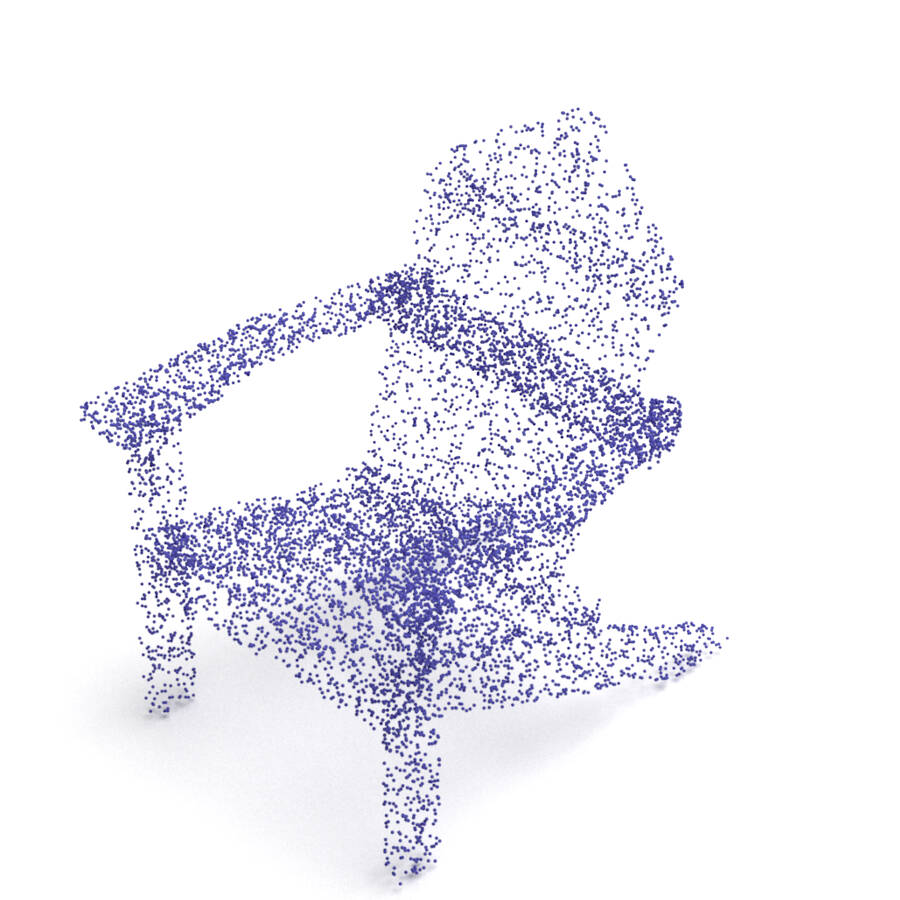}    &
    \includegraphics[width=21mm]{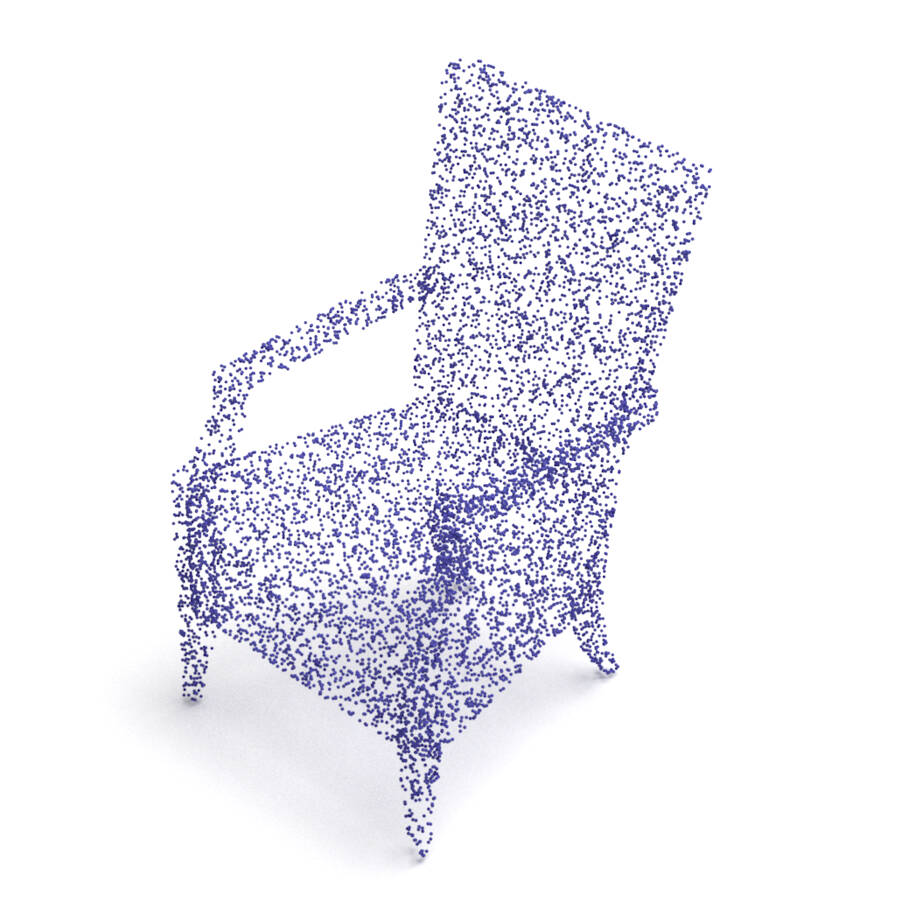}    &
    \includegraphics[width=21mm]{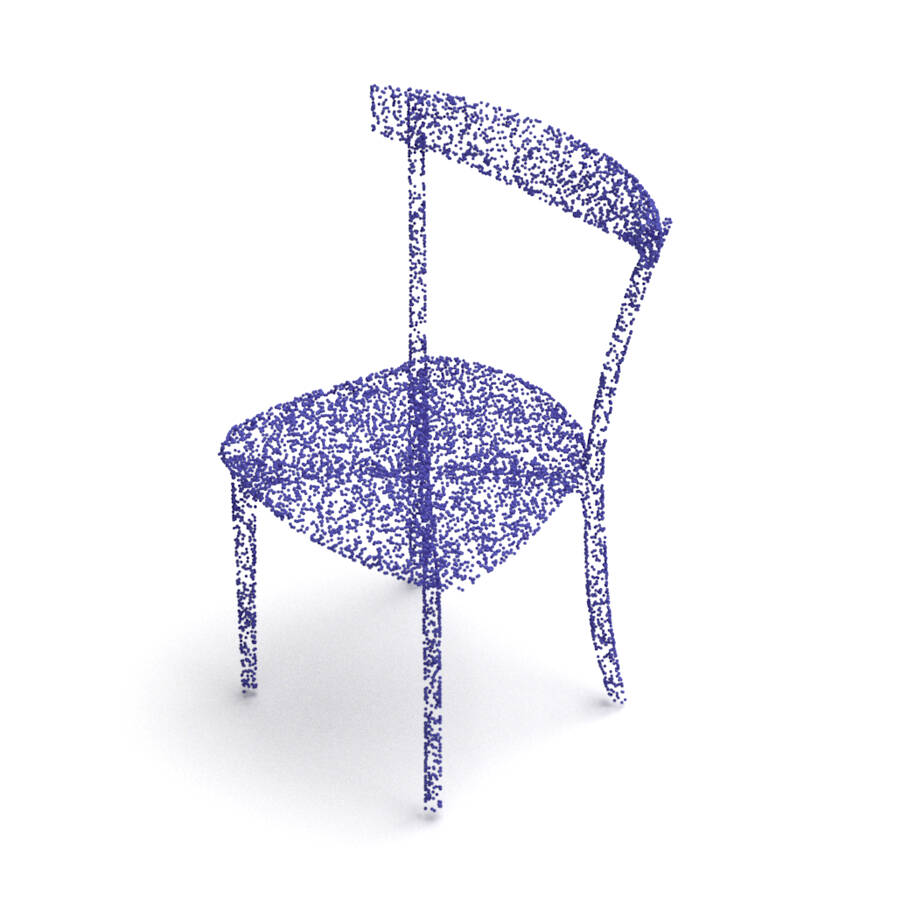}   &
    \includegraphics[width=21mm]{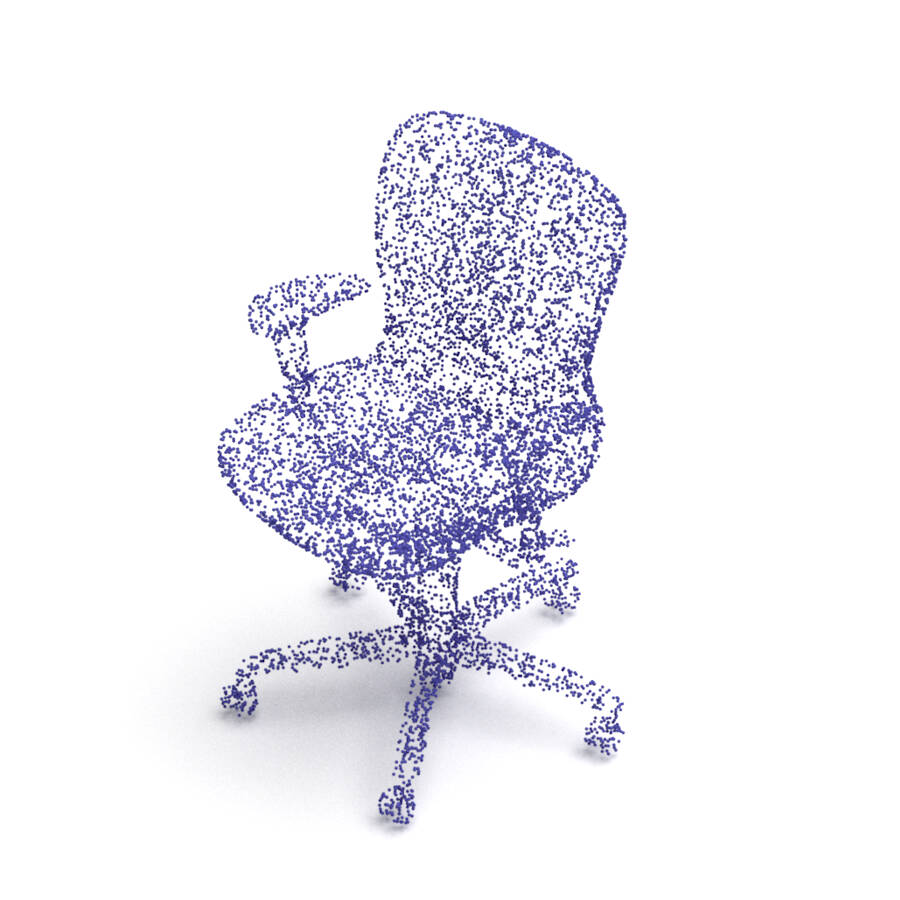}   &
    \includegraphics[width=21mm]{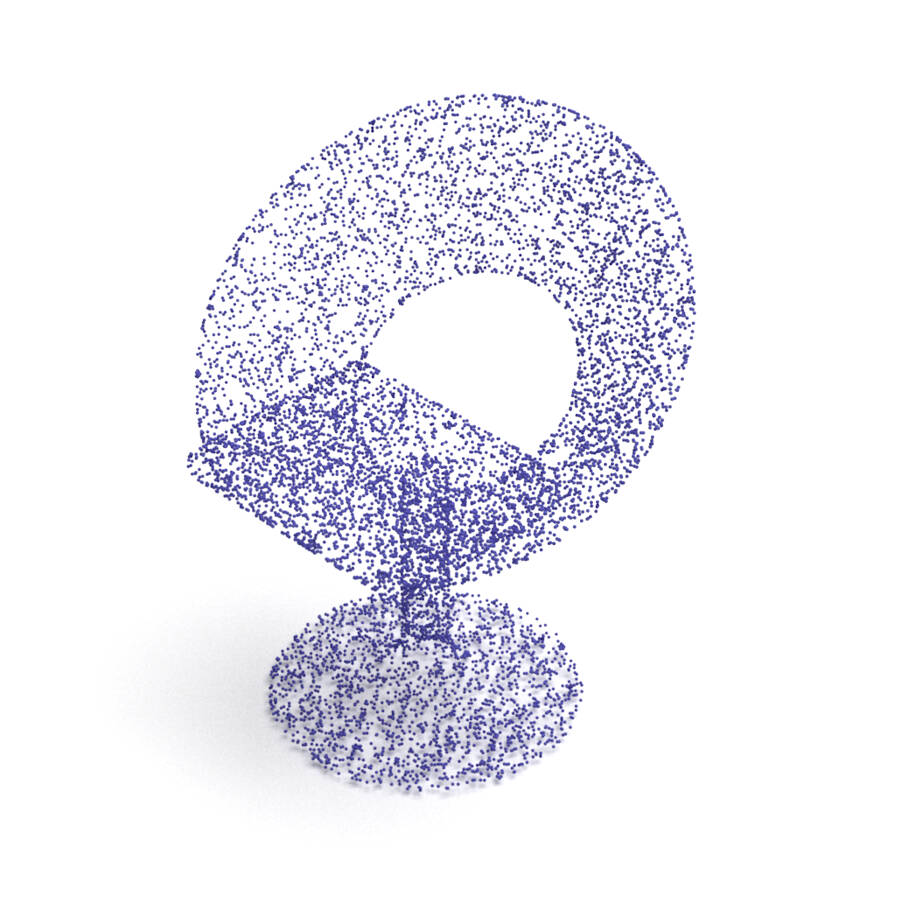}   &
    \includegraphics[width=21mm]{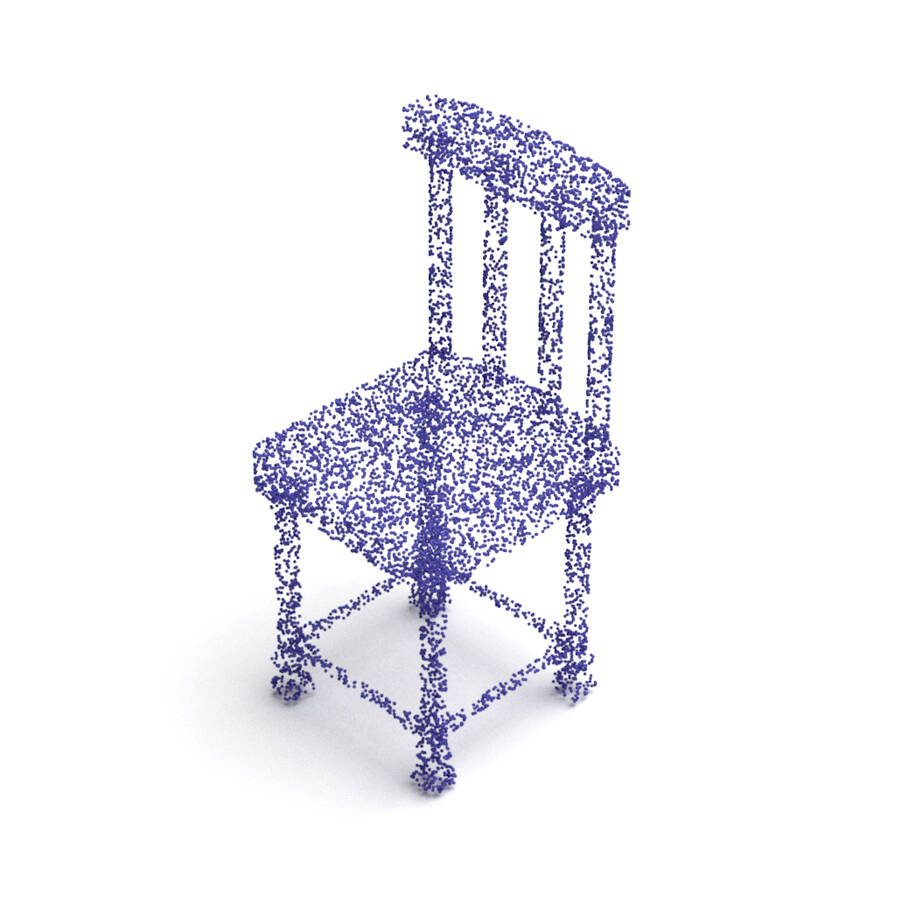}   &
    \includegraphics[width=21mm]{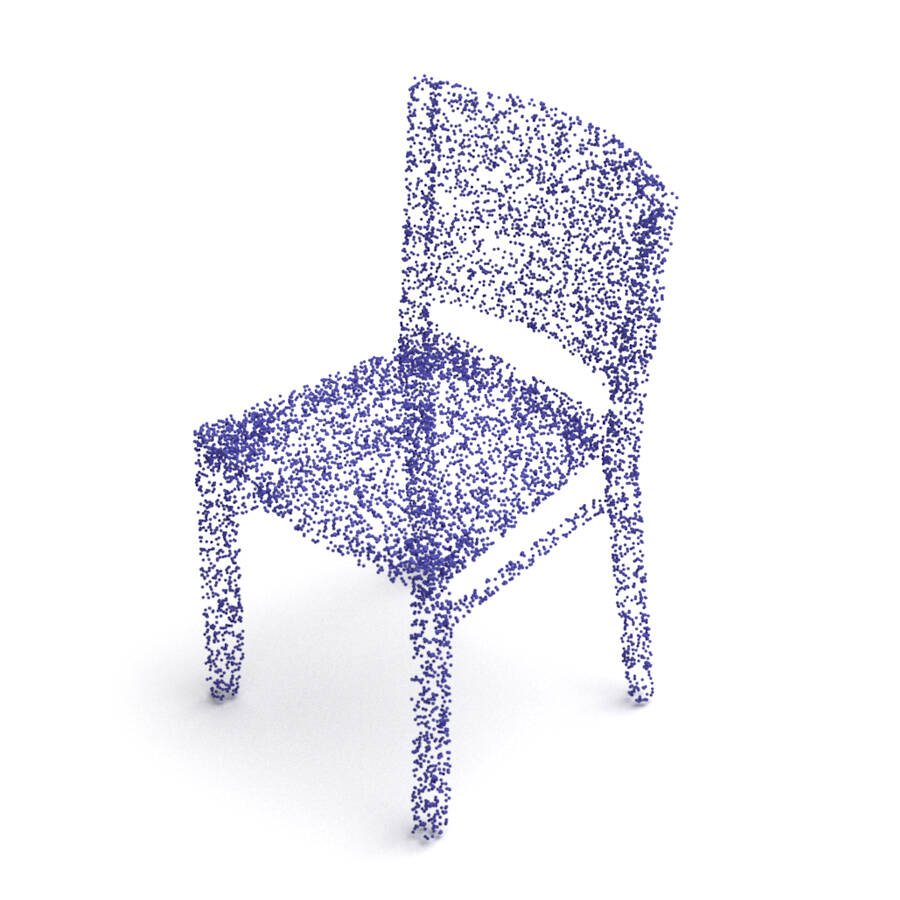}     \\
    \rotatetitle{18mm}{\textbf{ShapeGF}}                                                     &
    \includegraphics[width=21mm]{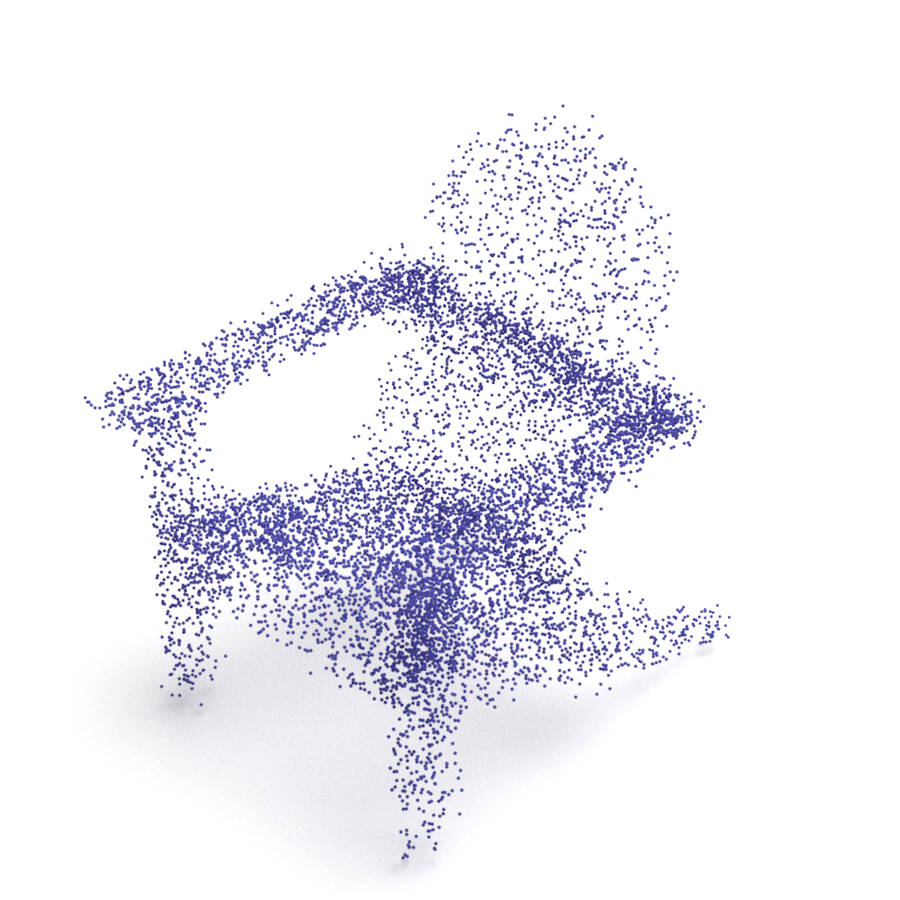} &
    \includegraphics[width=21mm]{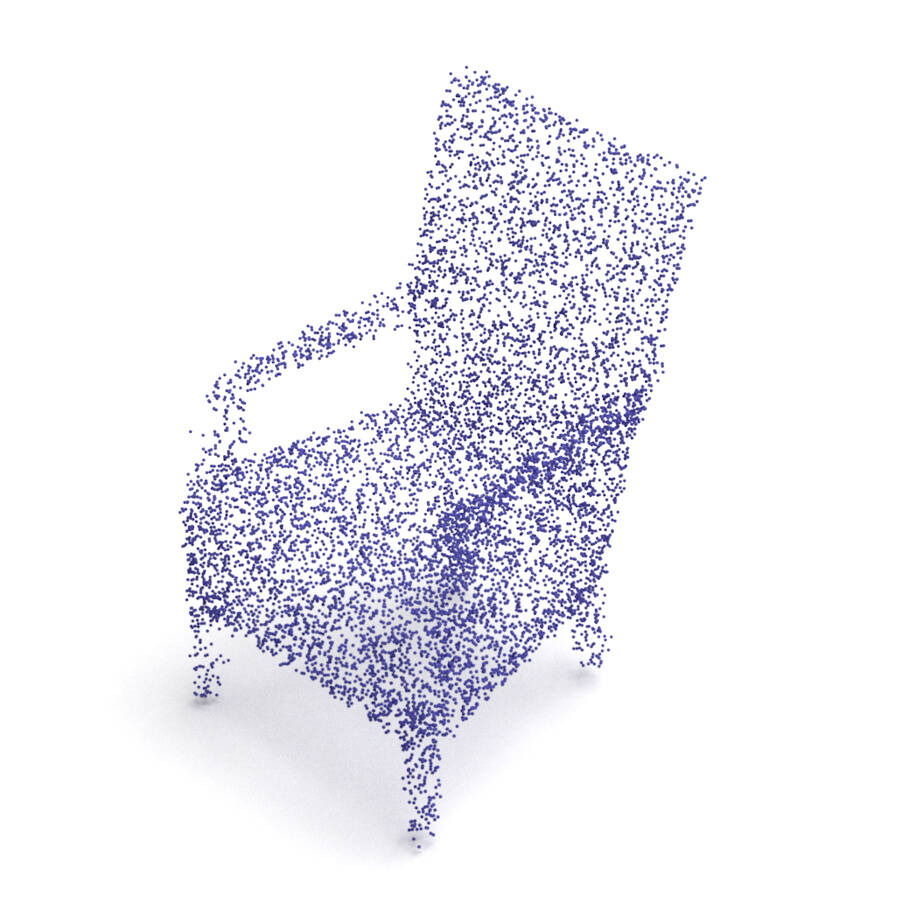} &
    \includegraphics[width=21mm]{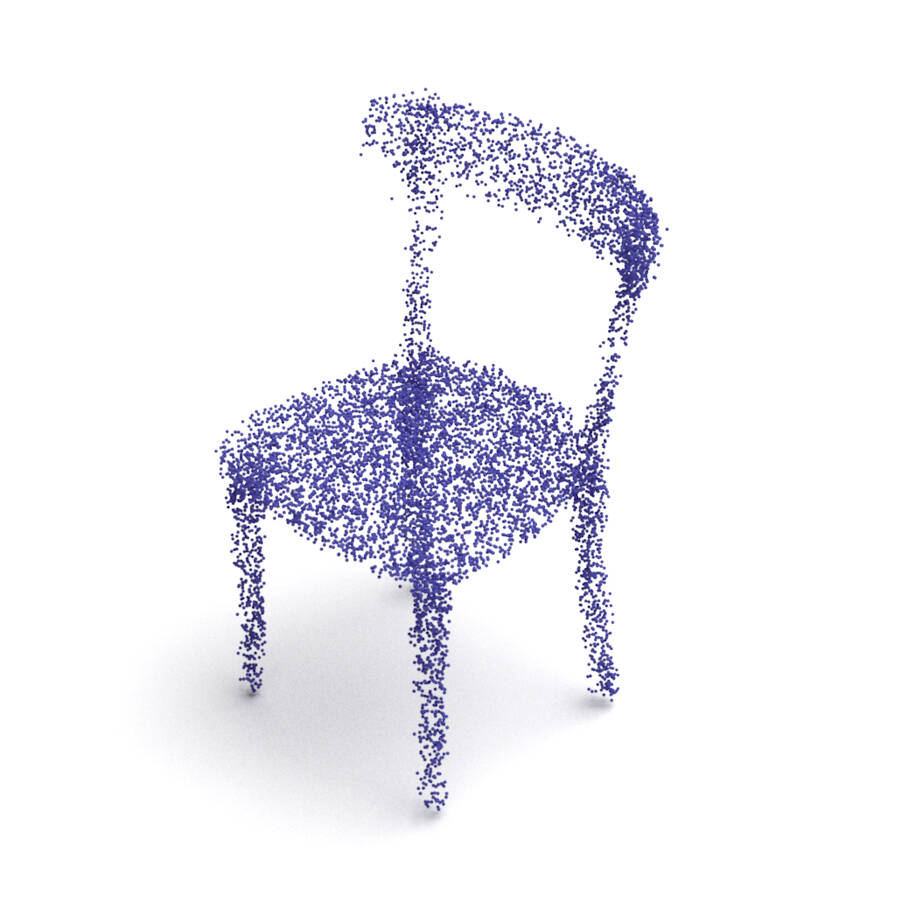} &
    \includegraphics[width=21mm]{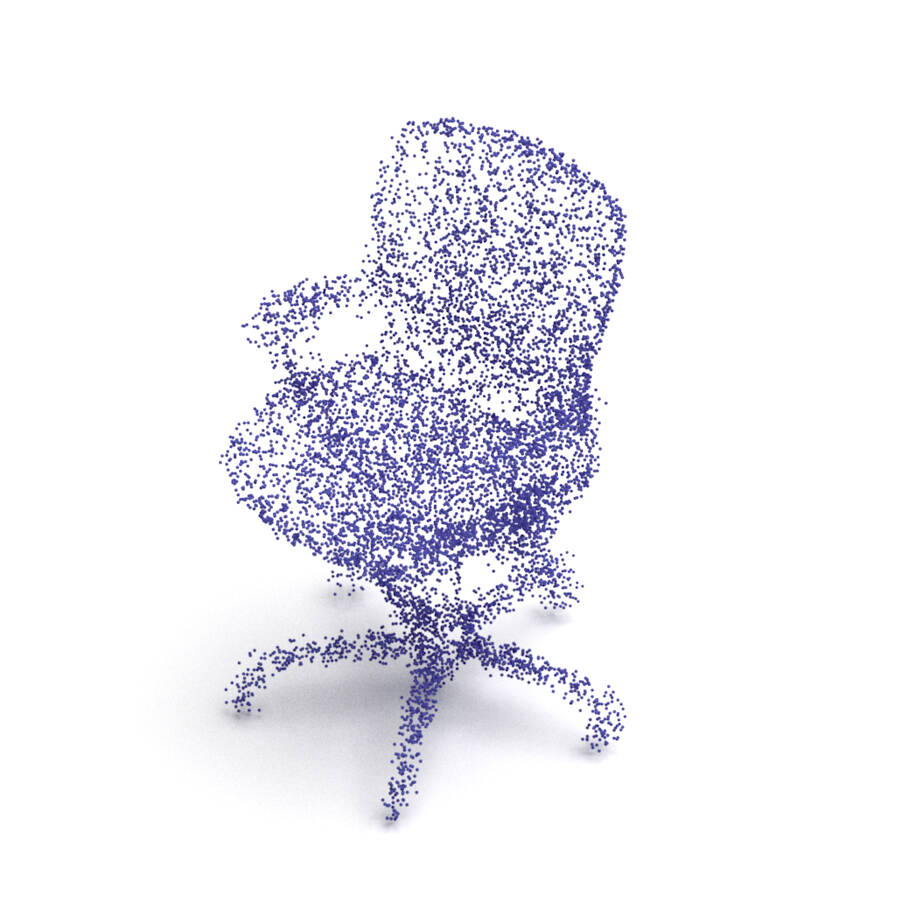} &
    \includegraphics[width=21mm]{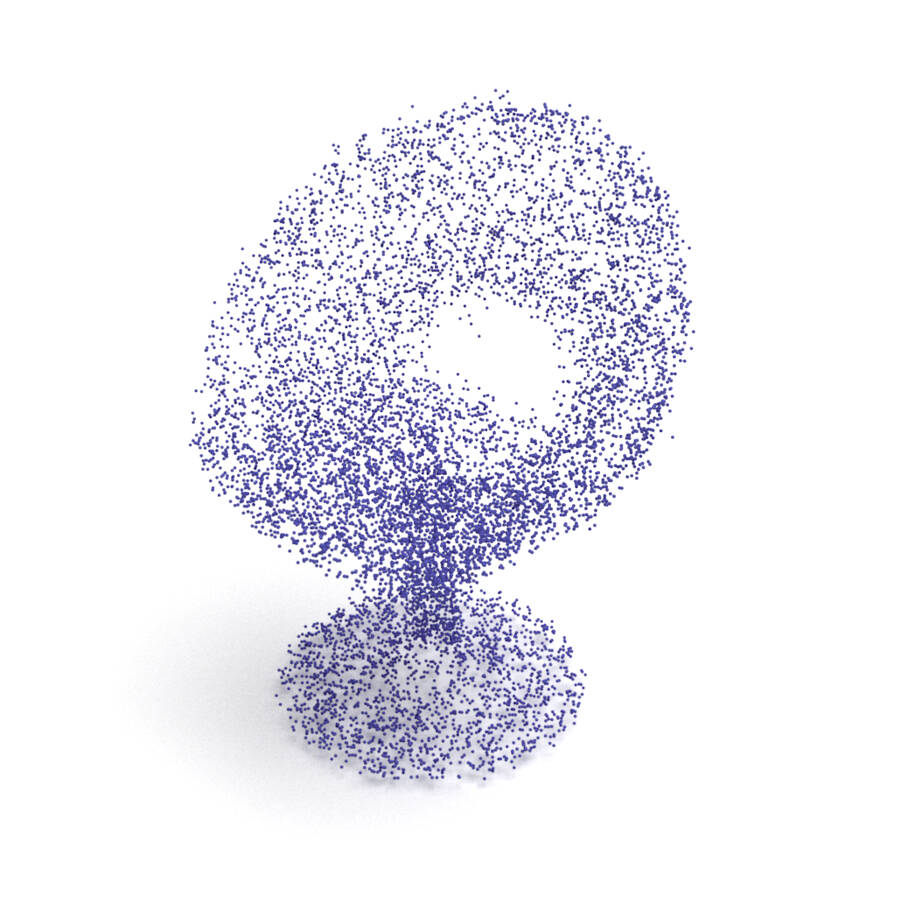} &
    \includegraphics[width=21mm]{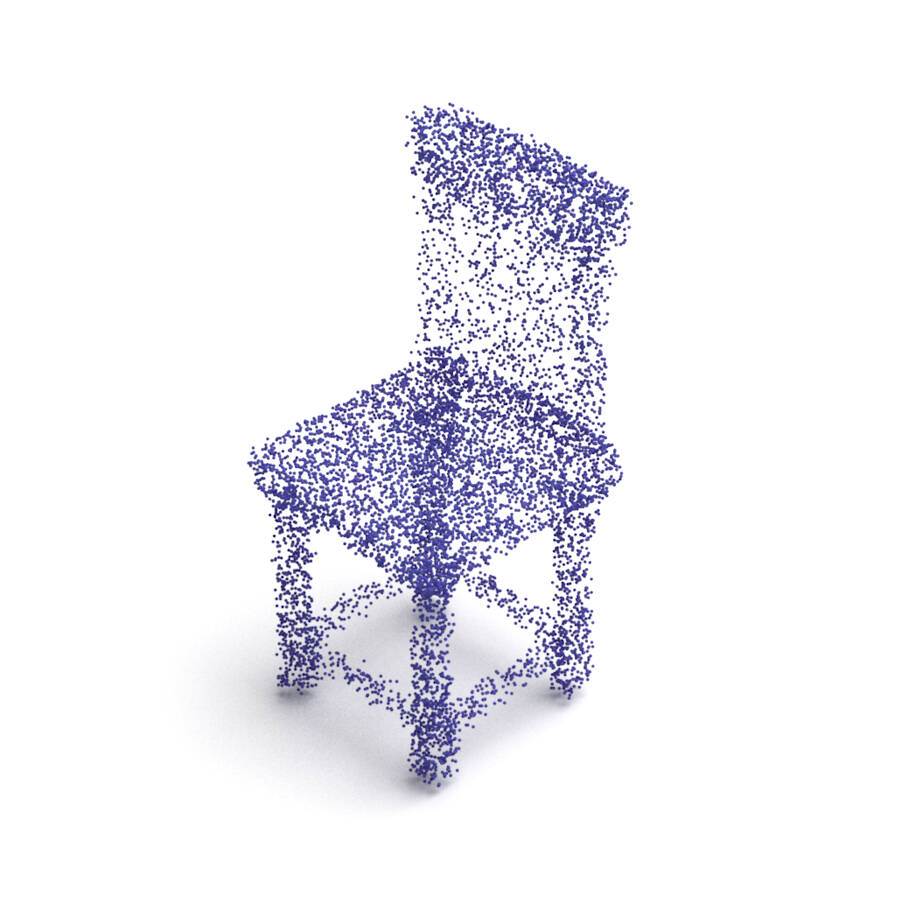} &
    \includegraphics[width=21mm]{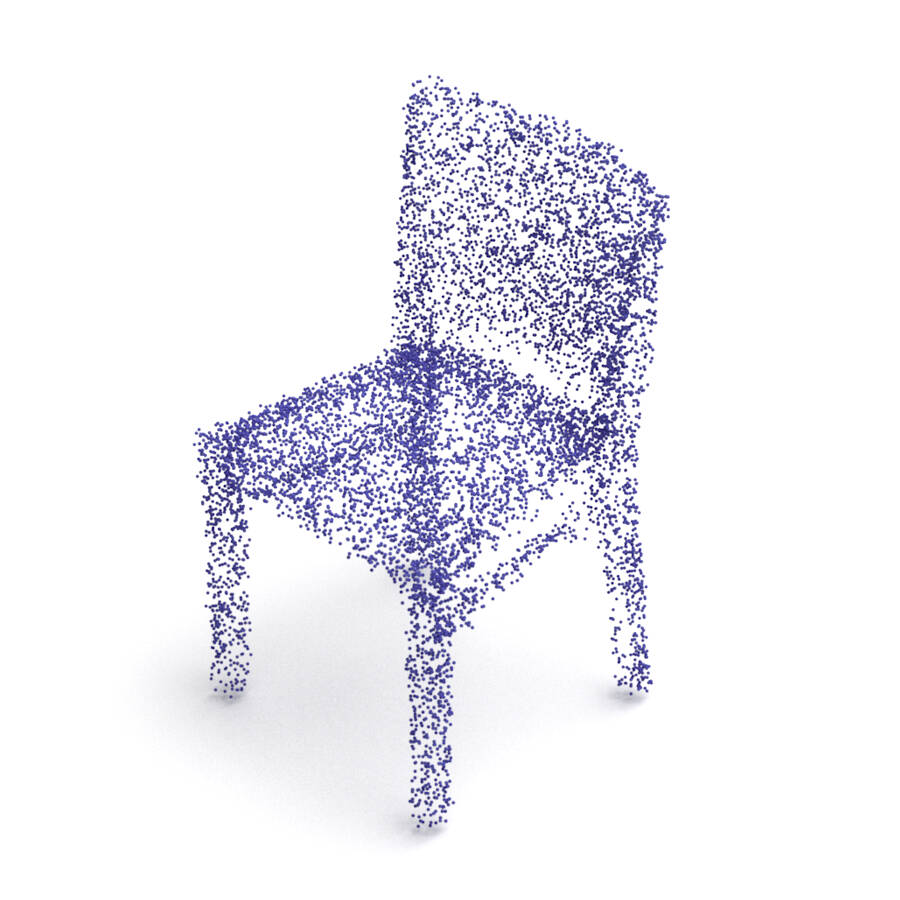}   \\
\end{tabular}
  \caption{Reconstruction examples of seen chairs (i.e., super-resolution).}
  \label{fig:appendix_reconstruct_seen_chair}
\end{figure*}

\begin{figure*}[t]
  \centering
  \setlength{\tabcolsep}{1mm}
  \begin{tabular}{lccccccc}
    \rotatetitle{18mm}{\textbf{Data}}                                                      &
    \includegraphics[width=21mm]{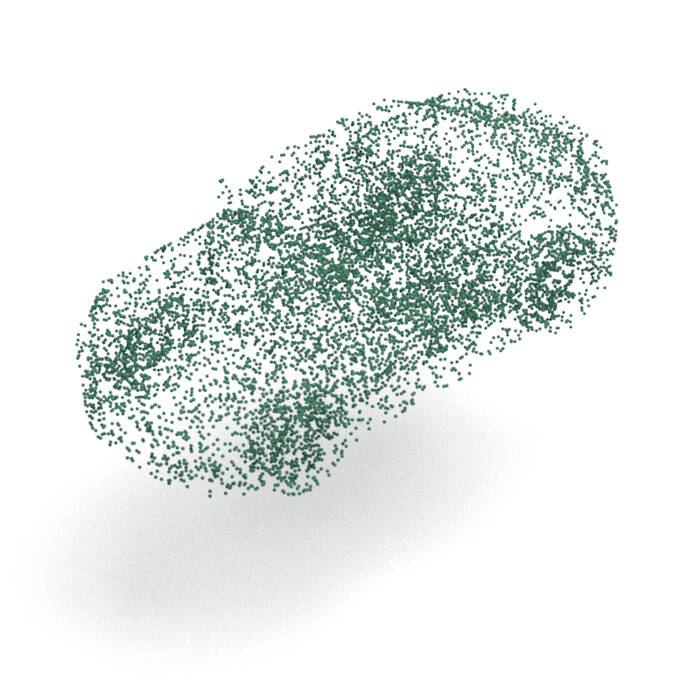}      &
    \includegraphics[width=21mm]{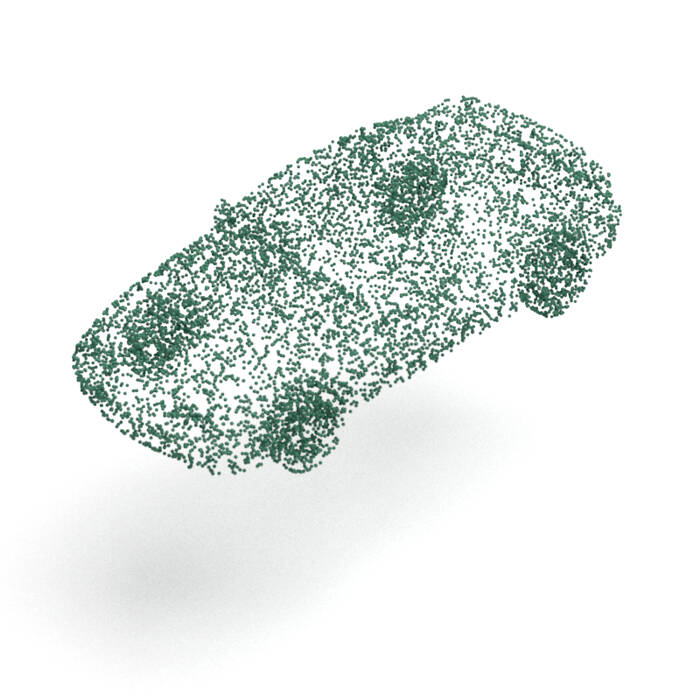}     &
    \includegraphics[width=21mm]{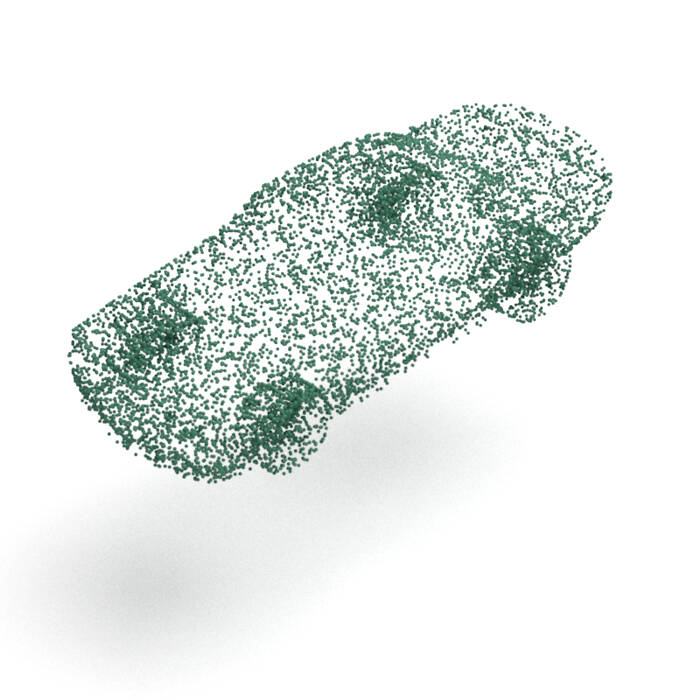}     &
    \includegraphics[width=21mm]{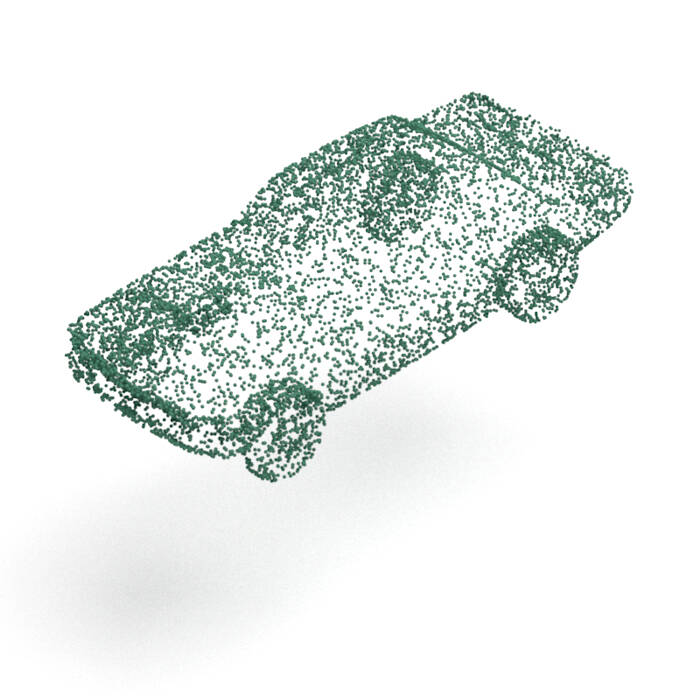}     &
    \includegraphics[width=21mm]{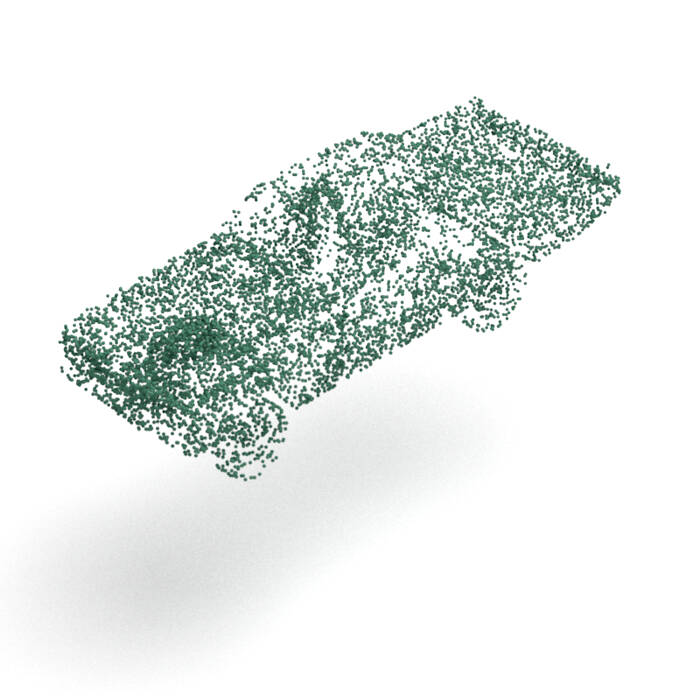}    &
    \includegraphics[width=21mm]{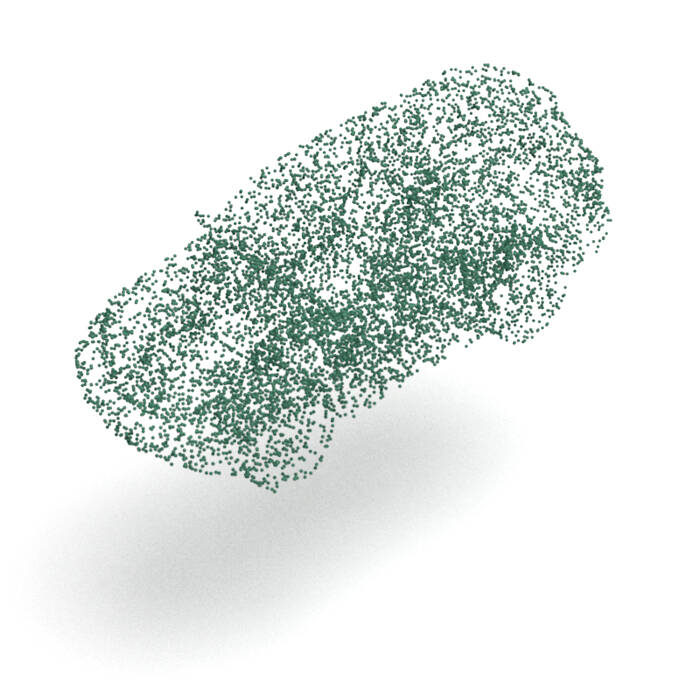}    &
    \includegraphics[width=21mm]{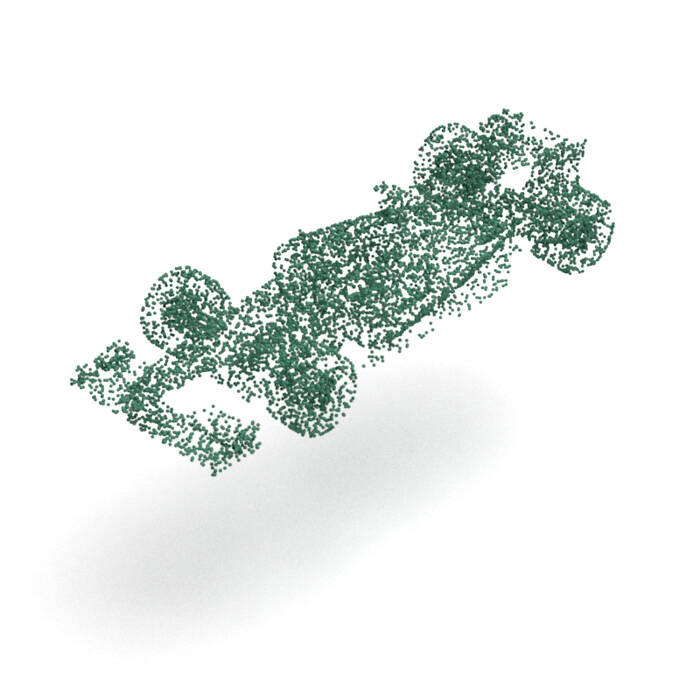}      \\
    \rotatetitle{18mm}{\textbf{ShapeGF}}                                                   &
    \includegraphics[width=21mm]{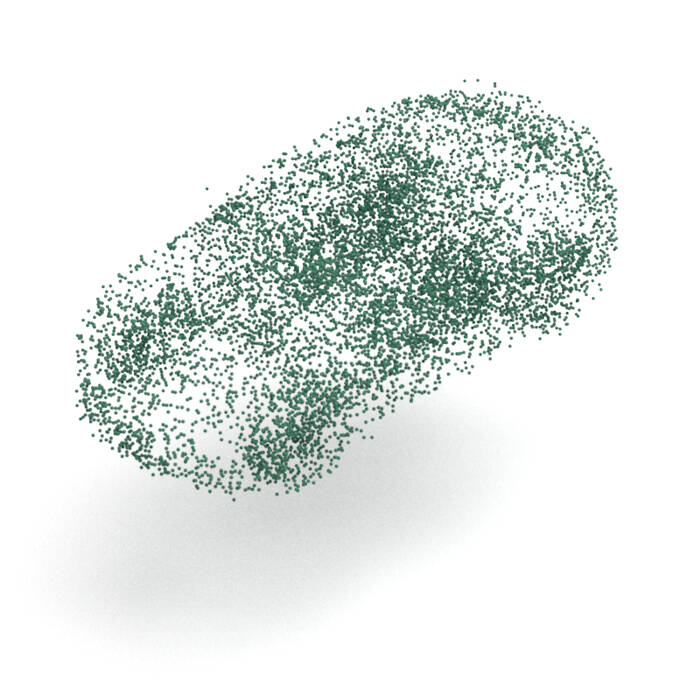} &
    \includegraphics[width=21mm]{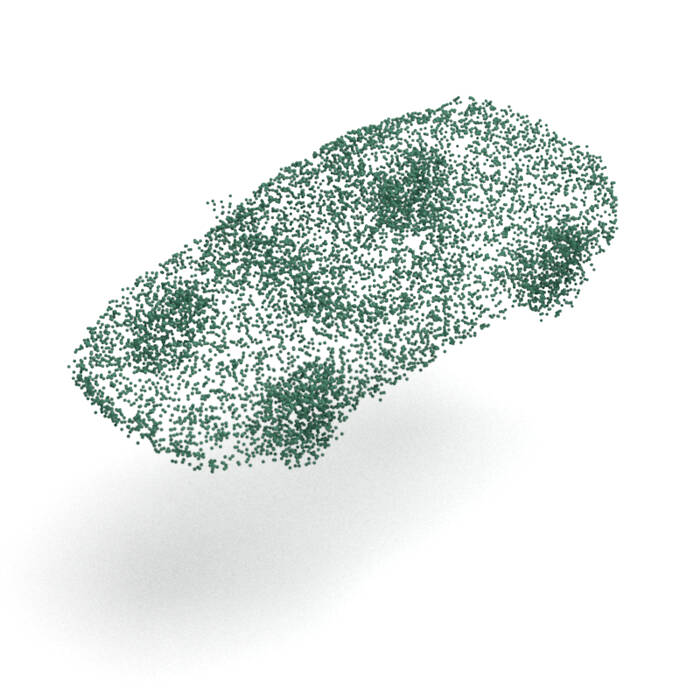} &
    \includegraphics[width=21mm]{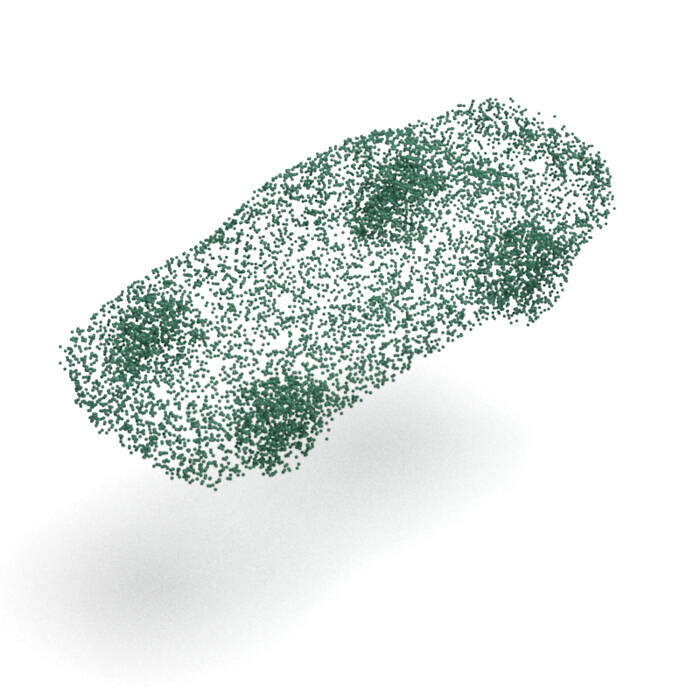} &
    \includegraphics[width=21mm]{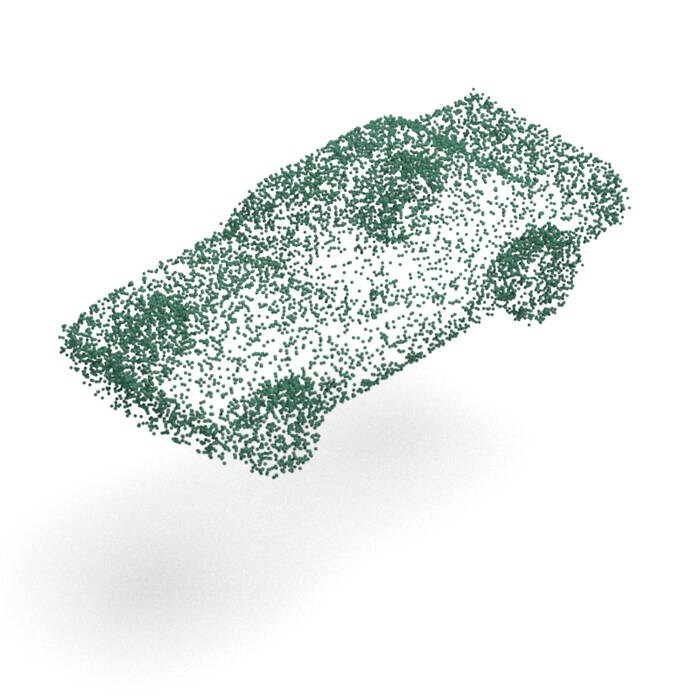} &
    \includegraphics[width=21mm]{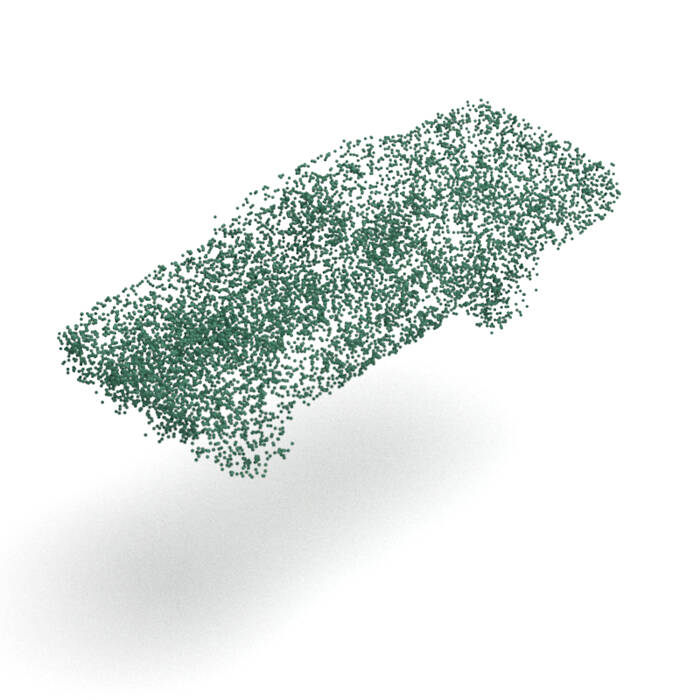} &
    \includegraphics[width=21mm]{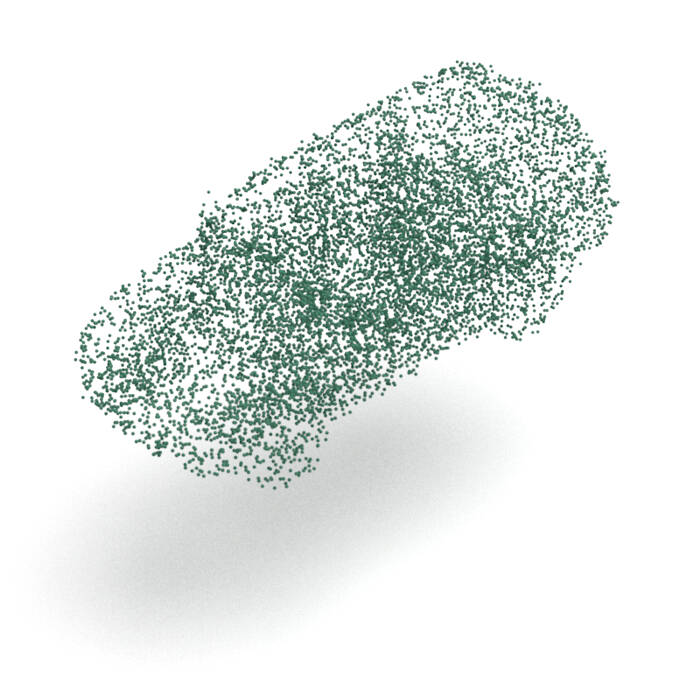} &
    \includegraphics[width=21mm]{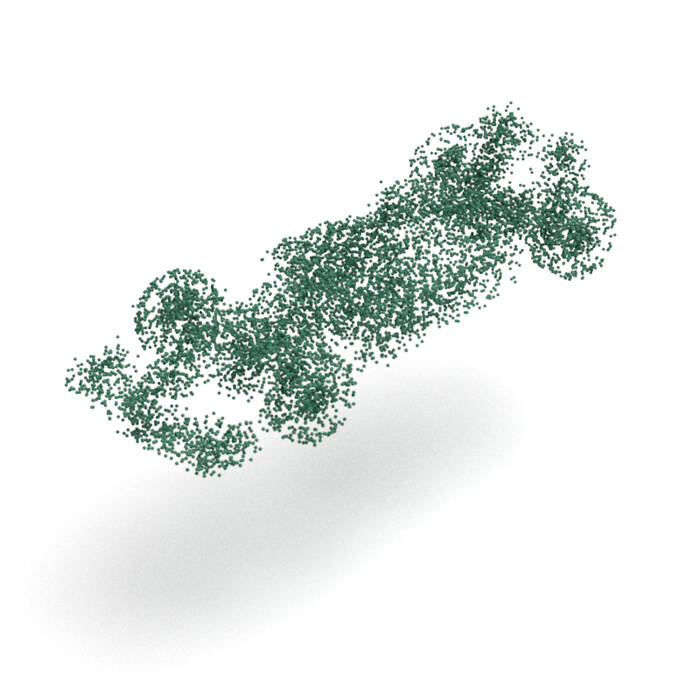}   \\
\end{tabular}
  \caption{Reconstruction examples of seen cars (i.e., super-resolution).}
  \label{fig:appendix_reconstruct_seen_car}
\end{figure*}

\begin{figure*}[t]
  \centering
  \setlength{\tabcolsep}{1mm}
  \begin{tabular}{lccccccc}
    \rotatetitle{18mm}{\textbf{Data}}                                                         &
    \includegraphics[width=21mm]{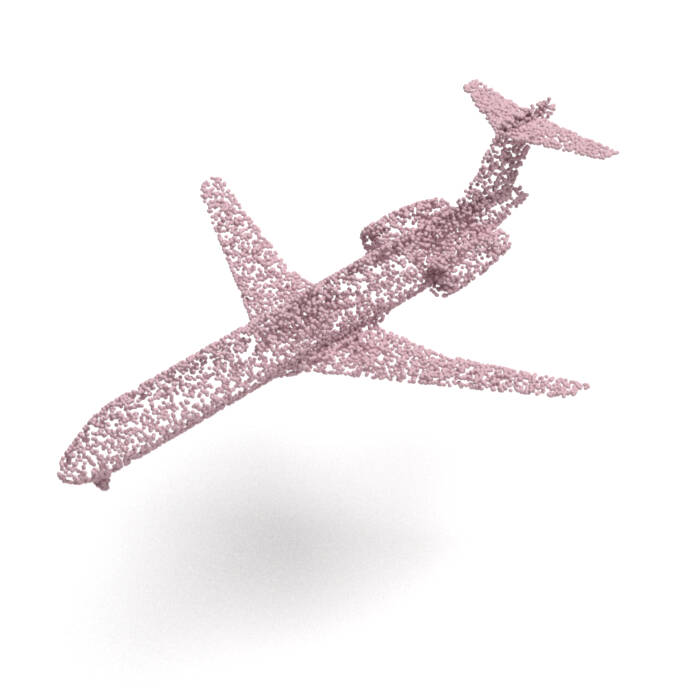}                &
    \includegraphics[width=21mm]{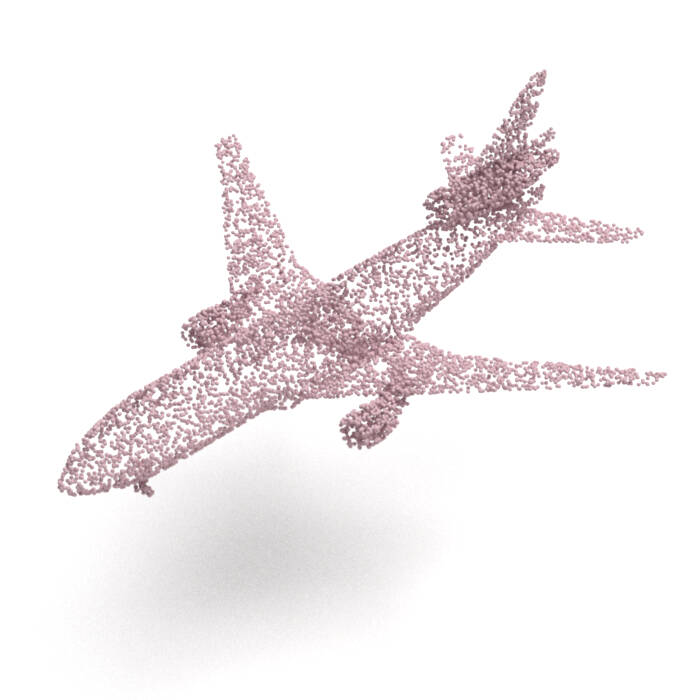}                &
    \includegraphics[width=21mm]{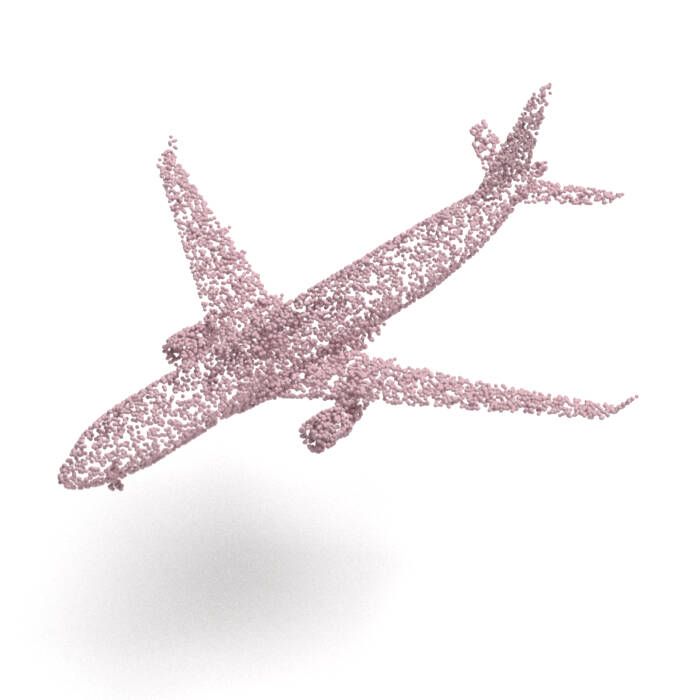}                &
    \includegraphics[width=21mm]{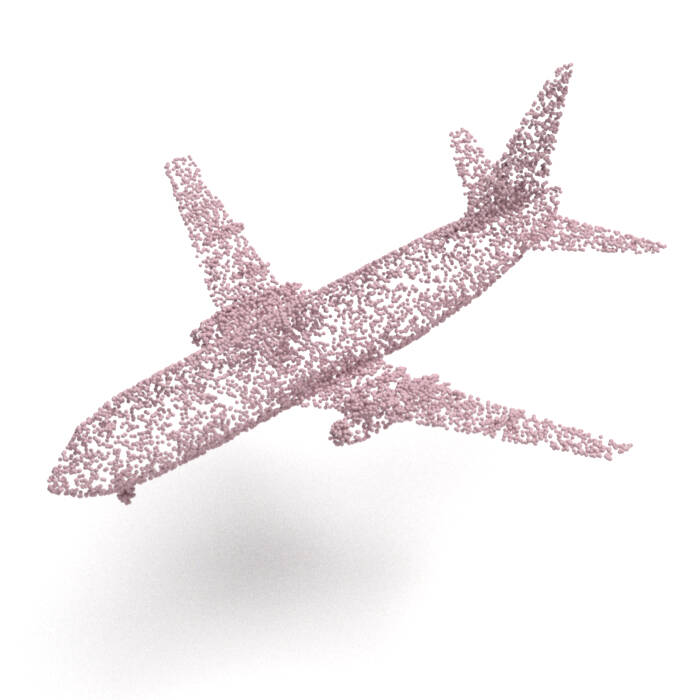}               &
    \includegraphics[width=21mm]{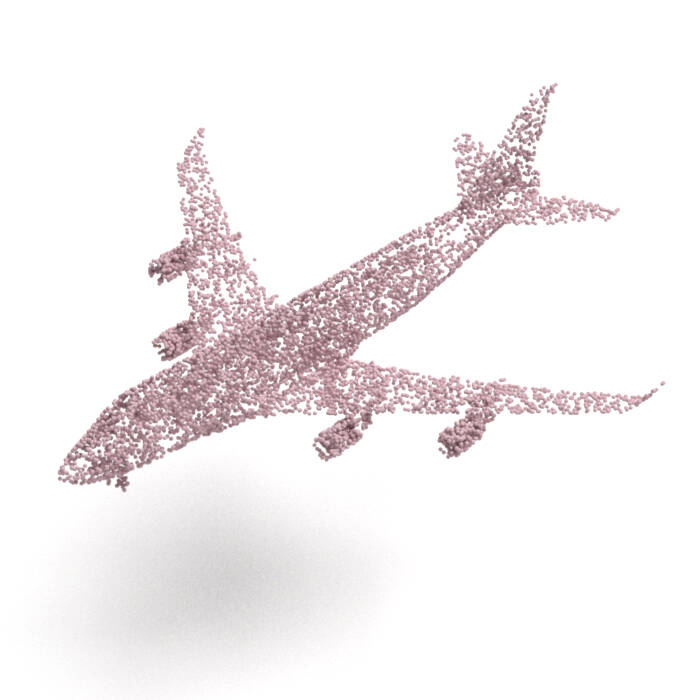}               &
    \includegraphics[width=21mm]{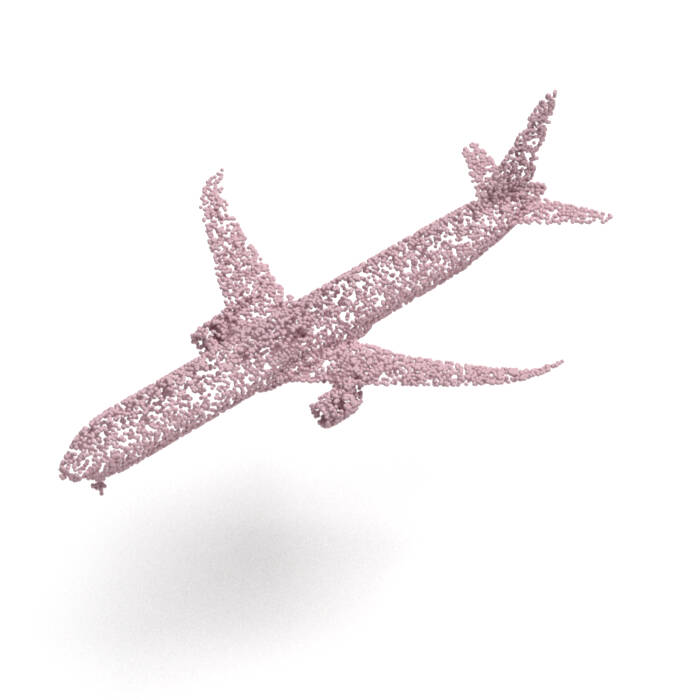}               &
    \includegraphics[width=21mm]{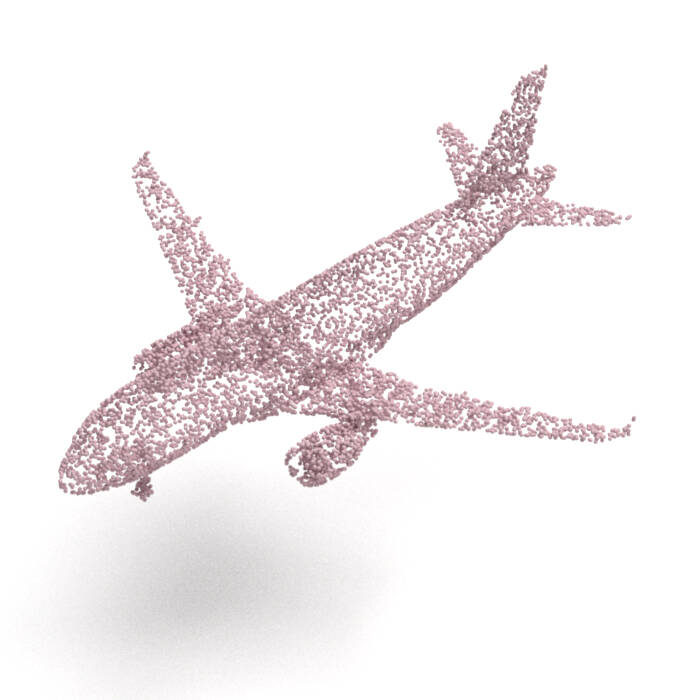}                 \\
    \rotatetitle{18mm}{\textbf{ShapeGF}}                                                      &
    \includegraphics[width=21mm]{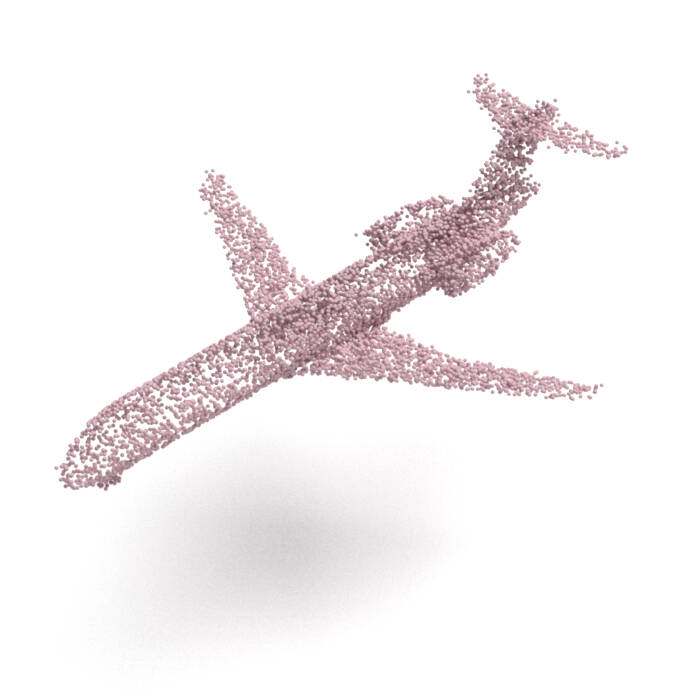}  &
    \includegraphics[width=21mm]{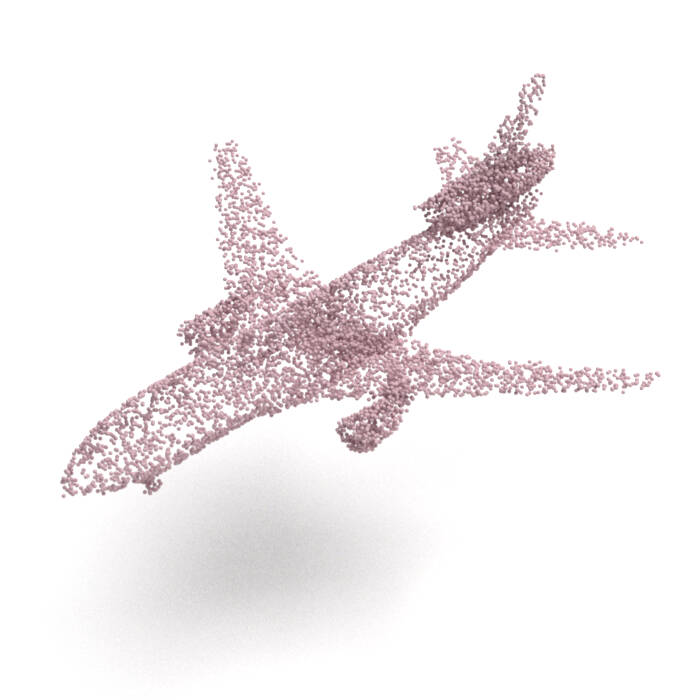}  &
    \includegraphics[width=21mm]{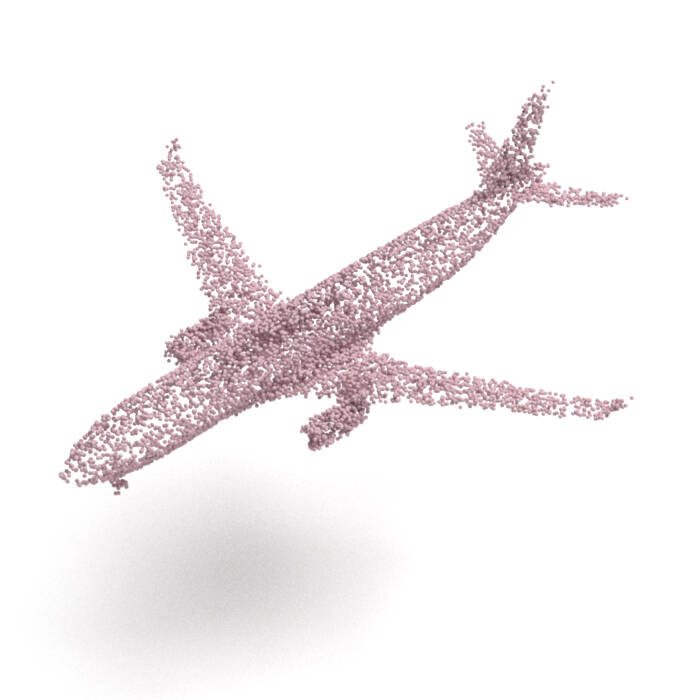} &
    \includegraphics[width=21mm]{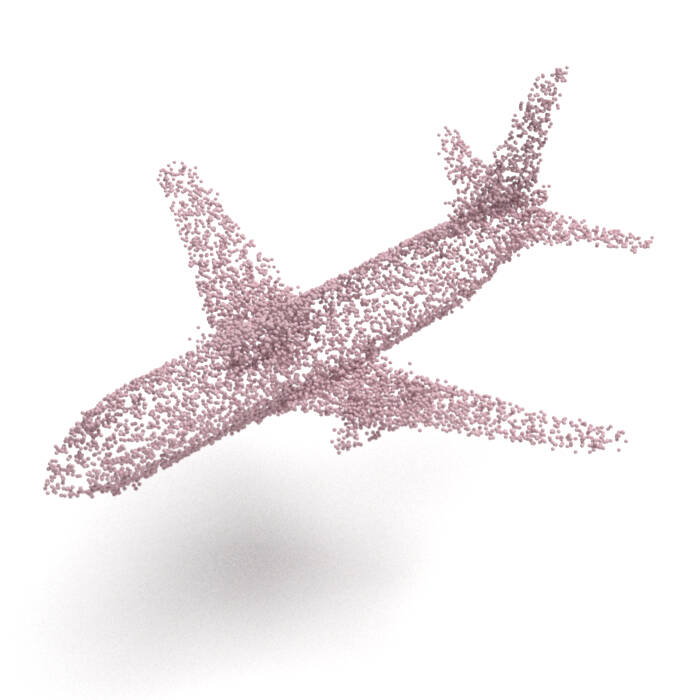} &
    \includegraphics[width=21mm]{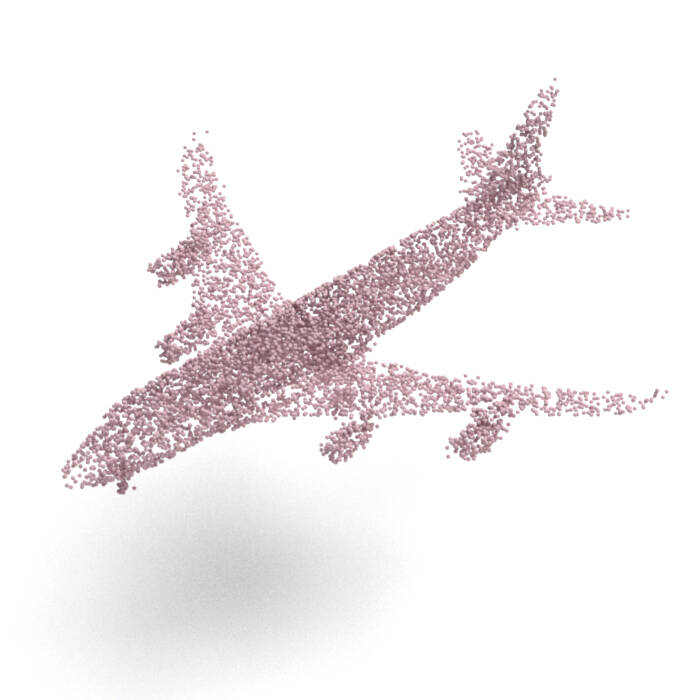} &
    \includegraphics[width=21mm]{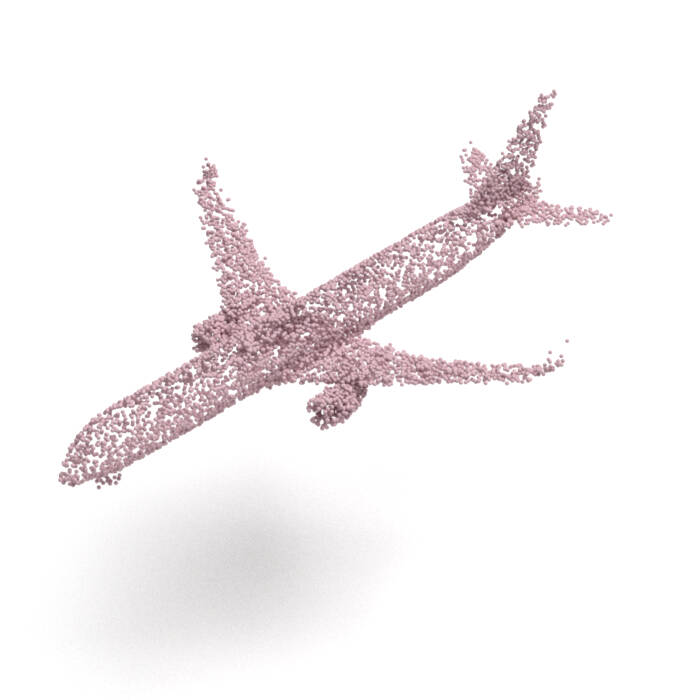} &
    \includegraphics[width=21mm]{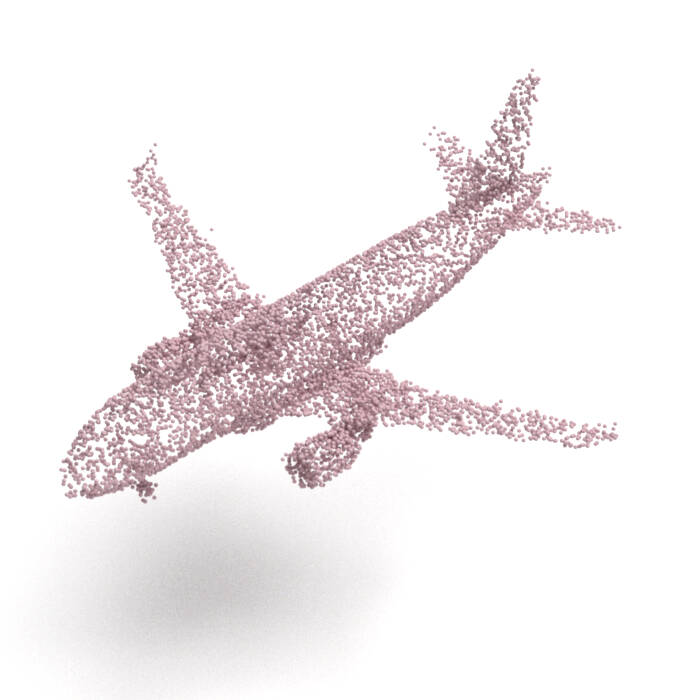}   \\
\end{tabular}
  \caption{Reconstruction examples of unseen airplanes.}
  \label{fig:appendix_reconstruct_unseen_airplane}
\end{figure*}

\begin{figure*}[t]
  \centering
  \setlength{\tabcolsep}{1mm}
  \begin{tabular}{lccccccc}
    \rotatetitle{18mm}{\textbf{Data}}                                                           &
    \includegraphics[width=21mm]{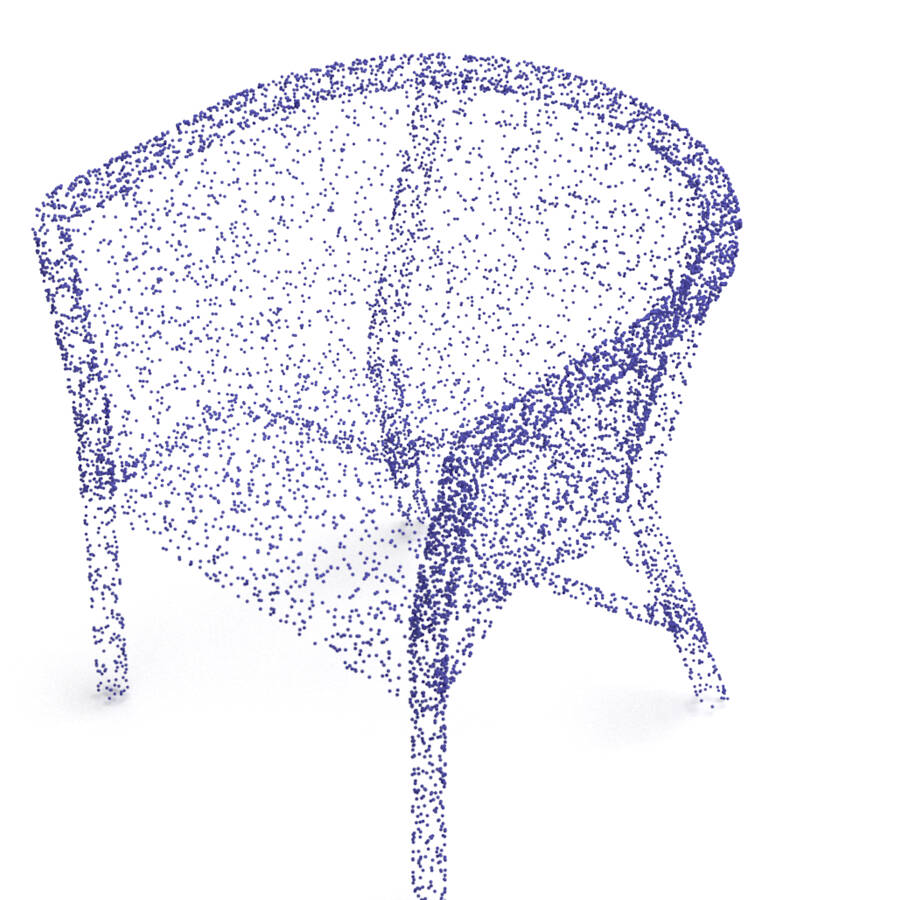}                &
    \includegraphics[width=21mm]{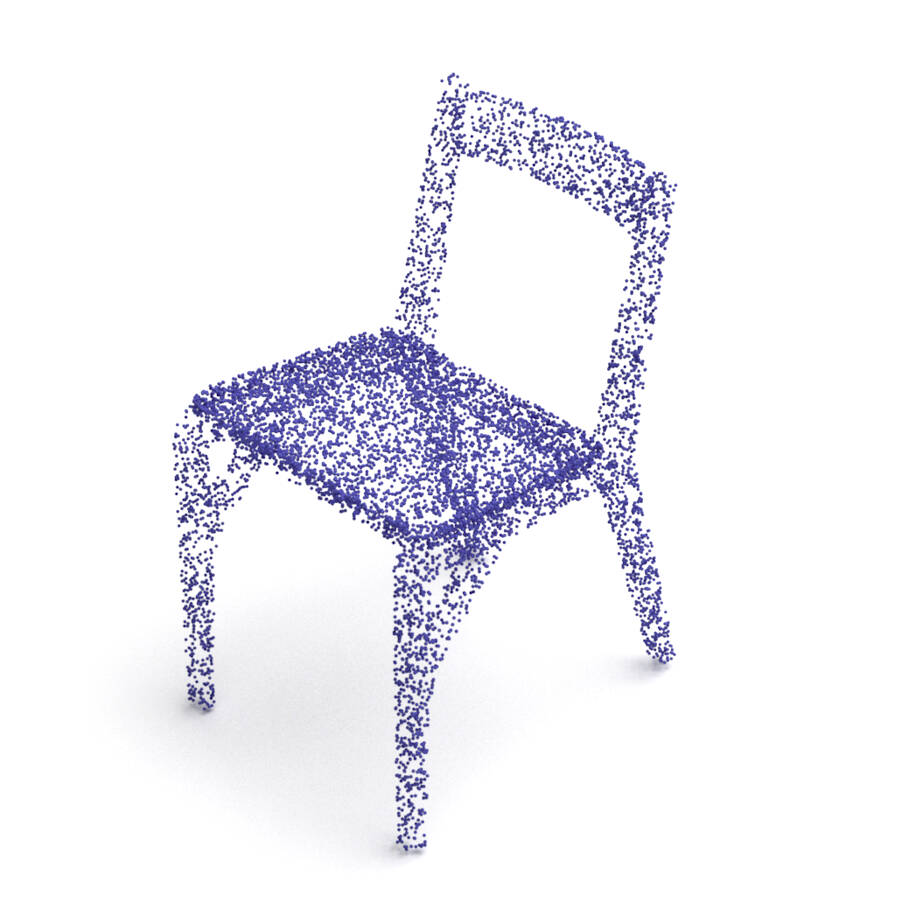}                &
    \includegraphics[width=21mm]{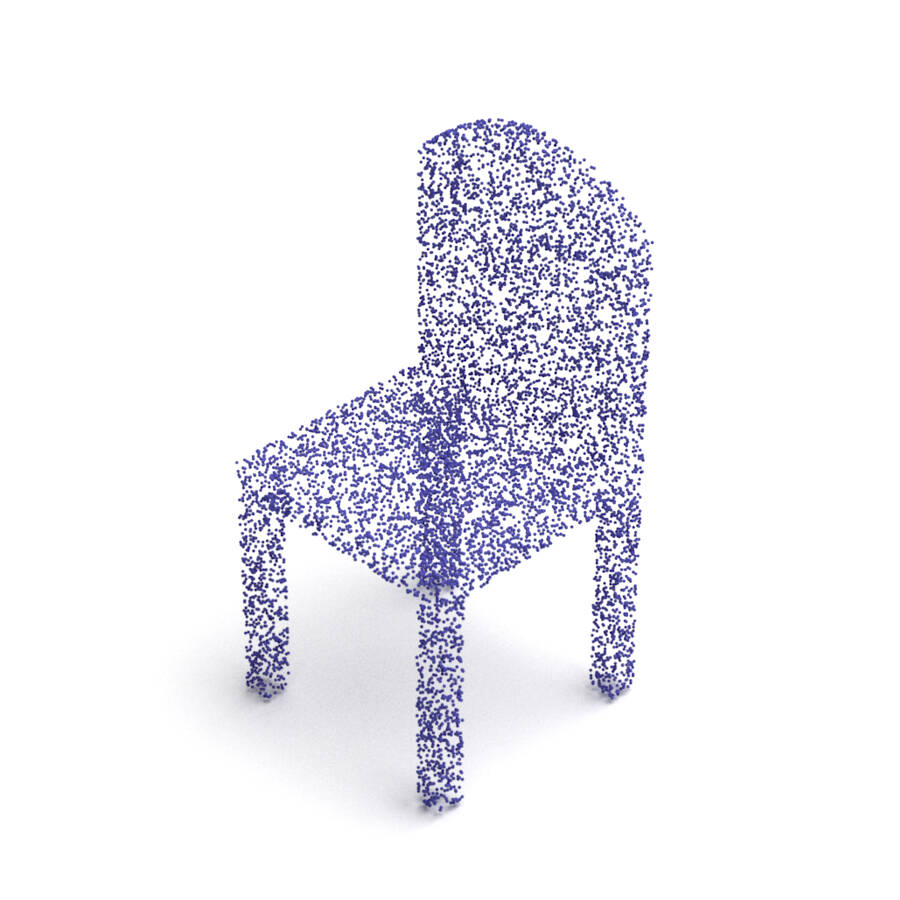}                &
    \includegraphics[width=21mm]{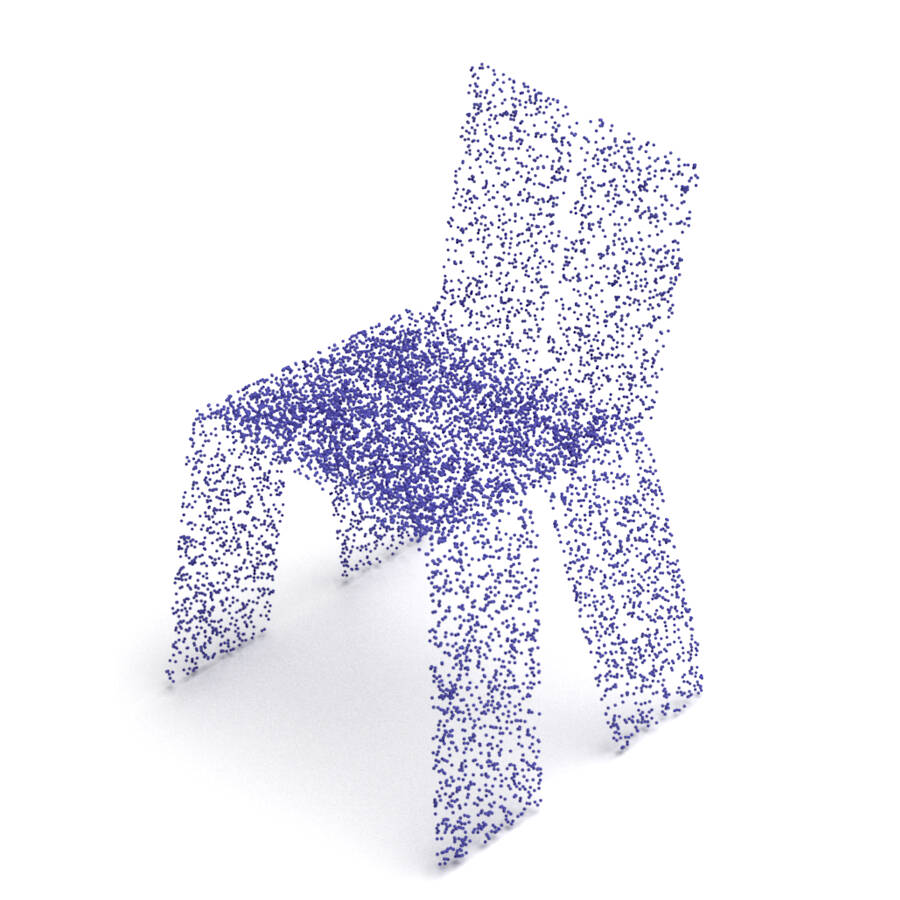}               &
    \includegraphics[width=21mm]{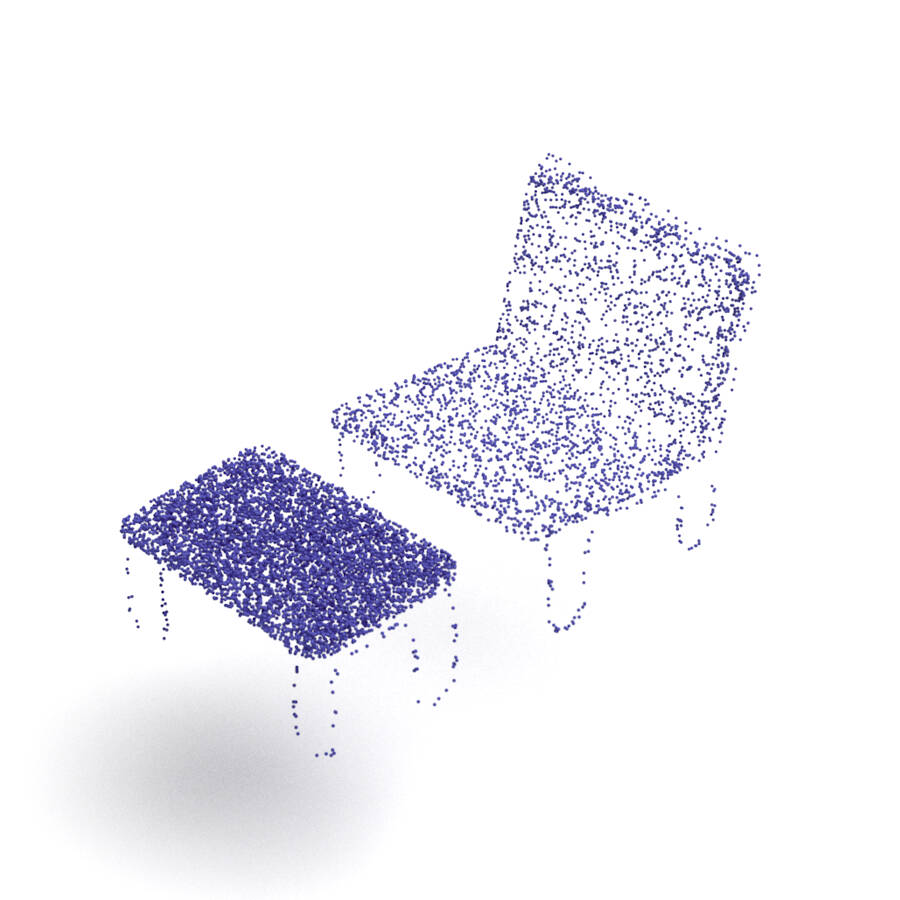}               &
    \includegraphics[width=21mm]{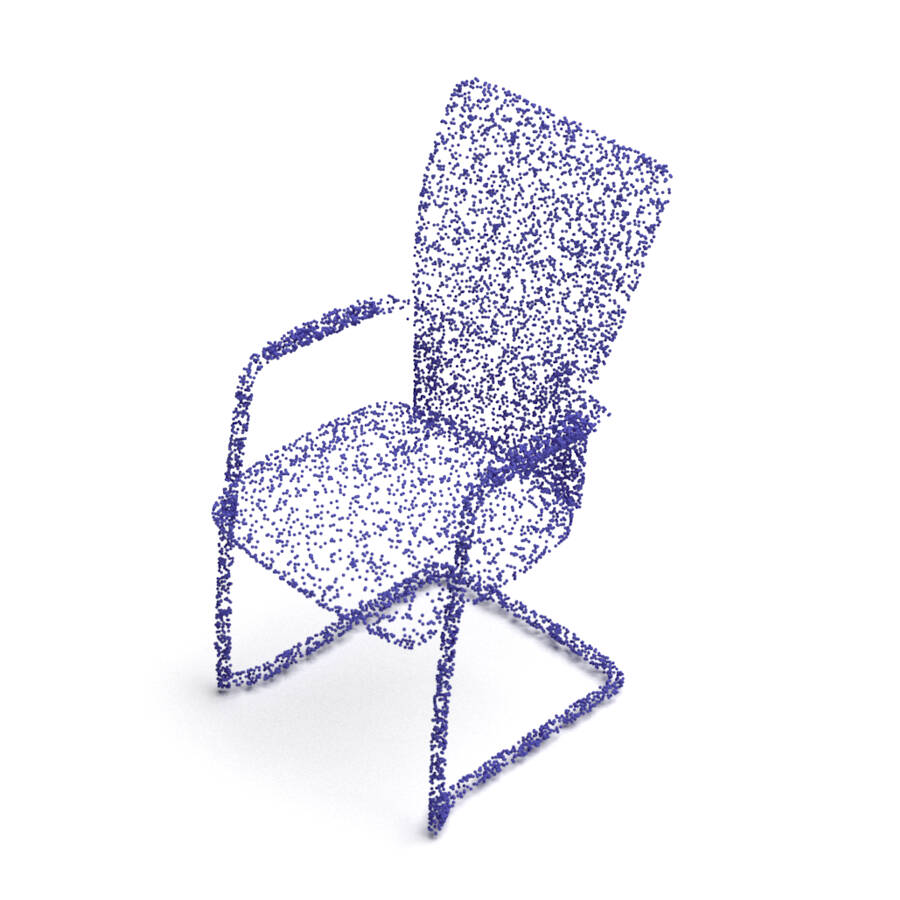}               &
    \includegraphics[width=21mm]{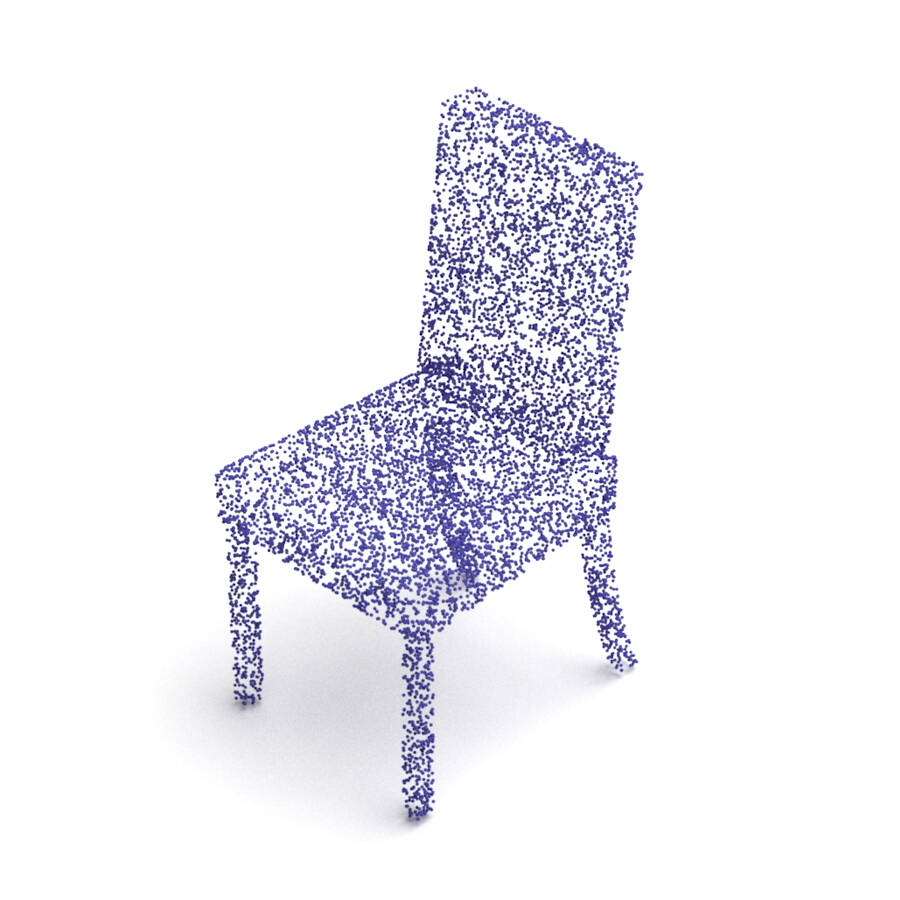}                 \\
    \rotatetitle{18mm}{\textbf{ShapeGF}}                                                        &
    \includegraphics[width=21mm]{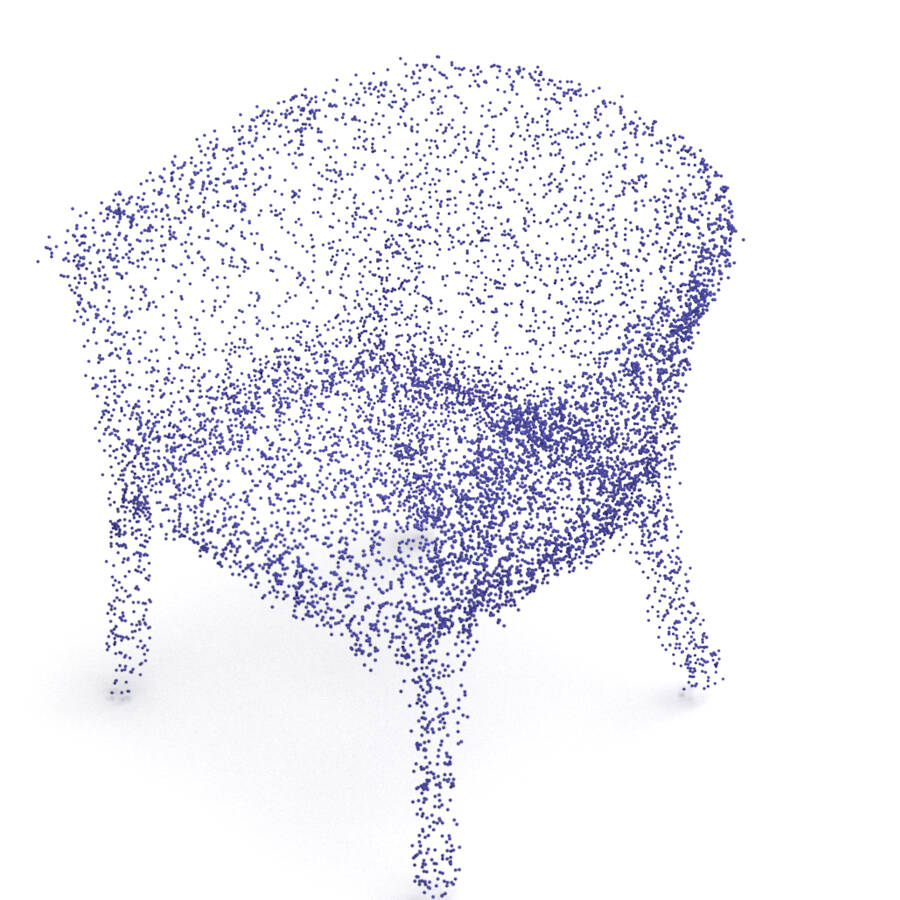} &
    \includegraphics[width=21mm]{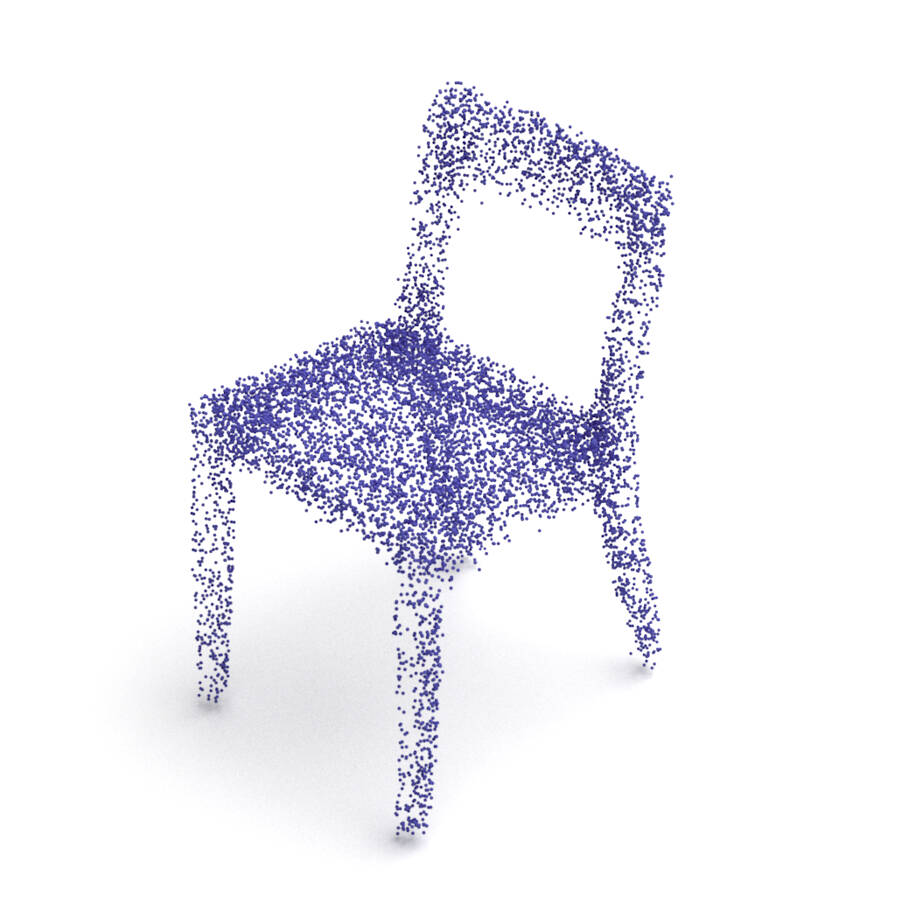}  &
    \includegraphics[width=21mm]{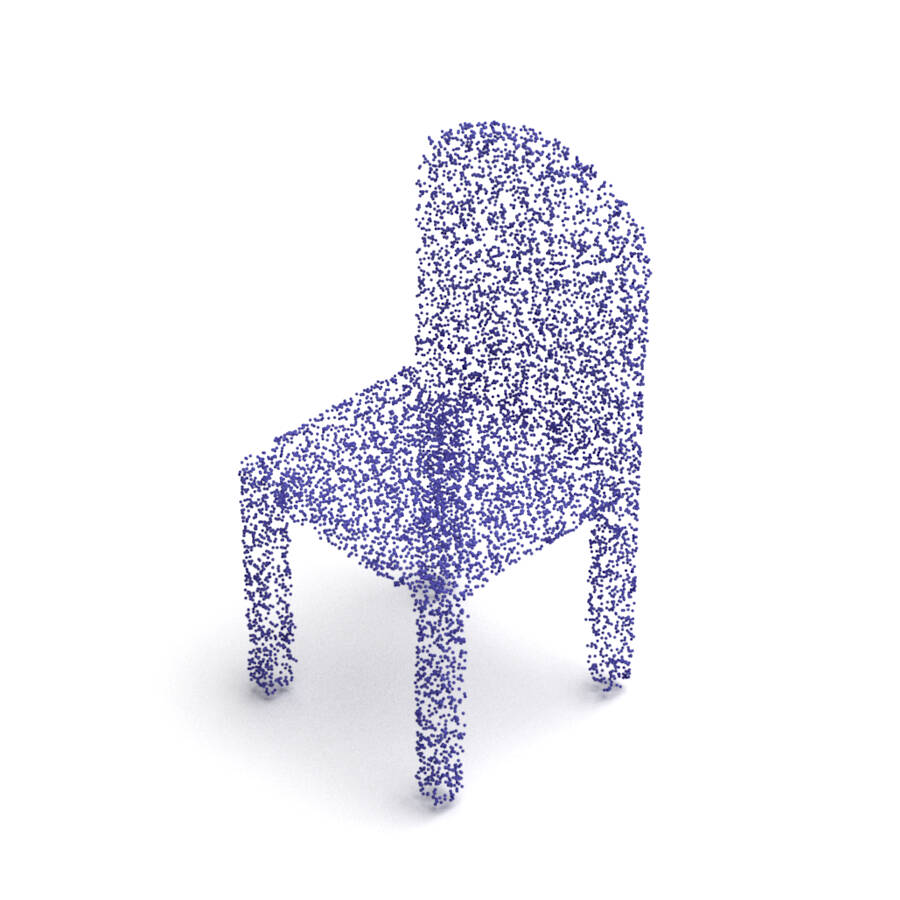}  &
    \includegraphics[width=21mm]{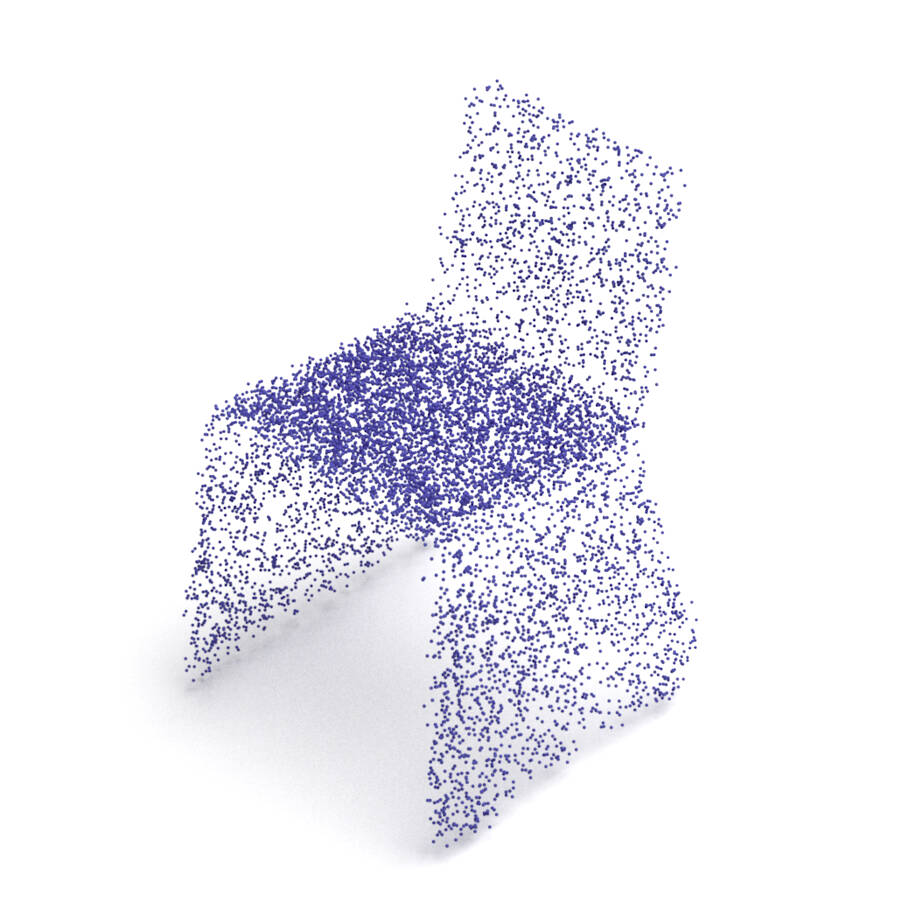} &
    \includegraphics[width=21mm]{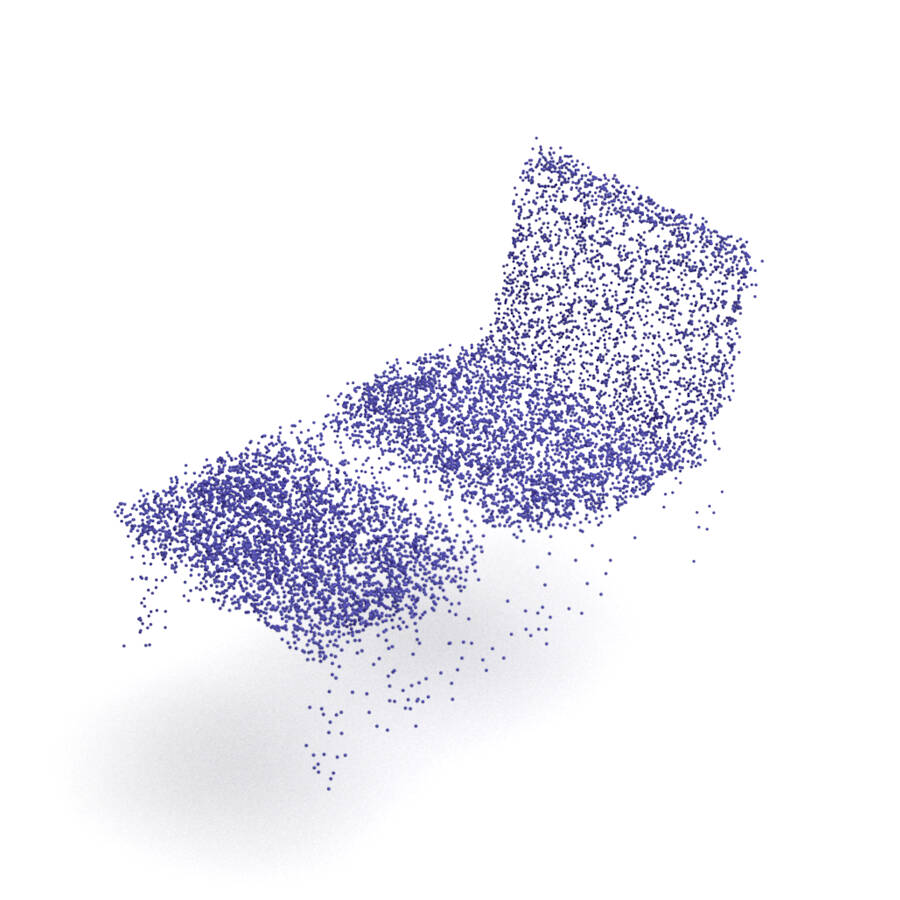} &
    \includegraphics[width=21mm]{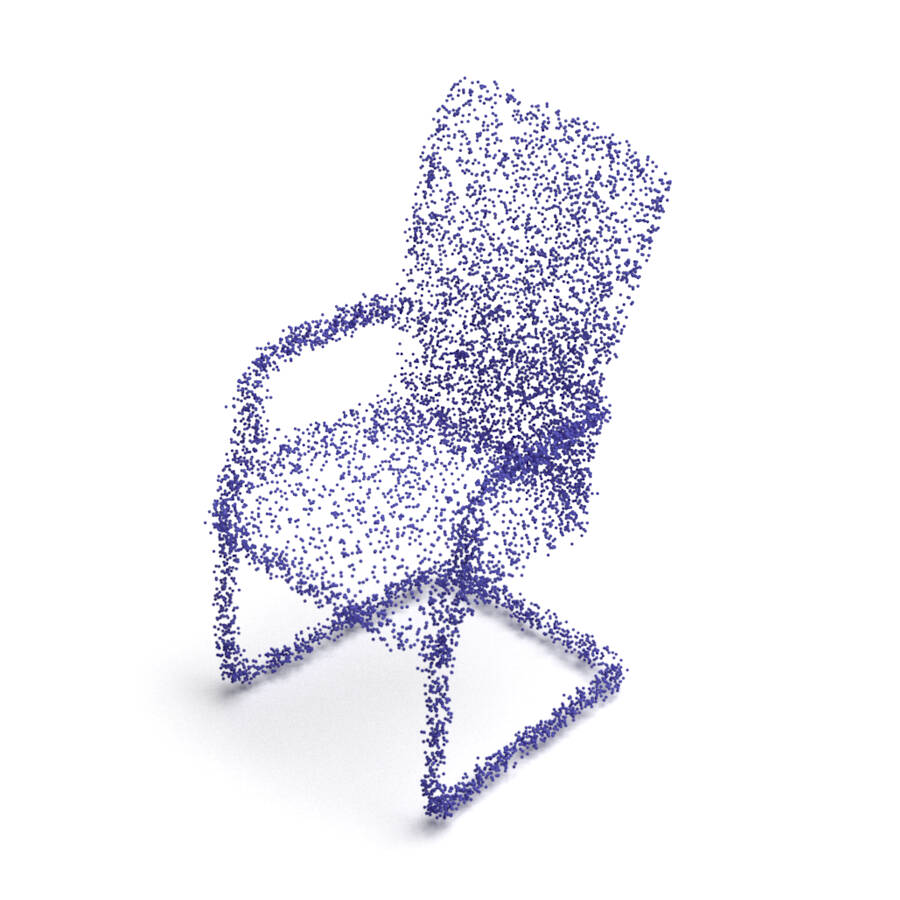} &
    \includegraphics[width=21mm]{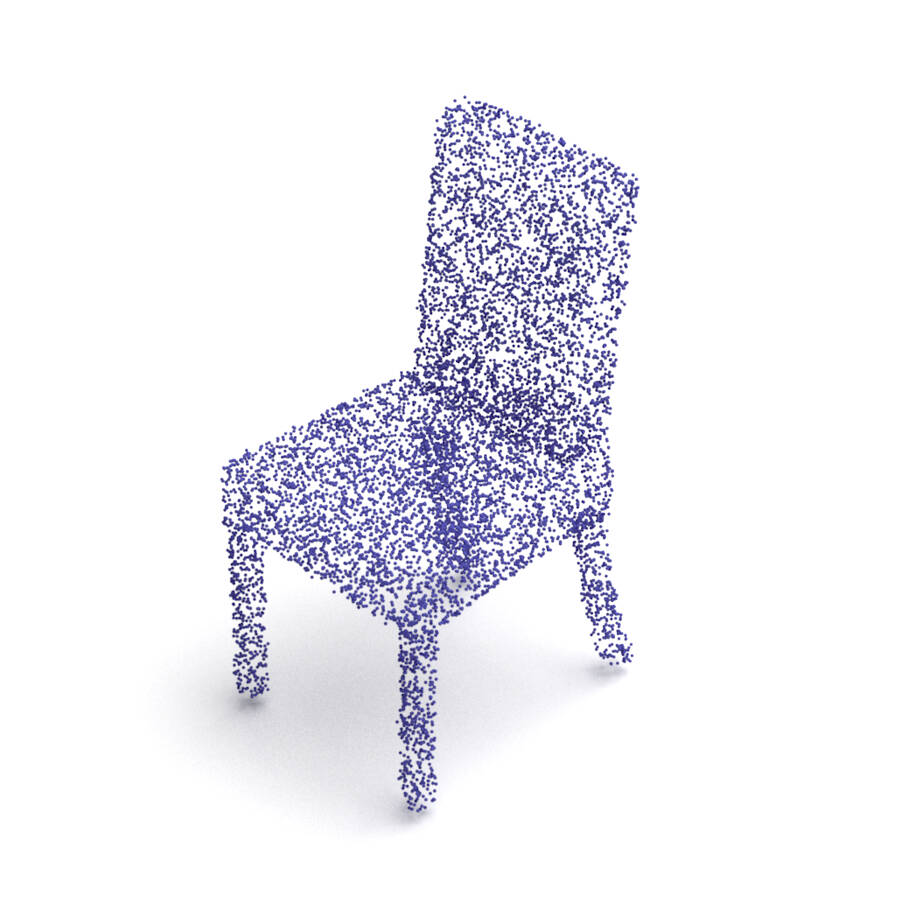}   \\
\end{tabular}
  \caption{Reconstruction examples of unseen chairs.}
  \label{fig:appendix_reconstruct_unseen_chair}
\end{figure*}

\begin{figure*}[t]
  \centering
  \setlength{\tabcolsep}{1mm}
  \begin{tabular}{lccccccc}
    \rotatetitle{18mm}{\textbf{Data}}                                                         &
    \includegraphics[width=21mm]{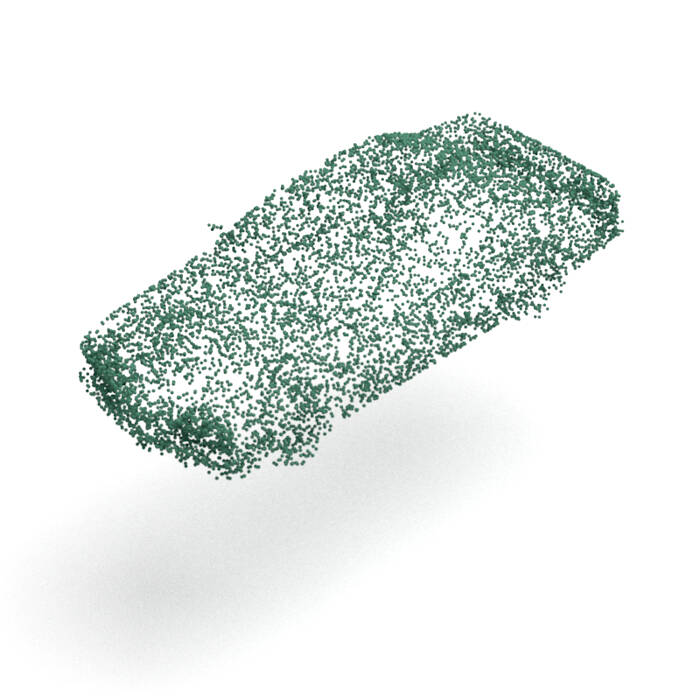}                &
    \includegraphics[width=21mm]{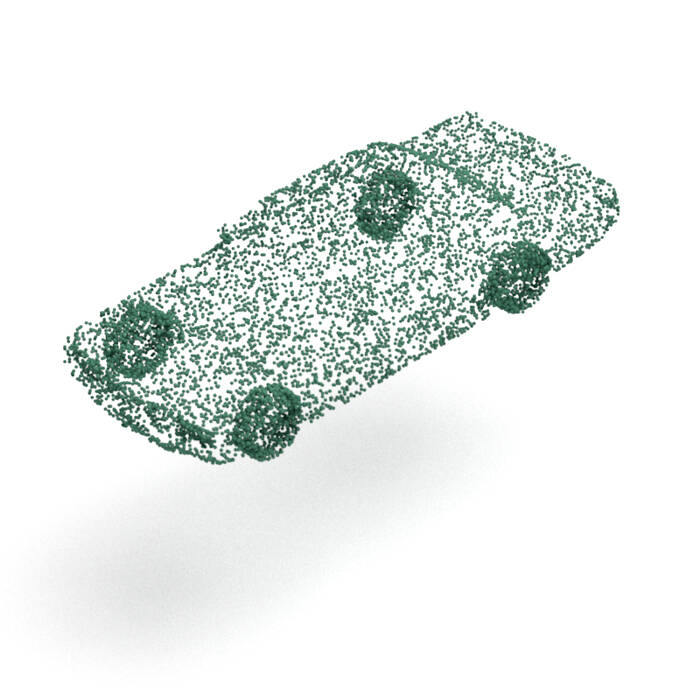}                &
    \includegraphics[width=21mm]{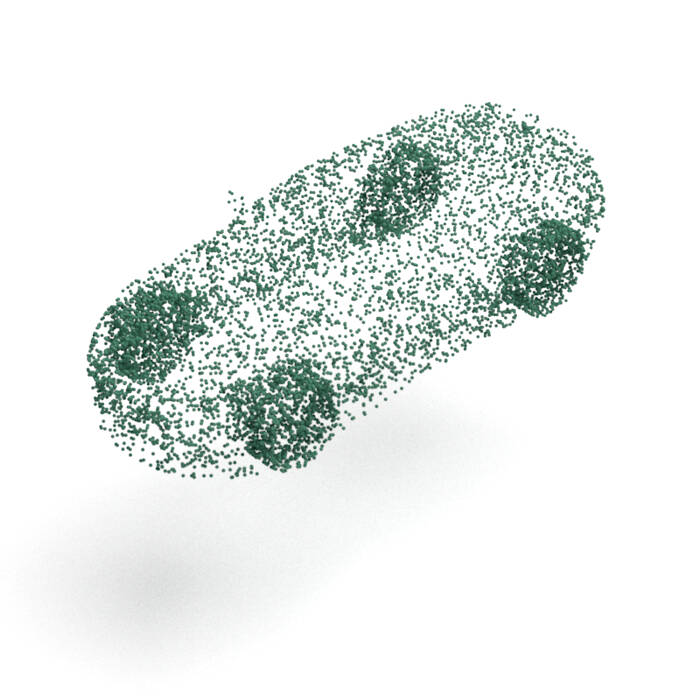}                &
    \includegraphics[width=21mm]{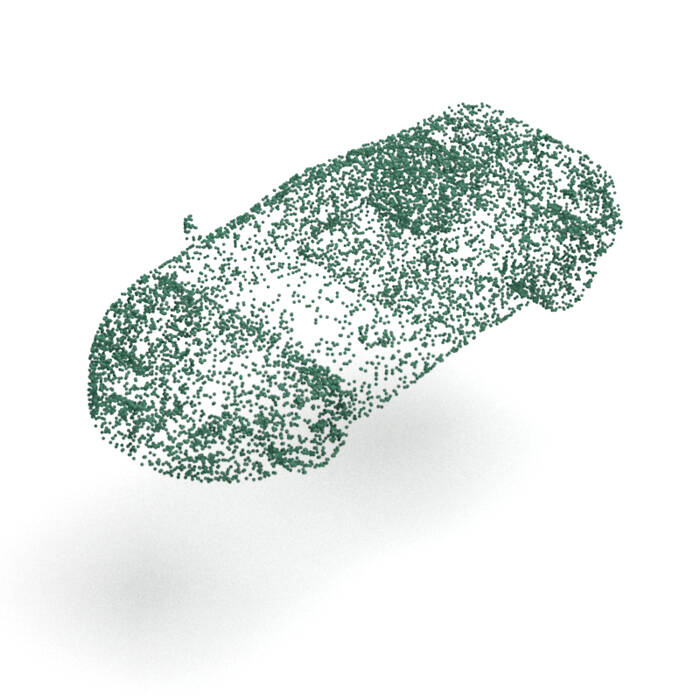}               &
    \includegraphics[width=21mm]{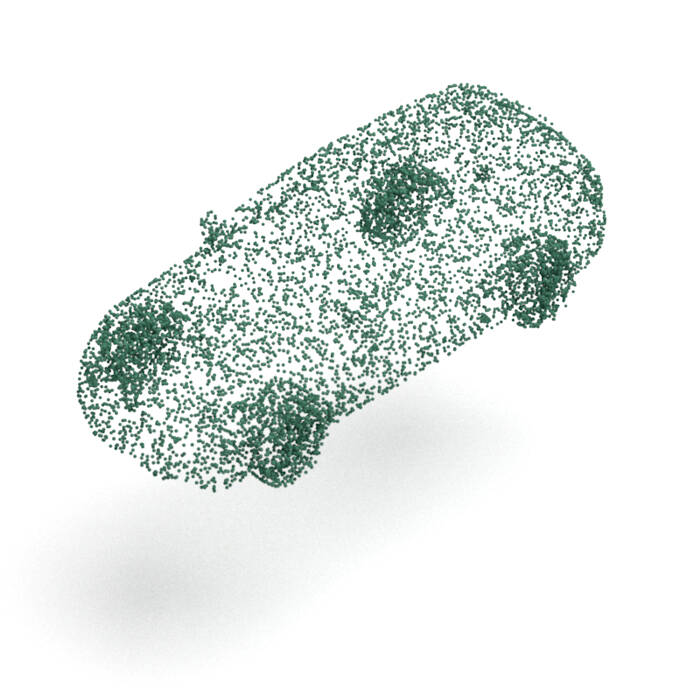}               &
    \includegraphics[width=21mm]{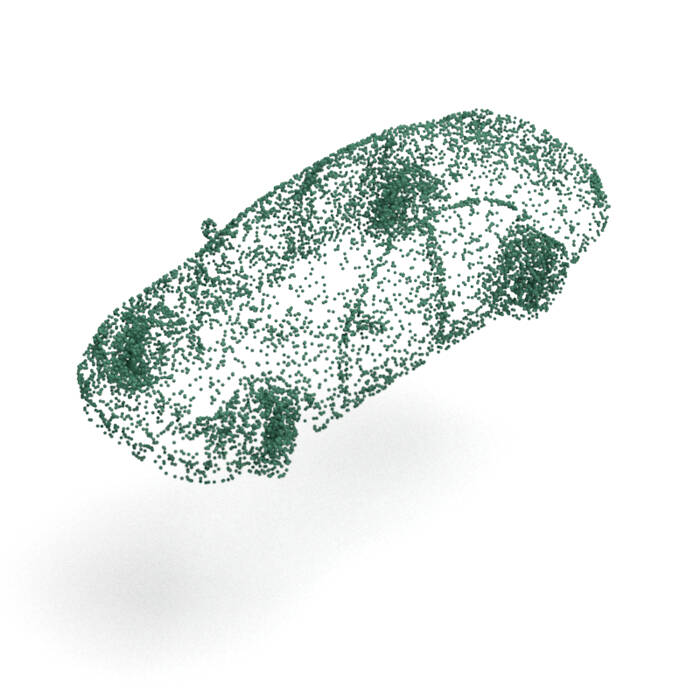}               &
    \includegraphics[width=21mm]{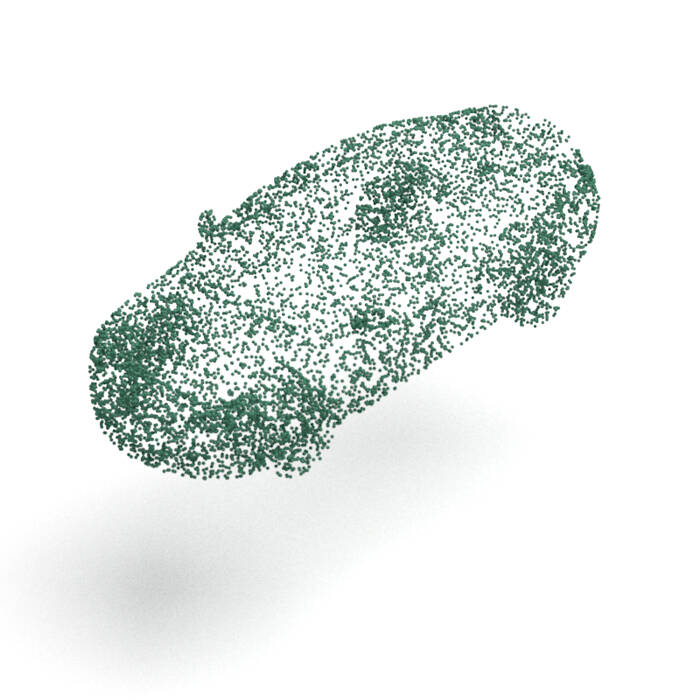}                 \\
    \rotatetitle{18mm}{\textbf{ShapeGF}}                                                      &
    \includegraphics[width=21mm]{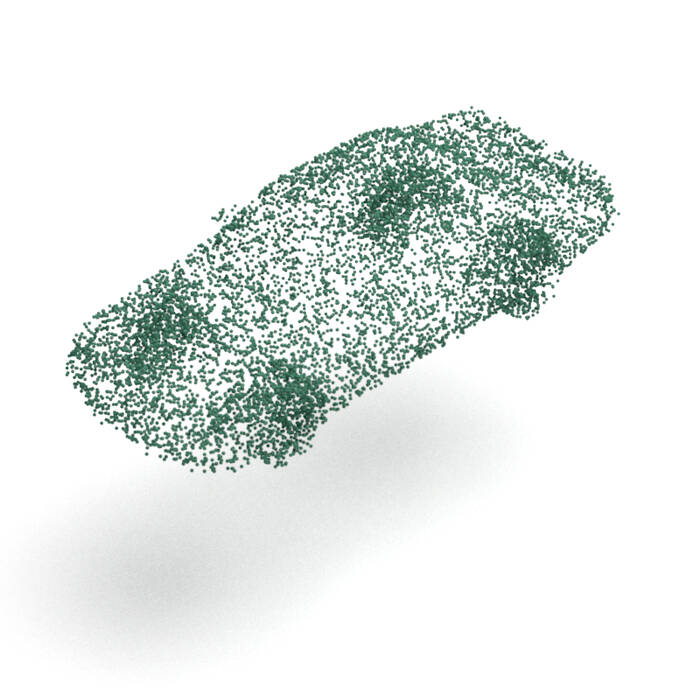}  &
    \includegraphics[width=21mm]{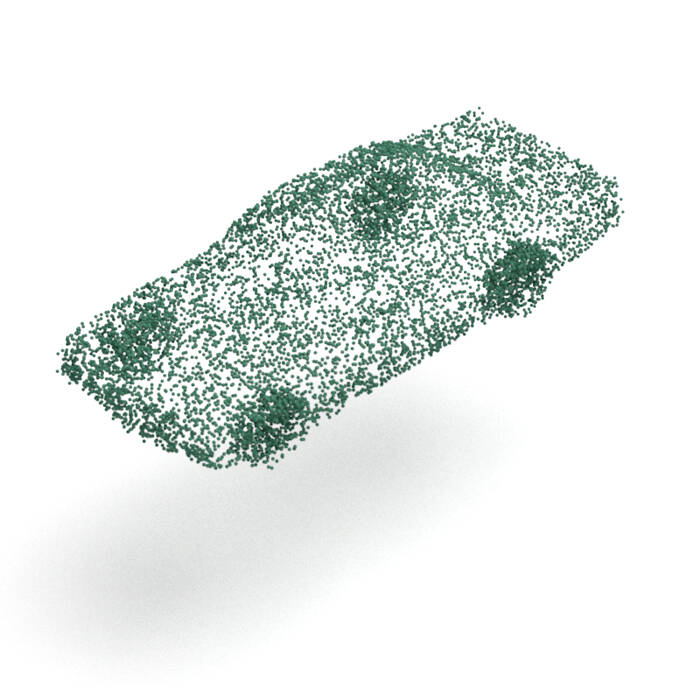}  &
    \includegraphics[width=21mm]{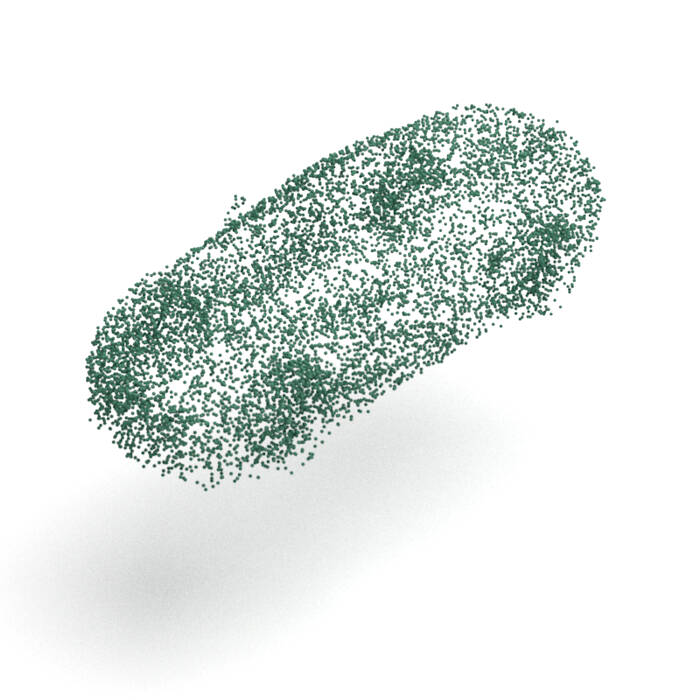} &
    \includegraphics[width=21mm]{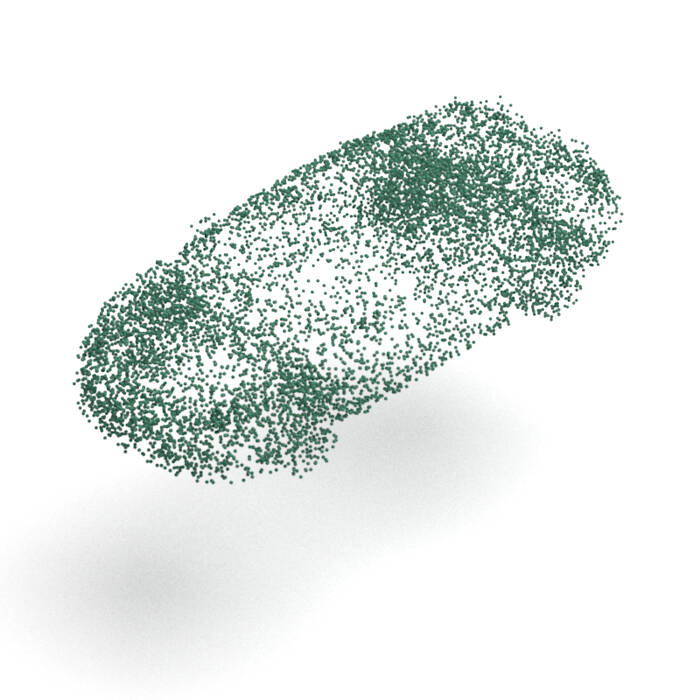} &
    \includegraphics[width=21mm]{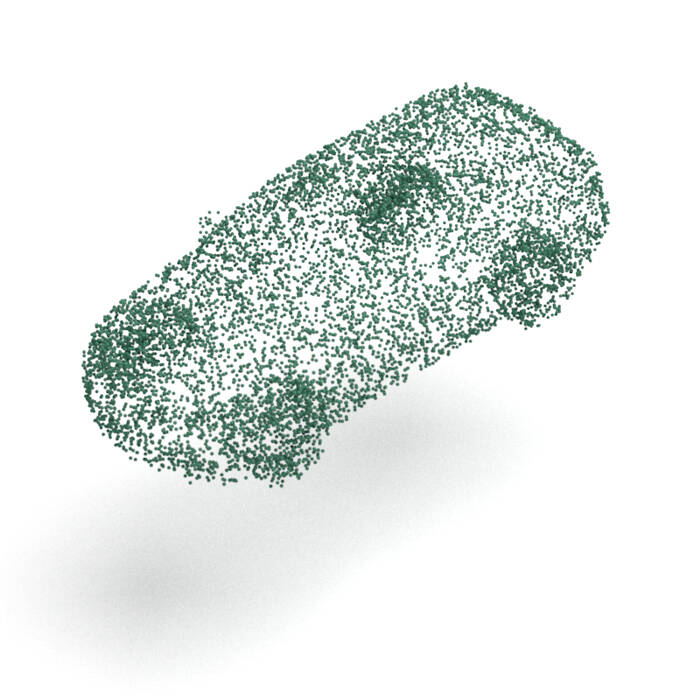} &
    \includegraphics[width=21mm]{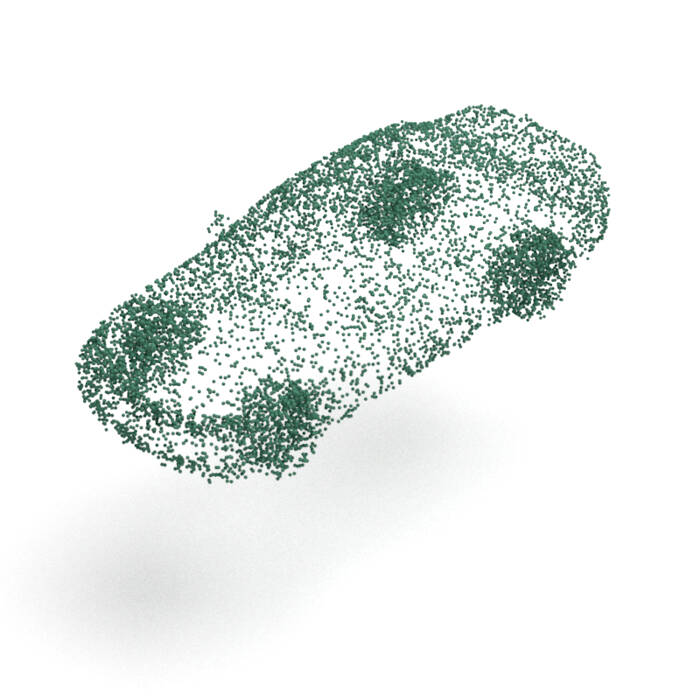} &
    \includegraphics[width=21mm]{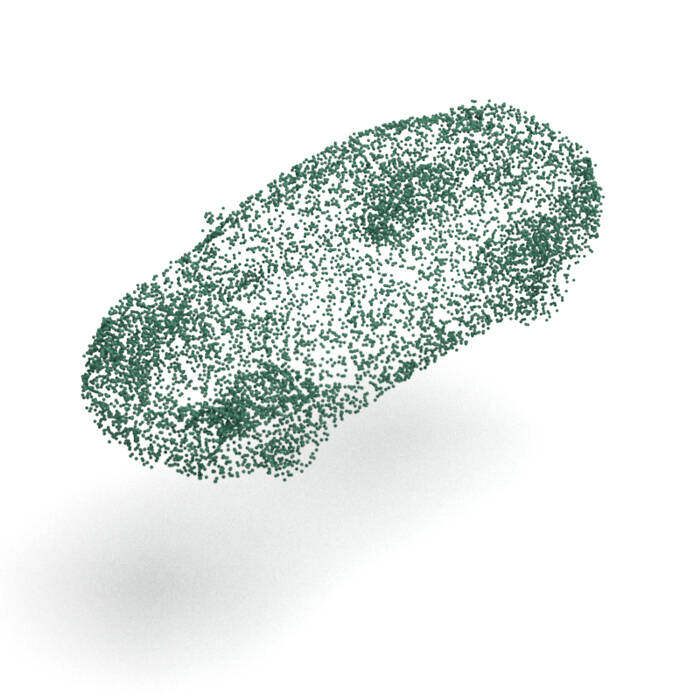}   \\
\end{tabular}
  \caption{Reconstruction examples of unseen cars.}
  \label{fig:appendix_reconstruct_unseen_car}
\end{figure*}

Figures~\ref{fig:appendix_reconstruct_seen_airplane}--\ref{fig:appendix_reconstruct_seen_car} summarize the reconstruction results of objects used for training (i.e., seen objects).
Figures~\ref{fig:appendix_reconstruct_unseen_airplane}--\ref{fig:appendix_reconstruct_unseen_car} summarize the reconstruction results of objects unused for training (i.e., unseen objects).
Due to the randomness of the point generator $F$, the reconstruction results are not completely the same as the original point clouds.

Recall that, in Fig.~\ref{fig:toydata}, PointFlow and SoftFlow generated blurred holes and intersections in the four-circle, whereas the result of ChartPointFlow is unblurred.
This tendency is true for chairs' holes in backrests, under armrests, and formed by legs in Figs.~\ref{fig:reconstruct}, \ref{fig:appendix_reconstruct_seen_chair}, and \ref{fig:appendix_reconstruct_unseen_car}.
Also in the 1st column of Fig.~\ref{fig:appendix_reconstruct_unseen_airplane}, ChartPointFlow generated rear engines of the airplane as hollow objects accurately, whereas PointFlow and SoftFlow generated rear engines as dense point clouds.
These results show that ChartPointFlow generated varying topological structures successfully.

In Fig.~\ref{fig:toydata}, PointFlow and SoftFlow generated the 2sines and double-moon suffering from string-shaped artifacts.
They generated similar artifacts near airplanes' wings in the 1st and 2nd columns of Fig.~\ref{fig:appendix_reconstruct_seen_airplane}, near chars' legs in the 4th column of Fig.~\ref{fig:appendix_reconstruct_seen_chair}, and in cars' side mirrors in the 4th and 6th columns of Fig.~\ref{fig:appendix_reconstruct_unseen_car}.
Conversely, ChartPointFlow did not.
These results show that ChartPointFlow generated protruding small subparts successfully.

\subsection{Additional Methods and Dataset}\label{app:partdataset}

Yang et al.~\cite{Yang2019} evaluated PointFlow as well as the previous works: r-GAN, l-GAN~\cite{Achlioptas2018}, and PC-GAN~\cite{Li2019}.
Kim et al.~\cite{Kim2020} ported PointFlow's codes to SoftFlow, and we did the same to ChartPointFlow and ShapeGF~\cite{Cai2020}.
Hence, the results in Tables~\ref{tab:generation_results} and \ref{tab:generation_results_all_metrics} are surely obtained under the same experimental settings.

GCN-GAN~\cite{Valsesia2019}, tree-GAN~\cite{Shu2019}, and Spectral-GAN~\cite{Ramasinghe2020} share experimental settings, which are different from those of the above-mentioned studies.
These studies employed the \texttt{PartDataset}~\cite{Yi2016} of ShapeNet for training and evaluation, did not use 1-NNA as a metric, and did not use the car category.
GCN-GAN and tree-GAN are GAN-based methods regarded as recursive super-resolutions.
Each method first generates a sparse point cloud, and then it adds more points recursively.
GCN-GAN assumed a graph structure among points and employed a graph convolution~\cite{Shu2019}.
Tree-GAN assumed a tree structure among points~\cite{Ramasinghe2020}.
Spectral-GAN is a GAN-based method that handles point clouds in the spectral domain~\cite{Ramasinghe2020}.
We also trained ChartPointFlow under the same experimental settings, and summarized the results in Table~\ref{tab:generation_results_all_metrics_partdata} when available.
ChartPointFlow outperformed there methods in terms of JSD, and MMD-EMD, and COV-EMD.
Recall that EMD is more reliable than CD; thus, ChartPointFlow is considered superior to these methods.

The experimental settings of PCGAN~\cite{Arshad2020} and PDGN~\cite{Hui2020} are unclear.
Taking their descriptions at face value, these studies compare methods evaluated using \texttt{Core} and methods evaluated using \texttt{PartDataset} in one table.
To avoid a confusing comparison, we omitted their results.

\begin{table*}[t]
  \begin{center}
\begin{tabular}{ccccccccc}
      \toprule
      \multicolumn{1}{c}{} & \multicolumn{1}{c}{} & \multicolumn{1}{c}{} & \multicolumn{2}{c}{MMD($\downarrow$)} & \multicolumn{2}{c}{COV($\%,\uparrow$)} & \multicolumn{2}{c}{1-NNA($\%$)}                                                    \\
      \cmidrule(l{.75em}r{.75em}){4-5}\cmidrule(l{.75em}r{.75em}){6-7}\cmidrule(l{.75em}r{.75em}){8-9}
      Category             & Number of Charts     & JSD($\downarrow$)    & CD                                    & EMD                                    & CD                              & EMD            & CD             & EMD            \\
      \midrule
                           & 1                    & 3.54                 & 0.221                                 & 3.15                                   & 49.63                           & \textbf{53.21} & 72.67          & 68.90          \\
                           & 4                    & 3.62                 & 0.220                                 & 3.11                                   & 48.89                           & 51.79          & 71.77          & 67.30          \\
                           & 8                    & 3.39                 & 0.217                                 & 3.08                                   & 49.66                           & 51.70          & 70.90          & 66.54          \\
                           & 12                   & 3.60                 & 0.213                                 & 3.06                                   & 48.40                           & 51.73          & 70.20          & 65.99          \\
      Airplane             & 16                   & 3.93                 & 0.215                                 & 3.07                                   & 49.52                           & 51.08          & 70.72          & 66.48          \\
\bottomrule
    \end{tabular}
  \end{center}
  \caption{Generation performances with different numbers of charts.
    The scores are multiplied by $10^{2}$ for JSD and MMD-EMD, and by $10^{3}$ for MMD-CD.
    $\uparrow$ denotes that a higher score is better.
    $\downarrow$ denotes that a lower score is better.}
  \label{tab:generation_results_chart}
\end{table*}

\begin{table}[t]
  \begin{center}
    \begin{tabular}{cccc}
      \toprule
      Category & Number of Charts & CD            & EMD           \\
      \midrule
               & 1                & 1.18          & 2.64          \\
               & 4                & 1.13          & 2.40          \\
               & 8                & 1.13          & 2.32          \\
               & 12               & 1.14          & 2.30          \\
      Airplane & 16               & 1.09          & 2.26          \\
               & 20               & 1.08          & 2.25          \\
               & 24               & \textbf{1.07} & \textbf{2.23} \\
\bottomrule
    \end{tabular}
  \end{center}
  \caption{Reconstruction performance evaluated through CD ($\times 10^{4}$) and EMD ($\times 10^{2}$).}
  \label{tab:reconstruction_results_chart}
\end{table}

\begin{table}[t]
  \begin{center}
    \begin{tabular}{cccc}
      \toprule
      Category & Number of Charts & NMI           & purity        \\
      \midrule
               & 4                & 0.29          & 0.63          \\
               & 8                & \textbf{0.33} & 0.76          \\
               & 12               & \textbf{0.33} & 0.79          \\
      Airplane & 16               & 0.31          & 0.79          \\
               & 20               & 0.31          & 0.80          \\
               & 24               & 0.30          & 0.80          \\
               & 28               & 0.30          & \textbf{0.81} \\
               & 32               & 0.29          & \textbf{0.81} \\
      \midrule
               & 4                & 0.23          & 0.65          \\
               & 8                & 0.31          & 0.71          \\
               & 12               & \textbf{0.39} & 0.85          \\
      Chair    & 16               & 0.35          & 0.84          \\
               & 20               & 0.36          & \textbf{0.86} \\
               & 24               & 0.35          & \textbf{0.86} \\
               & 28               & 0.34          & 0.85          \\
               & 32               & 0.32          & 0.84          \\
      \midrule
               & 4                & 0.10          & 0.71          \\
               & 8                & 0.15          & 0.71          \\
               & 12               & 0.18          & 0.72          \\
      Car      & 16               & \textbf{0.19} & 0.74          \\
               & 20               & 0.17          & 0.75          \\
               & 24               & 0.18          & \textbf{0.79} \\
               & 28               & 0.18          & 0.77          \\
               & 32               & 0.19          & \textbf{0.79} \\
      \bottomrule
    \end{tabular}
  \end{center}
  \caption{Segmentation performance evaluated through NMI and purity. Larger is better.}
  \label{tab:segmentation_results_chart}
\end{table}

\begin{table*}[t]
  \begin{center}
    \begin{tabular}{clccccccc}
      \toprule
      \multicolumn{1}{c}{} & \multicolumn{1}{c}{}              & \multicolumn{1}{c}{} & \multicolumn{2}{c}{MMD($\downarrow$)} & \multicolumn{2}{c}{COV($\%,\uparrow$)} & \multicolumn{2}{c}{1-NNA($\%$)}                                                    \\
      \cmidrule(l{.75em}r{.75em}){4-5}\cmidrule(l{.75em}r{.75em}){6-7}\cmidrule(l{.75em}r{.75em}){8-9}
      Category             & Model                             & JSD($\downarrow$)    & CD                                    & EMD                                    & CD                              & EMD            & CD             & EMD            \\
      \midrule
                           & r-GAN~\cite{Achlioptas2018}       & 7.44                 & 0.261                                 & 5.47                                   & 42.72                           & 18.02          & 93.58          & 99.51          \\
                           & l-GAN (CD)~\cite{Achlioptas2018}  & 4.62                 & 0.239                                 & 4.27                                   & 43.21                           & 21.23          & 86.30          & 97.28          \\
                           & l-GAN (EMD)~\cite{Achlioptas2018} & 3.61                 & 0.269                                 & 3.29                                   & 47.90                           & 50.62          & 87.65          & 85.68          \\
      Airplane             & PC-GAN~\cite{Li2019}              & 4.63                 & 0.287                                 & 3.57                                   & 36.46                           & 40.94          & 94.35          & 92.32          \\
                           & ShapeGF~\cite{Cai2020}            & 4.77                 & \textbf{0.214}                        & 3.29                                   & 47.64                           & 45.17          & 73.85          & 81.44          \\
\bottomrule
    \end{tabular}
  \end{center}
  \caption{Generation performances.
    The scores are multiplied by $10^{2}$ for JSD and MMD-EMD, and by $10^{3}$ for MMD-CD.
    $\uparrow$ denotes that a higher score is better.
    $\downarrow$ denotes that a lower score is better.}
  \label{tab:generation_results_all_metrics}
\end{table*}

\begin{table*}[t]
  \begin{center}
    \begin{tabular}{clccccc}
      \toprule
      \multicolumn{1}{c}{} & \multicolumn{1}{c}{}               & \multicolumn{1}{c}{}              & \multicolumn{2}{c}{MMD($\downarrow$)} & \multicolumn{2}{c}{COV($\%,\uparrow$)}                                                  \\
      \cmidrule(l{.75em}r{.75em}){4-5}\cmidrule(l{.75em}r{.75em}){6-7}
      Category             & Model                              & JSD($\downarrow$)                 & CD                                    & EMD                                    & CD                         & EMD               \\
      \midrule
                           & GCN-GAN~\cite{Valsesia2019}        & \hspace*{.46em}8.3\hspace*{.46em} & 0.8\hspace*{.46em}                    & \hspace*{.46em}7.1\hspace*{.46em}      & 31\hspace*{1.1em}          & 14\hspace*{1.1em} \\
                           & tree-GAN~\cite{Shu2019}            & \hspace*{.46em}9.7\hspace*{.46em} & 0.4\hspace*{.46em}                    & \hspace*{.46em}6.8\hspace*{.46em}      & \textbf{61}\hspace*{1.1em} & 20\hspace*{1.1em} \\
      Airplane             & Spectral-GAN~\cite{Ramasinghe2020} & ---                               & \textbf{0.2}\hspace*{.46em}           & \hspace*{.46em}5.7\hspace*{.46em}      & ---                        & ---               \\
      \cmidrule(){2-7}                                                                                                                                                                                                                \\[-4.4mm]
                           & ChartPointFlow (proposed)          & \hspace*{.46em}\textbf{3.14}      & 0.50                                  & \hspace*{.46em}\textbf{3.86}           & 44.56                      & \textbf{46.77}    \\
      \midrule
                           & GCN-GAN~\cite{Valsesia2019}        & 10.0                              & 2.9\hspace*{.46em}                    & \hspace*{.46em}9.7\hspace*{.46em}      & 30\hspace*{1.1em}          & 26\hspace*{1.1em} \\
\bottomrule
    \end{tabular}
  \end{center}
  \caption{Generation performances on PartDataset, ShapeNet.
    The scores are multiplied by $10^{2}$ for JSD and MMD-EMD, and by $10^{3}$ for MMD-CD.
    $\uparrow$ denotes that a higher score is better.
    $\downarrow$ denotes that a lower score is better.}
  \label{tab:generation_results_all_metrics_partdata}
\end{table*}

\begin{table}\centering
  \begin{tabular}{clcc}
    \toprule
    Category & Model                                & CD            & EMD           \\
    \midrule
             & ShapeGF~\cite{Cai2020}               & \textbf{0.98} & 2.55          \\
             & AtlasNet~\cite{Groueix2018}          & 1.01          & 2.95          \\
    Airplane & AtlasNet V2 (PD)~\cite{Deprelle2019} & 1.15          & 3.28          \\
             & AtlasNet V2 (PT)~\cite{Deprelle2019} & 1.01          & 3.57          \\
             & PointFlow~\cite{Yang2019}            & 1.21          & 2.77          \\
             & SoftFlow~\cite{Kim2020}              & 1.19          & 2.60          \\
    \cmidrule(){2-4}
             & ChartPointFlow                       & 1.07          & \textbf{2.23} \\
\bottomrule
  \end{tabular}
  \caption{Reconstruction performances evaluated through CD ($\times 10^{4}$) and EMD ($\times 10^{2}$).}
  \label{tab:reconstruction_results_all_metrics}
\end{table}

\clearpage

\end{document}